# DEVELOPMENT AND TESTING OF NOVEL SOFT SLEEVE ACTUATORS

A Thesis submitted

in partial fulfillment of the requirements for the degree of

Doctor of Philosophy in
Mechanical Engineering

By

Mohammed Abboodi

January 2025

Ottawa-Carleton Institute for
Mechanical and Aerospace Engineering

University of Ottawa
Ottawa, Ontario, Canada K1N 6N5

# ABSTRACT

As societies confront the challenges of aging populations and the growing prevalence of neurological and musculoskeletal disorders, the demand for mobility-assistive technologies that restore limb control or enhance physical abilities has become increasingly urgent. Many existing assistive devices rely on incompatible technological solutions, rendering them impractical and uncomfortable for users. This research seeks to address these limitations by advancing the development of soft and wearable actuators, which offer significant potential for creating functional and user-friendly mobility-assistive devices.

Traditional soft actuators are valued for their adaptability and compliance but face significant constraints in generating and transferring forces to the user. These limitations often necessitate complex interface mechanisms, which detract from both device effectiveness and user comfort. To overcome these challenges, this study introduces an innovative soft sleeve actuator design and presents three novel actuator models capable of performing linear, bending, and twisting motions while generating the required mechanical forces and moments. Additionally, the study unveils the Omnidirectional actuator, a breakthrough design that combines the motions of the original models.

The newly developed actuators feature an intricate design that requires a custom fused deposition manufacturing process. This tailored approach ensures precise production while effectively addressing the challenge of air leakage, a problem to resolve with traditional manufacturing methods. Moreover, customized testing setups are developed to assess the actuators' performance and explore how variations in geometric parameters and material properties affect the actuator’s ability to generate kinematic and kinetic outputs.

In summary, this research introduces a significant innovation to soft actuation, enabling the development of wearable devices capable of executing complex motions without relying on complex interface mechanisms. By achieving superior kinematic and kinetic performance, these advancements contribute significantly to the evolution of assistive mobility technologies, paving the way for more effective and user-centric solutions.

# ACKNOWLEDGEMENTS

First and foremost, I wish to express my deepest gratitude to my supervisor, Dr. Marc Doumit. His exceptional support, insightful guidance, and unwavering encouragement have been invaluable throughout this study. I am profoundly indebted to his mentorship, which has not only provided the inspiration to pursue this field of research but also significantly enriched my academic journey. His boundless generosity with his time and expertise has been instrumental in the successful completion of this work.

I would also like to extend my sincere appreciation to the University of Ottawa, particularly the Department of Mechanical Engineering, along with all its faculty members and support staff. Their assistance and encouragement over the years have been essential in helping me achieve my academic goals.

Finally, I express my heartfelt love and gratitude to my mother and father for their continuous and unwavering support and encouragement throughout the duration of my studies. Their belief in me has been a constant source of strength and motivation.

# TABLE OF CONTENTS

# LIST OF FIGURES

# LIST OF TABLES

# CHAPTER 1
# INTRODUCTION

## 1.1 Introduction

In recent decades, wearable mobility assistive devices (WMAD) have experienced rapid growth in applications aimed at enhancing human mobility and independence [1]. Originally developed for military and industrial uses, such as carrying heavy loads or reducing fatigue over long distances [2], these devices have since expanded to address a broad range of mobility challenges.

Most current WMAD are designed as rigid exoskeletons that interface with the user's body to support mobility. However, this reliance on rigid, robotic-inspired designs presents several limitations. The primary concern is the inherent discomfort and risk of injury caused by poor compatibility with biological joints [3]. Additionally, the bulk and inertia of rigid exoskeletons can interfere with natural gait patterns, leading to discomfort and fatigue [3]. These drawbacks contribute to a high user rejection rate, raising concerns about the overall effectiveness of current WMAD in improving mobility.

To overcome these challenges, there has been a shift towards developing soft WMAD that employ novel actuator technologies [3], [4], [5]. Unlike their rigid counterparts, soft WMAD uses flexible materials and soft interfaces for direct attachment to the body [6]. This design minimizes interference with the user's natural gait and enhances safety and comfort. However, the promise of soft WMAD remains largely unrealized due to the lack of an adequate actuation mechanism. The ideal actuator needs to be soft, safe, lightweight, yet powerful and energy efficient.

Current actuator technologies, such as shape memory alloys (SMAs) and electroactive polymers (EAP), fall short of these requirements. For instance, SMAs offer high force but are hampered by slow actuation speeds and limited recoverable deformation [7]. Although responsive, EAP suffer from low power-to-volume ratios [8]. Pneumatic actuators have shown promise in mitigating many of these limitations, but they come with their own set of challenges, including inefficient actuation due to radial expansion, low force production, and lack of precision in actuator movements [9].

This thesis introduces the concept of the soft sleeve actuator (SSA) and proposes four novel actuator designs: linear, bending, twisting and omnidirectional actuators. Unlike existing technologies, the SSAs are soft, wearable, and can be applied directly to the user's anatomy, eliminating the need for intermediary mechanisms such as braces or textile layers [10]. Additionally, the SSAs deliver adequate kinetic and kinematic performance while remaining lightweight, representing a significant advancement in the field of WMAD.

### 1.2 Research Objectives

The goal of this research is to conceptualize, design, fabricate, and evaluate novel SSAs capable of generating a range of motions and forces suitable for WMAD. The proposed actuator should be able to contract, extend, bend, and twist while meeting essential kinetic and kinematic requirements. Additionally, the research aims to achieve a comprehensive mechanical characterization of the newly proposed SSAs through comprehensive experimental testing and preliminary analytical modeling.

### 1.3 Methodology

This thesis employs a structured research methodology designed to address the study's primary objectives. The following is an outline of the sequential steps involved:

**Literature review**: The first phase entails a comprehensive review of existing literature on WMAD systems and soft actuator mechanisms. The goal is to identify unexplored areas of research and knowledge gaps, which will inform the scope and objectives of this study.

**Conceptual designs**: Upon identifying the research gaps, design requirements are identified, and various actuator concepts are brainstormed. The finalized concepts are developed through experimentation.

**Manufacturing process**: 3D printing technology is utilized to address and overcome the challenges encountered with various traditional manufacturing processes. The 3D printing optimization process is developed to achieve effective production of actuators with both effective airtightness and mechanical properties.

**Prototyping and testing**: Prototypes are manufactured using the optimized 3D printing process and then subjected to mechanical testing. Comprehensive experimental setups are developed in the laboratory to evaluate the kinematics and kinetics performance of the SSAs.

**Experimental characterization**: A systematic experimental study is undertaken to study the influence of design geometric parameters and material properties on the SSAs' performance.

**Analytical modeling**: Analytical models are developed to provide a theoretical framework for understanding the actuators' behavior under varying design parameters and operating variables.

**Discussion and conclusions**: Based on the outcomes of experimental characterization, the performance of the SSAs is evaluated, and a theoretical explanation is presented to clarify how variations in geometric parameters and material properties affect their functionality.

### 1.4 Contribution

The principal contributions of this thesis are fourfold, addressing a significant gap in the field of WMAD and soft actuation:

**Development of the first SSA:** This thesis introduces the first SSAs, marking an initial innovation in the field of soft actuation. The study presents a set of actuators that include linear, bending, and twisting. The study also proposes the Omnidirectional actuator, a breakthrough design that combines the motions of all original actuators. A Journal paper in IEEE Access [10] has been published covering the design and testing of the linear SSA. A journal paper covering the design and testing of the Omnidirectional actuator has been submitted and is currently under review.

**Development of manufacturing process:** Producing the SSA using 3D printing technology was a challenge and an innovation in this field. A comprehensive framework is introduced to optimize the manufacturing process, specifically addressing the challenges associated with flexible materials in Bowden extruder configurations while producing airtight pneumatic actuators. A journal paper covering the manufacturing optimization framework is in progress.

**Development of experimental testing platforms:** Testing the SSA prototypes demanded the custom design of an experimental testing platform that included the integration of complex sensing and operating functionalities. This platform has been developed in the laboratory by the candidate to evaluate the SSA prototypes under various types of actuations, providing a robust basis for experimental testing and investigation.

**Experimental mechanical characterization of SSA**: The thesis achieved a comprehensive experimental investigation into the impact of SSA design and operating parameters on actuator performance. This research offers valuable insights into how different materials and geometrical configurations influence the newly developed SSA behavior, contributing to the optimization of actuator design and functionality. A journal paper covering the mechanical characterization of SSA is in progress.

## 1.5 Thesis Outline

This thesis is organized into a series of chapters that address the key components of the research.

Chapter 1 introduces the study, outlining the motivation driving the research and highlighting its significant contributions to the field of WMAD and soft actuation.

Chapter 2 presents a comprehensive literature review, providing an overview of existing WMAD systems, soft actuator mechanisms, and the materials commonly used in their fabrication.

Chapter 3 delves into the design and development of SSA, explaining the operational principles underlying these mechanisms.

Chapter 4 begins by detailing the initial manufacturing techniques attempted and their challenges for producing SSAs. It then transitions to describing the final manufacturing 3D printing process developed for producing soft airtight actuators.

Chapter 5 presents the experimental platform developed for the evaluation of SSA, providing a detailed account of its design and functionalities.

Chapter 6 outlines the experimental setup and procedures used to test the SSA, followed by a discussion of the results, with a focus on how variations in geometric parameters and material stiffness influence actuator performance.

Chapter 7 introduces the geometrical and static models developed for the linear SSA, offering theoretical insights into its structural and functional attributes.

Chapter 8 showcases the dynamic abilities and accuracy of the SSA. The chapter also presents the Proportional-Integral-Derivative controller utilized in this study, detailing its role in enhancing the control and responsiveness of the linear SSA.

Chapter 9 provides a detailed evaluation of the sleeve actuators by integrating experimental findings from earlier chapters with analytical models. Through quantitative analysis, this chapter examines how variations in geometric parameters and material properties influence actuator performance.

Chapter 10 concludes the thesis by summarizing the primary findings and contributions and offering recommendations for future research directions.

### 1.6 Conclusion

In conclusion, this chapter has established the foundation for this research by emphasizing the critical need for more advanced WMAD and examining the constraints of current rigid and soft actuator technologies. In particular, the chapter introduced the primary goal of developing novel Soft Sleeve Actuators capable of generating substantial, precise motion while maintaining user comfort and safety. It also outlined the research objectives, explained the methodological approach, and highlighted the anticipated contributions stemming from the successful fulfillment of these objectives. The thesis structure provides a robust basis for the subsequent chapters, guiding the reader through the fundamental concepts, experimental analyses, and theoretical evaluations that collectively validate the feasibility and effectiveness of the proposed SSAs.

# CHAPTER 2

# LITERATURE REVIEW

## 2.1 Exoskeletons

Exoskeletons, or WMAD, are designed to augment human capabilities by interacting with the user's natural movements. Unlike prosthetics, which replace missing body parts, exoskeletons encapsulate the body's extremities, enhancing natural mobility and function. Constructed with mechanical systems that mimic human joint structures, they incorporate both active and passive actuators. Their multifaceted applications span from assisting the elderly and individuals with disabilities to aiding rehabilitation processes and elevating the performance of able-bodied individuals [11].

Historically, the concept of WMAD can be traced back to the late 19th century with Nicholas Yagn's design. His prototype harnessed leg-length bow springs to capture and utilize the body's downward energy [12]. By the 1960s, a significant leap in innovation was evidenced by General Electric's Hardiman, an industrial-focused powered exoskeleton. It had the potential to amplify users' strength, allowing them to lift 680 kg by employing a combination of electric and hydraulic actuation. However, its size, weight, stability issues, and excessive power consumption rendered it largely impractical for widespread use [12].

Advancements in actuation mechanisms have been pivotal in the contemporary era. The Berkeley Exoskeleton (BLEEX), equipped with linear hydraulic actuators, is particularly noteworthy. With its four hydraulic joints (two at the hips, one at the knee, and one at the ankle), it aims to enhance lower limb functionality in healthy users [13]. Similarly, Tsukuba University developed the Hybrid Assistive Limb (HAL), with designs accommodating varied needs. The HAL series, featuring servo motors and Harmonic Drive gears, assists in locomotion but is limited in operational duration and is not universally applicable [14]. Another notable example is Argo Medical Technologies' ReWalk, specifically designed for spinal cord injury survivors. Unlike HAL, it offers extended battery life after a single charge but lacks a balance mechanism [12].

Despite these advancements, the predominant exoskeleton designs still rely on rigid-based actuators. While effective, these actuators introduce challenges due to their weight, bulk, cost, and high energy consumption. The added weight alters the body's inertia, thereby increasing metabolic costs. Moreover, another significant challenge with these rigid exoskeletons is the need for precise alignment with the user's biological joints, which requires substantial adjustment time [12]. Although contemporary rigid exoskeletons showcase remarkable potential, they bear inherent limitations. The future trajectory suggests a shift towards soft exoskeletons, which hold the potential for heightened adaptability, reduced mass, and potentially diminished energy requirements.

### 2.2. Soft Exoskeletons

In the field of WMAD, a novel paradigm has emerged in the form of soft exoskeletons conceptualized to address the limitations inherent in their rigid predecessors [15]. These innovative constructs integrate pliable components and connect with the human body through textile-based interfaces [5], [16], [17]. Their operational efficacy is driven by a diverse array of actuators, including hydraulic [18], cable-driven [16], pneumatic artificial muscle [19], pneumatic network [20], and fiber-reinforced modalities. The literature highlights the significant advantages of soft exoskeletons, particularly in facilitating gait assistance. Compared to rigid models, they offer multiple benefits, notably demonstrating the potential to reduce metabolic expenditure while preserving the natural biomechanics of walking [21]. Their design ethos, characterized by minimal inertia, streamlined aesthetics, and rapid responsiveness, underscores their safety in human interaction [5]. Additionally, these exoskeletons harmonize effectively with the wearer's biological articulations [6].

A significant milestone in this domain was achieved by researchers from Harvard University who pioneered a soft ankle exoskeleton employing a Bowden cable system [6]. Integral to its design is a textile-based anatomical interface that harnesses the biomechanical robustness of the human skeletal structure for optimal load distribution. A salient feature involves the attachment of each leg's extremity to the Bowden cable's sheath, extending synchronously to the foot, resulting in the generation of the requisite tensile forces to elicit ankle torques. Its monoarticular design focuses actuation on a single joint. Weighing 10.1 kg, the system generates forces of up to 100 N at walking speeds of 1.5 m/s, while consuming approximately 59.2 W of power [6]. Building on the foundational concepts demonstrated by this initial prototype, the same team, led by Conor Walsh, developed a subsequent iteration capable of modulating both hip and ankle kinematics. A notable innovation involved the centralization of load distribution across both lower limbs via a unifying motor, facilitating torque propagation across multi-joint interfaces. This refined multiarticular design has a mass of 7.5 kg and can exert forces up to 200 N at velocities of 1.70 m/s. Notwithstanding their minimal gait perturbative effects [5], empirical validations regarding their potential to augment locomotion or attenuate metabolic expenditure remain inconclusive. In another significant stride, the XoSoft EU initiative engendered an exoskeleton with actuation capabilities at the hip and knee nexus, underpinned by a quasi-passive actuator system for enhanced energy efficiency and weight optimization [17]. Experimental validations with post-stroke patients underscored its potential in biomechanical modulation and metabolic consumption reduction by a significant 7.8% [17]. However, certain technical impediments persist. When extrapolated for knee extension functionalities, existing design paradigms pose risks due to potential actuator-joint incongruencies [22]. In response, Zhou et al. introduced the SOFTKEX framework—a continuum-structure-based knee exoskeleton [22]. However, its incorporation of semi-rigid constituents poses questions regarding the holistic flexibility of the system. The Myosuit stands out for its balance of mass, ergonomic comfort, and functionality, making it a leading example in the realm of soft exoskeletons. Its unique tri-layered design [4] and empirical evaluations underscore its proficiency in diverse biomechanical tasks.

However, the research trajectory delineating soft exoskeletons is punctuated with challenges. Extended wear can induce maladaptive shifts or deformations [21], with implications on gait modulation and metabolic demands. A limitation pertains to their inability to counteract compressive forces parallel to the anthropomorphic structure [23], with resultant shear forces potentially exacerbating wearer discomfort [16]. Traditional mitigation strategies, such as the tightening of mechanical conduits, often exacerbate user discomfort and have been hypothesized to compromise vascular dynamics [24]. Transmission efficacies of prevalent Bowden mechanisms have been documented to vary between 50-80% [23], [25], underscoring the exigency for research in optimizing transmission dynamics, especially for populations with health challenges. In conclusion, the trajectory of soft exoskeleton technologies, while promising, necessitates improvements to surmount inherent challenges and further the boundaries of biomechanical augmentation.

### 2.3 Soft Actuation Mechanisms

In the field of WMAD, actuators are quintessential components that facilitate the generation of the necessary mechanical force and motion. Historically, predominant actuation mechanisms have relied on rigid materials. While effective, these materials introduce potential safety risks during human-robot interaction and impose constraints on the adaptability and flexibility of wearable robotic systems [3]. Recognizing these inherent limitations, the research community has increasingly shifted its focus towards soft actuators, primarily composed of compliant materials. In recent years, a growing body of scholarly work has been dedicated to exploring and refining soft actuation mechanisms. Notable among these are shape memory alloy (SMA) actuators [7], [9], [26], [27], [28], [29], electroactive polymer (EAP) actuators [30], cable-driven actuators [31], and fluidic elastomer actuators [20], [32], [33], [34], [35].

However, it is important to note that the adoption of these mechanisms in wearable devices is not yet widespread. Certain mechanisms are subject to limited application, primarily due to concerns related to their actuation efficacy or the complexities associated with their fabrication processes. The following sections provide a comprehensive examination of these actuation mechanisms, elucidating their technical attributes and their applicability within the broader spectrum of WMAD.

#### 2.3.1 Shape Memory Alloy Actuators

Shape Memory Alloy (SMA) actuators represent a class of thermally activated materials known for their unique ability to "remember" and revert to their original pre-deformed configuration when subjected to specific thermal conditions. This distinctive behavior is primarily attributed to the material's capacity to undergo a phase transition, specifically the martensitic transformation between two distinct crystallographic structures.

At elevated temperatures, SMAs assume an austenite phase characterized by a body-centered cubic (BCC) lattice structure. In this phase, the material exhibits significant resistance to deformation. As the temperature decreases, the alloy transitions to the martensite phase, which is characterized by a face-centered cubic (FCC) lattice. This phase transition not only enhances the alloy's deformability but also enables it to retain the deformed state. Upon re-exposure to higher temperatures, the SMA reverts from the martensite phase back to the austenite phase, thereby returning to its original shape and generating actuation forces.

Although SMAs exhibit a relatively modest recoverable strain, typically around 8%, they can generate substantial stress, approaching 230 MPa [36]. To optimize their functional strain, innovative configurations, such as coil springs, have been developed [8]. Notably, pioneering research by Hironari utilized this configuration to design and develop artificial muscles [29]. Inspired by such innovations, Bundhoo and his team applied these concepts in their efforts to develop a prosthetic finger [29]. Extending the applicability of SMAs, Rodrigue and his team employed SMA-based coil springs to model wrist rotations [7].

However, despite their advantages, SMA actuators are not without limitations. One significant concern is their relatively high energy consumption. While these actuators can operate at reduced voltage levels, they require substantial current inputs. Unfortunately, only a small fraction of this consumed energy is converted into usable mechanical force [26]. Additionally, the actuation speed of SMAs presents a challenge. Although solutions such as water cooling can enhance actuation rates, they introduce further complexities, including increased bulk, weight, and energy demands. Furthermore, the thermally induced changes in wearable devices may lead to discomfort when in direct contact with the skin. The heat required for shape-memory activation poses potential hazards, particularly burn risks. Moreover, precise modulation of this thermally responsive transformation is problematic, especially considering the fluctuating ambient temperatures encountered by wearables. Given these challenges, the present study has elected to exclude SMA actuators from further consideration.

### 2.3.2 Twisted and Coiled Polymer Actuators

Twisted and Coiled Polymer (TCP) actuators, which utilize thermal expansion to drive actuation, are increasingly gaining prominence in the field of soft robotics. These actuators are constructed from polyamide filaments often sourced from inexpensive fishing line materials that are meticulously twisted into coils. Upon heating, the polymers undergo thermal expansion, resulting in a contraction in length and a simultaneous unwinding of the coil. This phenomenon is effectively harnessed for both contractile and torsional actuation. The structural configuration of the coils is critical, enabling the actuator to achieve significant actuation stress of up to 50 MPa or strain up to 49% [37], with an impressive power-to-weight ratio of 27.1 kW/kg [38]. TCP actuators also boast exceptional durability, with a lifespan exceeding 1 million cycles [39]. Furthermore, the low cost and simplicity of the materials used (e.g., fishing lines) contribute to their economic feasibility and widespread adoption.

However, despite these advantages, TCP actuators have inherent limitations that may constrain their broader application, particularly in wearable technologies. Their reliance on thermal actuation introduces challenges related to control precision and response speed [40]. The time required for the actuators to heat up and cool down can result in slower response times compared to other actuation systems [40], which may not be suitable for applications demanding rapid reflexive actions. Additionally, the necessity for a heat source to initiate actuation complicates integration into wearable devices, where maintaining a consistent and safe operating temperature is crucial [41]. Heat generation could affect not only the comfort and safety of the user but also the reliability and consistency of the actuator’s performance in varying environmental conditions.

### 2.3.3 Electroactive Polymer Actuators

Electroactive polymers (EAPs) have garnered significant attention in the field of advanced materials due to their unique ability to undergo dimensional transformations in response to electric fields. This distinctive property has spurred the development of soft actuators capable of controlled deformations upon electrical stimulation [42]. These actuators can be broadly classified into two primary categories: dielectric and ionic.

Within the realm of electroactive polymers, dielectric EAPs are particularly prominent [43], [44]. Their structure consists of a central dielectric elastomer membrane sandwiched between flexible conductive sheets. When an electric field is applied, these conductive layers are drawn together, exerting compressive forces on the dielectric elastomer, thereby inducing substantial structural deformation [38].

In contrast, ionic EAPs encompass three main types: ionic polymer-metal composites (IPMCs), carbon nanotubes (CNTs), and ionic polymer gels (IPGs). These materials’ function is based on the movement of ions within their polymer matrix [44]. When voltage is applied, ions shift positions, often accompanied by solvent movement. Depending on the direction of this movement, towards or away from components such as electrodes, the polymer can expand or contract, leading to precise and specific actuation behaviors. A significant advantage of ionic actuators is their ability to operate efficiently at much lower voltages than their dielectric counterparts. For human interface applications, this low-voltage operation, typically around 2 V, is particularly advantageous. However, potential drawbacks include slower response times and lower mechanical energy output.

The field of robotics has been significantly impacted by the integration of EAP actuators into various applications, such as soft grippers [45], ambulatory robots mimicking terrestrial and aquatic locomotion [46], and adaptable underwater robotic systems [47].

EAP actuators are characterized by several distinctive features, including their impressive actuation strains, often exceeding 200%. They also exhibit rapid response times, typically within 200 microseconds, and modest energy consumption [48]. Moreover, EAP actuators are economically appealing due to their cost-effectiveness, lightweight nature, and straightforward manufacturing processes, which collectively simplify production.

However, these advantages are accompanied by certain challenges. The power-to-volume ratio of EAP actuators is suboptimal [49], presenting a significant performance limitation. Additionally, concerns about durability persist, with instances of failure resulting from prolonged exposure to electric fields and the natural aging of the materials [42], [43]. In the context of wearable devices, these limitations underscore the need for further advancements in EAP materials before they can be fully integrated into actuation applications. Consequently, this study does not explore EAP actuators.

### 2.3.4 Cable-Driven Actuators

The cable-driven actuator is one of the most extensively utilized actuators in the domain of soft WMAD. These actuators employ a combination of flexible elements and cables connected to conventional motors, which allows for strategic placement of the actuation source. This design enhances the precision of assistance delivery. Such actuation mechanisms are prevalent in soft WMAD designs due to their inherent low inertia, rapid response time, and compatibility with human anatomy.

A fundamental distinction within cable-driven actuators is based on their transmission systems, categorizing them into open-ended and closed-loop cable configurations [23]. Specifically, open-ended cable systems exert force unidirectionally, necessitating an additional cable to initiate motion in the opposite direction. Key components of these open-ended systems include Bowden cables, motorized reels, pulley transmissions, and capstan transmissions [23].

The Bowden cable mechanism consists of an internal cable sheathed within a flexible conduit. One end of the cable is attached to a standard mechanical actuator, either rotary or linear, while the opposite end is anchored to a designated point [50]. A critical feature of this mechanism is the preloading component, often in the form of compressed springs, which maintains cable tension relative to its sheath [50]. The actuation process begins when the mechanical actuators retract the inner cable, generating the required tensile force. Due to its streamlined design, the Bowden cable mechanism is favored in wearable technology, often surpassing other cable-driven systems[25]. However, it is not without its limitations. The primary drawback is the inherent friction between the inner cable and its sheath, leading to increased power consumption [31]. Additionally, the mechanism requires pretension, further straining its energy requirements, and it offers limited displacement, constraining its range of motion [25]. Despite these challenges, the Bowden mechanism remains the dominant actuator in lower-extremity soft exoskeleton designs [50].

The motorized reel technique, which uses winches to manipulate cable lengths via a servo actuator, is more commonly adopted in rigid exoskeletons. However, its integration into soft wearable systems is limited. Two main reasons contribute to its restricted use: first, the need for maintaining high cable tension to prevent backlash, a challenge in soft exoskeleton designs [23], and second, the inherent bulk of the mechanism, which makes it less suitable for soft wearable applications [50]. While other transmission mechanisms exist, their application in soft exoskeleton designs is largely unexplored, and thus, they are not the focus of this review.

In soft actuation, cable-driven mechanisms translate motor-generated mechanical energy into movement within a compliant framework. However, integrating traditional rigid components, such as motors and bearings, may compromise the actuator's inherent flexibility, increasing its overall size [25]. This enlargement could jeopardize user comfort in wearable applications. Moreover, semi-flexible conduits may inhibit user mobility, introducing design complexities. Key challenges include efficient force distribution [31], friction, mechanical backlash [50], actuation precision, and potential misalignment between the skin and the material [5]. These limitations could restrict the applicability of cable-driven actuators in certain wearable devices [25]. Given these limitations, particularly concerning user comfort, system efficiency, and wearable design constraints, the decision was made to exclude this mechanism from the present investigation as it does not fulfill the primary objectives.

### 2.3.5 Soft Pneumatic Actuators

Soft pneumatic actuators have become a cornerstone technology in the field of soft robotics due to their distinct advantages and versatility. These actuators are not only lightweight and quick to respond, but they also prove highly useful across a wide range of applications, as evidenced by numerous studies [18], [35], [51]. Their capacity for significant deformation, coupled with built-in safety mechanisms and a high power-to-weight ratio, further underscores their functional importance [46], [52]. Additionally, their cost-effectiveness in production has led to increasing adoption in contemporary robotics research [53], [54]. Pneumatic actuators can be classified into several primary subtypes, including McKibben actuators or Pneumatic Artificial Muscles (PAMs), Pneumatic Network Elastomers (PneuNets), fiber-reinforced actuators, and soft bellow actuators. The following sections provide a detailed examination of these types.

#### 2.3.5.1 McKibben / Pneumatic Artificial Muscles

Emerging in the mid-20th century, the McKibben actuator, often referred to as the Pneumatic Artificial Muscle (PAM), represents a seminal contribution to the field of soft pneumatic actuation [55], [56]. Joseph L. Joseph L. McKibben's innovative work in the 1950s led to the creation of this pioneering actuator, which was initially applied as an orthotic tool for individuals with poliomyelitis [57]. Designed to closely emulate the biomechanics of natural muscles, McKibben actuators are constructed from flexible elastomeric tubes encased within a cylindrical braided sheath. Fluid pressurization causes these actuators to expand, resulting in a contraction like that of organic muscles, thus establishing their significance in the early development of soft robotics.

In the subsequent decades, particularly the 1980s, the Japanese company Bridgestone played a key role in the commercial introduction of PAMs. Schulte’s seminal research further elucidated the complex interplay between applied pressure, linear deformation, and the generated force of these actuators [57]. His work notably emphasized the critical influence of fiber quality within the braided sheath on overall actuator performance [57].

In early studies, researchers examined the relationships among geometric parameters, applied pressure, and resultant force using an energy-based theoretical framework, emphasizing the equilibrium conditions inherent in the actuator's design [55]. Notably, research posited that the actuator enters a state of stasis when the angle formed by the fibers approaches 35.3°. This finding underscores a critical insight: while increasing pressure enhances actuator rigidity, it simultaneously imposes a limit on its potential for deformation [58]. Over the past two decades, scholarly examinations of McKibben actuators have delved into the effects of fiber stretch dynamics and braiding patterns on the actuator's contraction capabilities [57].

The appeal of PAMs lies in their exceptional force-to-weight ratio, making them ideal candidates for augmenting human joints that require a combination of high torque and low bandwidth in specific physiological activities. However, their dynamic interaction with human skin can cause discomfort, a result of the continuous modulation of the actuator's surface area in contact with the body [56]. Furthermore, the large diameter and anchoring requirements of PAMs present significant challenges [52]. The necessity for extensive diameters to optimize force delivery necessitates a secure attachment to the user, which complicates their practical application. These complexities provide a compelling rationale for excluding PAMs from the current research focus.

**2.3.5.2 Pneumatic Network Actuator**

Within the expansive field of soft robotics, the Whitesides Research Group pioneered the development of the pneumatic network actuator. This actuator, intricately designed, consists of integrated air channels encased within elastomers [59]. The primary mechanism behind its bending motion involves the strategic layering of two elastomeric materials: an extensible layer, which contains interconnected air channels and chambers, and an inextensible layer, characterized by increased rigidity to impede linear motion, thereby facilitating bending or twisting actions. When pressurized, the difference in stiffness between the two layers induces the actuator's bending motion. However, Whitesides' prototypes face three notable limitations: reduced actuation speed, significant volume changes, and a limited operational lifespan [20].

In response to these challenges, Polygerinos et al. introduced an innovative variant known as Fast PneuNets [60]. This design mirrors Whitesides' framework but deviates in one significant aspect: the presence of interspatial gaps within the walls of each chamber in the extensible layer. Empirical research indicates that narrower interspatial gaps enhance the bending amplitude [60]. Unlike Whitesides' original design, the inextensible layer in Fast PneuNets is tri-layered, consisting of an elastic layer like the extensible layer, followed by a paper stratum [61], and culminating in a final, more rigid elastic layer. While the primary goal of PneuNet actuators is to facilitate bending motions, numerous modifications led by researchers have aimed to achieve twisting motion [62]. The actuator's performance metrics and kinetic properties can be adjusted by varying the geometric parameters of its air chambers and modifying the material stiffness [63]. For instance, increasing the number of chambers enhances the bending angle and resultant force while simultaneously reducing the required actuation pressure [61]. Similarly, changing chamber dimensions affects these parameters, and reducing wall thickness increases the actuator's agility, bending amplitude, and force output [60]. While more rigid materials require higher actuation pressures, they also correlate with faster bending kinetics and increased moments.

Building on the Fast PneuNets design, Polygerinos and his team developed a pneumatic glove for hand rehabilitation, incorporating five Fast PneuNets actuators [61]. This device was designed to mimic the natural curvature of human finger flexion, with the actuators affixed to a neoprene glove. However, a notable discrepancy exists between the circular profile of these actuators and the physiological flexion profile of human fingers. To address this, Hong et al. employed variable stiffness actuators to refine the motion profile and subsequently integrated them into the design of a hand-rehabilitation exoskeleton [64].

Nevertheless, PneuNet actuators are not without drawbacks. They are prone to a "ballooning" phenomenon, which increases friction and places excessive pressure on biological joints. This ballooning also diverts energy radially rather than directing it along the desired trajectory. Additionally, orchestrating complex motions remains a significant challenge, and the actuator's relatively bulky design, coupled with its high actuation pressure requirements, limits its practical applicability. Furthermore, the reduced actuation speed may constrain its use in certain applications [65]. An increased force output could also lead to structural compromise, which is undesirable in applications that require significant force or continuous cyclic motions, such as wearable assistance devices [60]. Due to constraints such as ballooning, excessive bulk, and high actuation pressures, pneumatic network actuators were omitted from this investigation because they do not meet the fundamental research goals.

#### 2.3.5.3 Fiber-Reinforced Actuators

Emerging from the seminal design of the McKibben actuator, fiber-reinforced actuators have since established a prominent position within actuation technologies. The inherent ability of elastic shells to convert air pressure into specific motions when subjected to mechanical constraints laid the foundation for this innovative actuation concept [66]. A significant advancement in understanding these actuators was achieved by Hirai et al., who proposed the first comprehensive modeling approach [66]. They conceptualized the actuators as fiber-constrained elastic bladders. Through subsequent modeling of various motions and detailed behavioral analyses, they provided an empirical exposition of the complex interplay between deformation and constraint topologies [66]. Unlike PneuNets actuators, the core of fiber-reinforced actuators is their singular cylindrical chamber, with motion dynamics primarily governed by the arrangement of constitutive fibers.

Building on this foundation, Connolly et al. made pioneering efforts to develop an analytical model that elucidates the relationship between the fiber's angular orientation and the resulting actuator motion [67]. Their innovative work led to the design of an actuator capable of extension, bending, and twisting by simply adjusting the fiber angles. Expanding on this concept, they combined these actuator segments in series, resulting in the creation of a worm-like robotic device [67] and a rehabilitative assistive glove [68].

A key advantage of the fiber-reinforced actuator lies in its adaptability. By merely adjusting the fiber angle, a wide range of motions can be achieved. For example, circumferential fibers positioned at 0° relative to the actuator's longitudinal axis induce maximum axial extension when the radial direction is constrained. Furthermore, Connolly et al. found that a dual-fiber alignment at +3° and -3° with respect to this axis could achieve the desired extension [69] . Conversely, axial fibers aligned at a 90° angle result in optimal radial expansion without axial elongation. For twisting motion, fibers should ideally be aligned at a 30° angle, while the maximum bending effect can be realized with fibers angled at ±5°.

However, various parameters influence actuator performance. Both actuator responsiveness and force generation are dependent on its cross-sectional morphology. Empirical studies suggest that a semicircular cross-section is ideal for bending motions and force generation, whereas a circular profile is better suited for linear and torsional motions [68]. The thickness of the air chamber wall is crucial in regulating actuator motion, actuation pressure, and force output. Thinning the wall increases actuator sensitivity, although it reduces force output. In configurations where reinforcements are densely integrated, an increase in pneumatic pressure is required [67], which enhances characteristics such as linearity, reliability, and robustness [67].

Comparatively, fiber-reinforced actuators exhibit superior durability and force generation capacity relative to PneuNets actuators. They enhance extension capabilities while minimizing the energy dissipation associated with the radial expansion of rubber elements. Nonetheless, challenges persist. Their fabrication process is complex, requiring significant time and resources [69]. Despite their versatile motion capabilities, controlling these actuators remains intricate [18], [69]. The elaborate architecture and specific material properties could limit the precision and accuracy essential for certain wearable technologies [70], thereby justifying their exclusion from some contemporary research initiatives

**2.3.5.4 Soft Bellow Actuators**

Soft bellow actuators, known for their ability to accommodate and induce movements through flexible structures, are critical components in the field of soft robotics [71]. Operating on the principle of volumetric deformation, these actuators enable precise manipulation by allowing controlled expansions and contractions through alterations in internal pressure. The design and material science underlying soft bellows are fundamental, directly influencing the functionality and suitability of actuators for specific applications.

The mechanical behaviors and design optimizations of bellow actuators have been extensively studied, with significant early contributions made by J.F. Wilson in 1984 [72]. Wilson's research provided foundational insights into the mechanical properties and optimization of bellow designs, critically evaluating the load-deflection responses of various geometries and proposing theoretical models corroborated by experimental data [72] . His work emphasized the profound impact of material properties and geometric configurations on the performance of bellows.

Bellow actuators leverage a variety of designs, each suited to specific applications and offering distinct performance characteristics. Circular bellows, engineered for tasks requiring straightforward, repetitive linear movements, are ideal for applications demanding simple expansion and contraction without directional bias due to their symmetrical cylindrical configuration [73], [74]. U-shaped bellows, which bend along the open part of their 'U' configuration under pressure, are well-suited for applications requiring angular adjustments, such as in articulated mechanisms or adaptive grippers in robotic systems [75].

Origami-inspired actuators offer diverse and complex motion capabilities through the art of paper folding. The Yoshimura pattern, composed of interconnected isosceles triangles, facilitates translational motion with notable axial stiffness, making it suitable for applications requiring strong linear actuation [76]. The accordion pattern, characterized by its zigzag folding, maximizes surface area relative to volume, enhancing the actuator's capacity for significant contraction and expansion [77]. The waterbomb pattern excels in volumetric transformations, offering excellent multi-axial compliance, making it ideal for tasks requiring complex bending and twisting [78]. Known for their ability to fold flat and deploy rapidly, the Miura-Ori pattern consists of parallelograms arranged in a zigzag pattern, providing structural stability and resistance to compression, particularly useful in space-efficient and rapid-deployment applications [79].

In the realm of wearable technology, bellow actuators are prominently featured in robotic orthoses, such as exoskeletons, which primarily utilize electric, hydraulic, and pneumatic mechanisms. For instance, soft robotic gloves and suits incorporate bellows to facilitate muscle movement and enhance joint stability, which is crucial in the rehabilitation of stroke survivors and individuals with mobility impairments [80]. Yi et al. developed a pioneering soft robotic glove utilizing a circular bellows actuator. This glove, notable for its lightweight design (weighing less than 50 grams) and use of soft materials, offers comfort and convenience for daily wear. It incorporates novel bidirectional linear soft actuators (BLSA) that enable independent finger movement, facilitating both flexion and extension. These actuators are designed to function effectively at low pressures, allowing the glove to achieve full-range finger motion and generate up to 40 N of force at the fingertips, sufficient for most grasping and interaction tasks in healthcare and service applications [71].

However, integrating actuation components into such devices often poses challenges. The glove's reliance on external Bowden cable systems for force transmission renders it bulky and restrictive, limiting its practicality and comfort for everyday use. To address these limitations, other researchers have integrated circular bellows with a guided mechanism into the fabric of gloves, enabling anisotropic kinematic bending optimal for specific rehabilitation exercises [81]. This design approach eliminates the need for external force transmission systems, enhancing the wearability and functionality of the gloves. Despite these improvements, the integrated actuators initially faced challenges due to their bulkiness, which impacted wearability and comfort. These actuators generate a force of only 20 N, which may not meet the requirements for tasks demanding higher strength. Innovations continue with the adoption of the Yoshimura origami pattern in bellows design. Yap et al. introduced another glove utilizing this pattern, allowing the actuators to bend efficiently with minimal pneumatic energy [82]. However, while these actuators operate at low pressures, the force they generate remains relatively modest (3.2 N for the thumb), potentially limiting their applicability for users requiring more robust rehabilitation exercises.

To further expand the application of bellows in wearable technology, researchers like Ang and Yeow have developed a soft exoskeleton that assists with wrist flexion and extension using U-shaped bellows [83]. This device aims to recover at least 70% of a healthy individual's range of motion, significantly facilitating activities of daily living (ADL). Similarly, Seo et al. have developed a wearable robot based on origami bellows to assist the shoulder joint, particularly in abduction movements [84]. This device, designed to mimic and support the natural scapulohumeral rhythm, generates significant torque (up to 12.6 N.m) at a 90-degree abduction angle, enhancing the strength and mobility of the shoulder joint for rehabilitation or enhancement purposes.

Given the enhanced functionality and safety provided by the bellows mechanisms, current research is focused on developing a novel sleeve mechanism based on this concept, anticipated to improve mobility and user comfort in wearable devices. This ongoing innovation underscores the transformative potential of bellows in the realm of wearable technology and soft robotics, highlighting their pivotal role in advancing medical rehabilitation and everyday functionality.

## 2.4 Materials for Soft Pneumatic Actuator

### 2.4.1 Silicone Rubbers

Silicone rubbers are fundamental components in the field of advanced elastomeric materials, renowned for their unparalleled flexibility and robust performance characteristics. These materials primarily consist of polydimethylsiloxane (PDMS), a silicone-based polymer characterized by a molecular backbone of alternating silicon and oxygen atoms, with each silicon atom connected to two methyl groups. This configuration enhances the material’s resistance to water and significantly improves its thermal stability.

The establishment of their key characteristics is controlled by a crosslinking mechanism, typically implemented at ambient temperatures using a binary system comprising a catalyst and a crosslinker. This vulcanization process is enhanced by heat or can be regulated using UV light in contemporary compositions [85], offering notable benefits in modern manufacturing processes like 3D printing. UV-reactive silicones particularly stand out for their extended moldability and rapid curing phases, crucial to produce detailed and accurate robotic components.

Within PDMS, the extended bond lengths and broad bond angles confer significant flexibility and elasticity to the material [86]. Impressively, silicone rubbers can endure elongations at break well beyond 1000% [85], illustrating their ability to merge extreme pliability with Shore A hardness ratings as low as 5 and substantial tensile strength up to 2.4 MPa [86]. Furthermore, their minimal viscoelastic behavior highlights their suitability in the field of soft robotics. These materials function across a vast temperature range, from -50°C to above 230°C, ideal for varied thermal environments. Their non-toxic and biocompatible properties are also essential for medical devices and wearable technology that interacts directly with human skin.

Processing silicone rubbers is straightforward, and their ability to withstand large deformations under pressure renders them indispensable for soft pneumatic actuators. Notable brands such as EcoFlex (Smooth-On, USA), DragonSkin (Smooth-On, USA), Sylgard (Dow Corning, USA), and Elastosil (Wacker Chemie, Germany) lead the market, each offering specialized grades tailored for specific soft robotic applications. For instance, EcoFlex is available in various Shore hardness ratings and is renowned for its ability to undergo significant deformation under minimal stress [87]. This makes it particularly effective for pneumatic or fluidic actuators that inflate deformable membrane structures. Its hyperelastic qualities allow it to stretch multiple times its original size without tearing, which is advantageous for applications such as prosthetic strain sensors and flexible wearable sensors for biomedical uses [87]. DragonSkin stands out due to its high tensile strength and exceptional elongation at break, suited for designs that anticipate a ballooning effect. When mixed with Slacker in precise ratios, it achieves a softer, tacky consistency, ideal for mimicking realistic soft human tissues [87]. Although Dragon Skin silicones have a higher Young's modulus, necessitating greater fluid pressure for actuation, they are capable of exerting larger forces, beneficial for applications requiring robust force output. Conversely, Sylgard 184 is noted for its optical clarity and low viscosity, making it less suitable for actuation but widely used in soft grippers and sensor applications [85]. Its average tear resistance and compatibility with plasma-activated surface bonding techniques render it a valuable material for specific applications in soft robotics.

### 2.4.2 Thermoplastic Polyurethane

Thermoplastic polyurethane (TPU) is an advanced block copolymer distinguished by alternating sequences of hard and soft segments, each imparting distinct physical properties. These segments are covalently bonded, forming a linear structure that critically influences the polymer’s mechanical behavior. The hard segments, typically composed of aromatic polyisocyanates such as Methylene Diphenyl Diisocyanate (MDI) or Hexamethylene Diisocyanate (HDI), are interspersed with chain extenders like butanediol. These hard segments provide structural rigidity and function as physical crosslinks, maintaining the polymer’s integrity and shape memory under stress. Conversely, the soft segments, primarily made up of polyols such as polyether polyol or polyester polyol, confer elasticity and flexibility. This dual-segment architecture allows TPU to exhibit properties similar to rubber at room temperature, while still being processable as a melt at elevated temperatures.

The glass transition temperature, *Tg*, of TPU can vary significantly, typically around -35°C for the soft segments, enabling the material to remain flexible and elastic under cold conditions. As temperatures exceed this *Tg*, the mobility of the soft segments increases, enhancing the material’s elastomeric properties. Moreover, TPU's melting point, generally around 216°C, marks the temperature at which the hard segments lose their structural efficacy. Above this melting point, the polymer can be reshaped and reprocessed, facilitating manufacturing processes such as 3D printing, injection molding, and extrusion [88]. This melt-processability makes TPU highly valuable in industries requiring versatile and durable materials.

The mechanical properties of TPU are highly tunable through the manipulation of the hard-to-soft segment ratio. A higher concentration of soft segments typically results in increased softness and flexibility, while a greater proportion of hard segments enhances tensile strength and thermal resistance. The elongation at break for TPU can range between 400% and 900%, indicating substantial deformability before rupture [88]. These characteristics make TPU ideal for applications demanding high durability and flexibility, such as in pneumatic actuators where repeated cycles of expansion and contraction occur [89].

In the context of thermoplastic elastomers (TPE), these materials are predominantly utilized in fused filament fabrication (FFF), a 3D printing technique particularly suited for creating soft and flexible structures[90]. TPEs, including various TPU formulations, offer diverse grades of flexibility and hardness, which are quantified by the Shore hardness scale. Notable TPU manufacturers include NinjaTek, Ultimaker, and Recreus, each providing products with specific mechanical profiles tailored to the needs of soft robotics[90]. For example, NinjaFlex 85 and Cheetah 98 by NinjaTek are designed for applications requiring exceptional elasticity and impact resistance, whereas products like Filaflex by Recreus offer different hardness levels suitable for precise engineering applications [90].

## 2.5 Conclusion

Despite significant actuation technology advancements, current soft actuators face several challenges in the field of WMAD that limit their effectiveness. Traditionally, these actuators are secured to the body using pads, straps, and textile layers. However, these interface components tend to stretch and loosen over time, compromising both the stability and comfort of the actuators.

Efforts to mitigate these issues by tightening the interface layers often result in discomfort and potential hazards for the wearer. Additionally, a considerable portion of the actuation energy is dissipated through the stretching of these interface components rather than being efficiently transmitted as useful torque to biological joints. This inefficiency is particularly problematic in soft WMAD, where the deformation of interface layers often leads to device malfunction. Energy losses due to radial expansion and surface friction further reduce the torque output, underscoring the need for improvements in both actuation speed and force for soft actuators to be viable in WMAD. The performance characteristics of various soft actuators are summarized in Table 2.1.

A key finding in the literature review is the notable gap in soft WMAD: the absence of a developed soft sleeve mechanism. While numerous actuator designs exist, a sleeve mechanism with the potential for diverse wearable applications is missing from the literature. Such a mechanism would minimize the need for attachment points, thereby avoiding the use of interface layers that are prone to stretching and energy dissipation. Unlike conventional designs, this sleeve mechanism would not rely on transmission mechanisms or additional attachment methods, which would reduce radial expansion and friction losses commonly observed in soft pneumatic actuators.

**Table 2.1:** Summary of current soft actuators advantages and limitations

| Type of Actuator | Power | Advantages | Limitations |
|---|---|---|---|
| **SMA** | Thermal | High power-to-weight ratio, low profile, lightweight | Limited displacement, high power consumption, slow response, safety |
| **EAP** | Electric | High strain, low profile, fast response, lightweight | Low power-to-volume ratio, high voltage |
| **Cable-Driven** | Mechanical | Fast response time, safe for human interaction | Rigidity, friction, limited displacement, backlash |
| **Pneumatic Network** | Pneumatic | Highly flexible, affordable, fast response | Friction, ballooning, bulky, high actuation pressure |
| **PAM** | Pneumatic | High force-to-weight ratio | Large diameter, slow response, affixed to the user |
| **Fiber-Reinforced** | Pneumatic | High actuating forces, fast response, affordable | Manufacturing, bulky, radial expansion, limited displacement |

# CHAPTER 3

# DEVELOPMENT OF SOFT SLEEVE ACTUATOR

The field of WMAD has seen remarkable progress in the development of actuators, yet a critical gap remains: the lack of a dedicated actuator tailored for wearable applications. Despite the diversity of actuator designs in the literature, none effectively address the challenges posed by conventional attachment and interface methods.

A SSA offers a transformative approach to wearable actuation by minimizing the need for external attachment points. This eliminates reliance on interface layers prone to stretching and energy dissipation, issues that undermine the performance and efficiency of current WMAD designs. Moreover, such a mechanism bypasses the need for complex transmission systems or additional attachment methods, reducing radial expansion and friction losses that frequently hinder the functionality of soft pneumatic actuators. This chapter introduces the conceptualization and development of the innovative SSAs, exploring its potential to redefine WMAD.

## 3.1 Design Requirements

In the development of wearable technology, particularly soft wearable mobility assistive devices (WMAD), several critical design requirements must be satisfied to ensure both functional performance and user comfort. First, the actuator should be flexible and wearable, adapting to diverse body shapes without hindering natural movement when not in use. To maintain comfort, it must be lightweight—ideally below 300 g—and compact, with a sleeve thickness of less than 25 mm. Biocompatible and hypoallergenic materials are crucial to prevent skin irritation, while durability and breathability ensure long-term usability under varied environmental conditions. Cost efficiency is also an important consideration, ensuring that the device remains accessible without compromising quality or safety. From a performance perspective, an operating pressure of 100–300 kPa strikes a balance between force output and material integrity, and a bandwidth of 1–2 Hz aligns with the typical human walking cadence. Linear actuators intended for lower-extremity support should generate 100–300 N of force and accommodate a 30–150 mm stroke range. Bending actuators should achieve up to 60° of bending and generate 10–30 Nm of torque, thereby reducing knee extensor demand by approximately 20–50%. Lastly, a twisting actuator should facilitate 30–40° of rotation to accommodate ankle inversion/eversion or internal-external knee rotation, while providing 2–15 Nm of torque for effective joint stabilization and control

### 3.2 Linear Actuator

Guided by the design requirements, the actuator's initial design is the linear soft sleeve actuator (LSSA). The design development and evolution of the LSSA is presented in this section.

#### 3.2.1 Initial Concept

The LSSA initial design is based on a triangular bellows mechanism, featuring two V-shaped walls, the internal and external walls, interconnected at their apex and base. The air connector links the LSSA to an air pressure source, as shown in Figure 3.1 (a). Although the bellows mechanisms have been widely used in conventional soft linear actuators, this study identified a limitation of this design. The limitation became apparent through the deformation depicted in Figure 3.1 (c), where inflation of the actuator resulted in an unexpected contraction of the internal wall, altering the actuator's shape from circular to elliptical. This issue is unresolved in the current literature, as no solutions for sleeve mechanisms have ever been proposed. Addressing this challenge necessitates a new design approach. For this study, three novel approaches were considered, namely the ring support, the lateral support, and the longitudinal support.

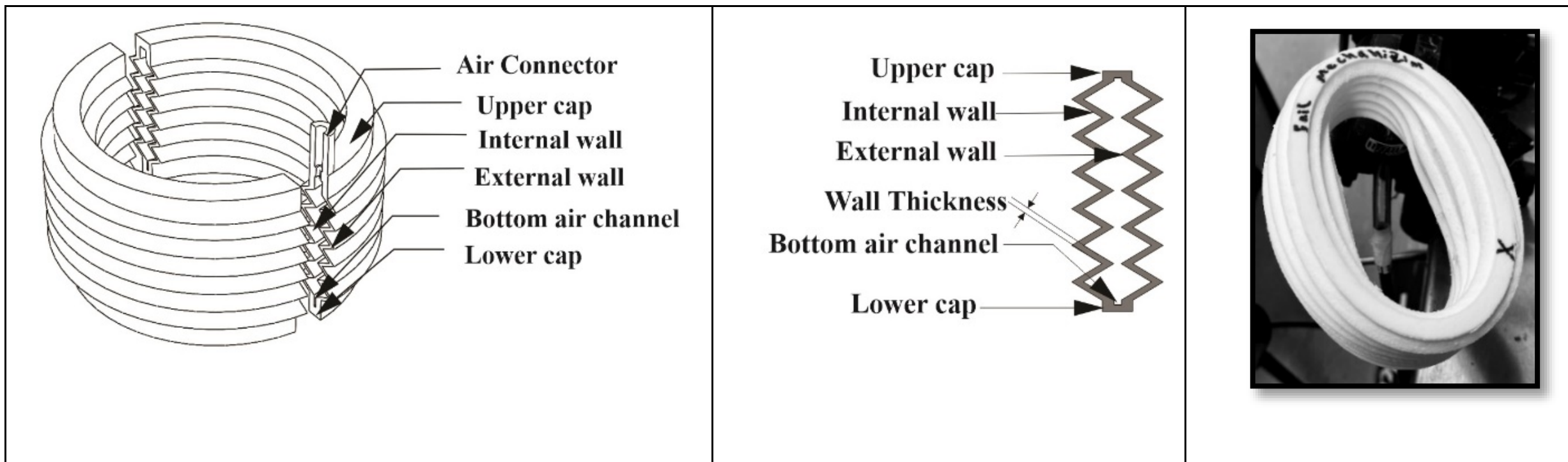

(a) (b) (c)

**Figure 3.1:** Initial concept (a) 3d representation (b) wall design (c) pressurized actuator

### 3.2.2 Ring Support Approach

The first concept, referred to as the ring support, was developed to counteract the undesirable contraction of the internal wall and the expansion of the external wall. This was achieved by incorporating a rigid structure. In this design, V-shaped rings were created to align with the contours of the valleys present in both the interior and exterior walls. These rings were then affixed to the actuator's walls, as illustrated in Figure 3.2. Upon testing, it was observed that this design successfully generated the desired linear motion and mitigated the contraction of the internal wall. However, despite the soundness of the approach, the incorporation of rigid V-shaped rings inherently reduced the actuator's flexibility. The actuator produced a limited extension and contraction of 15% and 5%, respectively.

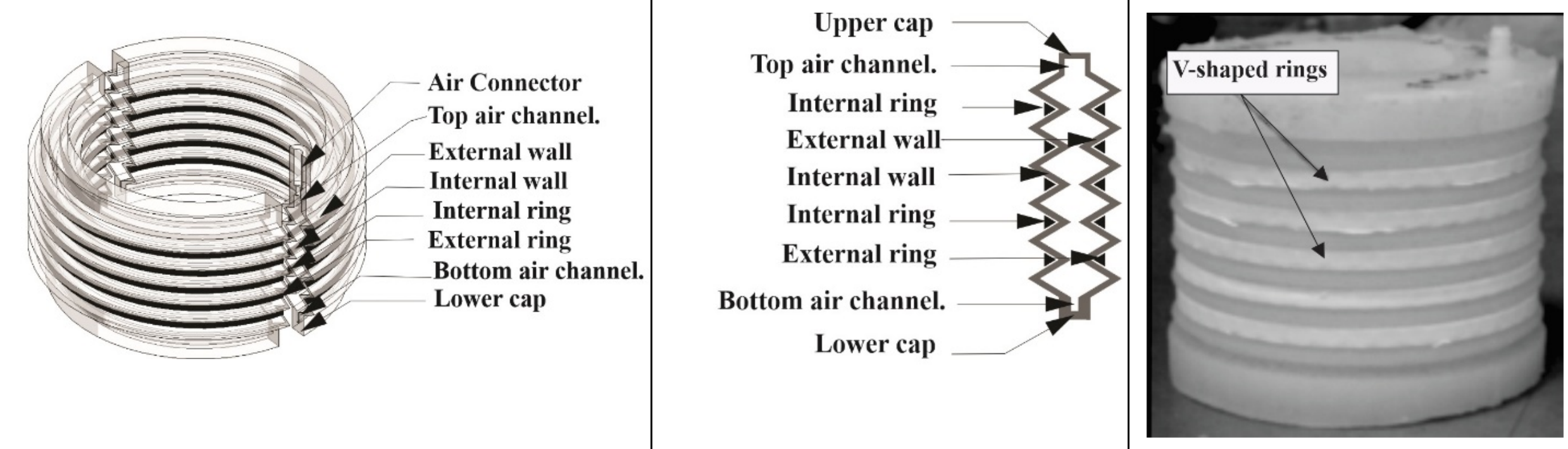

(a) (b) (c)

**Figure 3.2:** LSSA ring support concept (a) 3d representation (b) wall design (c) LSSA prototype

### 3.2.3 Lateral Support Approach

To address the challenges identified in previous designs, lateral tie-restraining layers were proposed to mitigate the simultaneous contraction of interior walls and the expansion of exterior walls of the LSSA. By minimizing unwanted radial movements, these layers enhance actuator flexibility and improve both the extension and contraction rates.

This conceptual framework led to the development of two distinct actuator models: a semicircular-based design and a triangular-based counterpart, as shown in Figures 3.3 (a) and 3.3 (c), respectively. Comparative evaluations revealed that while both models outperformed earlier designs, the triangular-based actuator demonstrated superior flexibility, and a notably higher rate of extension and contraction compared to its semicircular counterpart (Figures 3.3 (c) & (d)). As a result, the focus of this research shifted towards an exploration of the triangular model. However, with an extension rate limited to 20% and a contraction rate not exceeding 40%, the need for the development of a new, more effective model became evident.

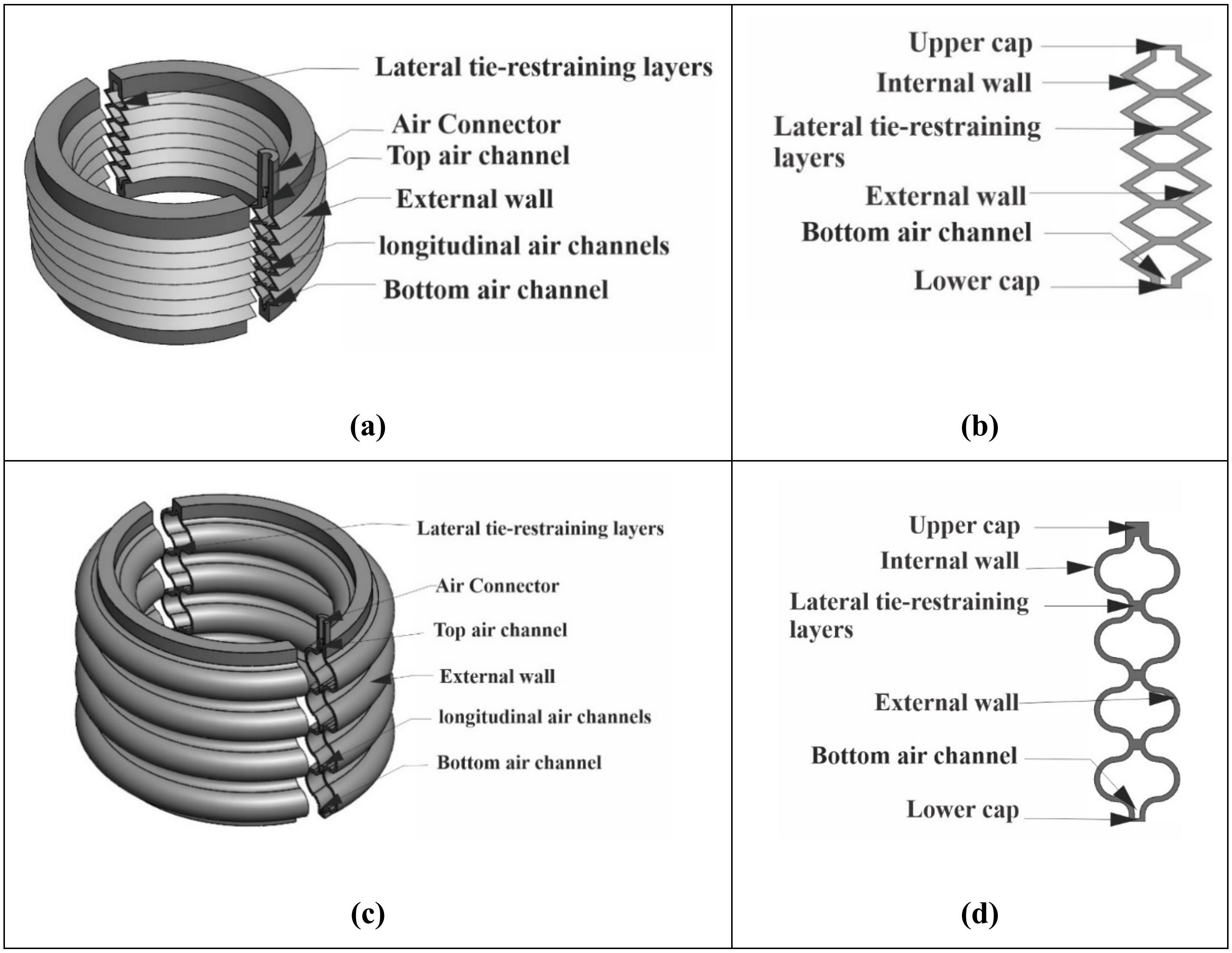

**Figure 3.3:** (a) LSSA triangular-based design 3d representation (b) a triangular-based wall design (c) semicircular-based 3d representation (d) semicircular-based wall design

### 3.2.4 Longitudinal Support Approach

To further enhance the linear displacement and flexibility of the LSSA, longitudinal tie-restraining layers were developed, as shown in Figure 3.5. These layers serve a dual purpose: they connect the internal and external walls of the actuator, maintaining a consistent distance between them and preventing radial displacement. By ensuring that the actuation force is directed linearly, these layers maximize the efficiency of motion transfer, thereby providing a stable framework for improved actuator performance. Building on this concept, the study further advances the design by introducing a folded mechanism, as shown in Figure 3.4. Unlike conventional bellows-based designs, which expand primarily through elastic deformation, this mechanism employs an origami-inspired folding mechanism activated by pneumatic pressure.

The fundamental operating principle of the new LSSA centers around the transformative dynamics enabled by its origami-inspired folded design. The corrugated walls are designed to harness applied pneumatic pressure and convert it into controlled mechanical motion. This is accomplished through an arrangement of folds that function as mechanical hinges, allowing the structure to expand linearly while minimizing lateral expansion. The design of the folds ensures that each segment unfolds in a coordinated sequence, enhancing the actuator's capacity to execute precise linear extensions and contractions.

To mitigate potential radial displacements that could compromise the actuator's performance, longitudinal tie-restraining layers have been integrated into the structure. These layers serve as a stabilizing element, maintaining the relative positions of the internal and external walls and ensuring uniform, controlled expansion in the intended direction of motion. The robustness of the tie-restraining layers is essential for preventing the walls from buckling under pressure, thereby preserving the integrity and effectiveness of the actuation mechanism. Additionally, the actuator incorporates an air channel architecture, as depicted in Figure 3.4, which incorporates multiple longitudinal channels spanning the full length of the actuator. This configuration ensures a consistent pressure gradient throughout the actuator. The integration of these components, the folded design with corrugated walls and the longitudinal tie-restraining layer, significantly enhances the flexibility and responsiveness of the actuator.

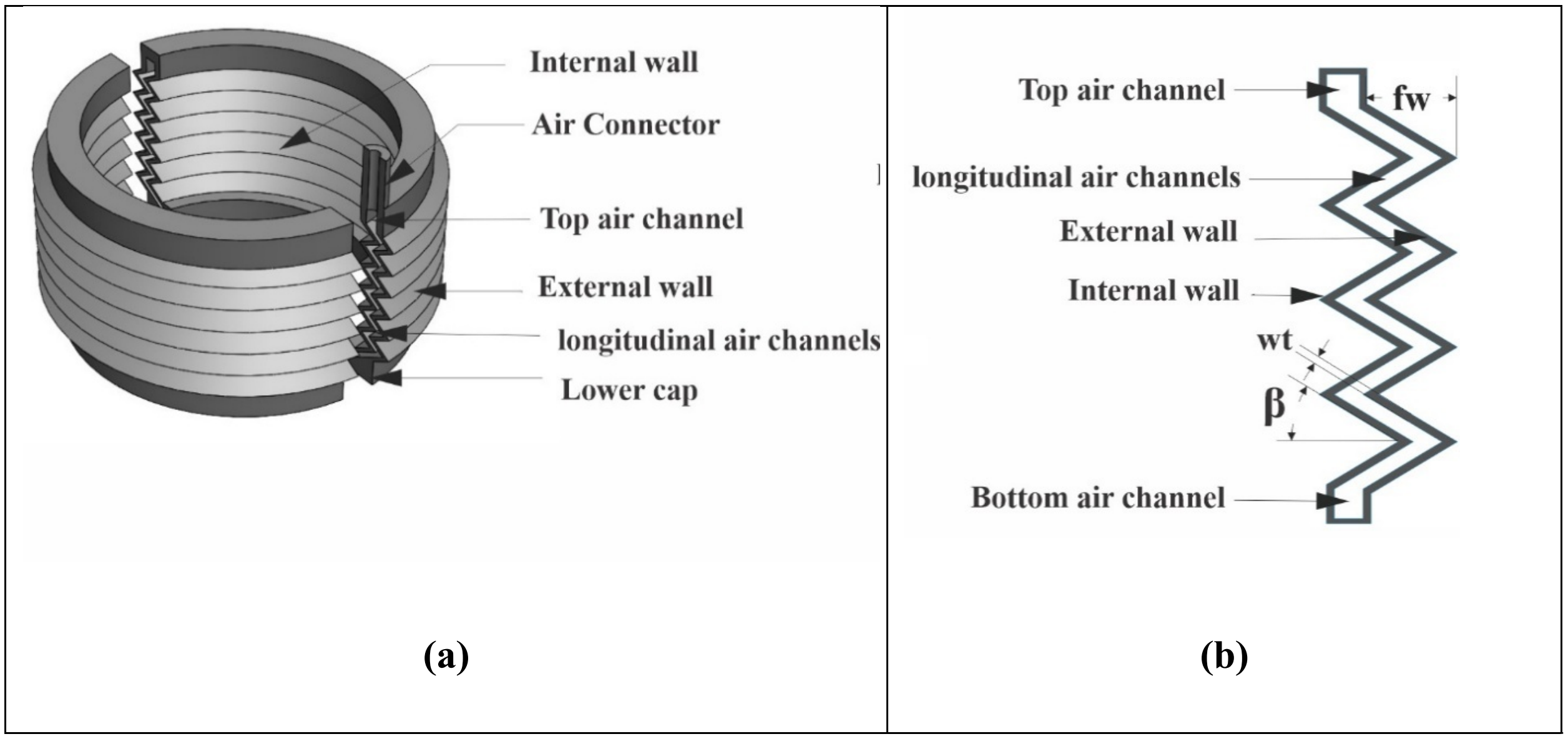

**Figure 3.4**: LSSA (a) 3d representation (b) wall design, fold Angle (*β*), wall thickness (*wt*), fold width (*fw)*

**LSSA Geometrical Parameters**

The critical geometric parameters that govern the performance of the LSSA include the fold angle *β*, number of tie-restraining layers *nr*, thickness of tie-restraining layers *tr*, wall thickness *wt*, and fold width *fw*. Each parameter plays a distinct role in defining the actuator's behavior and performance under varying operational conditions. Adjustments to these parameters can significantly alter the actuator's functionality, highlighting the importance of a comprehensive analysis. Figures 3.4 (b) and 3.5 illustrate these parameters.

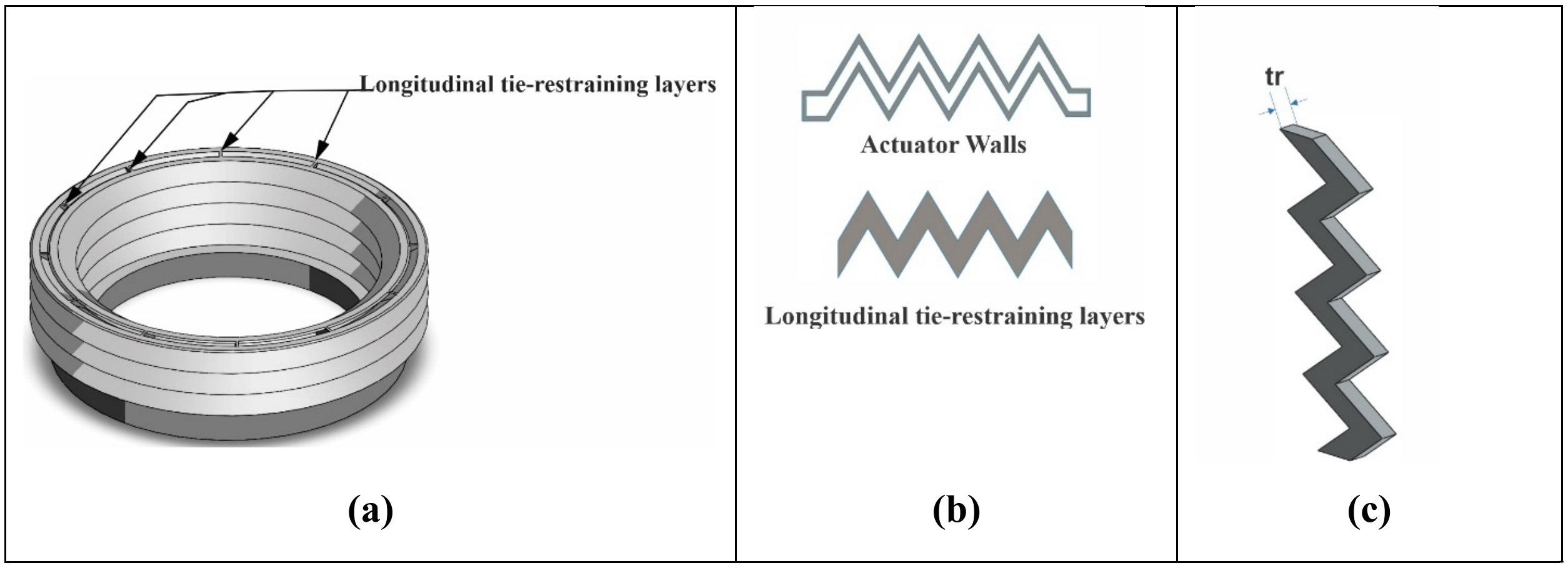

**Figure 3.5:** Longitudinal tie restraining layers (a) LSSA section view show these layers (b) restraining layers vs walls design (c) 3d representation of these layers

The fold angle defines the internal angle formed at the junction of two converging inclined sides of a fold. This angle is pivotal because it directly influences the actuator's ability to expand and contract. The tie-restraining layers function as the structural framework of the actuator, crucial for maintaining its designated shape. Two design parameters associated with these layers are the thickness and the number of layers. The thickness of the tie-restraining layers is critical in preserving the actuator's structural integrity across various operational pressures. The optimal thickness strikes a balance between durability and flexibility, preventing the layers from becoming excessively stiff, which could hinder the actuator's functional performance. The number of tie-restraining layers also impacts the actuator's durability and operational stability. Increasing the number of layers enhances structural integrity and reduces the undesirable ballooning effect under pressure. However, adding more layers may also increase the actuator's rigidity, affecting its flexibility and overall responsiveness to applied forces.

Wall thickness refers to the material thickness constituting the actuator's walls. This parameter is vital in determining both the actuator's flexibility and its capacity to handle various pressure levels. The fold width is the horizontal measurement between the peaks and valleys of the corrugations in the actuator's neutral (non-inflated) state. This measurement affects the actuator's overall ability to contract and expand.

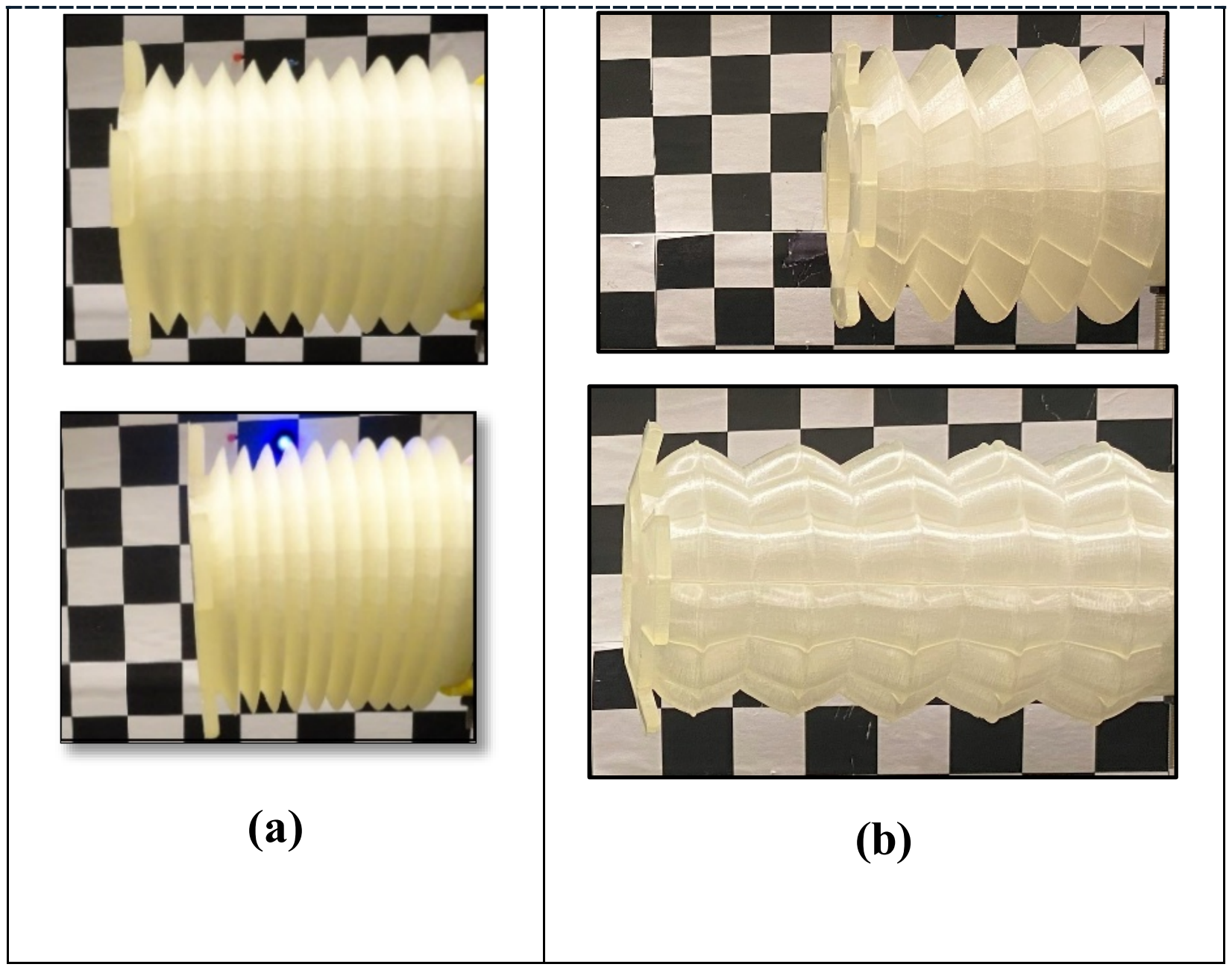

**Figure 3.6:** LSSA (a) contraction motion (b) extension motion

## 3.3 Bending Actuator

This section presents two novel concepts designed to induce bending of soft sleeve actuator (BSSA), namely, the extension variance approach, and constraining layer approach.

### 3.3.1 Extension Variance Approach

The extension variance approach employs a distinctive configuration of wall mechanics within the BSSA to facilitate bending motion. The outer wall is designed to be highly deformable, utilizing a corrugated structure, while the inner wall is designed to remain relatively rigid and undeformed. The differential response to pneumatic pressure between these two layers induces a bending motion. This concept was developed as illustrated in Figure 3.7. The prototype features an outer wall in a V-configuration contrasted with a flat inner wall. To maintain structural integrity, a lateral tie-restraining layer connects the inner and outer walls. The actuator comprises two hermetically sealed air chambers essential for its operation. Upon pressurization, the corrugated outer wall expands more than the non-corrugated inner wall due to its tailored flexibility and material properties. This asymmetric expansion results in the desired bending motion, as evidenced by prototype testing.

While successful, this design introduces a constraint: the flat inner wall inherently limits the actuator's flexibility. This limitation is particularly impactful in wearable applications, where adaptability and user comfort are critical. Additionally, while this approach effectively generates bending, it does so with a limited angle of less than 20 degrees.

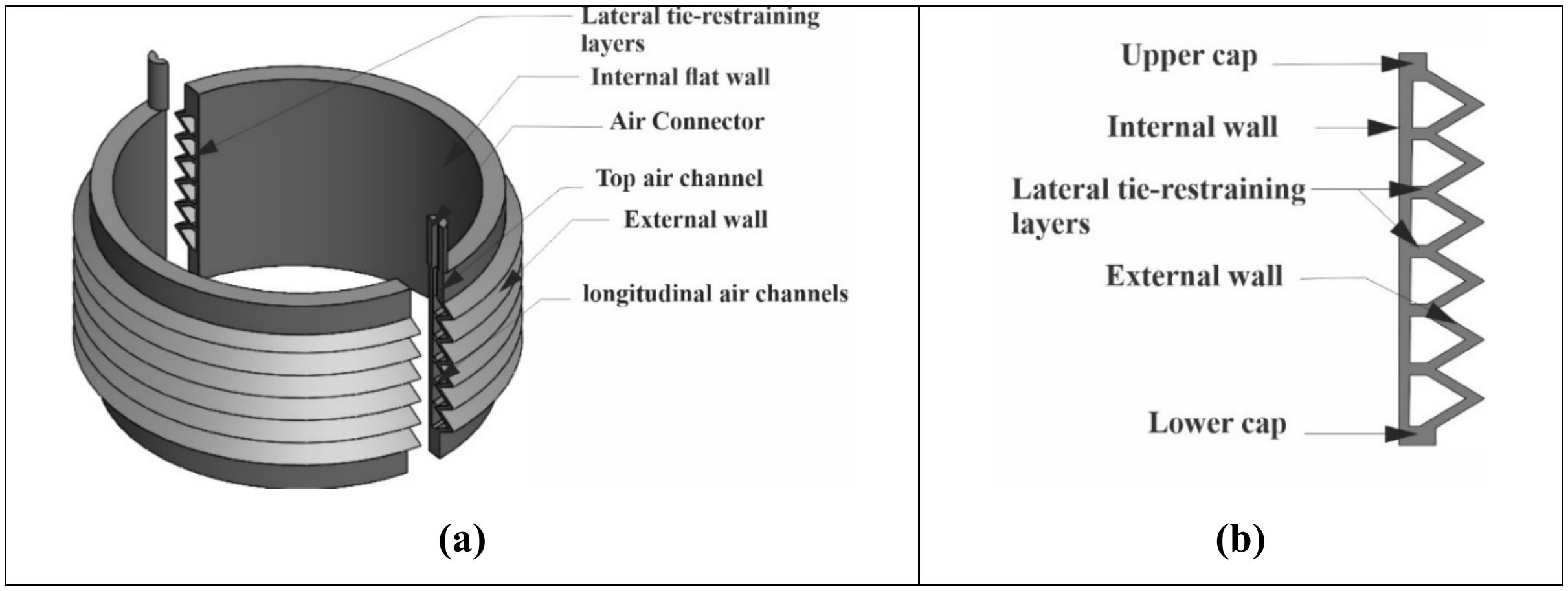

**Figure 3.7:** Bending sleeve actuator based on extension variance (a) 3d representation (b) wall design

### 3.3.2 Constraining Layer Approach

In this concept, the actuator is designed with one side restricted in movement by the inclusion of a non-flexible constraining layer. The main body of the actuator consists of a highly flexible framework, resembling a bellows structure, which allows for extension and contraction. In contrast, the constraining layers are made non-flexible and are placed to selectively inhibit motion. When pneumatic pressure is applied, the flexible side of the actuator expands more than the constrained side resulting in asymmetric expansion.

This asymmetric expansion causes the actuator to bend toward the side with the constraining layer, allowing for precise control over the direction of bending. For bidirectional actuator, constraining layers are positioned in the middle of the actuator. This placement facilitates balanced action on both sides, enabling the actuator to bend in either direction depending on the distribution of pressure across the chambers. Based on this approach, two prototypes were developed: the lateral support model and the longitudinal support model.

**Lateral Support Model**

This approach employs two V-shaped walls interconnected using lateral tie-restraining layers, as detailed in section 3.2.3. The sleeve main frame is designed for high flexibility, consisting of two hermetically sealed and isolated halves. Each half is connected to an air source via an air connector. Bending motion is achieved through the deployment of constraining layers, which divide the actuator into two segments and restrict linear movement at the midpoint, as illustrated in Figure 3.8. Compared to previous models, this design offers enhanced flexibility; however, while it demonstrates superior moment generation capabilities, it is limited by a maximum bending angle of less than 40 degrees. Additionally, this actuator exhibits increased rigidity.

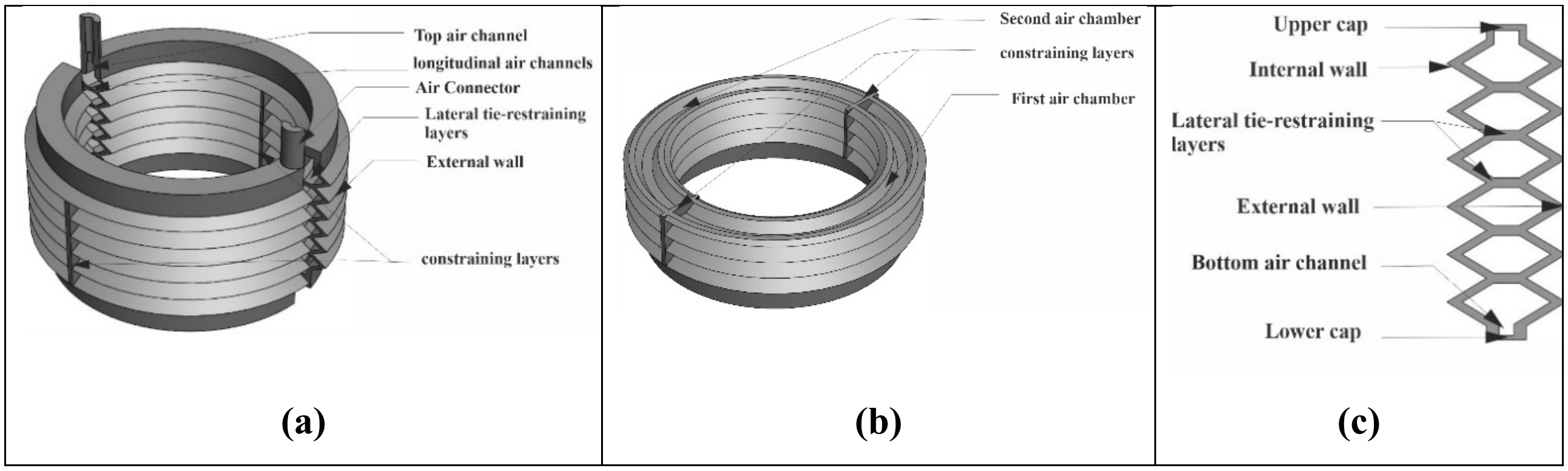

**Figure 3.8:** BSSA lateral support bending sleeve actuator (a) 3d representation (b) section view (c) wall design

**Longitudinal Support Model**

In the pursuit of enhancing actuator efficacy, a synergistic approach was employed, integrating the previously developed folded mechanism with a constraining method to generate bending motion. This approach features triangular folding walls connected by longitudinal tie-restraining layers, as detailed in Section 3.2.4. The design consists of two distinct halves, each forming a sealed air chamber, ensuring independent operation of each half. A key aspect of this design is the placement of custom constraining layers at the ends of each half, extending the full length of the actuator, as illustrated in Figure 3.9 (a). Within these halves is a designed air channel network comprising a series of longitudinal channels that span the entire length of the sleeve, with the top and bottom channels serving as nexus points to ensure connectivity between the chambers. A detailed representation of this bending actuator is provided in Figure 3.9.

The bi-directional capability of this actuator is particularly noteworthy, achieving a cumulative bending angle of 140 degrees, with individual bending reaching up to 70 degrees in either direction. The incorporation of the folded mechanism significantly improves its flexibility compared to its predecessors, making it suitable for integration into WMAD.

**BSSA Geometrical Parameters**

This section outlines the essential geometric parameters that define the performance and behavior of the BSSA, which shares several characteristics with the LSSA, as illustrated in Figures 3.9 (b) and 3.10 (c).

The fold angle is a key parameter that determines the internal angle at the junction of two converging inclined sides of a fold. Smaller fold angles have a significant impact on the bending angle, resulting in greater displacement and allowing for precise control over the actuator's bending motion. The tie-restraining layers function as the structural backbone of the actuator, ensuring that it maintains its intended shape. Without these layers, the actuator would lose its defined form, resembling a mere balloon (Figure 3.11 (d)). Two geometrical parameters associated with the tie-restraining layers are the thickness and the number of layers, which are explained in Section 3.2.6.

Wall thickness refers to the thickness of the actuator's material walls, which is vital for balancing flexibility with the ability to withstand varying pressure levels. Fold width is the horizontal distance between the peaks and valleys of the actuator's corrugations in its neutral state and plays a significant role in its ability to compress and expand. Constraining layers *tc* are non-flexible structures integrated within the actuator to restrict expansion on one side, thereby inducing bending motion. Their design and placement are critical for ensuring the actuator bends accurately and efficiently.

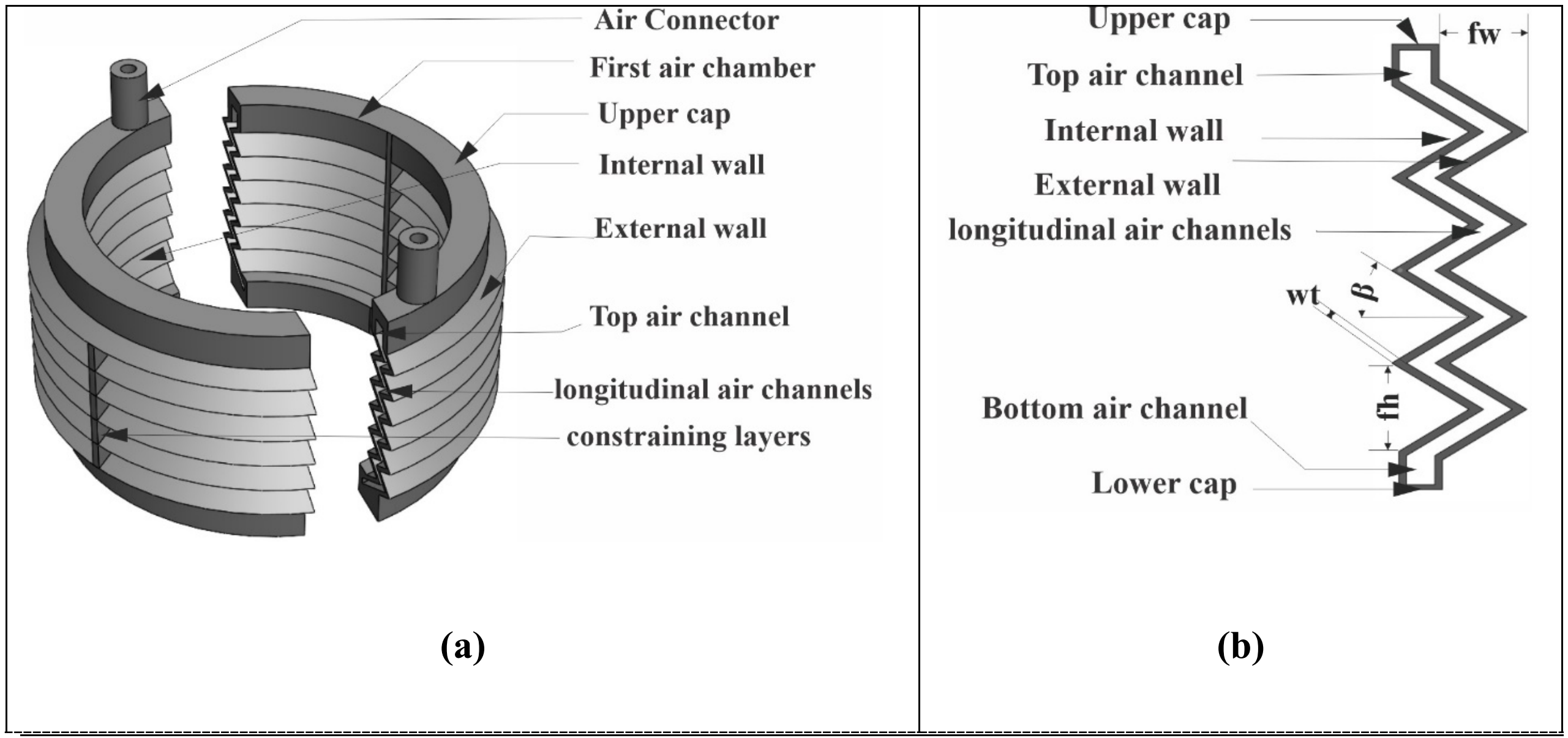

**Figure 3.9:** BSSA (a) 3d representation (b) wall design: fold angle (*β*), wall thickness (*wt*), fold width (*fw*)

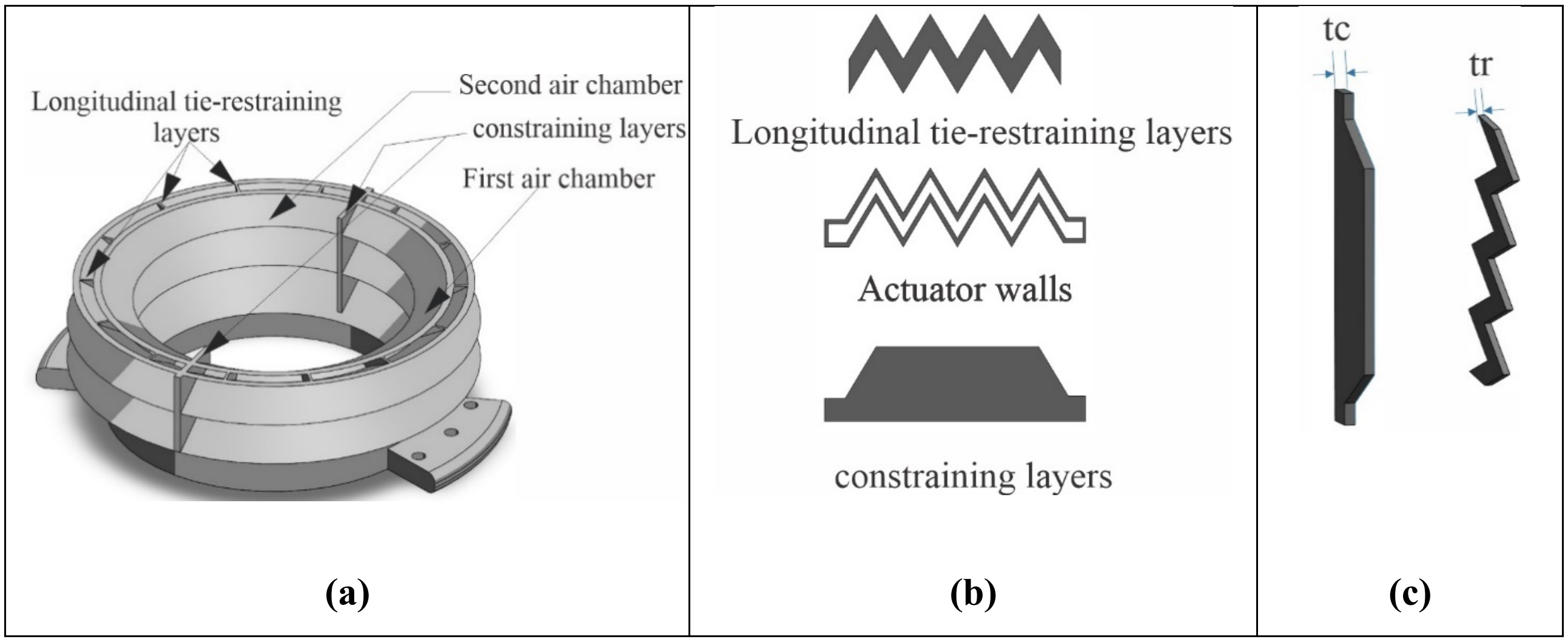

**Figure 3.10:** BSSA main components (a) section view (b) sketch of restraining layers, walls design and constraining layers (c) 3d representation of these layers

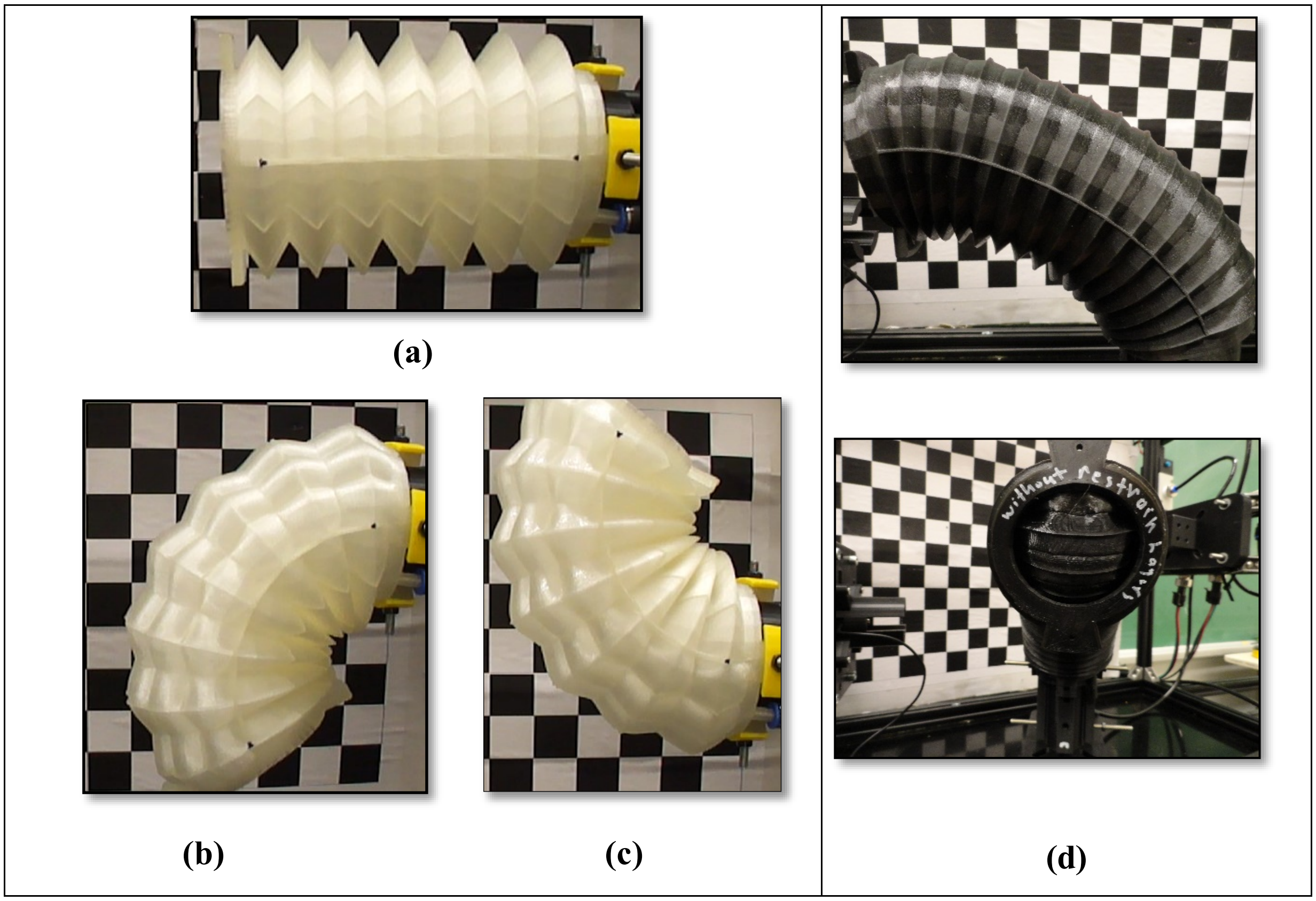

**Figure 3.11:** (a) BSSA prototype at relax mode (b) bending direction one (c) bending direction two (d) actuator without tie restraining layers

### 3.4 Twisting Actuator

WMAD often need to generate complex motions to accommodate multiple degrees of freedom. To meet these demands, a twisting soft sleeve actuator (TSSA) is developed. The central concept is based on the development of twist bellows, for generating rotational movements. These bellows are designed with a spiral configuration, suited to initiating and sustaining twisting motions. In its inactive state, each helical bellow maintains a wound configuration, like a coiled spring holding potential energy. This stable yet energized state establishes the foundation for a controlled actuation process. Upon the application of air pressure, each segment of the bellow expands. This expansion extends longitudinally along the spiral path of the bellow, adhering to its helical design. The natural tendency of the bellows to unwind under this expansion translates the linear force of the pneumatic pressure into rotational torque.

Conversely, when a vacuum is applied, the bellows contract. This contraction pulls the structure inward, effectively reversing the expansion forces. The bellows not only revert to their initial configuration but also actively rewind, generating a twisting motion in the opposite direction to that induced by inflation.

Building on the twisted bellow concept introduced in this study, two models were developed: the circular bellows-based architecture and the folded-bellow-based TSSA.

#### 3.4.1 Circular Bellows-based TSSA

The circular bellows-based architecture features a series of circular bellows arranged in a helical pattern along the actuator's length, facilitating uniform and controlled expansion. The bellows are interconnected by an upper air channel, which connects to a primary air connector attached to the pressure source. When air pressure is applied, each bellow expands and attempts to unwind, initiating a rotational motion. Several critical design variables influence the performance of this model:

The wall thickness significantly affects the actuator's flexibility and operational pressure. A minimum wall thickness of 1 mm is necessary to ensure airtightness while maintaining sufficient flexibility. The diameter of the circular bellows *rb* diameter directly impacts the twisting angle. While increasing the diameter can enhance the potential twisting angle, it also increases the sleeve's thickness, which may not be desirable for certain applications. The helical twist degree $\alpha$ is directly correlated with the twisting capability. A greater twist angle results in a larger generated twisting angle.

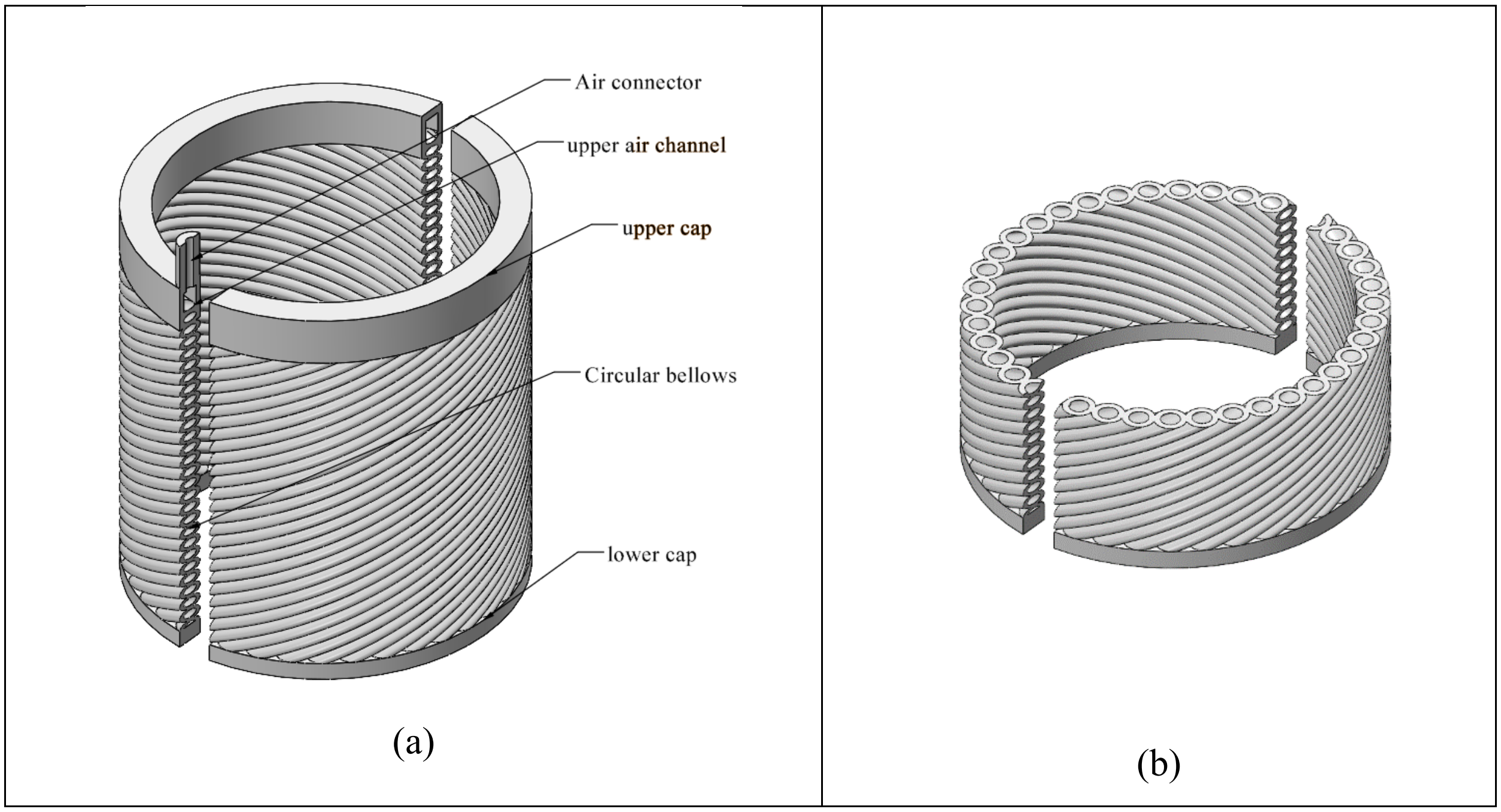

**Figure 3.12**: Circular bellows-based twisting sleeve actuator (a) 3d representation (b) section view

This model can achieve rotational angles exceeding 25 degrees. Notably, the actuator demonstrates superior performance under vacuum pressure compared to positive pressure. While the results are promising, a significant limitation of this model is its stiffness. Enhancements could be achieved by using materials with a lower Shore hardness, below 75 HA, to increase flexibility. Recognizing this limitation has led to the development of another model.

### 3.4.2 Folded-Bellow-based TSSA

Building on the concept of triangular folded bellows previously discussed, this model features bellows arranged in a helical configuration along the actuator's length (Figure 3.13). This geometrical design significantly enhances the actuator's flexibility, enabling a broader range of twisting motions by minimizing inherent material stiffness.

The folded bellows are sealed and isolated yet interconnected by a slim layer (S-layer) that links each bellow to its neighbors. This unique structure is critical for maintaining the integrity and uniformity of movement across the actuator. Air pressure distribution is managed through a top air channel that evenly disperses air across all bellows, while a bottom air channel ensures consistency at the base of the structure. Unlike the previous circular bellows-based model, the folded TSSA operates on the principle of unfolding rather than simple expansion. Upon the application of pneumatic pressure, each bellow segment extends longitudinally along its spiral path. This motion, driven by internal stresses and the unique folded geometry of the helix, efficiently transforms linear pneumatic force into rotational torque. Conversely, the application of a vacuum induces contraction, pulling the structure inward and reversing the expansion stresses, thereby enabling bidirectional rotational functionality. The performance of the folded TSSA is influenced by several key parameters:

Fold width is defined as the horizontal distance between the peaks and valleys of the actuator's corrugations. This parameter plays a critical role in determining the twisting angle achievable under positive pressure. The angle of the helix $\alpha$ is directly proportional to the extent of rotation, with a greater angle facilitating a larger twisting motion. Wall thickness is essential for ensuring airtight conditions and significantly impacts the actuator's flexibility.

This actuator demonstrates a significant advancement in flexibility compared to its predecessor, supporting a broader range of motion without compromising structural integrity. It retains the capability to rotate in specified directions depending on the applied pressure, positive pressure induces rotation in one direction, while negative pressure enables rotation in the opposite direction. Notably, the actuator demonstrates superior performance under positive pressure compared to vacuum pressure.

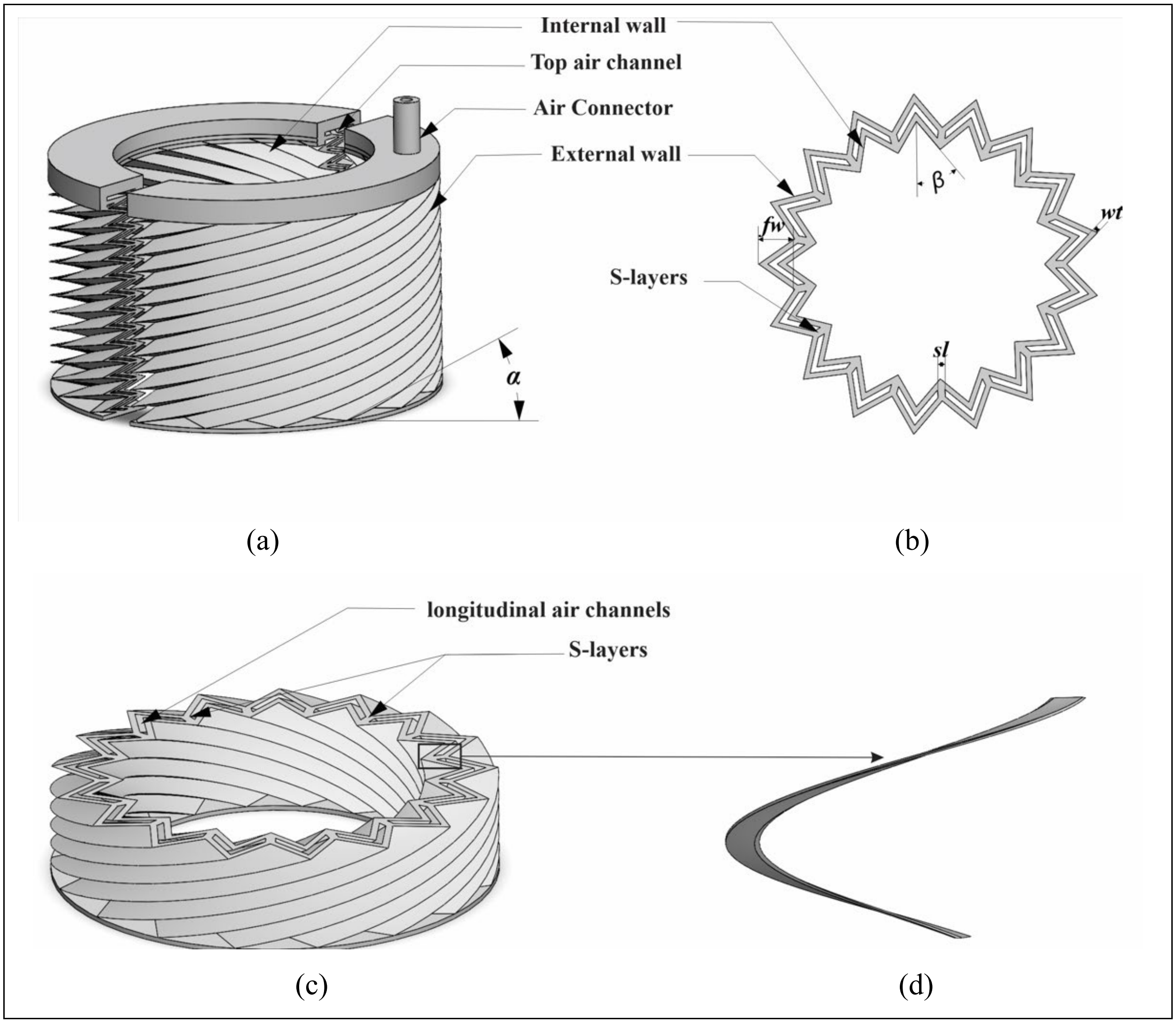

**Figure 3.13:** TSSA (a) 3d representation (b) wall design (c) section view (d) S-layer

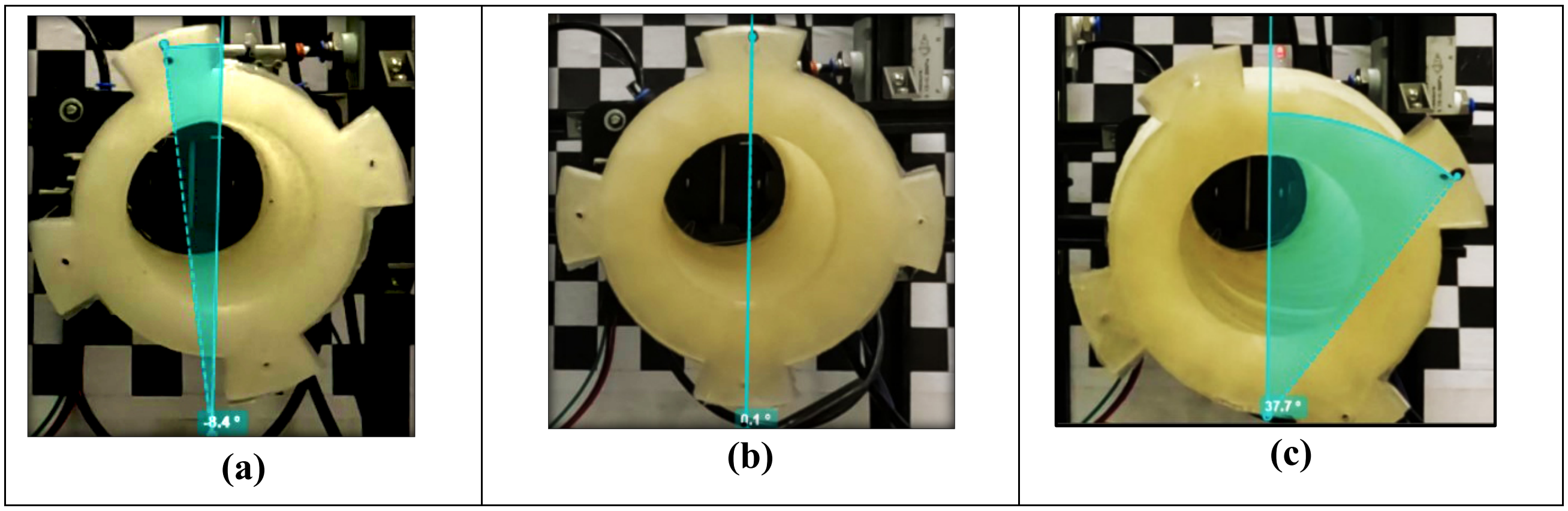

**Figure 3.14:** TSSA prototype (a) counterclockwise motion (b) resting status (c) clockwise motion

## 3.5 Differential Pressure Approach

While the constraining layer method effectively induces bending motion, it inherently limits the actuator's functionality to specific tasks. Many WMAD applications require actuators capable of performing wider ranges of motion to address diverse operational demands. To meet this need, this section introduces a differential pressure approach, which enables both linear and bending motions, facilitating the development of a versatile omnidirectional actuator.

This method utilizes actuators composed of highly flexible elastomeric bellows, specifically designed to endure significant deformation without sustaining permanent damage, an essential characteristic for repeated bending and straightening cycles. Multiple internal chambers are embedded within the actuator, with their arrangement to control the direction and degree of bending. Each chamber is independently controlled allowing precise manipulation of the actuator's configuration.

The core principle of this approach lies in the controlled application of differential pressure. Chambers on one side of the actuator are pressurized, causing the elastomeric material to expand while opposing chambers are subjected to reduced or negative pressure. This deliberate pressure imbalance induces asymmetrical deformation, resulting in bending toward the side with lower pressure. For purely linear motion, the pressures across all chambers are equalized, neutralizing bending effects and maintaining a straight configuration.

### 3.5.1 Omnidirectional Actuator

While the proposed LSSA and BSSA have demonstrated potential across a wide range of WMAD, there remains a need for an actuator capable of delivering a combination of different motion. To address these requirements, the omnidirectional soft sleeve actuator (OSSA) was developed as illustrated in Figure 3.15. The OSSA builds upon the operational principles of differential pressure, as previously detailed, utilizing a segmented structure with multiple air chambers aligned longitudinally along the actuator. These chambers are delineated by thin yet robust internal walls, effectively withstanding differential pressures without undesirable deformation.

When negative pressure (vacuum) is applied uniformly across these chambers, the actuator contracts. Conversely, applying positive pressure results in extension. To facilitate bending, certain chambers, designated as expansion chambers, are pressurized, while the opposite chambers, termed contraction chambers, are subjected to negative pressure. This arrangement creates an unbalanced force within the actuator’s structure, resulting in controlled bending motions.

Two novel OSSA models were developed: one using V-shaped bellows and the other featuring a folded bellows design. Each model leverages its unique geometry to enhance functionality, and both will be examined in detail in the following sections.

**OSSA V-shaped**

This model incorporates V-shaped bellows arranged into internal and external walls, interconnected via lateral tie-restraining layers, which enhance the structural integrity and functional alignment of the actuator. The design includes two distinct air chambers, each fully sealed and isolated from the other to prevent air transfer and ensure independent operation. Specialized sealed layers (S-layers) are employed to maintain airtightness between adjacent chambers.

While the V-shaped bellows model can achieve a contraction rate of 20% and an extension of up to 40%, it presents certain limitations primarily due to its structural design. The lateral support provided by the tie-restraining layers, although effective for maintaining shape and alignment, restricts the overall flexibility of the actuator. Additionally, the model's bending capacity is relatively limited, achieving a maximum bending angle of approximately 30 degrees. These limitations in flexibility and bending capacity prompted the development of an alternative model.

**OSSA Folded Bellows**

This actuator draws inspiration from the flexibility and agility of an elephant's trunk, translating these biological traits into a novel, bio-inspired device. The actuator comprises several air chambers, ranging from two to four, each configured as a separate, airtight compartment. To enhance flexibility, a folded bellows mechanism, previously discussed, has been integrated into the design. The isolation of the chambers is achieved through the incorporation of sealant layers, which effectively prevent cross-contamination and interaction between adjacent compartments (Figure 3.15). Additionally, each chamber is equipped with a dedicated air connector, allowing independent connection to an air source.

The operational principle of this actuator is based on the discussed differential pressure principle. For generating extension, all chambers are inflated, leading to the elongation of the actuator. Conversely, applying a vacuum across all chambers induces contraction. Bending motion is achieved by selectively pressurizing certain chambers while evacuating others, creating an imbalance that results in directional bending.

For applications requiring multidirectional bending, the design includes up to four air chambers (Figure 3.15). This configuration enables the actuator to bend in multiple directions (Figure 3.16), enhancing its utility in diverse settings. In operational terms, this actuator can achieve linear motion, up to 80% extension and 30% contraction. It can also perform bending motions to angles of +60 degrees and -60 degrees, demonstrating its versatility and effectiveness in achieving complex movements.

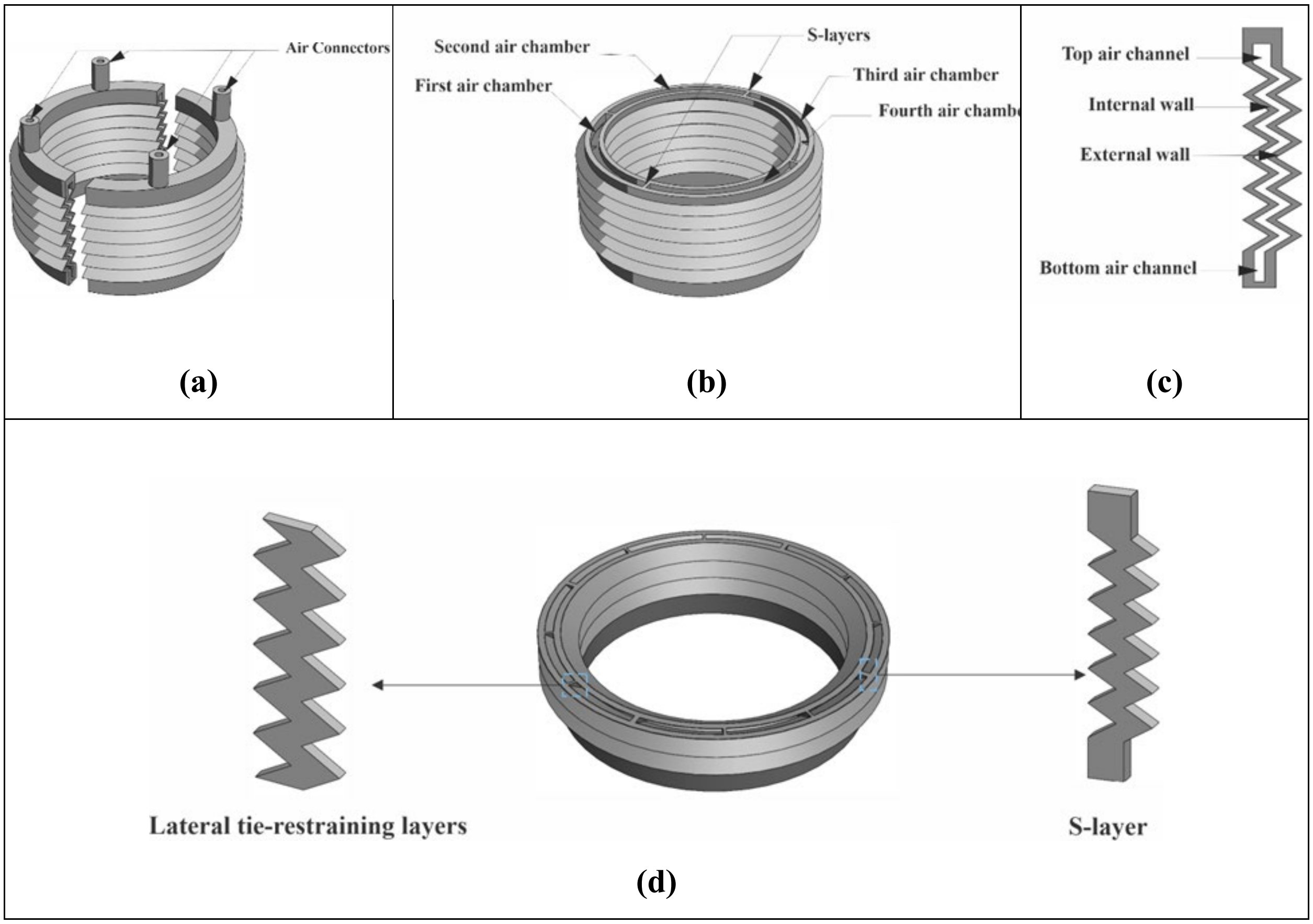

**Figure 3.15:** OSSA (a) 3d representation (b) section view (c) wall design (d) 3d representation of restraining layer and s-layers

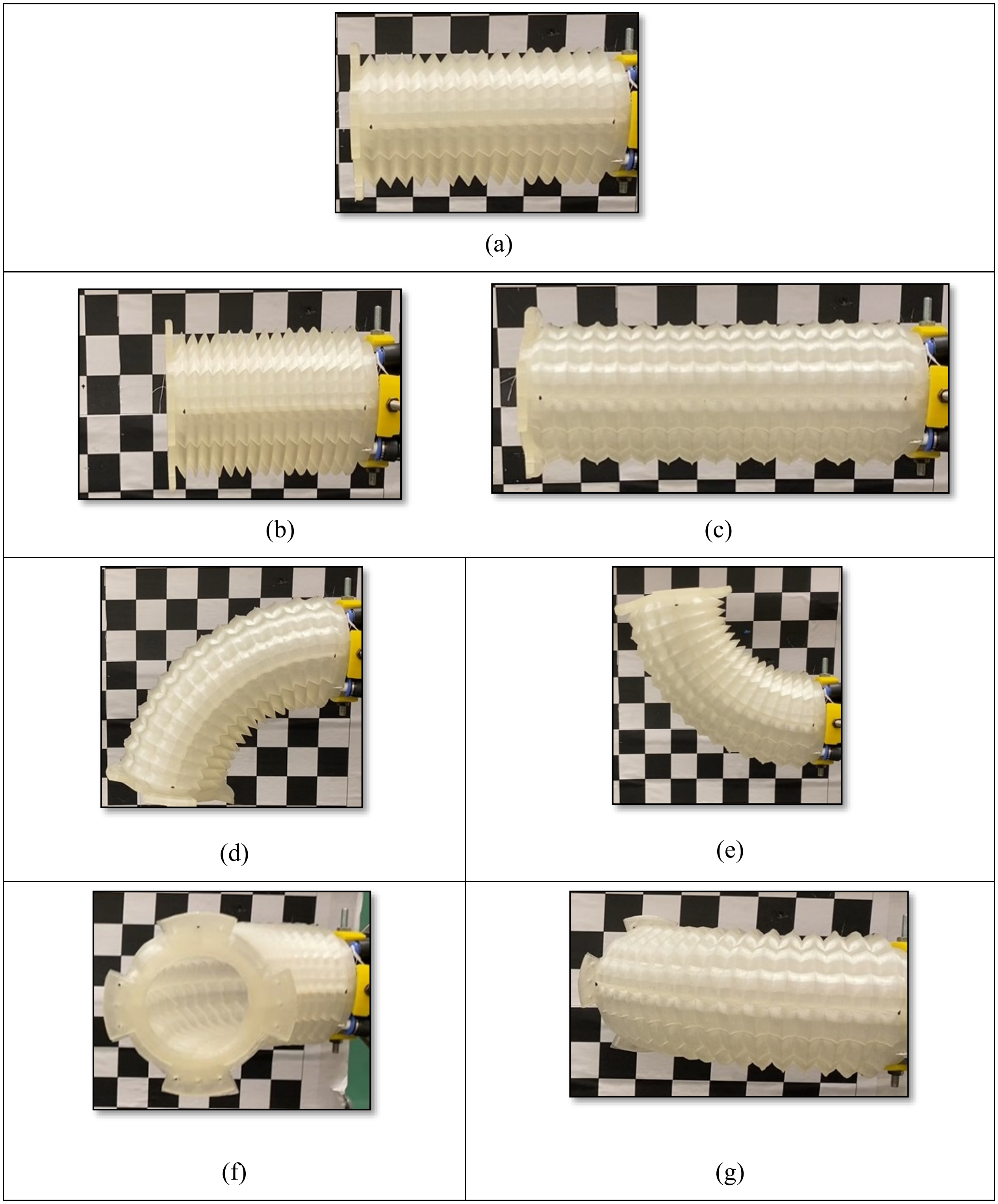

**Figure 3.16:** (a) OSSA prototype at relax mode (b) extension (c) contraction (d) bending direction one (e) bending direction two (f) bending direction three (g) bending direction four

### 3.6 Conclusion

This chapter presented the conceptualization, design, and development of the Soft Sleeve Actuators. It systematically introduced several innovative approaches to sleeve design, emphasizing their potential to enhance flexibility, motion, and adaptability while minimizing reliance on external attachment mechanisms.

The linear actuator designs progressed through iterative improvements, addressing inherent limitations such as radial expansion and limited displacement. The transition from ring-supported designs to longitudinally folded mechanism exemplifies how geometric innovation and the integration of tie-restraining layers can substantially improve linear motion efficiency while ensuring structural stability. Similarly, the bending actuators, developed through extension variance and constraining layer approaches, demonstrated significant advancements in generating controlled and directional bending motions. The incorporation of the folded constraining layer approach further established its suitability for complex applications by enabling bi-directional bending with enhanced flexibility and performance.

For twisting motion, the development of helical and folded-bellow actuators marked a significant advancement, offering enhanced rotational capabilities and superior performance under diverse pneumatic conditions. The differential pressure-based omnidirectional actuators demonstrated the transformative potential of integrating multiple motion modalities, enabling transitions between linear, bending, and rotational motions. These actuators showcased improved flexibility and operational efficiency, making them highly suitable for the diverse demands of wearable motion-assist devices (WMAD).

Table 3.1 provides a summary of the advantages and limitations of each actuator model, highlighting the progression from initial concepts to advanced designs. While each model addressed specific challenges, certain limitations persist, including increased manufacturing complexity and material constraints.

**Table 3.1:** Summary of SSA models

| Motion | Approach | Models | Advantages | Limitation |
|---|---|---|---|---|
| **Linear motion** | Ring support | Ring reinforced v-section | - Generates linear motion<br>- Relatively simple to manufacture | - Limited extension (15%) and contraction (5%)<br>- Rigid with reduced flexibility |
| | Lateral support | Semicircular cross-section | - Generates linear motion<br>- More flexible than ring support model | - Limited extension (24%) and contraction (13%)<br>- Rigid with reduced flexibility |
| | | Triangular cross-section | - Generates linear motion<br>- More flexible than circular bellows. | - Moderate extension (40%) and contraction (20%)<br>- Rigid with reduced flexibility<br>- Large profile |
| | Longitudinal support | Longitudinally folded | - Flexible<br>- Slim design<br>- Fast actuation | - Challenge to fabricate via conventional techniques. |
| **Bending motion** | Extension variance | Planar-base | - Generates bending motion | - Rigid with reduced flexibility<br>- High actuation pressure<br>- Limited bending angles ($20^{o}$) |
| | Constraining layer | Lateral reinforcement | - Generates bending motion<br>- Low actuation pressure | - Moderate bending angles ($40^{o}$)<br>- Moderate flexibility<br>- Large profile |
| | | Longitudinal support | - Flexible<br>- Low actuation pressure<br>- Bending angles ($140^{o}$)<br>- Compact profile | - Challenge to fabricate via conventional techniques |
| **Twisting motion** | Twisting bellows | Helical circular | - Generates rotary motion<br>- Twist angle ($25^{o}$) under vacuum | - Limited flexibility.<br>- Limited twist angle ($15^{o}$) under positive pressure.<br>- Slow actuation |
| | | Twisted folded | - Generates rotary motion<br>- Flexible<br>- Twist angle ($45^{o}$)<br>- Compact profile | - Limited twist angle ($10^{o}$)under vacuum. |
| **Omni-directional** | Differential pressure | V shaped | - Generates linear and bending motions | - Limited flexibility<br>- High actuation pressure<br>- Moderate displacement and bending angles.<br>- Large profile |
| | | Folded | - Flexible<br>- Low actuation pressure<br>- Superior displacement and bending angles<br>- Compact profile | - Challenge to fabricate via conventional techniques<br>- Dual-actuation mode |

# CHAPTER 4

# MANUFACTURING

The fabrication of intricate soft pneumatic actuators remains a developing field, facing considerable challenges when employing traditional manufacturing processes. These processes present advantages and limitations that influence the fabrication and performance of the final models. The following chapter explores various manufacturing techniques used for producing the proposed SSAs, emphasizing the challenges and constraints that made certain approaches unsuitable for this application. The final section details the developed 3D printing manufacturing process, outlining the methods employed to ensure the actuator's high quality and airtight construction.

## 4.1 Heat Shrink

The initial manufacturing method explored for producing SSAs involved the heat shrink technique, which uses heat-shrinkable tubing (3M heat-shrink tubing FP-301) to conform tightly around a designed 3D-printed mold when heated as shown in Figure 4.1. This approach enables rapid and consistent fabrication. The manufacturing process involves several critical steps, as outlined below:

**Mold design:** The geometry of the mold determines the actuator's final shape, rendering mold design a critical initial step. In this approach, the mold is first enclosed by relaxed heat-shrink tubing, which, upon heating, contracts around the mold. Initial attempts employed Acrylonitrile Butadiene Styrene (ABS) but localized internal temperatures exceeding its 105°C glass transition threshold led to deformation. To mitigate this issue, polycarbonate with a glass transition temperature of 150°C was subsequently chosen. This material selection preserved the mold's dimensional stability under the stresses imposed by the contracting tubing.

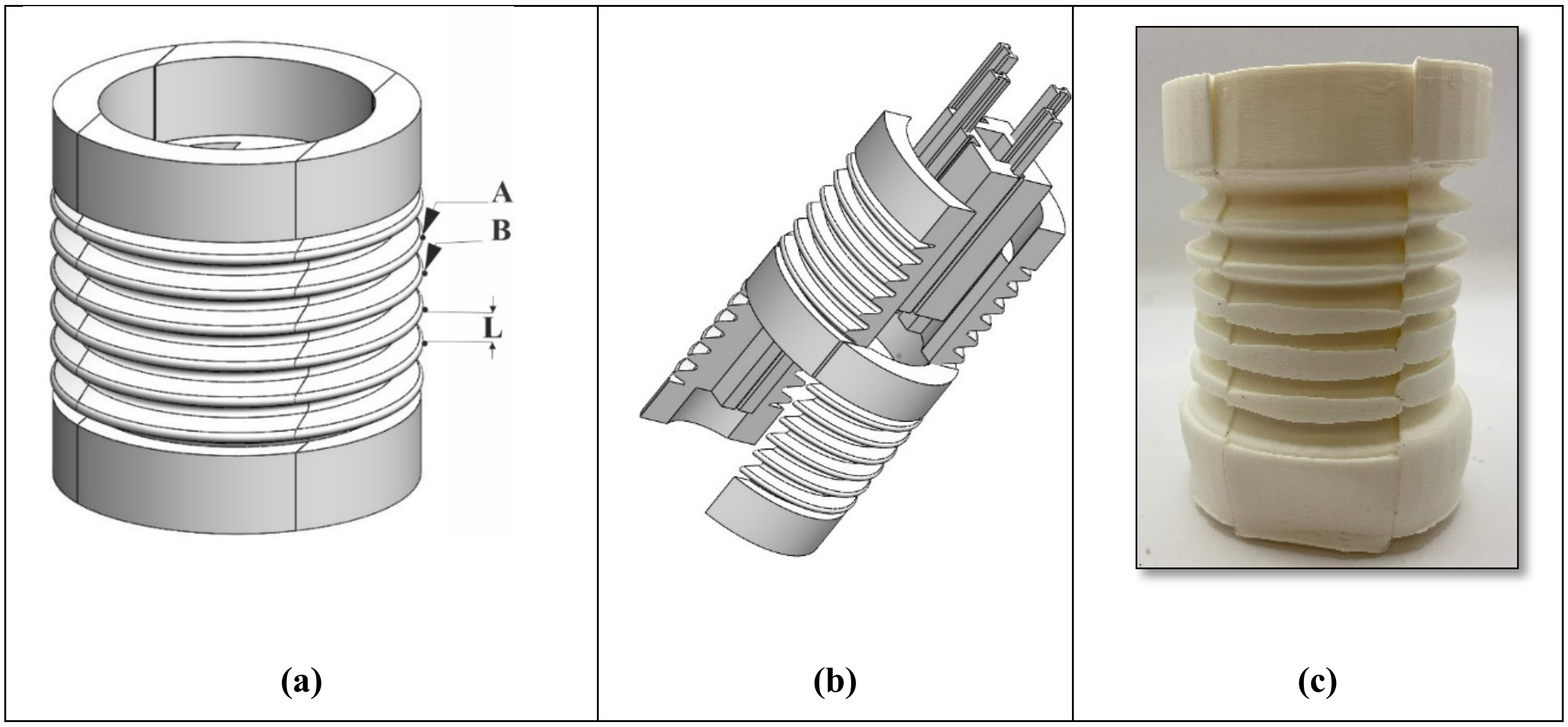

**Figure 4.1:** SSA heat-shrink mold (a) complete mold assembly, (b) disassembled mold parts (c) mold after heat exposure

**Heating:** The next step involved exposing the mold and heat-shrink assembly to controlled heat. Uniform and consistent heating is critical to achieve predictable and even contraction of the tubing around the mold and other components. Several heating methods, including heat guns, ovens, and boiling water, were assessed. Among these, boiling water was determined to be the most effective, providing uniform and predictable heat distribution.

**Mold removal:** Extracting the mold without damaging the thin actuator walls presented a challenge. To address this, the mold was redesigned as a modular assembly composed of interlocking segments. During removal, the central section was extracted first, allowing the peripheral sections to be drawn inward and removed (Figure 4.1 (b)). I shape keys were incorporated to ensure secure interconnections among the segments during the shaping process while enabling their removal without compromising the structural integrity of the actuator.

**Cap integration:** Finally, the actuator's upper and lower caps, fabricated from ABS, were integrated with the actuator's walls. Following this assembly, only the section containing the caps was exposed to heat. Under these localized heating conditions, the tubing contracted around the caps, ensuring a secure fit and preserving the actuator's structural integrity.

**Limitations of Heat Shrink Fabrication**

A key limitation of using 3M heat-shrink tubing FP-301 in the fabrication process is its insufficient contraction ratio, capped at 4:1, which fails to adequately form the valleys of the SSA design. During heating, the tubing anchors at the mold peaks, and subsequent contraction is limited to the span between these points. Attempts to force further contraction by applying additional heat led to tearing of the tubing. Additional drawbacks include the rigidity of the end caps, which reduces the actuator's overall flexibility, and frequent issues with incomplete airtightness, compromising the actuator's performance and efficiency.

## 4.2 Casting

The traditional casting method for fabricating SSA is performed by pouring silicone elastomer mixtures into custom-designed molds, allowing them to cure at room temperature. This straightforward approach, requiring minimal equipment, is well-suited for iterative designs and prototyping for the SSAs. The following steps describe the adopted manufacturing process.

**Mold design and fabrication:** Molds design consist of multiple components, internal and external walls, upper cap, and inner cap. These parts were 3D-printed using PLA material, as shown in Figure 4.2 (b).

**Material preparation:** Dragon Skin silicone rubber from Smooth-On (Easton, PA, USA) is selected for its elasticity and durability. The two-part silicone system is mixed and degassed in a vacuum chamber for 30 minutes to eliminate air bubbles and ensure optimal mechanical properties.

**Casting:** Silicone is poured into molds, with the internal and external walls cast first, followed by the upper and lower caps in a single step, using custom design molds.

**Curing and demolding:** The assembly cures for 16 hours. Molds, designed in four parts, simplify the demolding process to protect the actuator’s delicate features. Silicone sealant is applied at mold interfaces to prevent leakage during casting.

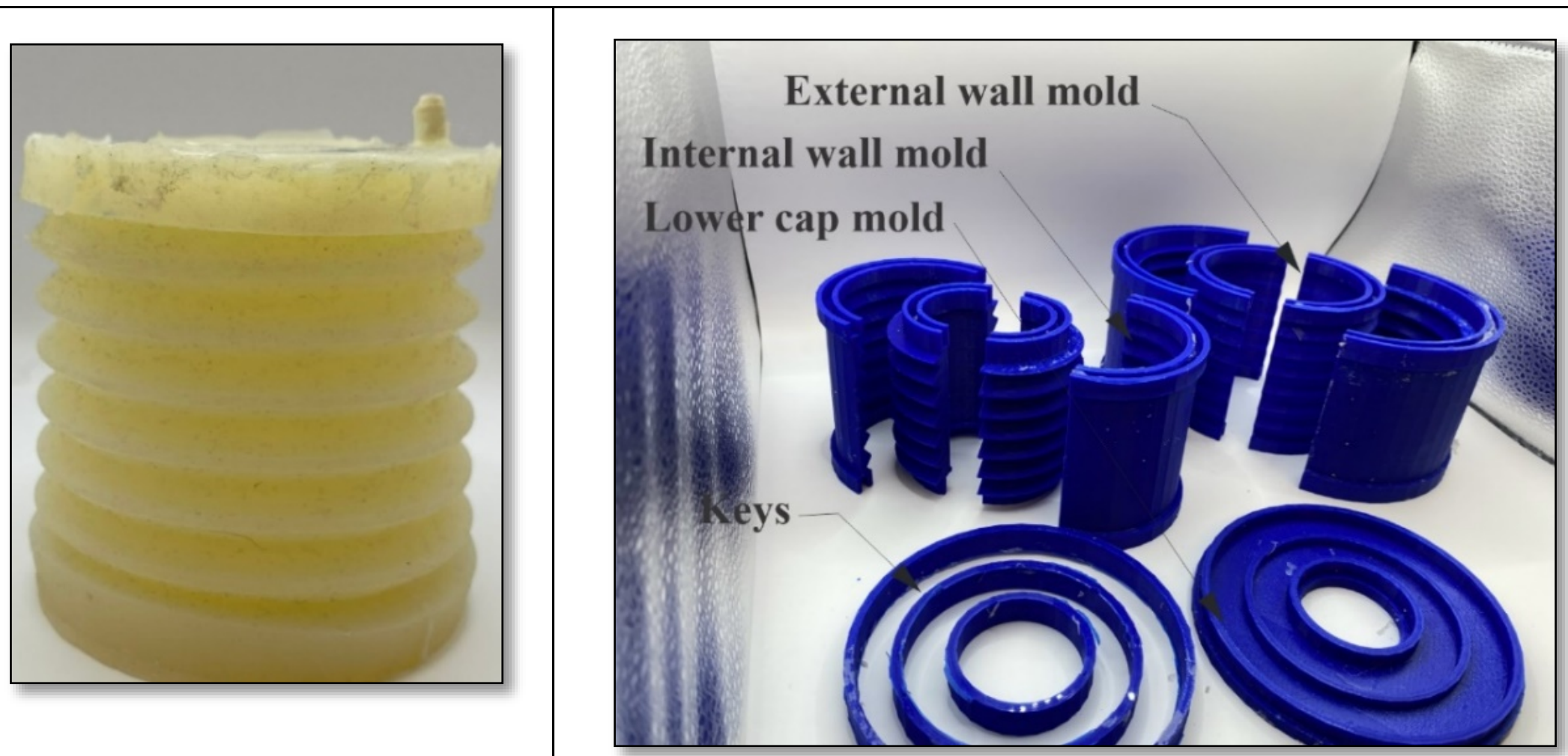

(a) (b)

**Figure 4.2:** SSA casting (a) prototype produced by casting (b) disassembled molds used in fabrication

### Limitations of Casting Fabrication

While the casting process is effective for fabricating basic components in soft robotics, it faces significant limitations for manufacturing intricate designs such as the SSAs. Among other limitations, mechanical integrity, airtightness and internal defects were among the primary challenges. Multi-part assemblies, like the SSA, are prone to mechanical failure under pressures above 200 kPa due to weak bonds at assembly junctions. The casting method often results in porous actuator walls with voids and holes, compromising air tightness. A wall thickness of at least 5 mm was required for successful fabrication, which made the actuator rigid and bulky. Moreover, gas entrapment during casting created voids and discontinuities in the walls. Vacuum application offers limited improvement. Due to these limitations, the casting method is more suited to simpler designs where precision and intricate detailing are less critical.

## 4.3 Additive Manufacturing

Additive manufacturing, characterized by the layer-by-layer deposition of material to construct components from 3D models, fundamentally differs from traditional subtractive methods. Its applicability in fabricating SSA is evident, but the inherent flexibility of the materials introduces unique challenges. This study evaluated the most suitable 3D printing technologies for SSA production, categorized by the physical state of their feedstock, namely, powder-based, liquid-based, and filament-based techniques.

### 4.3.1 Powder-Based

Powder-based additive manufacturing encompasses a variety of methods, each offering distinct capabilities. Within this category, Selective Laser Sintering (SLS) and Multi Jet Fusion (MJF) are currently the only techniques capable of processing flexible polymers. In contrast, other powder-based technologies such as Direct Metal Laser Sintering (DMLS), Selective Laser Melting (SLM), and Electron Beam Melting (EBM) are unsuitable for flexible components.

SLS utilizes a $CO_2$ laser to selectively sinter thermoplastic powder. Each layer is partially fused at temperatures below the polymer's melting point. The build platform is incrementally lowered, and new powder layers are deposited and preheated. This process enables the fabrication of complex geometries without requiring support structures. MJF applies binding agents to the powder bed and fuses the material using an infrared heat source. This approach allows precise control over mechanical properties, including graded flexibility, by selectively varying the binding agent and heat application.

Despite their capabilities, both SLS and MJF were found limited in their application to SSA fabrication. Residual powder removal from intricate internal cavities (1 mm) is challenging, making these techniques unsuitable for the complex geometries typical of SSAs.

### 4.3.2 Liquid-Based

Liquid-based additive manufacturing processes cure or solidify liquid resins or viscous inks to create 3D structures. Three principal technologies, Direct Ink Writing (DIW), PolyJet, Stereolithography (SLA), and Digital Light Processing (DLP) methods were considered for fabricating the SSA.

DIW extrudes shear-thinning inks to create multilayered structures at moderate pressures. While suitable for flexible materials, it faces issues such as structural deformation, nozzle clogging, slow printing speeds, and complex post-processing, limiting its practicality for SSA production. The PolyJet method deposits liquid photopolymers layer by layer, curing each with UV light. It supports complex geometries using soluble supports and multiple materials. However, the high cost of equipment and materials restricts its use for SSA fabrication. SLA and DLP, both technologies cure photosensitive resin using UV light, with DLP curing entire layers at once for faster production. These methods potentially excel in fabricating SSAs due to their ability to remove uncured resin from internal cavities, enabling intricate SSA designs.

Prototyping and preliminary investigations into 3D printing SSAs using DLP technology were conducted with the Elegoo Mars 2 printer and flexible resins, Liqcreate Premium Flex (Shore hardness 63A) and Flexible-X (Shore hardness 55A), as shown in Figure 4.3. The process included the following steps:

**Resin preparation:** The resin was shaken to ensure uniformity and then degassed in a vacuum chamber for 15 minutes to remove air bubbles that could compromise print quality.

**Optimizing print parameters:** Extensive testing determined the optimal settings for printing with these resins, as summarized in Table 4.1.

**Table 4.1:** Optimal printing parameters for SLA printer

| Resin | Layer thickness (μm) | Exposure time (s) | Base layer count | Base layer exposure (s) | Lift Height (mm) | Lift / retract speed (mm/min) | Light off-delay / bottom (s) |
|---|---|---|---|---|---|---|---|
| **Premium Flex** | 50 | 5.5 | 5 | 60 | 10 | 60 / 90 | 2/ 3 |
| **Premium Flex** | 50 | 17 | 2 | 60 | 10 | 90 / 120 | 1 / 3 |

**Orientation and preparation of hollow parts:** To address vacuum effects during printing, the parts were oriented at a 45° angle relative to the build plate, thereby reducing suction forces. Venting holes were also added to equalize internal pressures and prevent air entrapment. Post-printing, these holes were sealed with a custom mold and UV curing.

**Post-processing:** Printed actuators were washed in 95% ethanol to remove unreacted resin, followed by UV curing for 45 minutes to ensure stability and durability.

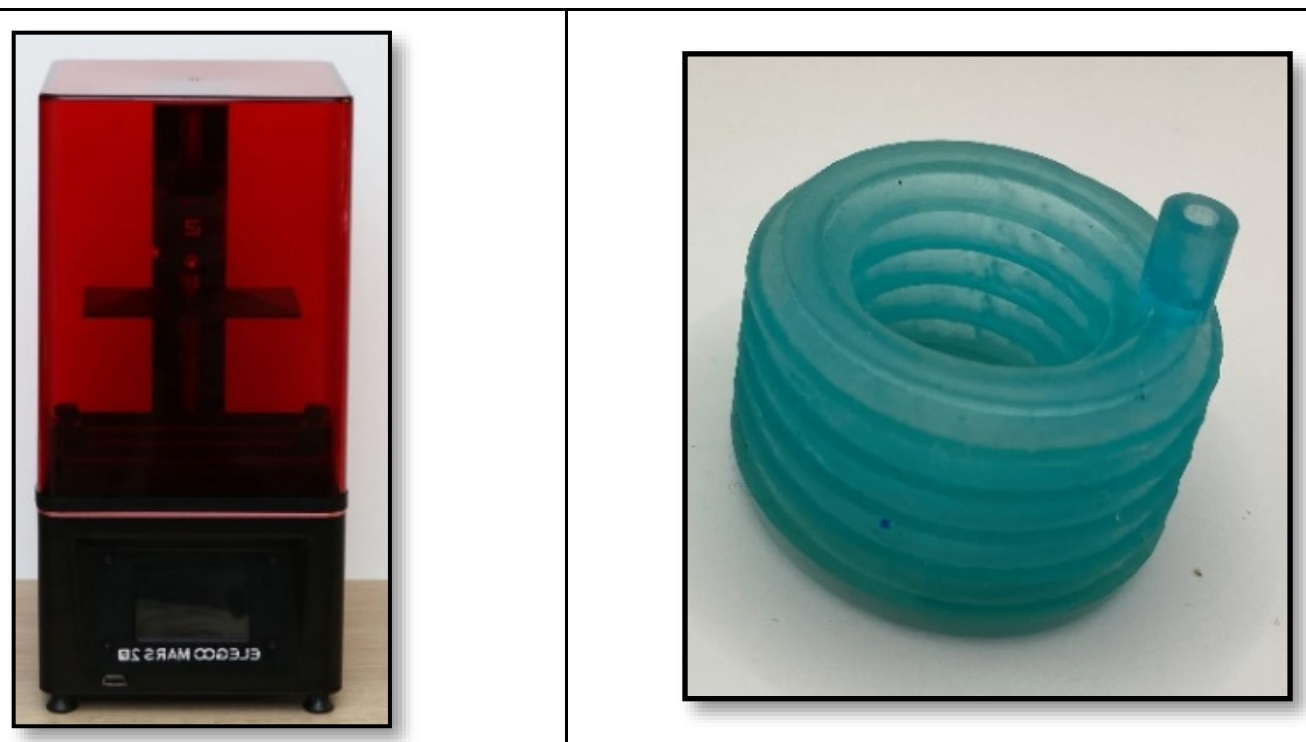

(a) (b)

**Figure 4.3:** (a) SLA printer used for the experiment (b) finished actuator produced by SLA.

**Limitations of DLP fabrication**

Despite the successful production of the SSA prototype, the use of DLP and flexible resins presents several challenges. The high cost of flexible resins compared to other additive materials, the emission of potentially harmful gases requiring specialized ventilation, and the limited elongation capacity of the resins (below 150%) are significant drawbacks. These factors make this method less viable as a final manufacturing solution for SSA.

### 4.3.3 Fused Deposition Modeling

Fused Deposition Modeling (FDM) is a widely used 3D printing technology that constructs objects through a layer-by-layer deposition of molten thermoplastic filaments. The process begins with solid filaments being heated above their glass transition temperature to form a viscous, flowable state. This molten filament is then extruded in a predefined path and solidifies upon cooling to create rigid structures. FDM's operational mechanics rely on material extrusion and thermal dynamics, ensuring that the filament flows precisely through a thermally regulated nozzle to build the object layer by layer.

When printing with flexible materials like Thermoplastic Polyurethane (TPU), additional challenges arise due to TPU's elasticity and flexibility. These properties require careful management of the extrusion process, balancing extrusion temperature, speed, and cooling rates to maintain material integrity and dimensional accuracy. The extrusion system in FDM printers can be classified into two main configurations: Direct Drive and Bowden extruders, each with unique advantages and challenges.

In Direct Drive systems, the extruder is mounted directly on the print head, providing a compact and efficient filament path that reduces the distance the filament travels before reaching the hot end. This design ensures more consistent flow and pressure during extrusion, which is critical when printing flexible materials like TPU. However, when using the Prusa i3 MK3S printer with a Direct Drive system to fabricate SSA, four major issues were identified. These include vibration, which causes layer displacement and misalignment for flexible materials; poor accuracy in the x and y directions due to vibration and the extruder's weight; filament entanglement, which disrupts long prints; and frequent nozzle clogs that are difficult to clear. These problems stem from the inherent design of Direct Drive extruders and could not be resolved with the Prusa i3 MK3S printer, leading to the exclusion of this setup in the study.

The Bowden extruder presents an alternative configuration in FDM systems. Here, the extruder motor is detached from the print head and mounted on the printer's frame, with a flexible tube guiding the filament to the print head (Figure 4.4). The primary benefit of this design is the reduced weight of the print head, which improves print speed and precision. Additionally, the Bowden system requires less maintenance for the print head. However, this setup also introduces challenges, particularly when printing with flexible materials like TPU. The extended travel distance of the filament increases friction, the potential for filament buckling, and delayed responses to extruder commands. These issues, coupled with the properties of flexible filaments, necessitate careful tuning of extrusion parameters to prevent jams and ensure consistent material flow.

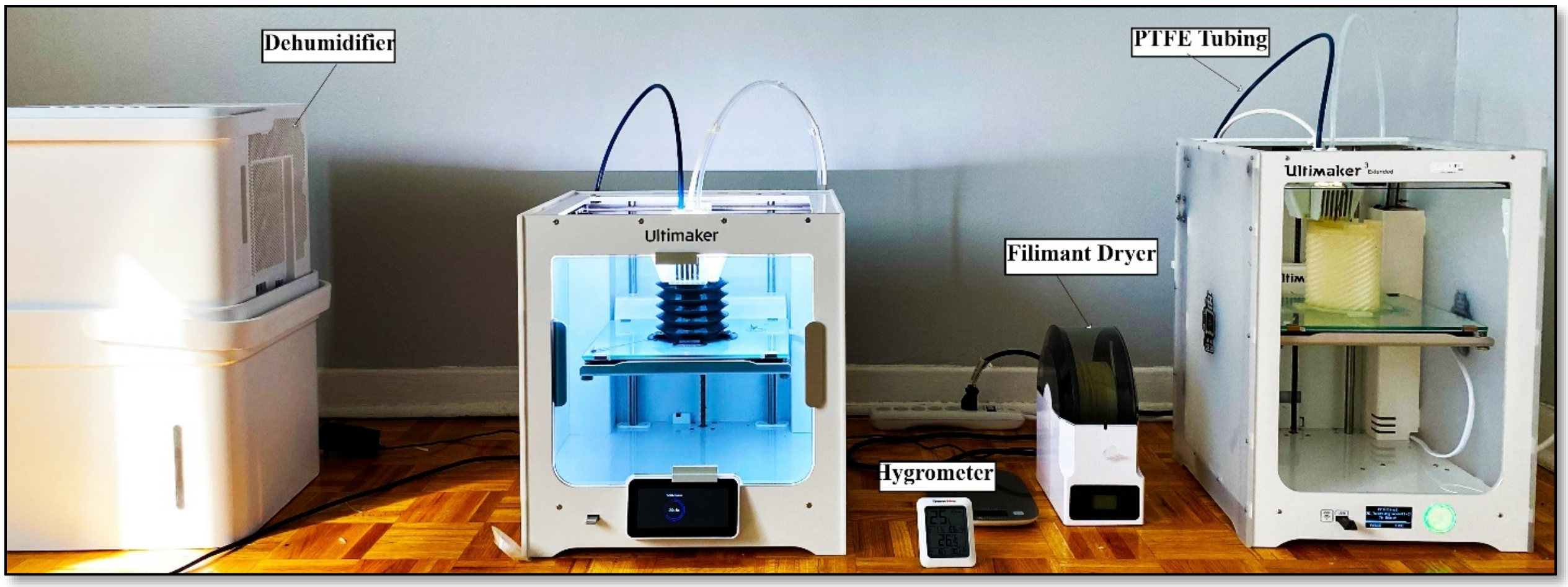

**Figure 4.4:** Controlled-environment additive manufacturing setup for SSA production

**FDM Bowden-Based fabrication**

Current literature reveals a gap in the use of Bowden printers for manufacturing airtight actuators, likely due to technical limitations. To address this, the study introduces a novel framework aimed at optimizing the production of the SSA, particularly focusing on the challenges of working with flexible materials like TPU in Bowden extruder configurations. The framework was successfully validated using two Bowden-based printers, the Ultimaker 3 Extended and Ultimaker S3, leading to the production of over 100 SSA models with various designs. This success demonstrates the framework's adaptability and potential to advance flexible 3D printing in actuator manufacturing. The principal steps involved in this framework are outlined below.

**Minimize the negative effect of humidity:** TPU's hygroscopic nature leads to moisture absorption, causing filament swelling, changes in material properties (such as reduced elasticity and increased brittleness), and steam formation during extrusion, which disrupts material flow and weakens the print. To mitigate these issues, the framework applied two strategies: controlling ambient humidity with a dehumidifier to maintain levels below 20% and pre-treating the filament by drying it for 5 hours at 50°C using a filament dryer with silica gel packets. These measures help maintain consistent print quality and material properties.

**Resolve material extrusion issues:** Extruding flexible TPUs causes the filament to buckle, kink, or coil inside the extruder, leading to inconsistent flow or complete stoppages. A four-step methodology was developed to mitigate these issues. Firstly, a proper tension is essential for consistent filament flow. Too much tension can deform the filament, causing blockages, while too little can result in under-extrusion. The methodology involves adjusting tension incrementally to find an optimal balance for smooth extrusion. Secondly, the reduction of friction and buckling in the Bowden Tube by replacing the standard tube with a short PTFE tube with ultra-low friction reduces resistance, minimizing buckling and preventing filament kinking or coiling. Thirdly, enhancing nozzle cleaning since flexible filaments are sensitive to clogs due to their low viscosity. The framework uses two cleaning methods: cold pulls (to remove residue) and hot cleaning (to clear tougher obstructions), ensuring the nozzle remains clear and functional. Lastly, retraction management, due to the viscoelastic nature of flexible filaments, retraction can cause issues like filament stretching, inconsistent pressure, and wear. In this framework, the researcher recommends disabling retraction when printing with flexible materials to avoid these issues, while optimizing parameters such as temperature and speed to mitigate stringing or oozing.

**Eliminate support material:** In this study, employing polyvinyl alcohol (PVA) as a soluble support material with TPU in SSA fabrication highlighted critical compatibility issues. The chemical interactions between PVA and TPU during the printing process adversely impacted TPU's interlayer adhesion, compromising the structural integrity of the material. Notably, the removal of PVA supports frequently resulted in gaps within the actuator walls, increasing susceptibility to air leakage. Furthermore, the differing thermal properties of PVA and TPU exacerbated TPU's tendency to ooze and form strings. These temperature disparities caused extensive stringing, where extraneous TPU strands formed between print sections, significantly reducing the print quality of the SSA. Subsequent experiments using TPU-based support materials yielded slight improvements in print quality but introduced new challenges. These supports were difficult to remove and often damaged the actuator walls during post-processing. The removal process frequently caused surface micro-abrasions and deformities, negatively affecting the operation of the actuators. Moreover, the voids left by these supports critically compromised the airtightness essential for pneumatic functionality, highlighting the persistent challenges associated with using support materials in actuator fabrication.

To address these challenges, this study introduces the 30-degree rule to eliminate the need for external support materials by optimizing print orientation and geometry design. The methodology lies in determining the minimum overhang angle at which structures can be printed independently without requiring support. Empirical analysis conducted in this study identified that overhangs with angles below 30 degrees result in unacceptable drooping and structural weaknesses. By adhering to this threshold, the layer-by-layer deposition process is optimized, ensuring that each new layer is adequately supported by the one beneath it. This approach not only minimizes reliance on support materials but also reduces material waste. Furthermore, printing at a 30-degree angle enhances the cooling and solidification rates of each layer, improving the structural integrity and surface finish of overhangs.

**Bed adhesion challenges:** Resolving bed adhesion challenges is essential for achieving reliable 3D printing results. The flexibility and low stiffness of TPU often lead to inadequate adhesion to the build plate, causing warping or detachment during the printing process. These issues are further complicated by TPU's variable thermal expansion under different temperature gradients. To enhance bed adhesion and ensure consistent results, several steps were implemented. The application of adhesives, such as glue sticks or 3D printing adhesives, significantly improved the initial adhesion and stability of TPU during printing. Among these, Elmer’s glue sticks proved particularly effective, providing strong bonding for flexible materials while being easily washable and preserving the cleanliness of the build plate. Precise control of the build plate temperature further enhanced TPU adhesion. Through testing, a bed temperature of 50°C was identified as optimal, ensuring strong adhesion and reducing the risk of warping or detachment. Additionally, the use of adhesion assistants, specifically the “brim” technique, provided an effective mechanism for increasing attachment to the build plate. By expanding the object’s footprint on the build surface, the brim added stability and minimized detachment risks.

**Optimize extrusion temperature:** Optimizing extrusion temperature is essential for ensuring high-quality prints and maintaining the mechanical integrity of the SSA. This study employed a systematic protocol to determine the optimal extrusion temperature for a newly introduced filament brand. The process began with a baseline temperature of 260°C, adjusted in 5°C increments based on observed printing behavior. If excessive fluidity manifested as stringing or oozing was detected, the temperature was reduced in 5°C decrements. Conversely, if the filament displayed brittleness or inadequate layer adhesion, the temperature was incrementally increased. Following this iterative approach, an optimal extrusion temperature of 235°C was identified as ideal for producing high quality SSA models.

Temperatures below 225°C adversely affected the printing process. At these lower temperatures, the filament exhibited excessive viscosity, impeding smooth flow through the nozzle and increasing the likelihood of clogging and under-extrusion. Insufficient thermal energy hindered proper interlayer bonding, resulting in weak mechanical integrity. This weak bonding led to the formation of micro-voids and discontinuities within the actuator’s walls, compromising airtightness and significantly diminishing pneumatic performance.

In contrast, temperatures exceeding 240°C caused overheating, which induced excessive fluidity and associated issues. Pronounced oozing and stringing negatively impacted surface finish, dimensional accuracy, and the complexity of intricate designs. Additionally, elevated temperatures delayed the cooling and solidification processes, increasing the risk of layer deformation, particularly in overhangs or unsupported regions.

**Optimize print speed:** Optimizing print speed is vital for achieving high-quality fabrication of SSA. In this study, a protocol was developed, beginning with an initial speed of 40 mm/s and systematically reducing it in 5 mm/s increments to address observed print deficiencies. Through this iterative approach, an optimal speed of 15 mm/s was identified, striking a balance between precision and efficiency. Deviations from this benchmark speed resulted in several adverse effects.

At speeds below 10 mm/s, prolonged thermal exposure within the hot extruder led to polymer chain degradation, compromising essential mechanical properties such as tensile strength and elasticity. This degradation undermined the actuator's structural integrity and functional longevity. Additionally, slower extrusion rates caused material accumulation at the nozzle tip, resulting in surface irregularities, voids, and discontinuities that impaired the overall quality and performance of the actuators.

In contrast, speeds exceeding 15 mm/s introduced inertial effects that misaligned filament placement, particularly in complex geometries, thereby reducing dimensional accuracy and airtightness. Rapid extrusion also shortened the interaction time between successive layers, preventing adequate thermal bonding. Insufficient interlayer fusion increased the risk of delamination under mechanical or pneumatic stress and introduced micro-gaps and weak points. These defects significantly compromised pneumatic functionality, causing air leakage and reducing the actuator's operational reliability.

**Optimize flow rate:** The flow rate, often referred to as the extrusion multiplier in 3D printing, is a critical parameter that determines the volume of material extruded in relation to the movement of the printer's nozzle. Precise calibration of this parameter is essential for achieving optimal print quality, especially in the fabrication of SSA. This study initially set the extrusion multiplier to 100% flow rate, followed by iterative adjustments based on observed print quality issues. Incremental testing identified 110% as the optimal flow rate, effectively balancing material deposition and overall performance. Deviations from this optimal value introduced specific challenges, as detailed below.

At flow rates of 105% or lower, under-extrusion becomes a significant issue. Insufficient material deposition results in thin walls and weak points within the structure, critically undermining the structural integrity required for pneumatic actuators. This deficiency increases porosity in the SSA walls, potentially forming channels that allow air leakage, thereby compromising airtightness.

Conversely, setting the flow rate above 125% leads to over-extrusion, where excessive material deposition causes surface defects such as blobs and zits. These irregularities negatively impact both the actuator's aesthetic and functional characteristics by creating uneven surfaces that may impede the movement of adjacent components. Furthermore, over-extrusion exacerbates oozing, where material leaks from the nozzle during non-printing movements, and stringing, where thin plastic strands form between separate parts of the print. These defects degrade the surface quality, reduce airtightness, and impair the precision of the actuator.

**Optimal layer height:** Selecting the appropriate layer height is critical in the fabrication of SSA to balance print resolution, structural integrity, and production efficiency. This study identified 0.1 mm as the optimal layer height, offering a favorable compromise between detail accuracy, mechanical strength, and processing time. Deviations from this standard introduce specific challenges that affect both the quality and efficiency of production.

Using a layer height smaller than 0.1 mm significantly increases overall print time, rendering it impractical for scaled manufacturing. Printing flexible materials such as TPU at finer layer heights also elevates the risk of nozzle clogging. TPU's elasticity can cause it to adhere within the nozzle, leading to frequent print failures.

Conversely, selecting a layer height above 0.1 mm compromises the precision necessary for fabricating actuators with completely sealed structures. Thicker layers hinder effective interlayer adhesion, leading to weaker bonds between layers. This reduced bonding strength increases the likelihood of delamination under mechanical stress or during cyclic loading and pressure variations, significantly impacting the actuator's durability and reliability.

**Optimal Overlap Percentage:** The overlap parameter, which defines the degree to which each printed layer extends over adjacent layers, is a critical factor in ensuring the structural integrity and airtightness of SSA. Proper management of overlap directly impacts the performance and reliability of the actuator, balancing material efficiency with mechanical robustness. This study identifies an optimal overlap of 20–25%, ensuring sufficient layer fusion without compromising the actuator's flexibility or functionality.

When the overlap falls below 15%, bonding between layers is inadequate, resulting in structural weaknesses that become pronounced under operational mechanical stresses. Additionally, gaps caused by insufficient overlap can compromise the actuator’s airtightness, allowing air to escape and significantly reducing its efficiency and performance.

Exceeding 30% overlap introduces inefficiencies, including unnecessary material usage, which increases production costs and reduces the overall efficiency of the printing process. Furthermore, excessive overlap can lead to defects such as oozing or blobbing, where surplus material accumulates. These defects degrade surface smoothness, impair dimensional accuracy, and compromise the effectiveness of the actuator’s seals, hindering its performance.

**Nozzle size:** In this study, a nozzle diameter of 0.4 mm was determined to be optimal for printing with TPU in the fabrication of SSA. Attempts with smaller nozzles, such as 0.25 mm, revealed significant operational challenges, including frequent clogging. The increased resistance and pressure required to extrude TPU through such a narrow aperture disrupted material flow, resulting in under-extrusion and inadequate interlayer adhesion, both essential for ensuring the structural integrity of the actuators. Conversely, larger nozzles, such as 0.8 mm, facilitated improved material flow but compromised the precision necessary for achieving finer geometric details and surface smoothness, which are critical for the functional performance of pneumatic actuators.

**Layer width:** Layer width, also referred to as extrusion width, is a critical parameter in 3D printing that dictates the thickness of the filament extruded by the nozzle. For a 0.4 mm nozzle, the optimal layer width typically ranges between 0.32 mm (80% of the nozzle diameter) and 0.48 mm (120% of the nozzle diameter). In this study, a layer width of 0.32 mm was specifically selected for printing SSAs. This narrower width enhances the flexibility of the actuator walls and reduces their thickness, thereby improving the actuators' ability to deform under pressure, which is crucial for their efficient operation.

**Number of layers:** The selection of the number of layers in the fabrication of SSA plays a pivotal role in determining their airtightness. This study highlights that the airtightness of actuators depends more significantly on the number of layers than on wall thickness alone. For example, an actuator with a total wall thickness of 1.6 mm, composed of two layers of 0.8 mm each, exhibited inferior airtightness compared to an actuator with a total thickness of 0.96 mm distributed across three layers of 0.32 mm each. Further experimentation demonstrated that configurations with only one or two layers were insufficient to achieve the required airtightness. Conversely, configurations incorporating three to five layers consistently provided reliable airtight seals. However, the use of four or five layers compromised the actuator's flexibility and increased production time and material consumption without yielding proportional performance enhancements. Based on these findings, the study recommends fabricating SSA with three layers, as this configuration optimally balances airtightness, material efficiency, and flexibility, ensuring operational effectiveness.

**Optimal fan speed:** Fan speed is a critical parameter in the 3D printing process, particularly when fabricating SSA. This setting governs the cooling rate of the material post-extrusion, which directly affects the solidification of thermoplastic materials. The cooling rate influences several key factors, including interlayer adhesion, surface quality, and the mechanical properties of the final product. Empirical testing identified an optimal fan speed range of 30% to 40% of its maximum capacity, which effectively balances rapid cooling, enhancing dimensional accuracy and slower cooling, which prevents thermal shock that can lead to brittleness or warping.

Deviations from this optimal range result in significant print quality and performance issues. Fan speeds below 30% provide insufficient cooling, delaying the solidification of the extruded material. This slow cooling process can cause deformation and sagging, particularly in areas requiring high precision, such as overhangs or intricate details. Furthermore, the prolonged malleability of the material increases the risk of warping or shifts during printing, compromising the structural and functional integrity of the pneumatic actuator. Such defects reduce the mechanical performance of the final product, limiting its effectiveness.

Conversely, fan speeds exceeding 40% cause excessively rapid cooling and solidification, potentially preventing adequate thermal bonding between successive layers. Proper bonding is crucial for maintaining mechanical strength and airtightness in pneumatic actuators. Excessive cooling may lead to layer delamination, where the rapid transition from molten to solid does not allow sufficient time for the layers to fuse effectively. This weakens structural cohesion and compromises the airtightness necessary for optimal actuator functionality. Additionally, high fan speeds may physically displace layers that are not yet fully solidified, particularly in lightweight or finely detailed features, distorting the intended geometry and reducing print accuracy.

### 4.4 Conclusion

This section provided a comprehensive framework to address the challenges of fabricating the SSA using FDM with TPU material. Through systematic optimization of key printing parameters, such as extrusion temperature, print speed, flow rate, and fan speed, as well as strategies for mitigating issues like bed adhesion, support material compatibility, and nozzle selection, the study demonstrates significant advancements in achieving airtight and mechanically robust actuators. The introduction of methodologies like the 30-degree rule and tailored approaches for minimizing humidity, improving extrusion consistency, and enhancing layer adhesion underscores the potential of Bowden extruder systems for flexible material applications. Table 4.2 provides the final specification for this optimized 3D-printed process. These findings establish a foundation for overcoming technical limitations and enabling scalable production of high-performance SSAs.

**Table 4.2:** Optimized printing settings

| Parameter | Value | Unit | Parameter | Value | Unit |
|---|---|---|---|---|---|
| **Quality** | | | | | |
| Nozzle diameter | 0.4 | mm | Build plate adhesion type | Brim | - |
| Layer height | 0.1 | mm | Brim width | 10 | mm |
| Initial layer height | 0.27 | mm | Brim only outside | Enable | - |
| Line width | 0.32 | mm | Brim speed | 20 | mm/s |
| Wall line count | 3 | - | Z hope speed | 20 | mm/s |
| **Infill** | | - | Optimize printing order | disable | **-** |
| Infill density | 100% | - | Alternat extra wall | disable | - |
| Infill Pattern | Zigzag | - | Infill pattern | line | - |
| **Material** | | | Acceleration control | Enable | - |
| Printing Temperature | 235 | C | Infill Acceleration | 1500 | $mm/s^2$ |
| Initial layer printing temperature | 257 | C | Wall ordering | Insite to outside | - |
| Build pate temp | 40 | C | Wall Acceleration | 750 | $mm/s^2$ |
| Flow | 110 % | - | Jerk control | Enable | - |
| Infill Flow | 125 | - | Retraction | Disable | - |
| **Speed** | | | Support material | Disable | - |
| Print speed | 15 | mm/s | Avoid printed part when travel | Enable | - |
| Initial layer print speed | 11 | mm/s | Jerk control | Enable | - |
| Top/ Bottom speed | 12 | mm/s | Retraction | Disable | - |
| Fan speed | 30% | - | | | |
| Travel speed | 200 | mm/s | | | |
| Initial Travel speed | 130 | mm/s | | | |
| Build plate adhesion type | Brim | - | | | |
| Brim width | 9 | mm | | | |

# CHAPTER 5

# EXPERIMENTAL PLATFORM

This chapter introduces an experimental platform that is custom developed in the laboratory by the candidate to evaluate the performance of the newly developed SSAs. The platform, as shown in Figure 5.1 is designed to measure both the force and displacement characteristics of SSAs, enabling a comprehensive analysis of their performance. Its adaptable design allows for compatibility with various SSA models, making it a versatile tool for experimental research. Additionally, the platform provides precise control over actuator motion and internal pressures, ensuring accurate and reliable scientific investigations.

The main frame of the platform is constructed using 2020 V-slot aluminum extrusion, with aluminum profile corner brackets, 3-way corners, and L-shaped connectors used to link the aluminum links together. This configuration allows the platform to be easily modified and adjusted, with additional components added as needed based on the experiment's requirements.

The experimental platform is composed of several key subsystems, including the airflow control system, force measurement module, position measurement module, actuator mounting module, and data collection module. Each of these units plays a vital role in the platform's operation and will be discussed in greater detail in subsequent sections.

The experiment utilizes both positive and negative (vacuum) pressure modes. An air compressor is used for positive pressure. Two methods were employed to generate negative pressure: a vacuum pump and a vacuum ejector. The vacuum ejector demonstrated superior performance in terms of flow, making it a more efficient choice for experiments requiring rapid and consistent vacuum generation.

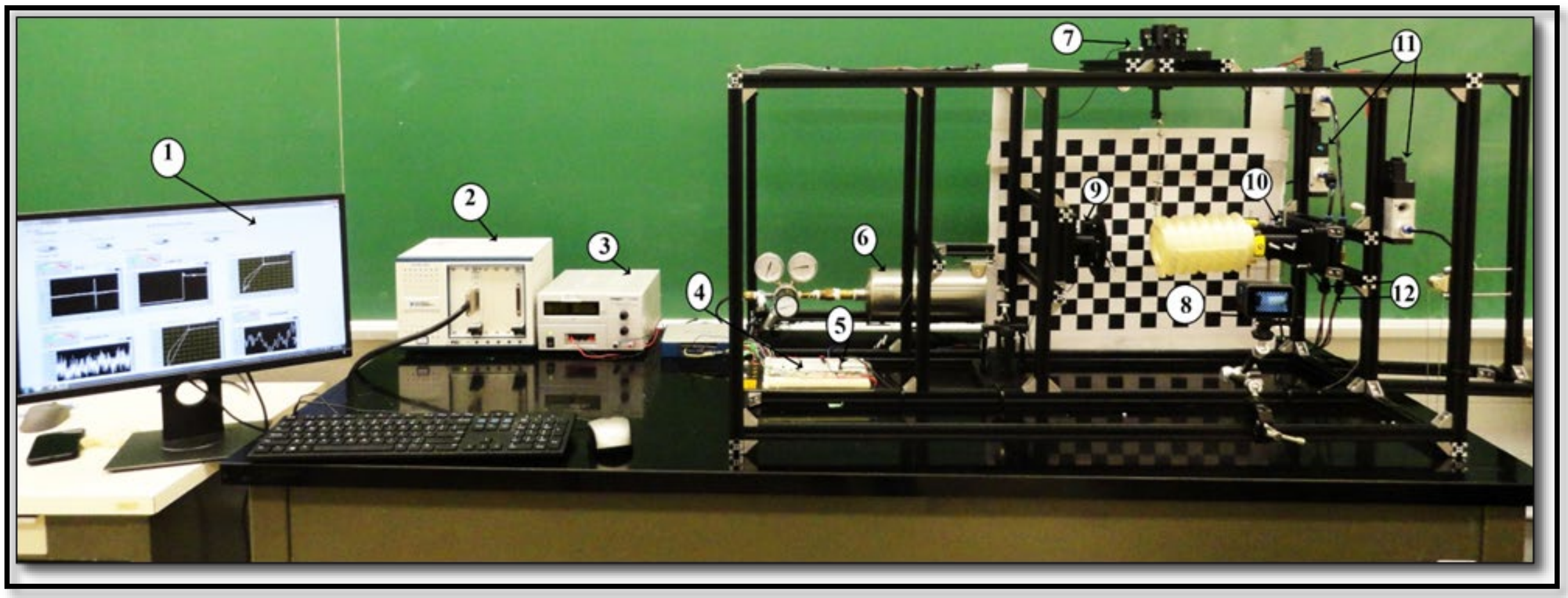

**Figure 5.1:** Primary components of the experimental platform: (1) graphical user interface (2) data acquisition system (3) power supply (4) electric circuit (5) vacuum sensor (6) air buffer tank (7) force measurement unit for bending actuator (8) 4k camera (9) force measurement unit for linear actuator (10) Fixture (11) solenoid valves (12) pressure transducers.

### 5.1 Airflow Control System

The airflow control system is essential in modulating the pneumatic conditions necessary for accurately testing and evaluating soft actuators. This system provides fine control over air pressure and flow, which are critical parameters in the actuation and control of SSA. The main components of this system are:

**Air buffer tank:** A tank (Figure 5.1 (6)) is used to maintain a stable and consistent air supply pressure to the actuators. Its primary function is to store compressed air, mitigating pressure fluctuations arising from the experiment's demands. This ensures that the supply pressure fed to the valves remain steady, which is essential for achieving reliable and repeatable test results.

**Valves:** Air valves play a vital role in regulating airflow and pressure for the SSA. The platform uses two types of valves: solenoid valves and proportional valves. Three solenoid valves Mettleair-4V410 (Mettleair, USA) were used which are electrically operated with rapid response times, making them ideal for fast switching. These valves can handle pressures up to 800 MPa with a flow rate of 2 SCFM, suitable for quickly pressurizing or depressurizing the SSA. However, their binary operation (fully open or closed) limits their precision in applications requiring fine-tuned airflow control. To address this limitation, the platform incorporates proportional valves, which allow continuous modulation of airflow. The QBX Industrial Electronic Pressure Regulator serves as the primary proportional valve, offering precise control with a flow rate of 1.2 SCFM and a maximum operating pressure of 1.2 MPa. It responds to electronic signals, enabling high-precision adjustments necessary for experiments demanding fine control over the actuator's movements.

**Pressure sensors:** Accurate pressure measurement is critical for evaluating SSA, enabling the analysis of force and displacement at varying pressure levels. Two types of sensors are used in this study. Positive pressure sensors from Texas Instruments 61CP0220100 measures pressures from 0 to 690 kPa with an accuracy of ±0.75%. It operates with a supply voltage of 4.5–5.5 VDC and outputs a voltage range of 0.5–4.5 VDC, providing reliable data for moderate to high-pressure pneumatic systems. Vacuum Sensors from Panasonic ADP5110 measure vacuum pressures from 0 to -100 kPa with a precision of ±1.25%, ensuring accurate monitoring of negative pressure for actuators relying on vacuum operation. These sensors ensure precise pressure data across different operating ranges.

### 5.2 Position Measurement Module

The position measurement module is designed to measure the displacement generated by the SSAs, including both linear and angular displacement. This module is essential for evaluating displacement at specific pressure levels, allowing for precise characterization of the actuator's performance. The displacement data obtained from this module can also be utilized as feedback to control the position of the actuator in various applications. The position measurement module consists of two key components.

The first component of the module is dedicated to measuring linear displacement where a Linear Variable Differential Transformer (LVDT) is employed. The selected LVDT has a working range of 0 to 150 mm, with infinite resolution of ± 0.5% linearity. This configuration is suitable for most testing within the experimental platform but due to its mechanical nature, the LVDT is not ideal for applications of high-speed. To address the need for high-speed and highly accurate linear displacement measurements, an industrial displacement laser sensor, specifically the OPTEX-H2CD (OPTEX, Japan), is employed. This laser sensor offers superior accuracy with a measurement precision of ±0.02 mm and a rapid sampling period of 1 millisecond. The OPTEX-H2CD sensor is particularly advantageous in dynamic testing scenarios where rapid changes in actuator position must be captured and analyzed in real-time.

The second component of the position measurement module focuses on measuring angular displacement, which includes both twisting and bending motions of the soft actuators. For this purpose, a high-resolution camera, the GoPro Hero11, is utilized. This camera is capable of capturing video at a 5.3K resolution with a frame rate of 240 frames per second (FPS), making it practical for capturing detailed and high-speed motion. Kinovea software was used to analyze the recorded video data. Kinovea is widely recognized for its application in measuring angular motion in soft robotics and provides a robust tool for accurately analyzing the angular displacement captured by the GoPro camera [61], [91].

### 5.3 Force Measurement Unit

The force measurement unit (FMU) is another component of the experimental platform designed to measure the forces exerted by SSAs during testing. This unit provides essential data for evaluating actuator performance under various pressure conditions, ensuring that force measurements align precisely with the specific type of tested motion. The force measurement unit comprises a force sensor integrated into a custom assembly. This assembly is designed to securely hold the force sensor, ensuring precise alignment with the actuator to capture the forces generated during testing. The assembly is composed of several key components:

**Mounting structure:** The mounting structure serves as the base for attaching the force measurement unit to the main experimental frame. It is designed to securely anchor the force sensor while incorporating features that allow for easy adjustment in three directions. This capability ensures that the force sensor can be precisely aligned with the direction of the force generated by the actuator, which is critical for accurate measurement.

**Sensor housing:** The sensor housing is designed to securely enclose the force sensor within the mounting structure. This housing ensures the sensor remains stable and correctly positioned throughout the experiment, minimizing potential errors caused by misalignment or movement. The stability provided by the sensor housing is essential for obtaining consistent and reliable data.

**Coupling structure:** The coupling structure is responsible for linking the actuator to the force sensor, ensuring that the forces exerted by the actuator are accurately transferred to the sensor. This structure is composed of several interconnected parts that work together to ensure that the force is transmitted directly from the actuator to the sensor without any loss or distortion.

The entire assembly is manufactured using a 3D printer with PLA material, chosen for its durability. This material ensures that the assembly can withstand the forces generated during testing while maintaining the necessary alignment and stability. Three specialized assemblies have been developed to address the different types of forces generated by various actuator motions, as illustrated in Figure 5.2. Each assembly is optimized for precise alignment with the direction of force application.

The force sensors used in this study include the FC22310100, FC22310050, and FC22310025 models (TE Connectivity, Switzerland), selected based on the specific force ranges required for different tests. These sensors have an accuracy of 1% of the full-scale span, providing reliable data across the various force levels encountered during experiments. The sensors operate with an excitation voltage of 5V.

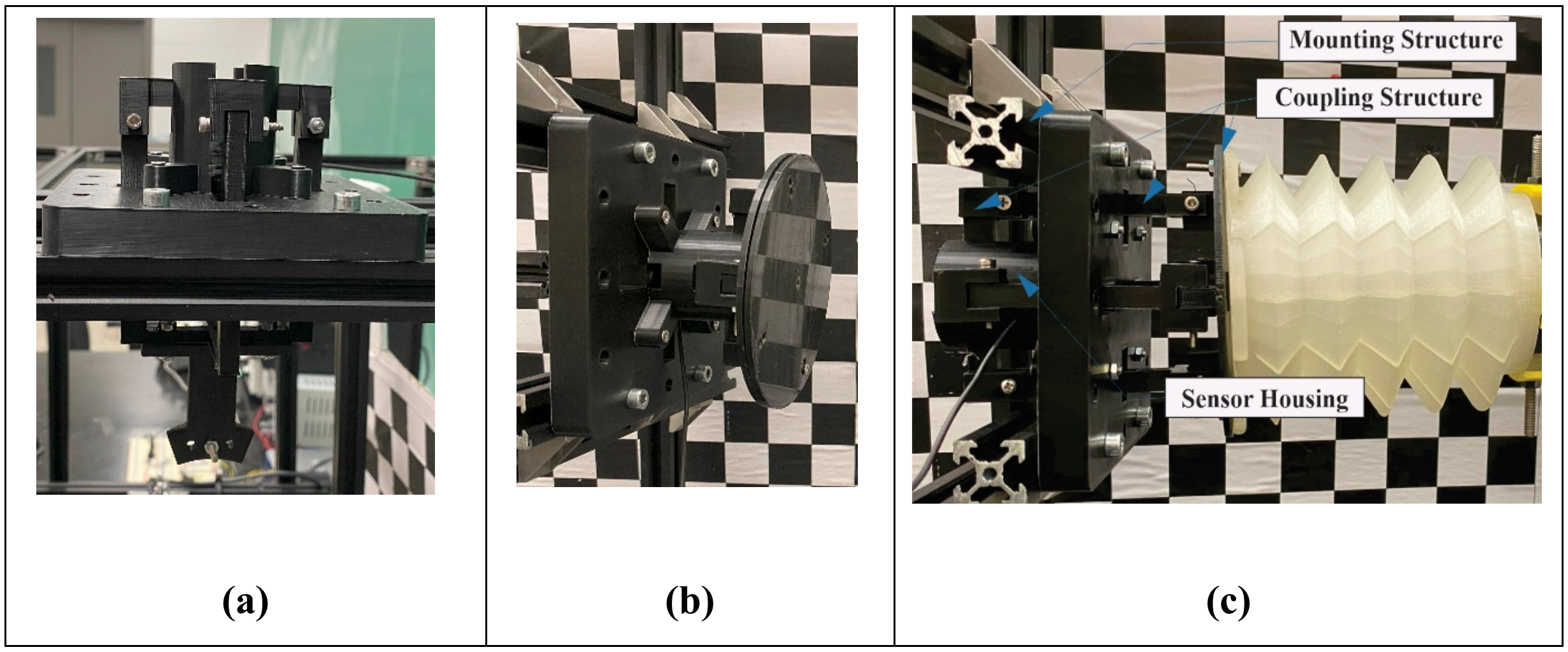

**Figure 5.2:** Force measurement module for different motions: (a) bending (b) extension (c) contraction

### 5.4 Actuator Mounting Module

The actuator mounting module is designed to hold the actuator during testing securely. This module plays a critical role in ensuring that the actuator remains fixed at one end while allowing the other end to move freely. Preventing unnecessary motion at the fixed end is vital, as such motion could lead to inaccurate measurements. Several designs were developed during this study, but only one successfully met the objectives, as shown in Figure 5.3. This design allows the actuator mounting module to move in two directions, providing flexibility while maintaining stability. The Actuator Mounting Module consists of two main parts: the mounting structure and the actuator fixture.

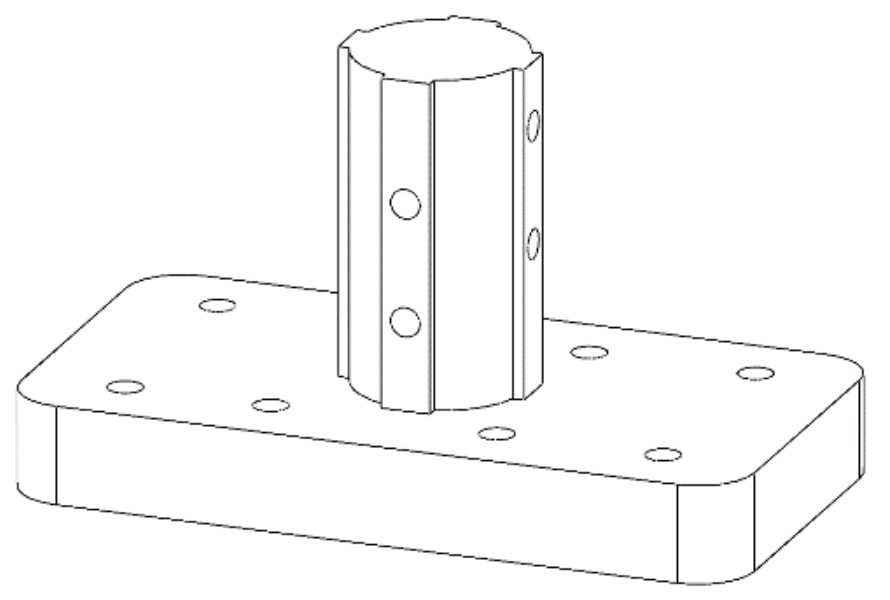
(a)

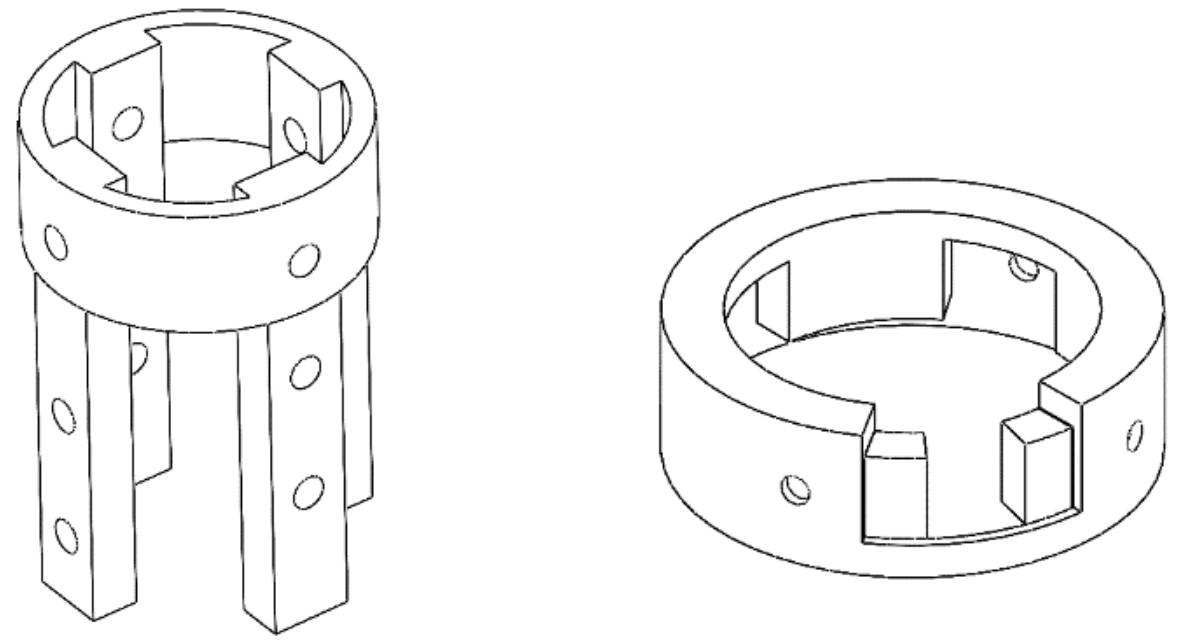
(b)

**Figure 5.3:** (a) Mounting bracket (b) actuator fixture

**Mounting structure:** The mounting structure is responsible for securing the Actuator Mounting Module to the experimental platform and allows for the precise adjustment of the actuator's position. For this, a bracket is used to attach the actuator fixture to the experimental platform which ensures that the actuator is held securely and can be adjusted as needed to align with the experimental setup. Furthermore, an aluminum assembly is used to attach the Mounting Bracket to the experimental platform. This assembly provides the necessary support and rigidity, ensuring that the actuator remains fixed in place during testing.

**Actuator fixture:** The Actuator fixture shown in Figure 5.3 (b) is designed to secure and attach the actuator to the mounting structure. The fixture ensures that the actuator is held firmly in position, preventing any movement that could interfere with the accuracy of the displacement and force measurements. Most parts of the Actuator Mounting Module were manufactured using a 3D printer with Tough PLA material. This choice of material was made for its durability, ensuring that the module can withstand the forces exerted by the actuator while maintaining the necessary alignment and stability.

### 5.5 Data Acquisition and Control Module

This module is designed to convert the signals generated by various sensors into meaningful data that can be analyzed to assess the performance of soft actuators. Additionally, this module generates control signals to operate the valves, facilitating precise control over the experimental conditions. The Data Collection Module is composed of three essential parts:

**Electric circuit:** The electric circuit, as shown in Figure 5.4, powers the sensors, valves, and other components within the experimental platform. It ensures that each element receives the voltage and current for optimal operation. This circuit also serves as the interface between the sensors and the data acquisition system, transmitting the raw signals generated by the sensors for further processing.

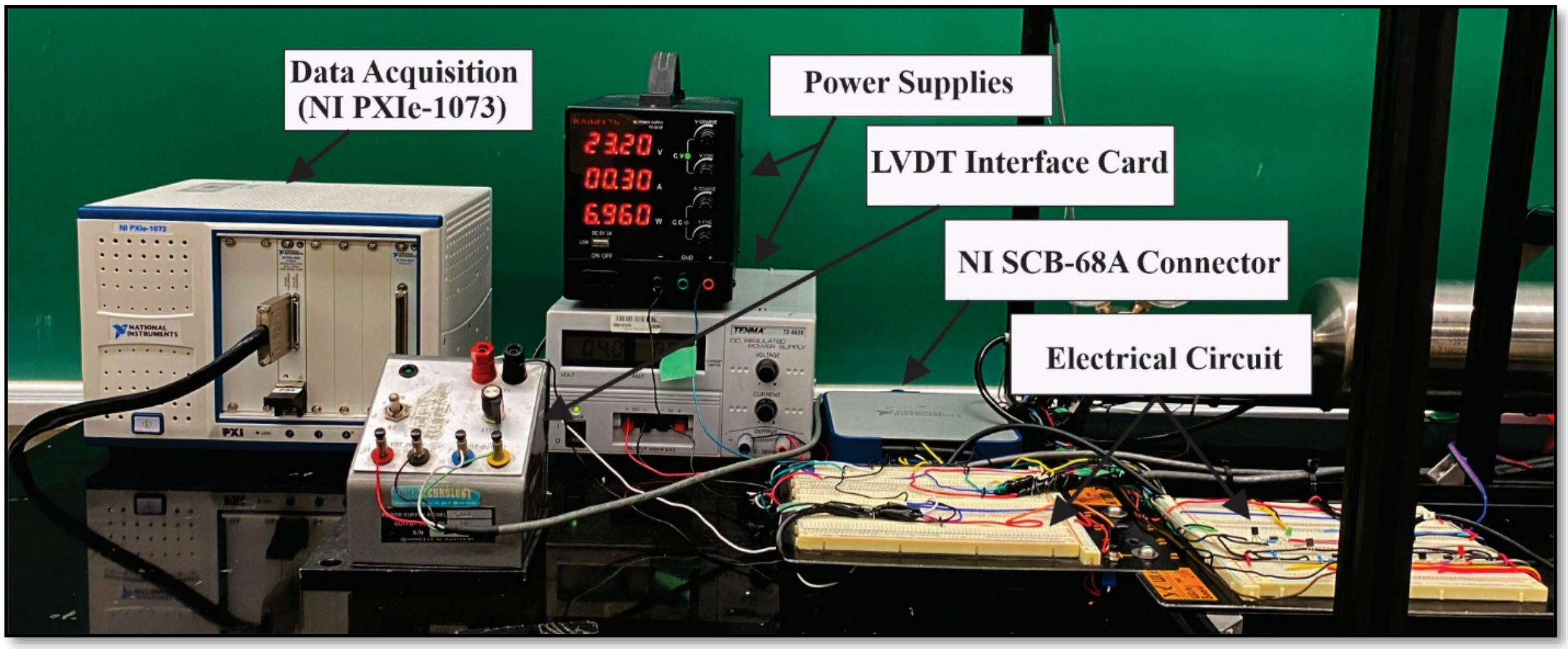

**Figure 5.4:** Key Electronic Components of the Experimental Setup

**National Instruments Data Acquisition System:** The data acquisition system, provided by National Instruments (NI PXle-1073), plays a central role in the Data Collection Module. It converts the analog signals from the sensors into meaningful data, which can then be processed and analyzed. This system translates the raw sensor outputs into equivalent measurement metrics, such as force, displacement, or pressure, providing a quantifiable basis for evaluating the actuator's performance.

**Graphical interface:** The graphical interface, illustrated in Figure 5.5, is developed using LabVIEW and serves as the user interface for the experimental platform. It enables communication with the main experimental system and translates the data collected by the sensors into visual formats, such as graphs and charts. This visualization allows to monitor the actuator's performance in real-time and make informed decisions during the experimentation process.

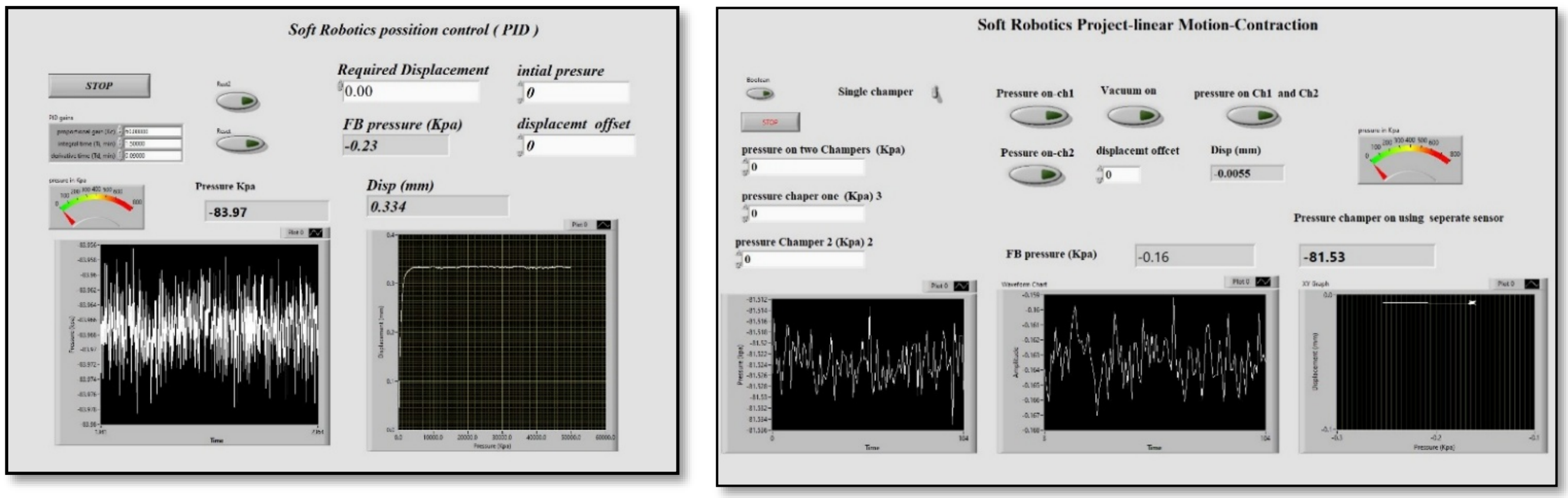

**Figure 5.5:** Graphical interface developed using LabVIEW for two experiments

## 5.6 Conclusion

In conclusion, this chapter has presented a concise overview of the specialized experimental platform devised to assess the performance of the SSAs. This platform, which integrates high-precision measurement systems and comprehensive data acquisition tools, accommodates various actuation modes and supports in-depth analyses of both kinematic and kinetic parameters. By implementing robust testing protocols and verifying component reliability, it ensures the reproducibility, accuracy, and robustness required for subsequent experimental studies. Ultimately, this sophisticated setup underpins the validation of the SSAs' mechanical efficacy and provides a foundational basis for the investigations detailed in the following chapters.

# CHAPTER 6

# RESULTS

This chapter presents the results of a comprehensive experimental investigation into the design and operational factors that influence the performance of newly developed SSA models, namely the LSSA, BSSA, TSSA and OSSA. By examining how variations in materials, geometrical configurations, and operating conditions affect actuator behavior and performance, these results will provide valuable insights for understanding and optimizing SSA models.

## 6.1 LSSA

The performance of the LSSA depends on both material stiffness and key geometric parameters, as described in Section 3.2.4 and illustrated in Figures 3.4 and 3.5. These parameters, including radius (*r*), actuator length (*l*), fold width (*fw*), fold angle (*β*), number of restraining layers (*nr*), thickness of restraining layers (*tr*), wall thickness (*wt*), and Shore hardness (*sh*), constitute the principal design variables governing the actuator's force output and motion. Their specific values were selected to meet both performance requirements and manufacturing constraints. For example, the minimum feasible fold angle 30º, while the lowest printable Shore hardness is 85 Ha. Ensuring airtight integrity requires a wall thickness of at least 0.96 mm, and a nozzle size of 0.4 mm, dictates the minimum restraining layer thickness 0.4 mm. Furthermore, a minimum of ten restraining layers can be used to withstand internal pressures of up to 200 kPa.

To assess the impact of these parameters on LSSA performance, a series of models was developed and evaluated using the experimental platform described previously. Table 6.1 provides an overview of these models, highlighting the selected geometric and material parameters. The actuator's ability to generate force and linear motion was examined under two primary modes of operation: extension, driven by positive internal pressure, and contraction, achieved through vacuum. This comprehensive evaluation illuminates how variations in geometry and material properties collectively influence the LSSA's overall performance in real-world conditions.

**Table 6.1:** LSSA models (all dimensions in mm; all angles in degrees)

| Model | *r* | *l* | *fw* | *β* | *tr* | *wt* | *nr* | *sh* |
|---|---|---|---|---|---|---|---|---|
| ***L1*** | 30 | 80 | 16 | 30 | 0.4 | 1.2 | 18 | 85 |
| ***L2*** | 30 | 80 | 16 | 30 | 0.4 | 0.96 | 18 | 85 |
| ***L3*** | 30 | 80 | 16 | 30 | 0.4 | 1.6 | 18 | 85 |
| ***L4*** | 30 | 80 | 16 | 30 | 0.4 | 0.96 | 16 | 85 |
| ***L5*** | 30 | 80 | 12 | 30 | 0.4 | 0.96 | 18 | 85 |
| ***L6*** | 30 | 80 | 8 | 30 | 0.4 | 0.96 | 18 | 85 |
| ***L7*** | 30 | 80 | 16 | 30 | 0.8 | 0.96 | 18 | 85 |
| ***L8*** | 30 | 80 | 4 | 30 | 0.4 | 0.96 | 18 | 85 |
| ***L9*** | 30 | 80 | 12 | 35 | 0.4 | 0.96 | 18 | 85 |
| ***L10*** | 30 | 80 | 12 | 40 | 0.4 | 0.96 | 18 | 85 |
| ***L11*** | 30 | 80 | 12 | 30 | 0.4 | 0.96 | 18 | 95 |
| ***L12*** | 30 | 80 | 12 | 30 | 0.4 | 0.96 | 14 | 85 |
| ***L13*** | 30 | 80 | 16 | 30 | 0.8 | 0.96 | 12 | 85 |

### 6.1.1 LSSA Force Performance during Extension

The evaluation of the forces generated by the LSSA was conducted using the experimental setup depicted in Figure 6.1 (a). In this configuration, one end of the LSSA was securely fixed to the fixture on the experimental platform, while the opposite end of the LSSA was free to move. The FMU was positioned to contact the free-moving end of the actuator, enabling force measurement. The LSSA was then subjected to controlled pressurization, with pressures incrementally increased to a maximum of 200 kPa.

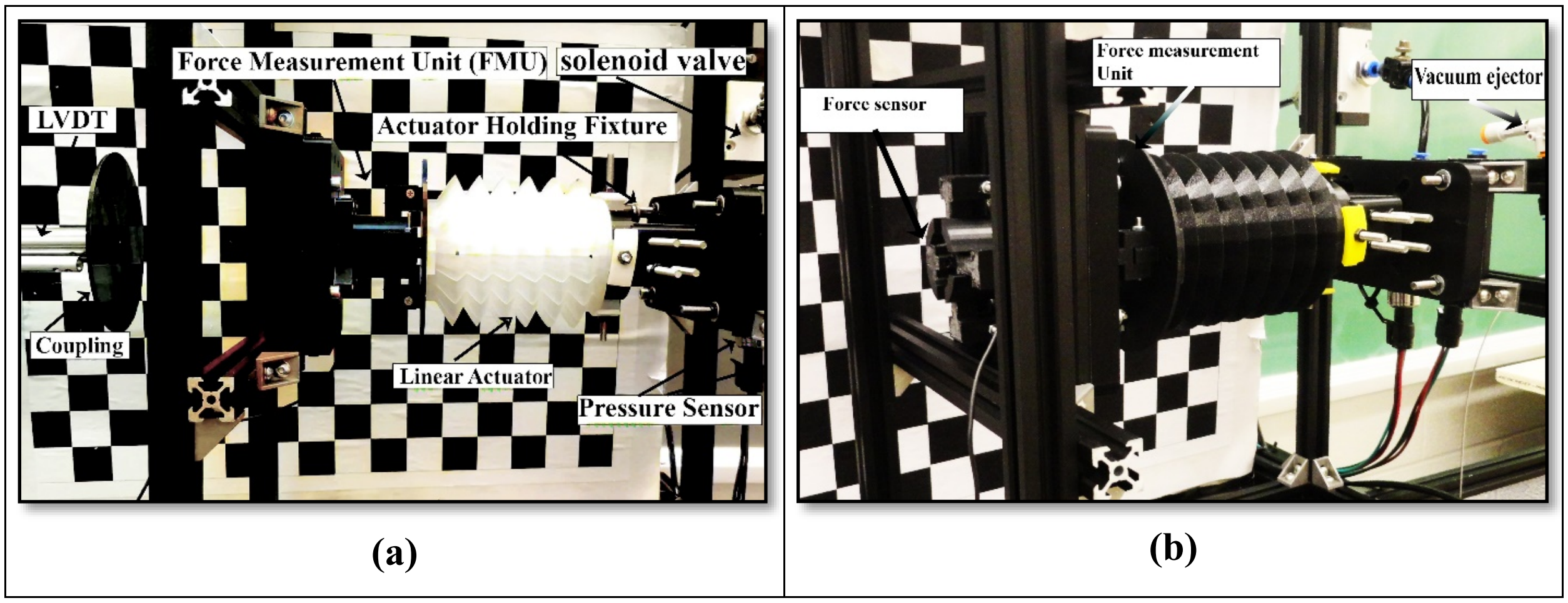

**Figure 6.1:** Experimental setup for measuring forces in LSSA (a) extension (b) contraction

The results for the impact of geometric parameters and material stiffness on the force-generating capability of the LSSA models are summarized in Figure 6.2. The following subsections provide an in-depth analysis of the impact of each parameter.

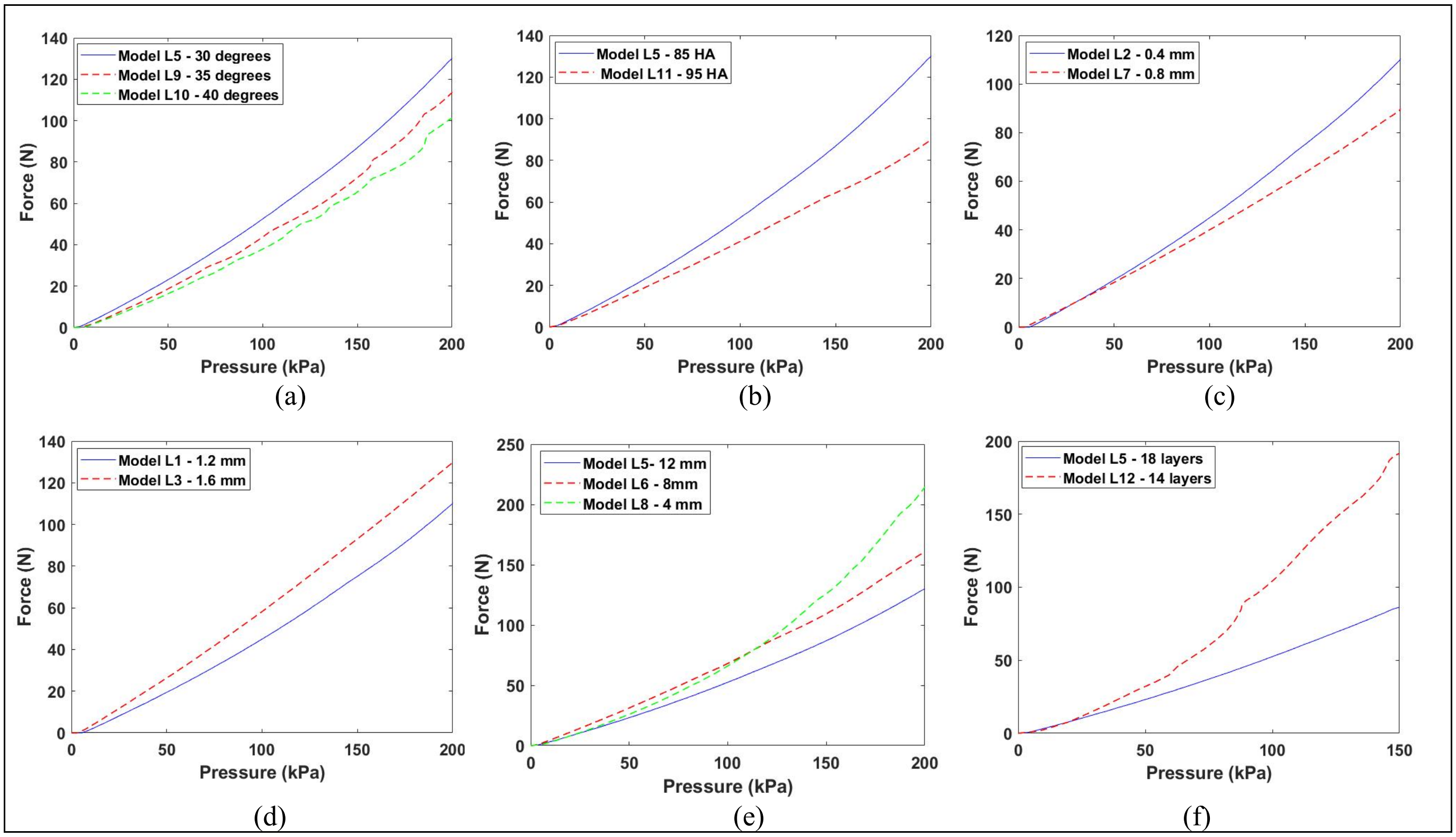

**Figure 6.2**: Impact of geometrical parameters and material stiffness on generated force of LSSA in extension motion (a) effect of fold angle (b) effect of material stiffness (c) effect of the tie restraining layers thickness (d) effect of wall thickness (e) effect of fold width (f) effect of the number of tie-restraining layers.

**Effect of fold angle**: As illustrated in Figure 6.2 (a), increasing the fold angle from 30° to 40° reduced peak force, decreasing from 130 N to 101 N. This inverse relationship suggests larger fold angles reduce the LSSA's ability to generate force during extension.

**Effect of material stiffness:** The LSSA models were fabricated using two grades of TPU: a more flexible grade (TPU 85A, Ninjaflex) and a less flexible grade (TPU 95A, Ultimaker). As shown in Figure 6.2 (b), increasing the material stiffness from 85A to 95A led to a marked decrease in peak force, from 130 N to 89.9 N, indicating that higher stiffness materials negatively impact force generation during extension.

**Effect of tie-restraining layer thickness:** An examination of the tie-restraining layer thickness revealed that increasing the layer thickness from 0.4 mm to 0.8 mm decreased force generation, as shown in Figure 6.2 (c), with a reduction from 110 N to 95 N. This suggests that thicker restraining layers constrain the LSSA's ability to generate force.

**Effect of wall thickness:** As depicted in Figure 6.2 (d), an increase in wall thickness from 1.2 mm to 1.6 mm resulted in enhanced force output, with the force increasing from 110.5 N to 130.2 N during extension. This positive correlation suggests that thicker walls may contribute to greater force generation.

**Effect of fold width**: The data presented in Figure 6.2 (e) show that a decrease in fold width from 12 mm to 4 mm significantly enhanced the force generated during extension, with an increase from 130 N to 214.5 N. This finding indicates that narrower fold widths may improve the LSSA's efficiency in generating force.

**Effect of the number of tie-restraining layers:** Figure 6.2 (f) demonstrates that increasing the number of restraining layers from 14 to 18 significantly diminished the peak force output, with a reduction from 191 N to 86 N during extension. This indicates that a higher number of restraining layers inhibits the LSSA's performance by adding resistance.

### 6.1.2 LSSA Displacement Performance during Extension

The influence of geometric parameters and material stiffness on the displacement-generating capability of the LSSA was investigated by securing the actuator to the experimental platform and then subjected to controlled pressurization up to a maximum of 200 kPa. The extension of the LSSA was measured using an LVDT, as shown in Figure 6.3 (a). The models detailed in Table 6.1 were employed to examine how variations in geometric parameters and material stiffness influenced the actuator's performance. The results of these investigations are summarized in Figure 6.4.

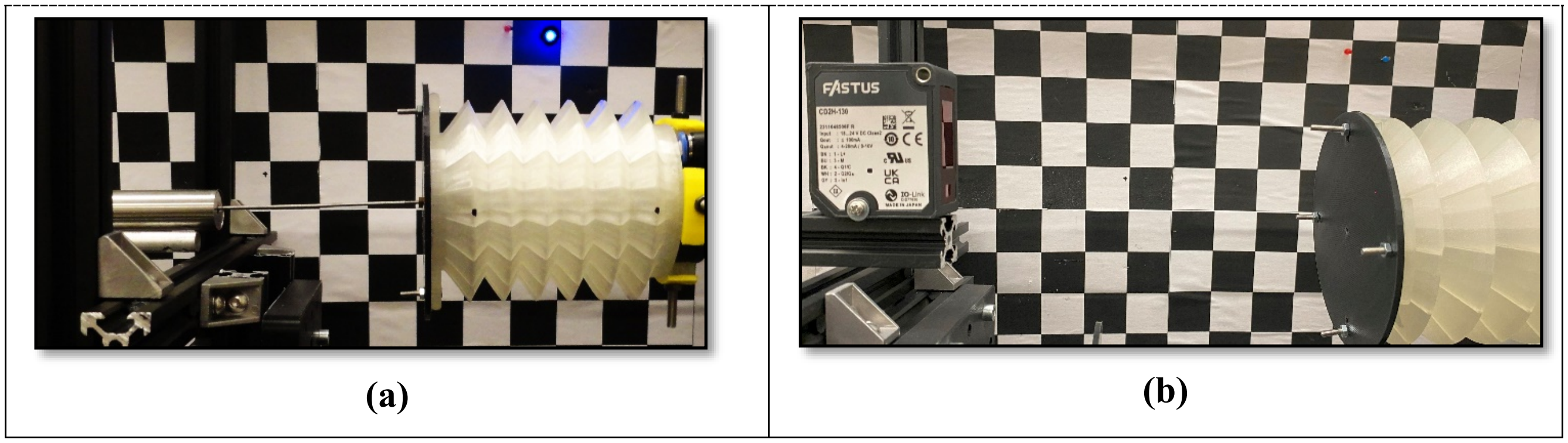

**Figure 6.3**: Measuring linear displacement (a) LVDT-based experimental setup (b) Laser-based experimental setup

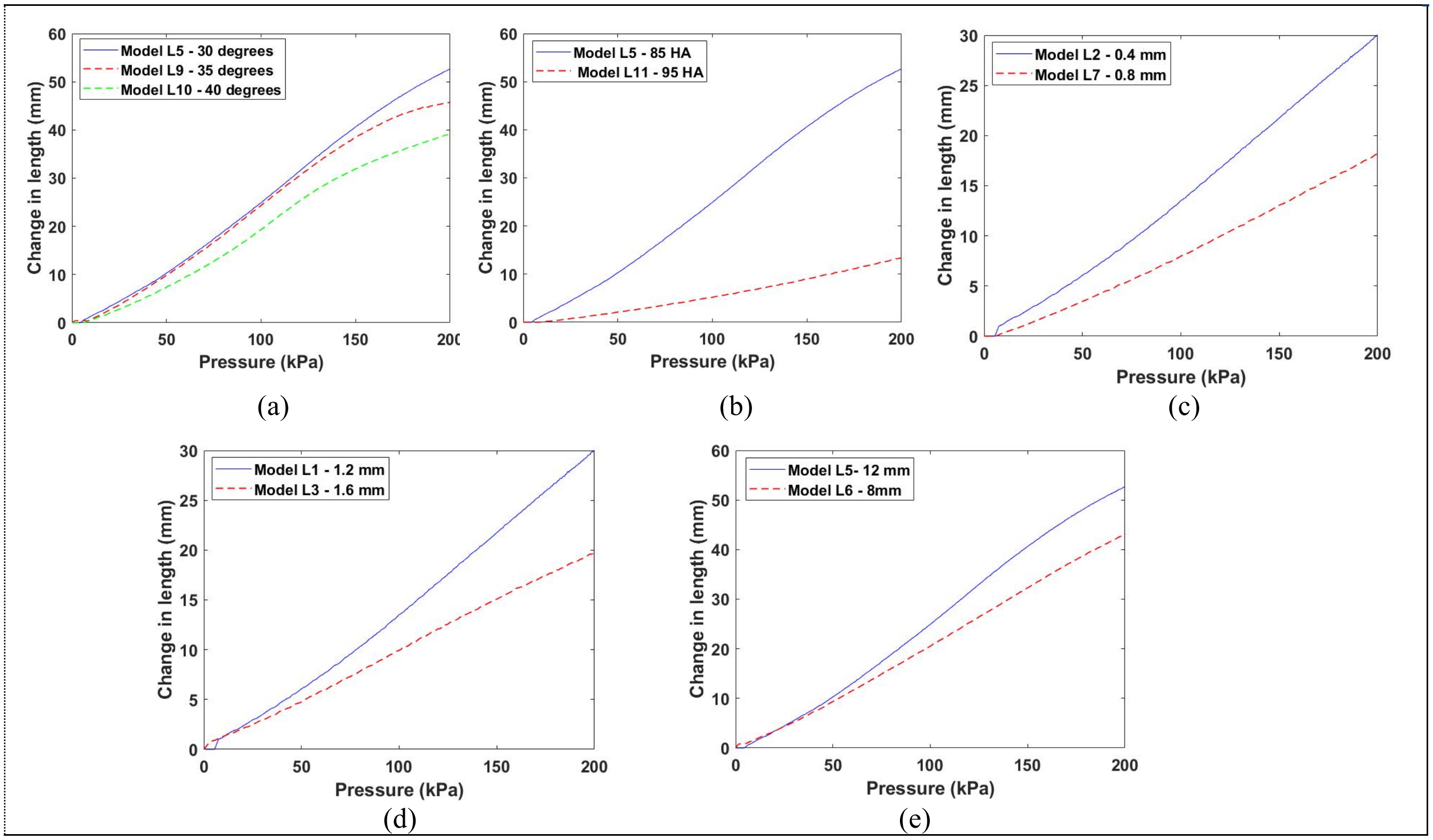

**Figure 6.4:** Impact of geometrical parameters and material stiffness on displacement of LSSA in extension motion (a) effect of fold angle (b) effect of material stiffness (c) effect of the tie restraining layers thickness (d) effect of wall thickness (e) effect of fold width.

**Effect of fold angle:** As illustrated in Figure 6.4 (a), increasing the fold angle from 30º to 40º negatively impacts the extension length. This increase is associated with a reduction in displacement during extension, from 52.7 mm to 39.3 mm. This observation suggests that larger fold angles constrain the LSSA's ability to extend.

**Effect of material stiffness:** Figure 6.4 (b) presents the experimental results for two actuator prototypes fabricated with different material stiffness values, specifically 85A and 95A. The findings indicate that increased material stiffness leads to a significant decrease in displacement during extension, with a reduction from 52.7 mm to 13.4 mm. These results highlight the inverse relationship between material stiffness and LSSA extension, suggesting that softer materials facilitate greater extension.

**Effect of tie-restraining layer thickness:** An increase in the thickness of the restraining layers from 0.4 mm to 0.8 mm was observed to have a notable effect on the LSSA's performance. As depicted in Figure 6.4 (c), this increase in layer thickness led to a decrease in extension displacement from 30 mm to 18 mm. This outcome indicates that thicker restraining layers may inhibit the LSSA's extension capability.

**Effect of wall thickness:** The influence of wall thickness on LSSA displacement is evident, with an increase from 1.2 mm to 1.6 mm resulting in a significant reduction in extension displacement. As shown in Figure 6.4 (d), this increase in wall thickness corresponds to a decrease in extension displacement from 30 mm to 20 mm. This finding suggests that a thicker wall structure restricts the LSSA's ability to extend.

**Effect of fold width:** As depicted in Figure 6.4 (e), an increase in fold width from 8 mm to 12 mm resulted in a decrease in the extension rate. Specifically, the extension displacement diminished from 53 mm to 43 mm as the fold width increased. This trend underscores the inverse relationship between fold width and extension capability, indicating that narrower folds enhance the LSSA's extension performance.

### 6.1.3 LSSA Force Performance during Contraction

The evaluation of the forces generated by the LSSA during contraction was conducted using the experimental setup shown in Figure 6.1 (b). In this configuration, the LSSA was securely affixed to the experimental platform and subjected to controlled vacuum pressure, with levels incrementally increasing to a maximum of -95 kPa. The force generated at the terminal end of the actuator was recorded. The impact of geometric parameters and material stiffness on the force-generating capability of the LSSA during contraction was investigated using the models presented in Table 6.1. The results of these tests are summarized in Figure 6.5.

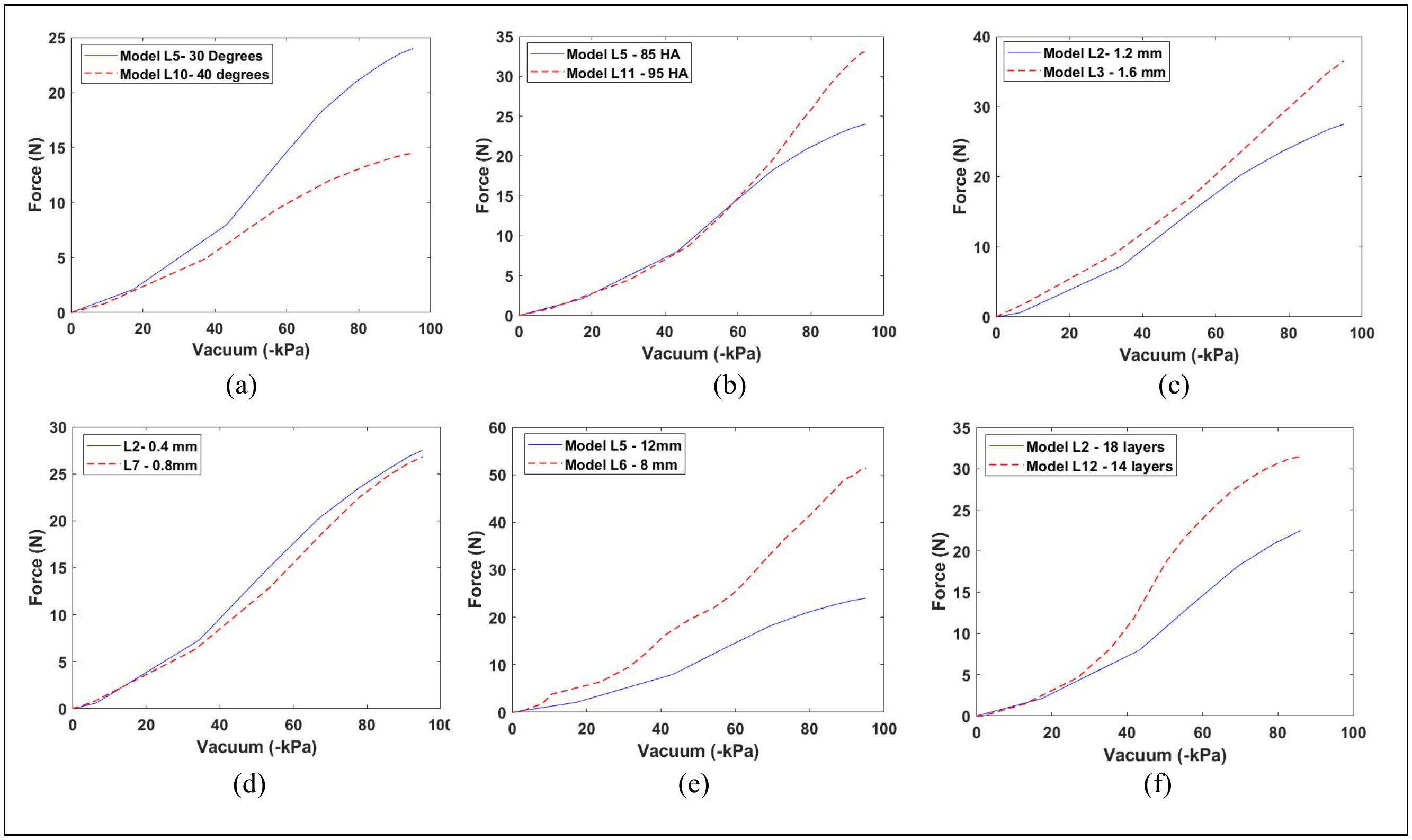

**Figure 6.5**: Impact of geometrical parameters and material stiffness on generated force of LSSA in contraction motion (a) effect of fold angle (b) effect of material stiffness (c) effect of wall thickness (d) effect of the tie restraining layers thickness (e) effect of fold width (f) effect of the number of tie-restraining layers.

**Effect of fold angle:** As demonstrated in Figure 6.5 (a), increasing the fold angle from 30° to 40° leads to a decrease in force generation from 24 N to 14.5 N. This indicates that larger fold angles negatively impact the LSSA's ability to generate force during contraction.

**Effect of material stiffness:** Figure 6.5 (b) illustrates the effect of material stiffness on force generation during contraction. An increase in stiffness from 85A to 95A increased in force output from 24 N to 33 N, suggesting stiffer materials enhance the LSSA's force-generating capacity under vacuum conditions.

**Effect of tie-restraining layer thickness:** The impact of tie-restraining layer thickness on force output was found to be minimal. As shown in Figure 6.5 (d), increasing the layer thickness from 0.4 mm to 0.8 mm had a negligible effect on force generation during contraction, indicating that this parameter may not significantly influence the actuator's performance.

**Effect of the number of tie-restraining layers:** The number of restraining layers was found to inversely affect force generation, as depicted in Figure 6.5 (f). Increasing the number of restraining layers from 14 to 18 led to a reduction in force, with a decrease from 35 N to 20 N. This suggests that additional restraining layers impede the LSSA's ability to exert force during contraction.

**Effect of wall thickness:** As presented in Figure 6.5 (c), an increase in wall thickness was associated with a significant enhancement in force output. A marginal increment in wall thickness of 0.4 mm resulted in an increase in force from 27.5 N to 36.5 N, indicating that thicker walls contribute to greater force generation during contraction.

**Effect of fold width:** The influence of fold width on force output was pronounced. As shown in Figure 6.5 (e), a reduction in fold width from 12 mm to 8 mm led to a substantial increase in force generation, with the force doubling from 24 N to 51.5 N. This demonstrates that narrower fold widths significantly enhance the LSSA's force output during contraction.

### 6.1.4 LSSA Displacement Performance during Contraction

The influence of geometric parameters and material stiffness on the contraction displacement of the LSSA was investigated using the same experimental setup and procedure described in Section 6.1.2 for extension motion. However, in this case, the LSSA was subjected to controlled vacuum pressures, incrementally increasing up to a maximum of -95 kPa. The LSSA models used for this analysis are detailed in Table 6.1, and the results are summarized in Figure 6.6.

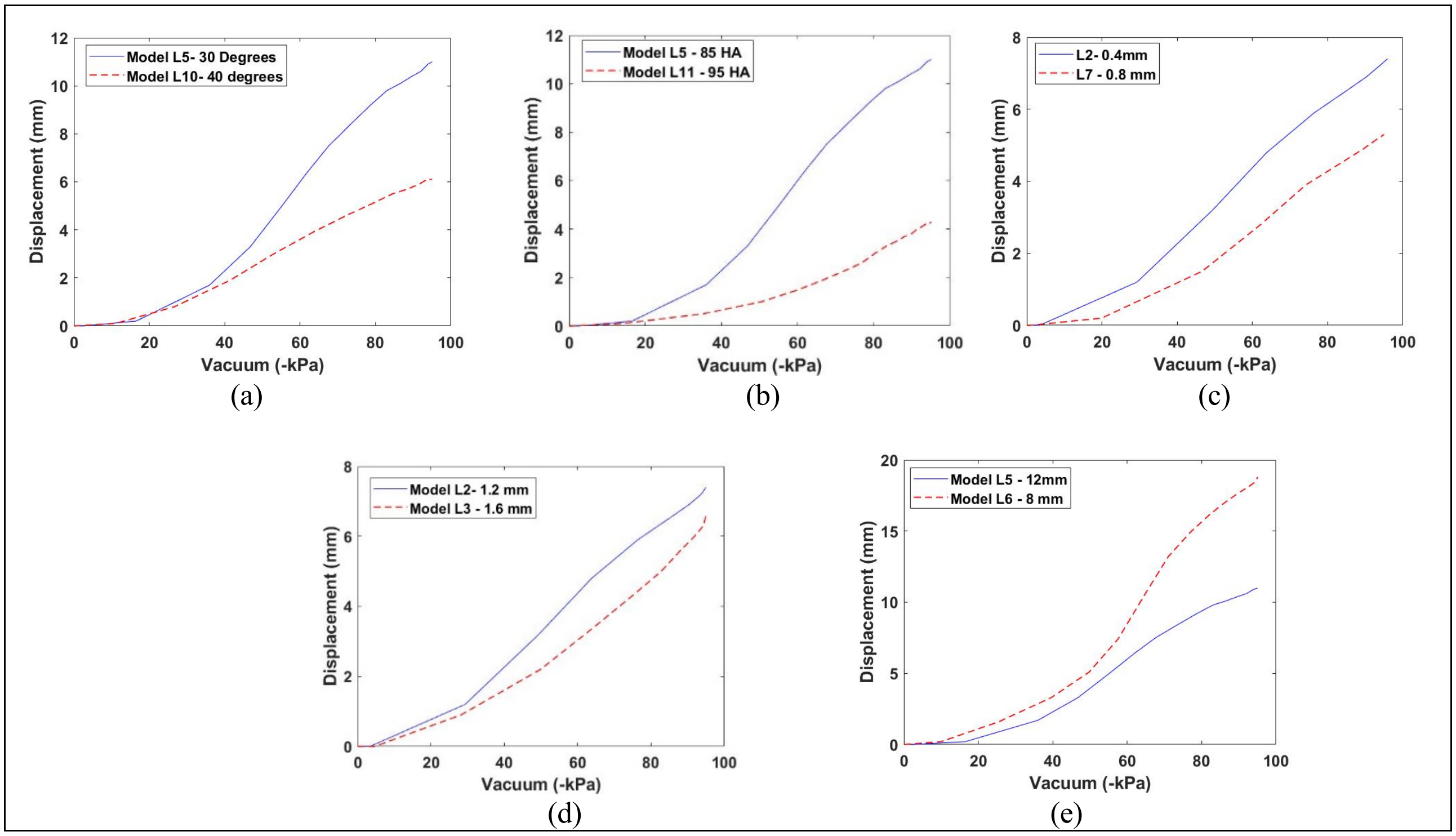

**Figure 6.6:** Impact of geometrical parameters and material stiffness on the displacement of linear SSA in contraction motion (a) effect of fold angle (b) effect of material stiffness (c) effect of the thickness of the tie-restraining layer (d) effect of wall thickness (e) effect of fold width.

**Effect of fold angle:** As demonstrated in Figure 6.6 (a), increasing the fold angle from 30º to 40º results in a reduction in contraction displacement, decreasing from 11 mm to 6 mm. This suggests larger fold angles may hinder the LSSA's ability to contract effectively.

**Effect of material stiffness:** The impact of material stiffness on contraction displacement is illustrated in Figure 6.6 (b). An increase in material stiffness from 85A to 95A leads to a significant reduction in displacement, with the contraction distance decreasing from 11 mm to 4.3 mm. This finding indicates that stiffer materials limit the LSSA’s capacity to contract.

**Effect of tie-restraining layer thickness:** As shown in Figure 6.6 (c), an increase in the thickness of restraining layers from 0.4 mm to 0.8 mm reduces contraction displacement from 7 mm to 5 mm. This suggests that thicker restraining layers restrict the LSSA’s ability to achieve full contraction.

**Effect of wall thickness:** The effect of wall thickness on contraction displacement is depicted in Figure 6.6 (d). An increase in wall thickness from 1.2 mm to 1.6 mm leads to a decrease in contraction displacement, reducing from 7 mm to 6 mm. This suggests that thicker walls may impede the LSSA's contraction capability.

**Effect of fold width:** As illustrated in Figure 6.6 (e), a decrease in fold width by 4 mm enhances the contraction displacement. Specifically, reducing the fold width increases the contraction distance from 11 mm to 19 mm. This finding highlights the positive influence of narrower fold widths on the LSSA's contraction performance.

### 6.1.5 LSSA Extension Performance versus Load

To evaluate the LSSA extension performance when subjected to an additional load, two experiments were conducted. The first experiment aimed to assess how an external load influences the LSSA's force generation using the experimental setup illustrated in Figure 6.7 (a). In this setup, a stainless-steel cable was employed to transfer designated loads to the proximal end of the LSSA. This load transfer was facilitated by a coupling unit, which connected one end to the LSSA and the other to the force measurement unit. A key objective in designing this experimental setup was to ensure that the applied load was centrally aligned with the LSSA. This was achieved through a redesign of the fixture, which centralized the load and allowed the stainless-steel cable to move freely and without friction, minimizing potential interference during the evaluation phase.

To analyze the LSSA's performance, LSSA Model L6 was subjected to three different masses: 1 kg, 2 kg, and 3.5 kg. The results, presented in Figure 6.7 (b), indicate an inverse relationship between the applied load and the force generated by the LSSA; as the applied load increased, there was a corresponding decrease in the actuator's force output.

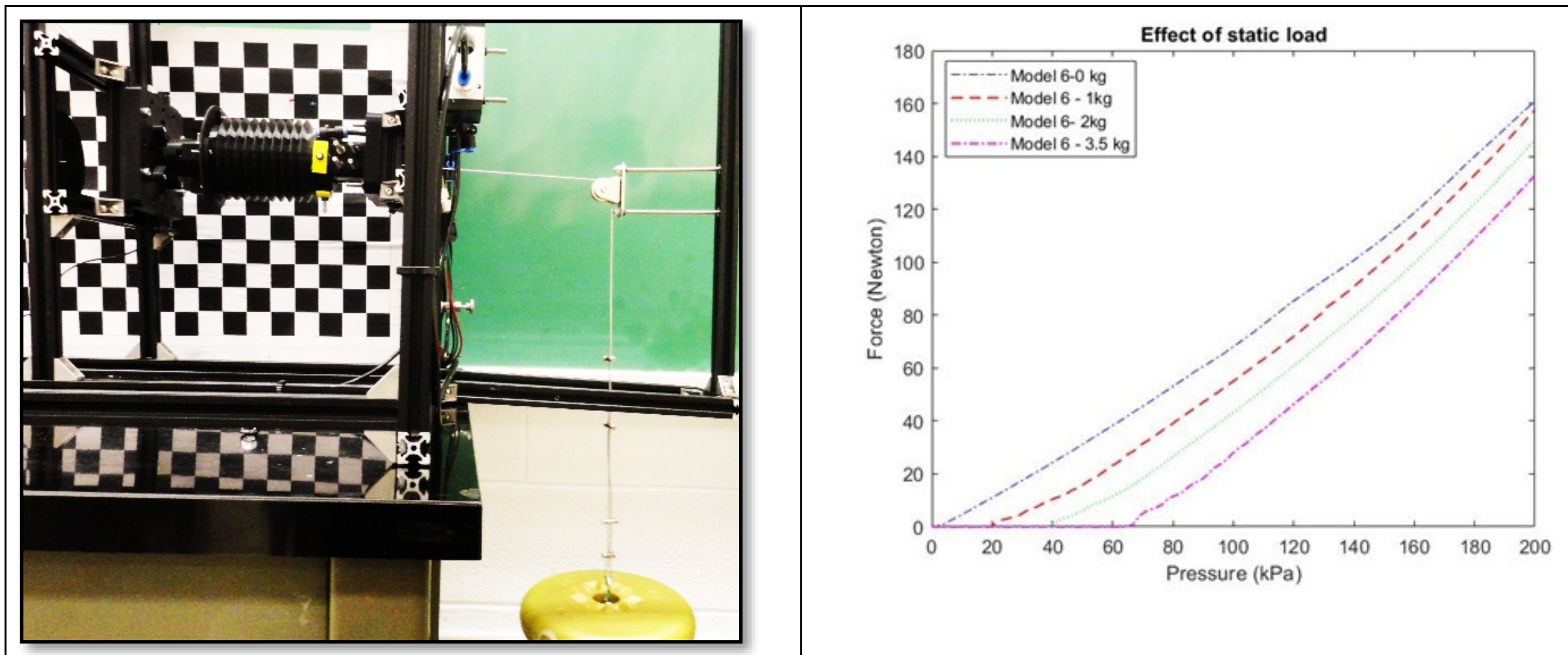

(a) (b)

**Figure 6.7:** Effect of static load on generated force during extension motion (a) experimental setup (b) results

The second experiment focused on evaluating the effect of static load on the LSSA's displacement. As depicted in Figure 6.8 (a), the experimental setup again utilized a stainless-steel cable to connect the load to a designed coupling unit. This coupling unit served a dual purpose: it transferred the load from the cable to the actuator's end and established a connection between the LSSA and the LVDT, as shown in Figure 6.8 (a). A pulley system was incorporated into the setup to ensure that the force applied was centralized on the actuator. Additionally, a support table was used to ensure that no load was applied to the actuator when inactive, thereby preventing unintended contraction.

To investigate the impact of static mass on the LSSA's movement, the LSSA Model L5 was tested with two different static masses: 3 kg and 6 kg. During these tests, the LSSA was exposed to pressures up to 200 kPa, and its movement was recorded and analyzed. The results, as shown in Figure 6.8(b), indicate that the addition of a large mass (6 kg) reduced the actuator's movement by 7 mm. This reduction is considered to have a marginal influence on the overall operational efficacy of the LSSA.

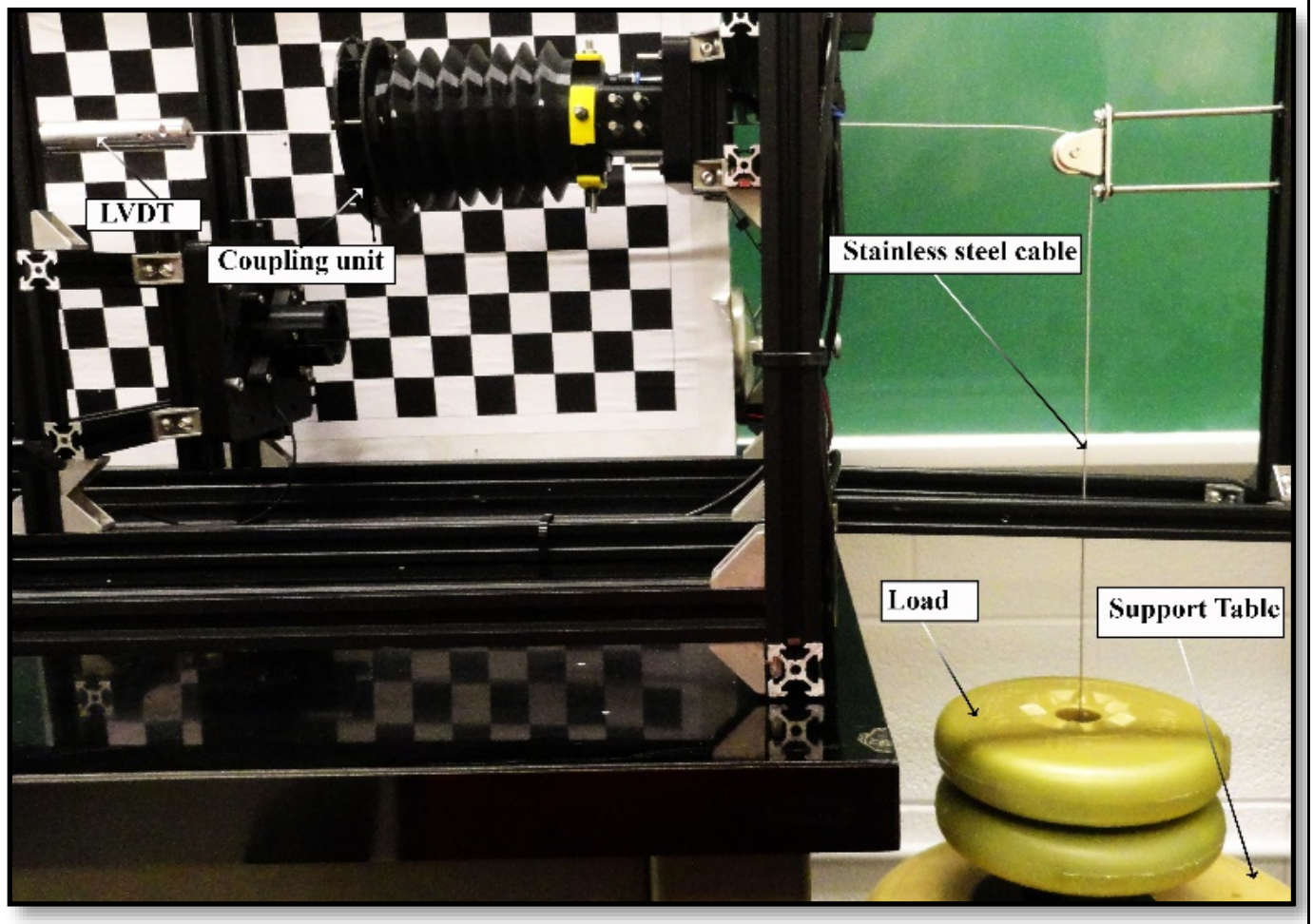

**(a)**

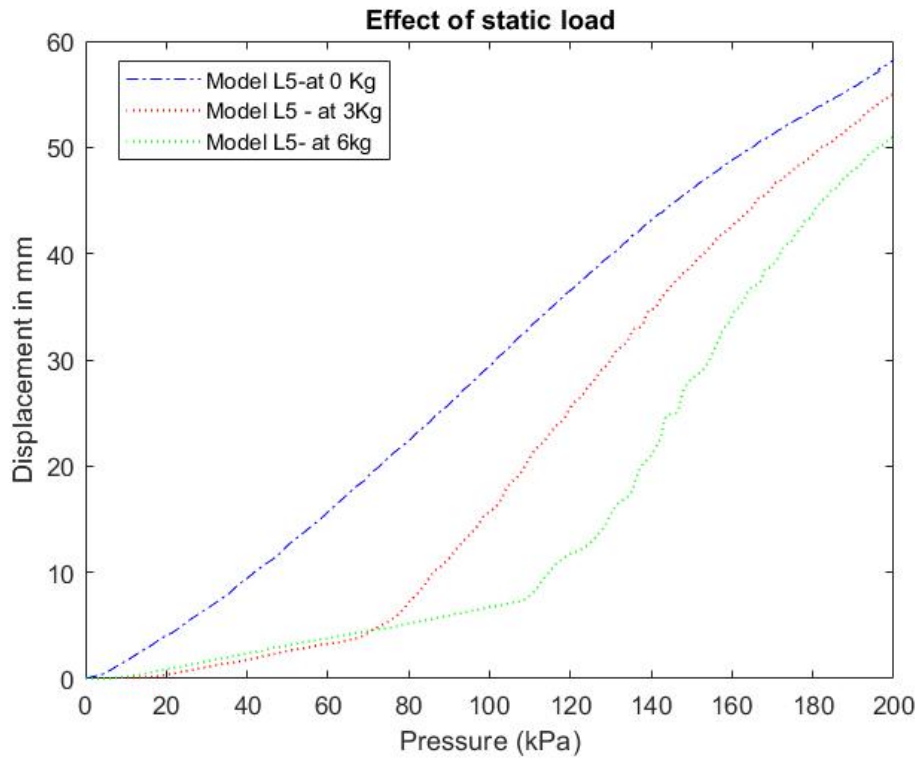

**(b)**

**Figure 6.8:** Effect of static load on generated displacement during extension motion (a) experimental setup (b) test results

### 6.1.6 LSSA Contraction Performance versus Load

The study investigated the impact of static load on the force generation of the LSSA during its contraction motion. The experimental setup, depicted in Figure 6.9 (a), involved applying specific loads to the LSSA's proximal end using a stainless-steel cable, while a coupling unit ensured precise load application and accurate force measurement.

The response of the LSSA to varying static masses was assessed through experimental testing conducted on Model L6, utilizing two distinct masses of 1 kg and 2 kg. The results, illustrated in Figure 6.9 (b), demonstrated an inverse relationship between the applied load and the force generated by the actuator. Specifically, as the static load increased, the LSSA's force output decreased proportionally, highlighting the impact of increased load on the LSSA's contraction performance.

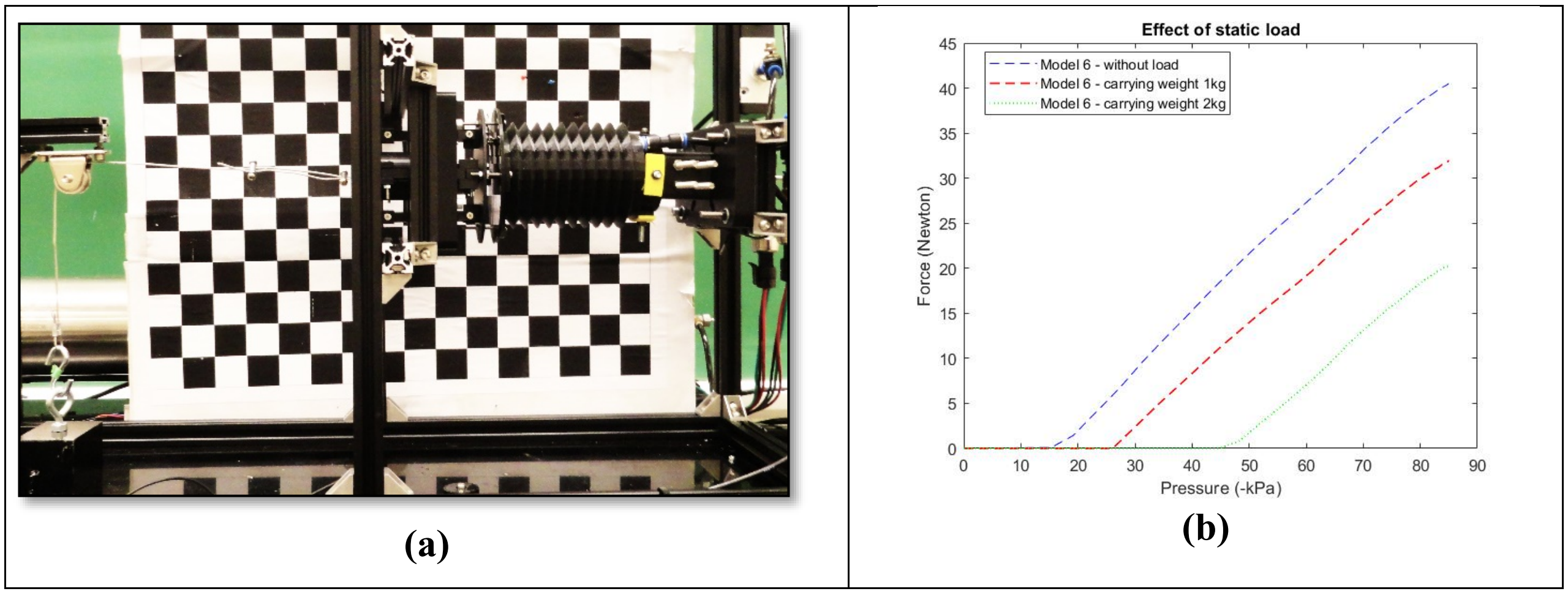

**Figure 6.9:** Effect of static load on generated force during contraction motion (a) experimental setup (b) test results

A subsequent experiment was conducted to evaluate the effect of static load on contraction displacement. The experimental setup was modified, as shown in Figure 6.10 (a), with the integration of a LVDT positioned at the rear of a specialized fixture. This fixture was designed to allow unobstructed movement of the LVDT, which was connected to the LSSA via a coupling unit. The actuator was loaded using a stainless-steel cable, and a pulley system was incorporated to ensure that the applied force was aligned with the LSSA's line of action.

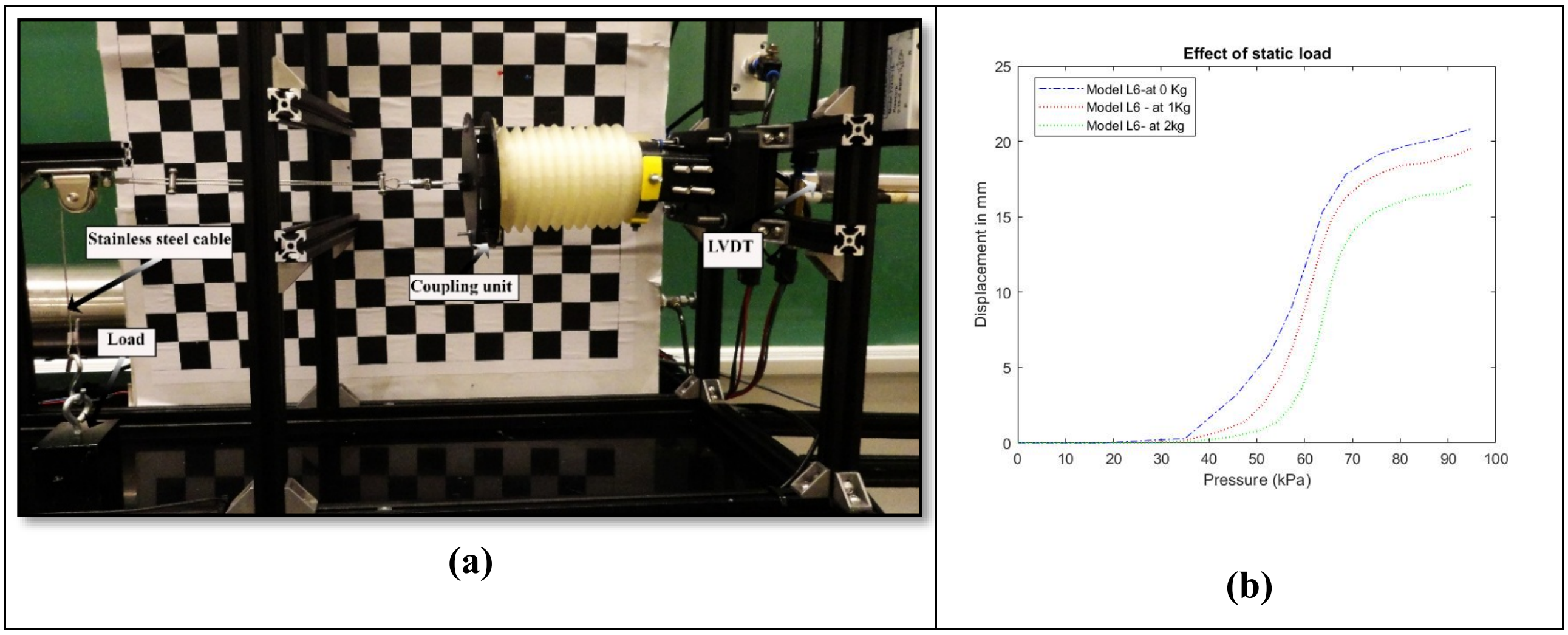

**Figure 6.10:** Effect of static load on generated displacement during contraction motion (a) experimental setup (b) test results

The data presented in Figure 6.10 (b) revealed that increasing the mass from 1 to 2 kg reduced the actuator's displacement from 20.8 mm to 17.2 mm.

### 6.1.7 LSSA Axial Stiffness Force

For LSSA, axial stiffness force is an important characteristic, representing the resistance along the actuator’s longitudinal axis when subjected to internal pressure. This force arises from the actuator's inherent material properties and structural configuration, providing a counteracting force that maintains shape stability during deformation. Understanding the factors influencing axial stiffness force is essential for optimizing the actuator’s performance.

A custom setup was developed to test the actuator at atmospheric pressure in the Instron testing machine to examine the influence of geometric parameters and material stiffness on axial stiffness force. During testing, LSSA models were extended while the tensile force was recorded at each displacement increment. The design parameters presented in Section 3.2.4 were independently varied to investigate their individual effects on axial stiffness. The experimental findings, shown in Figure 6.11, provide insights into how material stiffness and geometric factors contribute to axial stiffness force in SSAs.

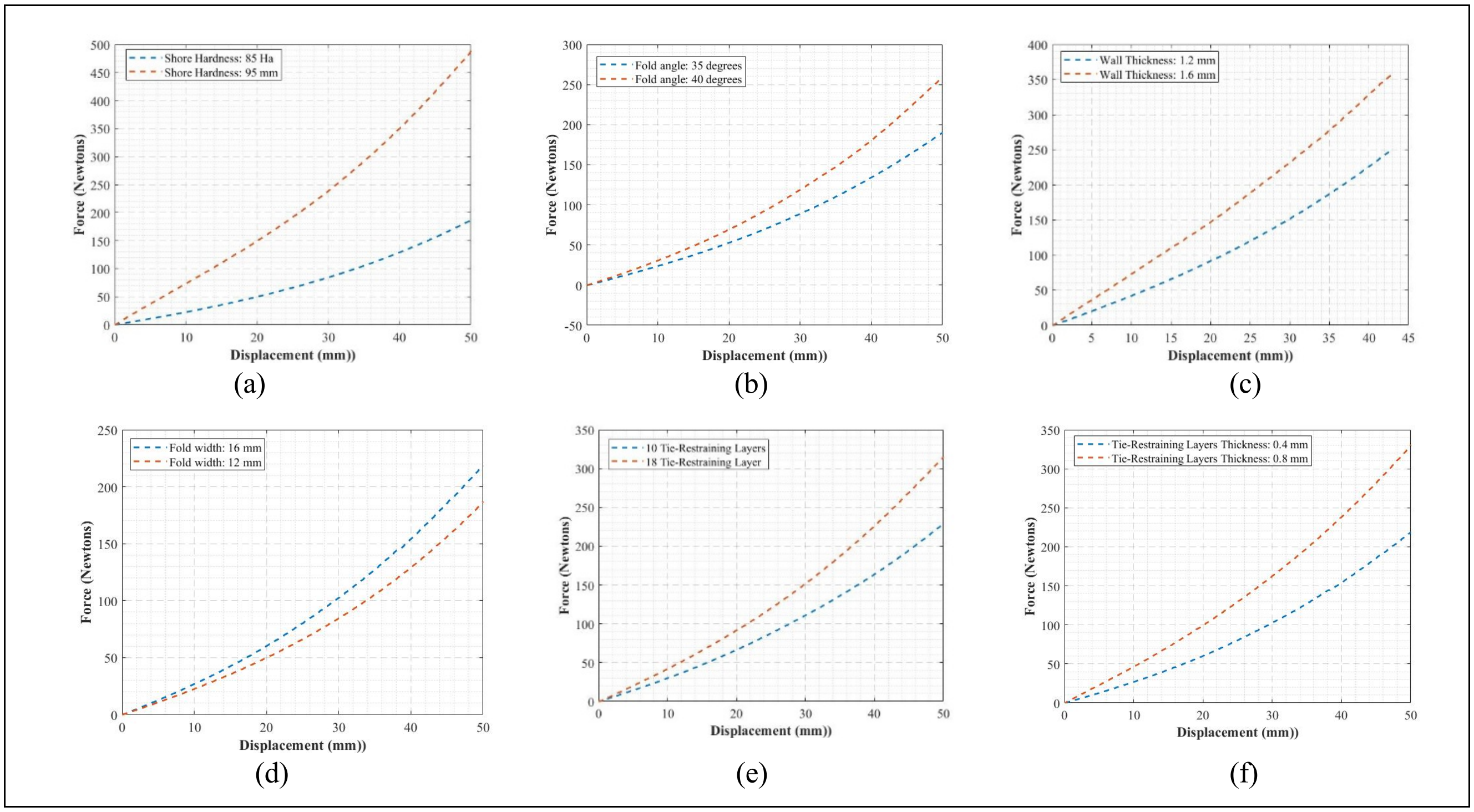

**Figure 6.11:** Impact of geometrical parameters on the axial stiffness (a) effect of fold angle (b) effect of constrained layers (c) effect of material stiffness (d) effect of wall thickness (e) effect of the tie restraining layers thickness (f) effect of the number of tie-restraining layers.

**Effect of material stiffness:** As illustrated in Figure 6.11 (a), increasing the Shore hardness of the material from 85A to 95A resulted in a rise in force from 187 N to 489 N. This substantial increase is attributable to the higher elastic modulus of stiffer materials, which resist deformation.

**Effect of fold angle:** Figure 6.11 (b) shows that increasing the fold angle from 35° to 40° led to an increase in axial stiffness force from 190 N to 260 N. A larger fold angle alters the actuator's geometry, increasing its structural rigidity along the axial direction.

**Effect of wall thickness:** As depicted in Figure 6.11 (c), increasing wall thickness from 1.2 mm to 1.6 mm elevated the axial stiffness force from 251 N to 359 N. Thicker walls add structural rigidity, reducing the actuator's flexibility and thereby necessitating a greater force for equivalent displacement.

**Effect of fold width:** As shown in Figure 6.11 (d), increasing the fold width from 12 mm to 16 mm resulted in a minor increase in axial stiffness force. Although fold width influences stiffness by altering geometry, its impact is less pronounced than that of other parameters.

**Effect of tie-restraining layer thickness:** Figure 6.11 (e) demonstrates that increasing tie-restraining layer thickness from 0.4 mm to 0.8 mm resulted in an axial stiffness force increase from 228 N to 315 N. Thicker tie-restraining layers enhance resistance to axial deformation by providing greater constraint to the actuator's movement.

**Effect of number of tie-restraining layers:** As shown in Figure 6.11 (f), increasing the number of restraining layers elevates the axial stiffness force, as each additional layer contributes to the actuator's resistance to axial deformation.

The experimental results reveal that the axial stiffness force of SSAs is strongly influenced by both material stiffness and specific geometric parameters. Material stiffness has the most substantial effect due to its direct relationship with the elastic modulus. Geometric parameters such as wall thickness, number of restraining layers, and layer thickness also play critical roles by affecting the cross-sectional area and structural configuration of the actuator. In contrast, parameters like fold angle and fold width have a comparatively lesser effect on axial stiffness force. The study further demonstrates that axial stiffness force is displacement-dependent, exhibiting minimal impact at small displacements but increasing sharply at higher displacements due to the nonlinear behavior of the LSSA models.

### 6.1.8 LSSA Stiffness versus Pressure

To further evaluate the stiffness of the actuator, an experimental setup, as depicted in Figure 6.12 (c), was developed. This setup integrated the LSSA with an Instron testing machine through a custom-designed actuator fixture (Figure 6.12 (c)). The primary objective of this setup was to assess the actuator's stiffness as a function of its displacement and internal pressure. A stable and controlled air pressure source for the actuator was provided by a compressor and buffer tank.

The experiments were conducted under five distinct constant pressure conditions: 25, 50, 75, 100, and 125 kPa. A pressure regulator (QBX pressure control valve, Proportion-Air, Inc.) was used to maintain these pressure levels within the actuator, ensuring consistency and accuracy during the stiffness measurements.

The test protocol involved securing Model L13 one end of the actuator to a fixed fixture while the opposite end was connected to a coupling unit attached to the Instron machine's head, as shown in Figure 6.12 (c). The Instron head displaced the actuator by 40 mm at a controlled speed of 50 mm/min, with data acquisition facilitated through LabVIEW software.

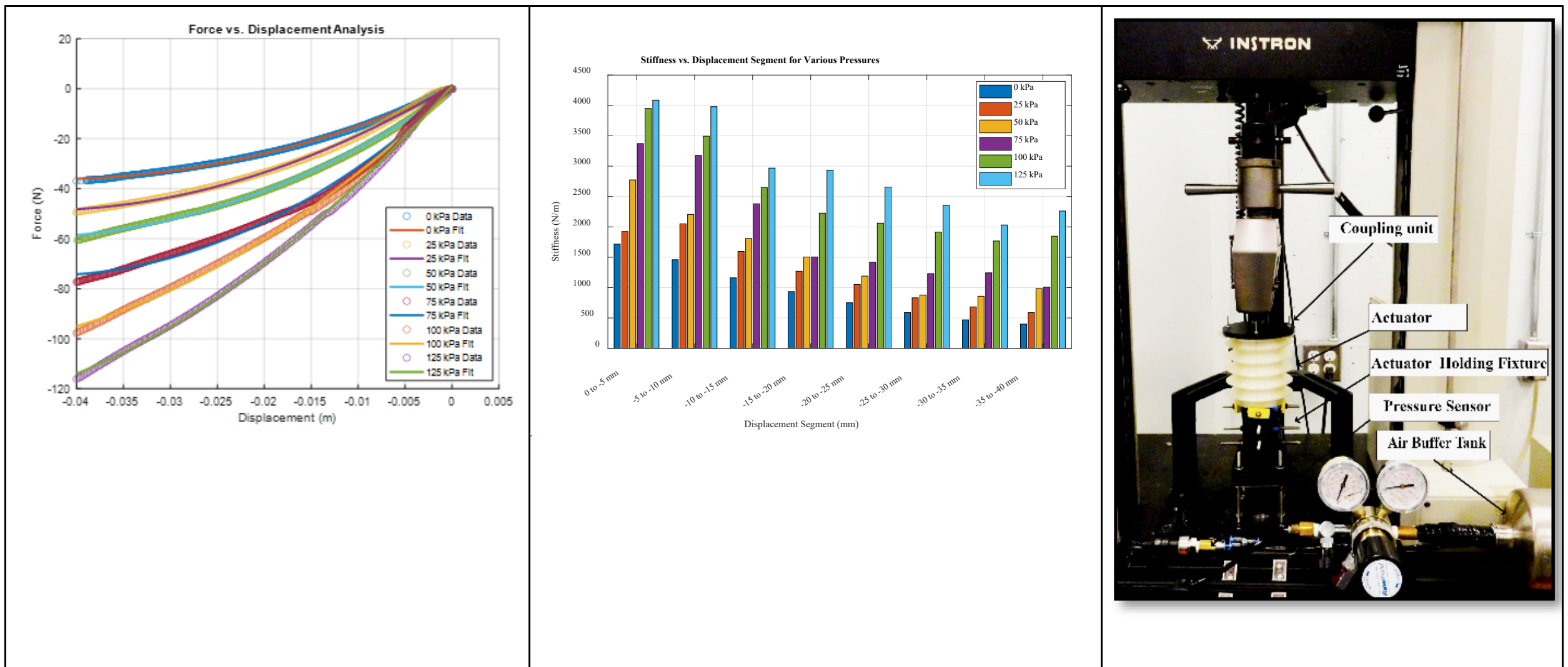

(a) (b) (c)

**Figure 6.12:** LSSA Model L13 variable stiffness experiment (a) experimental and fitted force-displacement data (b) Variation of actuator stiffness with displacement under different pressure conditions (c) experimental Setup

The results in Figure 6.12(a), illustrate the actuator's response across the five pressure settings. As the internal pressure within the actuator increased, the force required to induce displacement also increased, indicating a direct relationship between pressure and stiffness. The force-displacement relationship exhibited by the LSSA was nonlinear, and a polynomial model was applied to approximate the force-displacement curve.

Further analysis of the actuator's stiffness, defined as the force-to-displacement ratio, was conducted across eight discrete measurement intervals ranging from 0 to 40 mm in 5 mm increments. This analysis revealed a clear dependence of stiffness on internal pressure variations. Notably, at 0 kPa, the stiffness measured within the 0 to 5 mm displacement range was 1716.42 N/m, which increased significantly to 4087.55 N/m at 125 kPa for the same displacement range. This trend demonstrates a strong correlation between increased internal pressure and enhanced stiffness within the actuator. Figure 6.12 (b) graphically depicts the relationship between internal pressure and stiffness, highlighting the critical influence of pressure adjustments on the actuator's mechanical properties.

### 6.1.9 LSSA Force versus Displacement

This experiment aimed to measure the force output of a LSSA at different displacement points during extension, offering critical insights into the actuator's force-displacement characteristics and understanding how force varies as the LSSA moves through its range of motion.

The experimental setup, depicted in Figure 6.12 (c), is like that used for stiffness testing in section (6.1.8) with specific modifications to focus on force-displacement measurements. The procedure began by securely mounting the LSSA to an Instron testing machine, pressurizing the actuator to 125 kPa, and initiating the testing phase. During the experiment, the Instron head was gradually raised, allowing the LSSA to extend, and the force output at each displacement point was recorded. This process continued until the actuator reached its maximum extension, at which point the force generated diminished to zero. The test results for Model L13, illustrated in Figure 6.13, show a progressive decrease in force as displacement increased, ultimately reaching zero at the LSSA's full extension.

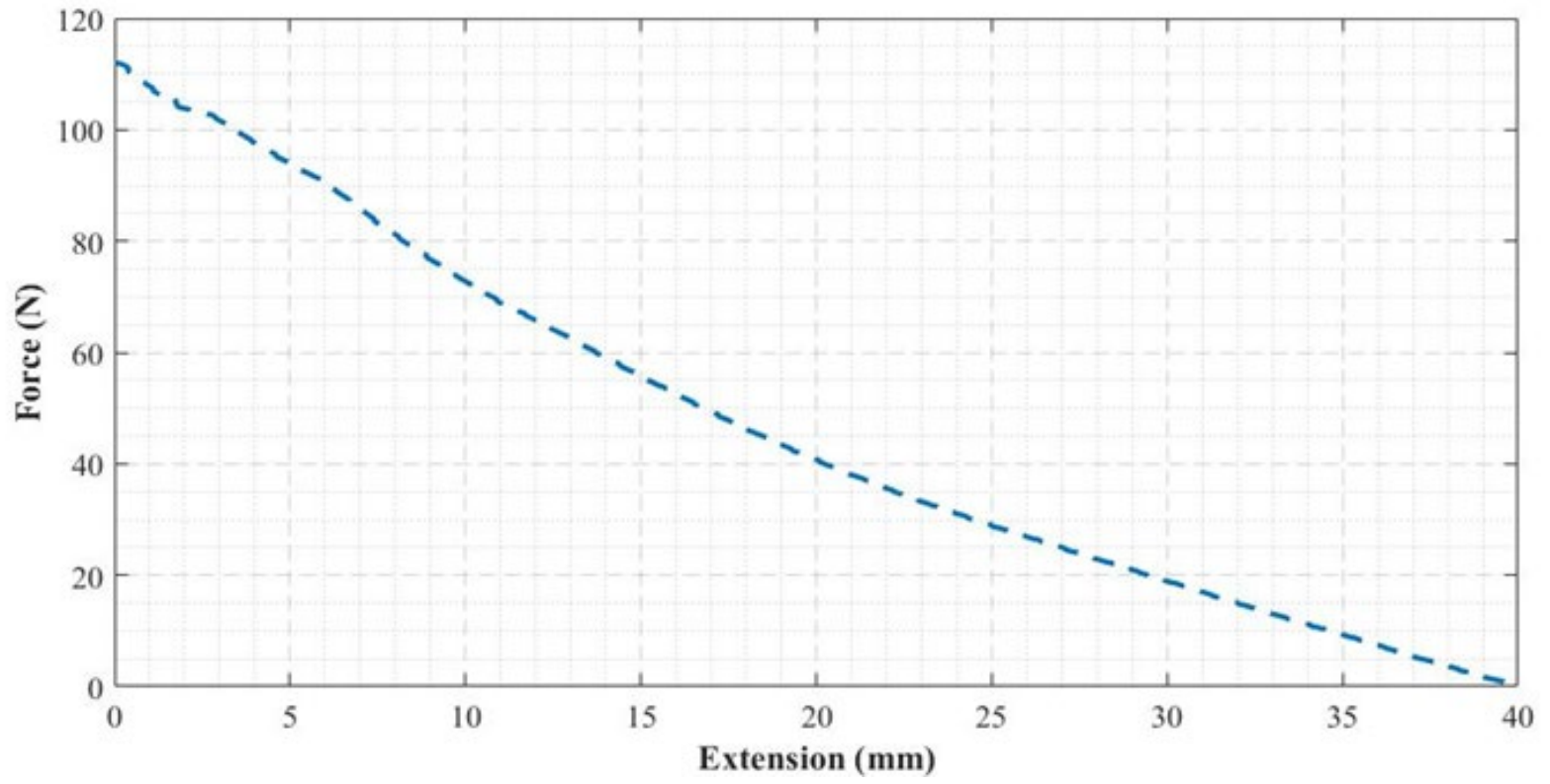

**Figure 6.13:** Force–displacement test results for model L13

The observed decline in force with increasing displacement is primarily attributed to the LSSA's rising axial stiffness. As the LSSA extends, its stiffness increases, which inversely affects its ability to generate force. This phenomenon highlights the inherent trade-off between displacement and force generation, particularly at the upper limits of the LSSA's extension range.

## 6.2 BSSA

This section presents the experimental results for the Bending Soft Sleeve Actuator (BSSA), whose performance depends on both material stiffness and key geometric parameters. To evaluate the influence of these parameters, a series of models was developed, as summarized in Table 6.2. The table provides a comprehensive overview of the geometric parameters, including radius (*r*), actuator length (*l*), fold width (*fw*), fold angle (*β*), number of restraining layers (*nr*), thickness of restraining layers (*tr*), wall thickness (*wt*), constraining layer thickness (*tc*), and Shore hardness (*sh*), as illustrated in Figures 3.4 and 3.5.

The specific values for these parameters were selected to meet both performance requirements and manufacturing constraints. For example, the minimum feasible fold angle is 30º, while the lowest printable Shore hardness is 85 Ha. Ensuring airtight integrity requires a wall thickness of at least 0.96 mm. A nozzle size of 0.4 mm dictates the minimum restraining layer thickness of 0.4 mm, and a minimum of ten restraining layers is needed to withstand internal pressures up to 200 kPa. The minimum constraining layer thickness is 1.6 mm to maintain airtight chambers.

By systematically varying these parameters, it is possible to examine their combined impact on the BSSA's overall performance. The following subsections detail the experimental results obtained from these models, focusing on how different geometric and material configurations influence bending behavior and force output.

**TABLE 6.2:** BSSA Models (all dimensions in mm; all angles in degrees)

| Model | *r* | *l* | *fw* | *β* | *tr* | *wt* | *nr* | *tc* | *sh* |
|---|---|---|---|---|---|---|---|---|---|
| ***B1*** | 30 | 120 | 16 | 30 | 0.4 | 0.96 | 18 | 16 | 85 |
| ***B2*** | 30 | 120 | 16 | 30 | 0.4 | 1.2 | 18 | 1.6 | 85 |
| ***B3*** | 30 | 120 | 16 | 30 | 0.4 | 1.6 | 18 | 1.6 | 85 |
| ***B4*** | 30 | 120 | 16 | 30 | 0.8 | 1.2 | 18 | 1.6 | 85 |
| ***B5*** | 30 | 120 | 16 | 30 | 0.4 | 1.2 | 18 | 1.6 | 95 |
| ***B6*** | 30 | 120 | 16 | 30 | 0.4 | 1.2 | 18 | 3.2 | 85 |
| ***B7*** | 30 | 120 | 16 | 30 | 0.4 | 1.2 | 10 | 1.6 | 85 |
| ***B8*** | 30 | 120 | 14 | 30 | 0.4 | 1.2 | 10 | 1.6 | 85 |
| ***B9*** | 30 | 120 | 12 | 30 | 0.4 | 0.96 | 18 | 1.6 | 85 |
| ***B10*** | 30 | 120 | 8 | 30 | 0.4 | 0.96 | 18 | 1.6 | 85 |
| ***B11*** | 30 | 120 | 16 | 45 | 0.4 | 1.2 | 18 | 1.6 | 85 |
| ***B12*** | 30 | 120 | 16 | 30 | 0.8 | 0.96 | 12 | 1.6 | 85 |
| ***B13*** | 30 | 120 | 16 | 30 | 0.8 | 0.96 | 10 | 1.6 | 85 |

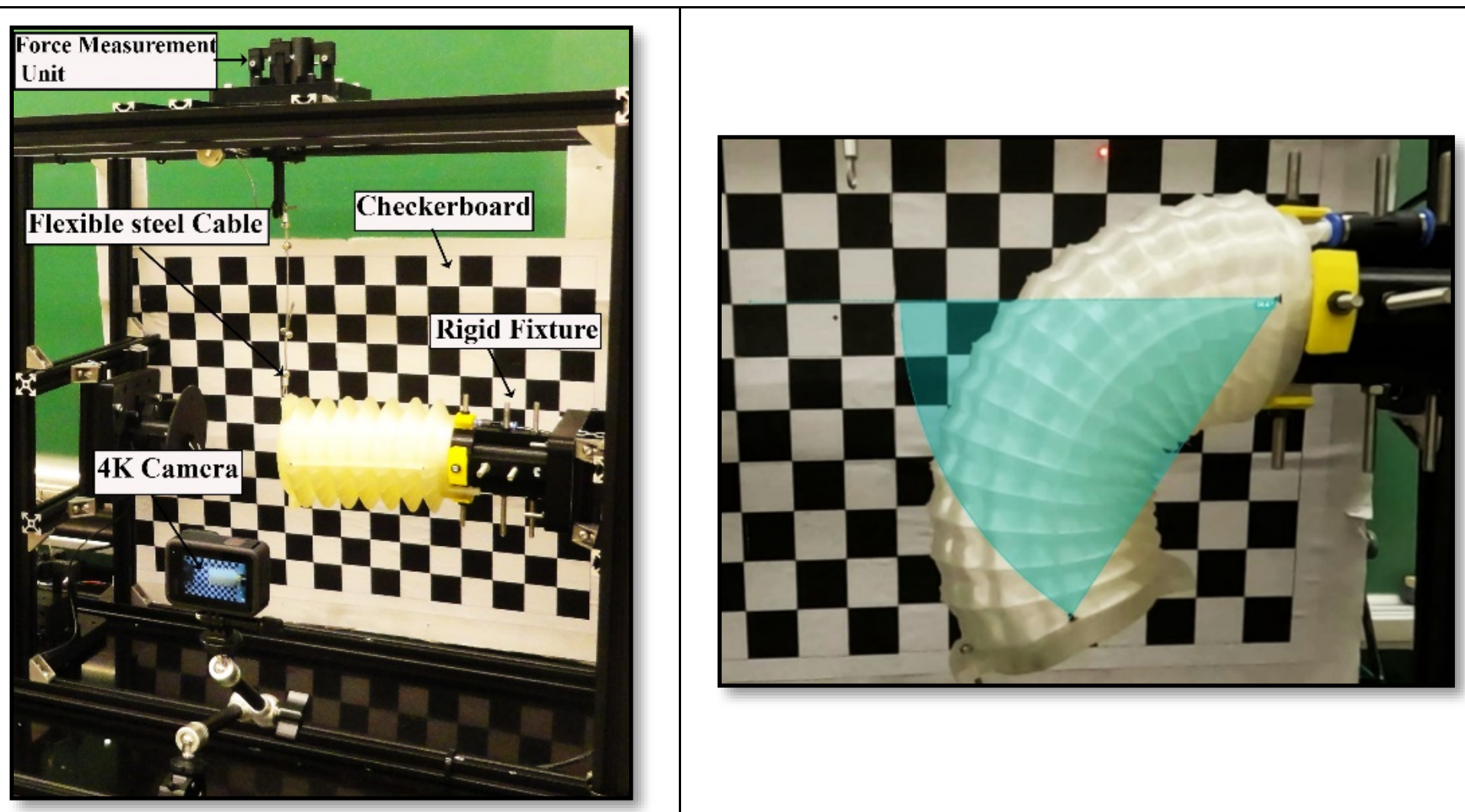

(a) (b)

**Figure 6.14:** (a) Experimental setup for measuring force during bending motion (b) experimental setup for measuring angle during bending motion

### 6.2.1 BSSA Force Performance during Bending

The evaluation of the forces generated by the BSSA was conducted using the experimental setup depicted in Figure 6.14 (a). This setup includes a rigid fixture that anchors the BSSA at one end, leaving the opposite end free to move. The BSSA was subjected to controlled pressurization, with pressure levels incrementally increased up to a maximum of 200 kPa. The impact of geometric parameters and material stiffness on the force-generating capability of the BSSA was investigated using the models presented in Table 6.2. The results of these tests are summarized in Figure 6.15.

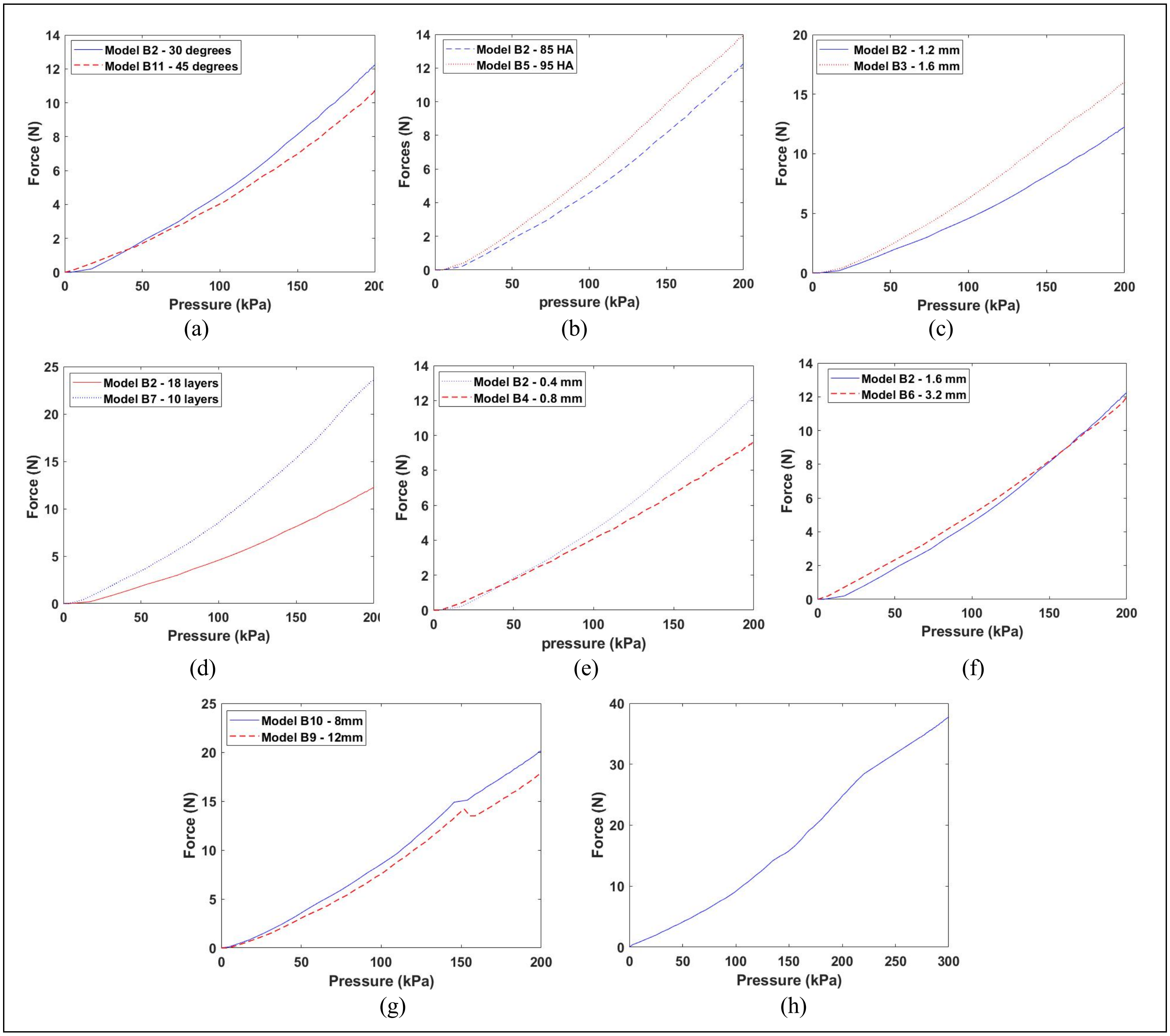

**Figure 6.15:** Impact of geometrical parameters and material stiffness on generated force of BSSA. (a) effect of fold angle (b) effect of material stiffness (c) effect of wall thickness (d) effect of the number of tie-restraining layers (e) effect of the tie restraining layers thickness (f) effect of constrained layers (g) effect of fold width (h) generated forces for model B13.

**Effect of fold angle:** As shown in Figure 6.15 (a), increasing the fold angle from 30º to 45º resulted in a reduction in force from 12.3 N to 10.8 N, indicating an inverse relationship between fold angle and force generation.

**Effect of material stiffness:** The experimental results in Figure 6.15 (b) demonstrate that increasing the material stiffness from 85A to 95A slightly increases force output, with values rising from 12.3 N to 14N.

**Effect of wall thickness:** Figure 6.15 (c) reveals that an increase in wall thickness from 1.2 mm to 1.6 mm leads to enhanced force generation, with force increasing from 12.3 N to 16 N. This suggests that greater wall thickness positively contributes to force output.

**Effect of the number of tie-restraining layers:** Testing results shown in Figure 6.15 (d) indicate that reducing the number of tie-restraining layers from 18 to 10 leads to a significant increase in force output, from 12.3 N to 23.5 N, suggesting that fewer restraining layers enhance force generation.

**Effect of the thickness of tie-restraining layers:** As depicted in Figure 6.15 (e), increasing the thickness of tie-restraining layers by 0.4 mm reduces the force from 12.3 N to 9.3 N, indicating a negative correlation between layer thickness and force generation.

**Effect of constrained layers:** Figure 6.15 (f) illustrates that increasing the thickness of constrained layers from 1.6 mm to 3.2 mm has a negligible impact on the forces generated by the BSSA.

**Effect of fold width:** The data presented in Figure 6.15 (g) indicates that increasing the fold width from 8 mm to 12 mm results in a force increase of 3 N, suggesting that wider folds contribute to enhanced force generation.

### 6.2.2 BSSA Displacement Performance during Bending

The measurement of the bending angle during bending motion was conducted according to a testing protocol established by previous studies [61], [92]. In the experimental setup, the BSSA's proximal end was securely clamped within an actuator fixture, while the distal end remained unrestrained to allow for free bending motion.

To accurately capture and analyze the BSSA's motion, a high-speed 4K camera (GoPro Hero 11) was positioned to align with a checkered background, enhancing the precision of the measurements. The recorded footage was processed using the open-source software Kinovea, which was utilized to calculate the bending angle, consistent with the approaches used in similar studies [61], [91], [92]. The impact of geometric parameters and material stiffness on the bending angle generation capability of the BSSA was investigated using the models detailed in Table 6.2. The results of these investigations are summarized in Figure 6.16.

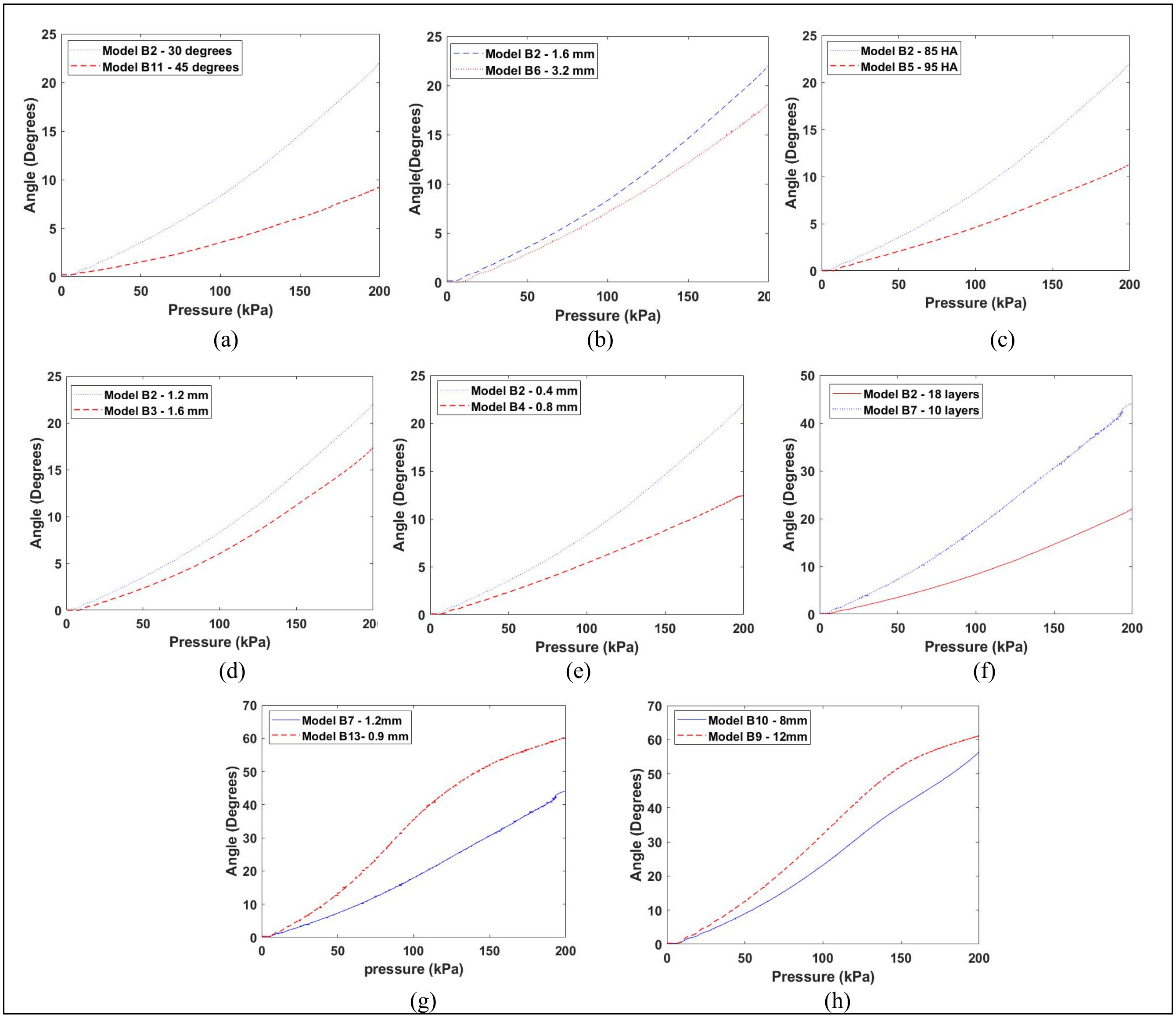

**Figure 6.16:** Impact of geometrical parameters and material stiffness on bending angle of BSSA (a) effect of fold angle (b) effect of constrained layers (c) effect of material stiffness (d) effect of wall thickness (e) effect of the tie restraining layers thickness (f) effect of the number of tie-restraining layers (g) improve the bending angle by reducing wall thickness (h) effect of fold width.

**Effect of fold angle:** As depicted in Figure 6.16 (a), increasing the fold angle from 30º to 45º significantly reduces the bending angle, decreasing from 22º to 9.2º. This suggests that larger fold angles limit the BSSA's bending capability.

**Effect of material stiffness:** The results presented in Figure 6.16 (c) demonstrate a substantial decrease in the bending angle, from 22º to 12º, as material stiffness increases. This indicates an inverse relationship between material stiffness and bending flexibility, showing that stiffer materials reduce the BSSA's ability to bend.

**Effect of wall thickness:** As illustrated in Figure 6.16 (d), increasing the wall thickness from 1.2 mm to 1.6 mm reduces the bending angle from 22º to 17.3º. This suggests that thicker walls hinder the BSSA's bending motion.

**Effect of the number of tie-restraining layers:** Testing results depicted in Figure 6.16 (e) reveal that reducing the number of tie-restraining layers from 18 to 10 leads to a significant increase in the bending angle, from 22º to 44º. This finding suggests that fewer restraining layers enhance the BSSA's bending capability.

**Effect of the thickness of tie-restraining layers:** Figure 6.16 (f) shows that increasing the thickness of the tie-restraining layers by 0.4 mm results in a decrease in the bending angle from 22º to 12º. This suggests that thicker restraining layers reduce the bending capability of the BSSA.

**Effect of constrained layers:** As shown in Figure 6.16 (b), increasing the thickness of the constrained layers from 1.6 mm to 3.2 mm leads to a slight reduction in the bending angle, from 22º to 18º. This suggests thicker constrained layers may marginally restrict the BSSA's bending motion.

**Effect of fold width:** Regarding bending performance, Figure 6.16 (h) demonstrates that increasing the fold width from 8 mm to 12 mm increases the bending angle from 56º to 61º. This indicates that wider folds enhance the BSSA's bending capability.

### 6.2.3 BSSA Displacement versus Load

The effect of static load on the bending angle of a BSSA was investigated using the experimental setup shown in Figure 6.17 (a). In this configuration, a stainless-steel cable was employed to apply the static load to the actuator. The cable was routed through a dual-pulley system to ensure that a vertical force was consistently applied to the proximal end of the actuator. This arrangement was carefully designed to maintain the alignment and integrity of the applied force, ensuring that the static load was uniformly distributed across the actuator's structure.

To preserve the BSSA's initial bending angle, the tension in the cable was carefully calibrated. This calibration was crucial for maintaining the actuator's quiescent state, free from any unintended influence by the applied load.

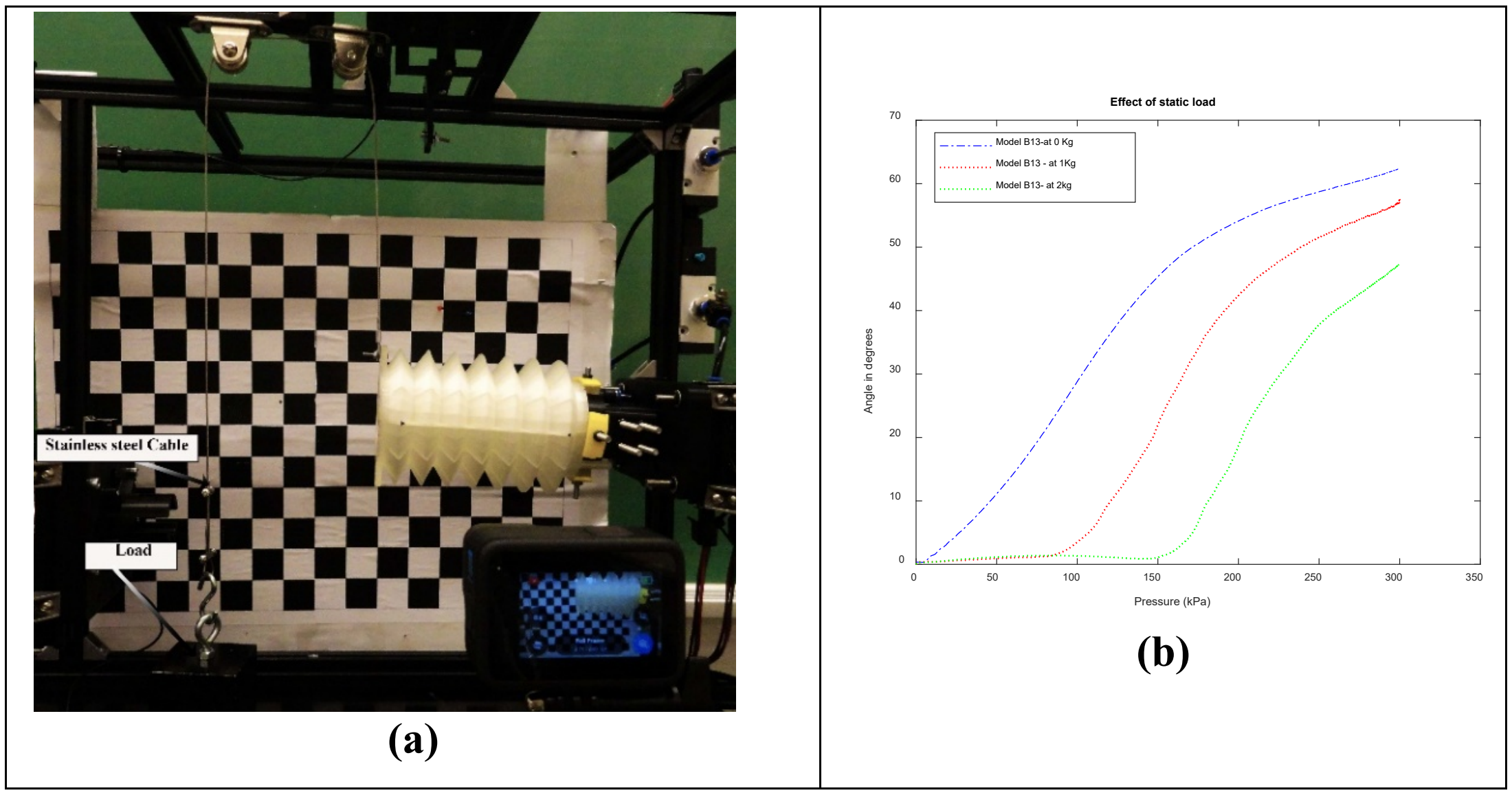

**Figure 6.17:** Effect of static load on generated force during bending motion (a) Experimental setup (b) test results for B13

For this test, the BSSA model b13 was subjected to controlled pressurization up to 300 kPa. The bending angles produced by the BSSA were analyzed using Kinovea, including a 4K resolution camera. This high-resolution imaging enabled precise and detailed measurement of the BSSA's response to the applied static load, capturing the bending angle.

The experimental results, presented in Figure 6.17 (b), demonstrate a clear reduction in the bending angle as the static load increased. Specifically, the bending angle decreased from 62.3º to 47.5º as the static mass increased from 1 kg to 2 kg. This observed decrease in the bending angle under higher static loads highlights the significant impact of external loads on the BSSA's performance.

### 6.2.4 BSSA Force versus Bending Angles

This experiment was conducted to investigate the force generated by the BSSA at different bending angles. Utilizing an experimental setup similar to section 6.2.1, this study aimed to quantify the force output corresponding to various angular positions of the actuator.

In this experimental framework, the BSSA was first maneuvered to achieve the target bending angle. Once the desired angle was attained, the BSSA was connected to the force measurement unit to accurately record the corresponding force output. This method ensured precise measurement of the force generated at specific bending angles.

The data presented in Figure 6.18 reveal a trend of force reduction as the bending angle increases. Specifically, the force generated by the BSSA model B13 decreased from 39.3 N in its initial state to 34.1 N when bent to a 40° angle. This observation highlights the inverse relationship between bending angle and force generation, providing insights into the performance characteristics of the actuator under varying bending conditions.

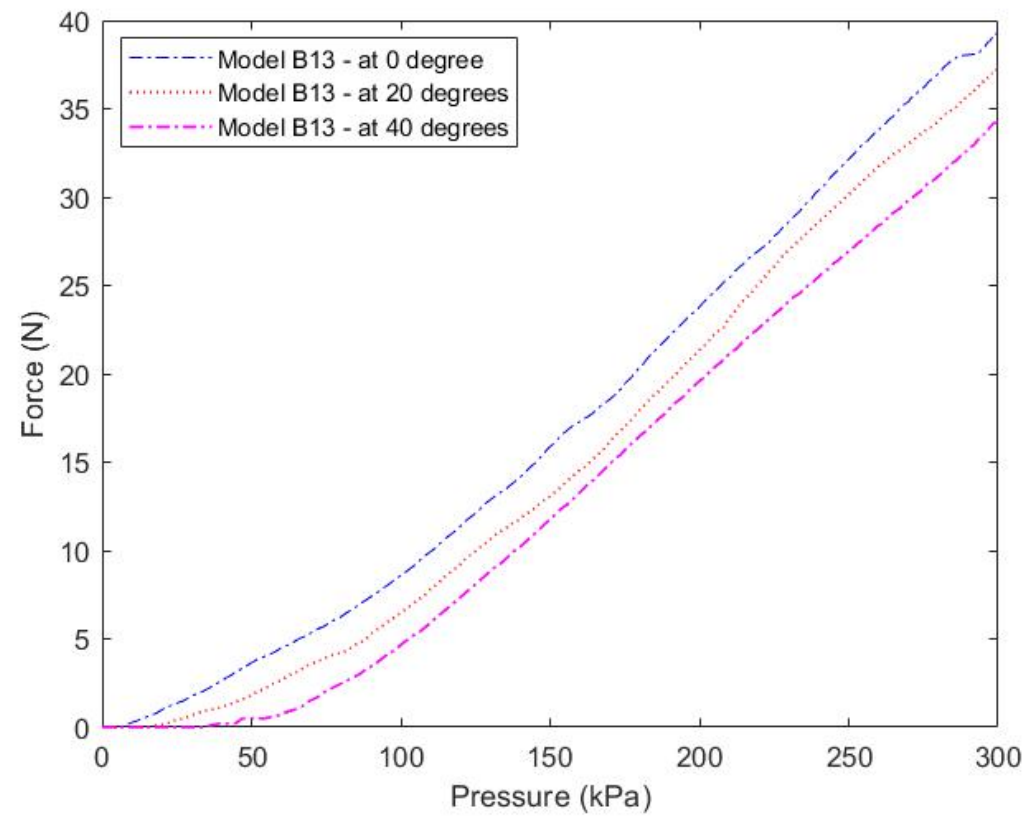

**Figure 6.18:** Force generated at various bending Angles

### 6.3 OSSA

This section presents the experimental results for the OSSA. The proposed OSSA was examined to assess its capacity to generate force and displacement, including both linear displacement and bending angles. Its performance depends on material stiffness as well as key geometric parameters. To investigate the influence of these parameters, a series of OSSA models was designed and evaluated using the previously described experimental platform. This approach enabled the isolation and detailed analysis of each parameter, offering insights into how they affect the actuator's force output and displacement characteristics.

Table 6.3 provides a comprehensive overview of the tested models, specifying critical geometric parameters such as radius (*r*), actuator length (*l*), fold width (*fw*), fold angle (*β*), number of restraining layers (*nr*), thickness of restraining layers (*rt*), wall thickness (*wt*), number of chambers (*nc*), and Shore hardness (*sh*). These parameters, illustrated in Figure 3.15, were carefully selected to address both performance objectives and manufacturing constraints. For example, the minimum feasible fold angle is 30º, while the lowest printable Shore hardness is 85 Ha. Achieving airtight integrity requires a wall thickness of at least 0.96 mm, and a nozzle size of 0.4 dictates a minimum restraining layer thickness of 0.4 mm. Moreover, at least ten restraining layers are needed to withstand internal pressures of up to 200. By varying these parameters, the effects on the OSSA's overall performance—encompassing both linear and bending motion.

**Table 6.3:** OSSA Models (all dimensions in mm; all angles in degrees)

| Model | *r* | *l* | *fw* | *β* | *rt* | *wt* | *nr* | *nc* | *sh* |
|---|---|---|---|---|---|---|---|---|---|
| ***Omni 1*** | 30 | 120 | 16 | 30 | 0.4 | 0.96 | 18 | 2 | 85 |
| ***Omni 2*** | 30 | 120 | 16 | 30 | 0.4 | 1.2 | 18 | 2 | 85 |
| ***Omni 3*** | 30 | 120 | 12 | 30 | 0.4 | 0.96 | 18 | 2 | 85 |
| ***Omni 4*** | 30 | 120 | 16 | 30 | 0.8 | 1.2 | 12 | 2 | 85 |
| ***Omni 5*** | 30 | 120 | 16 | 30 | 0.8 | 1.2 | 18 | 2 | 85 |
| ***Omni 6*** | 30 | 120 | 8 | 30 | 0.4 | 0.96 | 10 | 2 | 85 |
| ***Omni 7*** | 30 | 120 | 12 | 35 | 0.4 | 0.96 | 10 | 2 | 85 |
| ***Omni 8*** | 30 | 120 | 12 | 40 | 0.4 | 0.96 | 18 | 2 | 85 |
| ***Omni 9*** | 30 | 120 | 12 | 30 | 0.4 | 0.96 | 18 | 2 | 95 |
| ***Omni10*** | 30 | 120 | 8 | 30 | 0.4 | 0.96 | 10 | 3 | 85 |
| ***Omni11*** | 30 | 120 | 8 | 30 | 0.4 | 0.96 | 10 | 4 | 85 |

### 6.3.1 OSSA Force Performance during Extension

The forces generated during the extension motion of the OSSA were evaluated using the experimental setup depicted in Figure 6.19 (a). In this configuration, one end of the OSSA was securely fixed to a fixture on the platform, while the opposite end was left free to move. The FMU was positioned to contact the free-moving end of the OSSA, enabling measurement of the forces exerted during extension. All chambers of the OSSA were subjected to controlled pressurization up to a maximum of 130 kPa. The force generated at the terminal end of the OSSA was recorded throughout the experiment.

To investigate the impact of geometric parameters and material stiffness on the force-generating capability of the OSSA during extension, a series of models, as detailed in Table 6.3, were tested. The results, summarized in Figure 6.20, provide a comprehensive understanding of how each parameter influences the OSSA's performance. The following subsections offer an in-depth analysis of these effects, highlighting the critical factors that contribute to the actuator's force generation during extension.

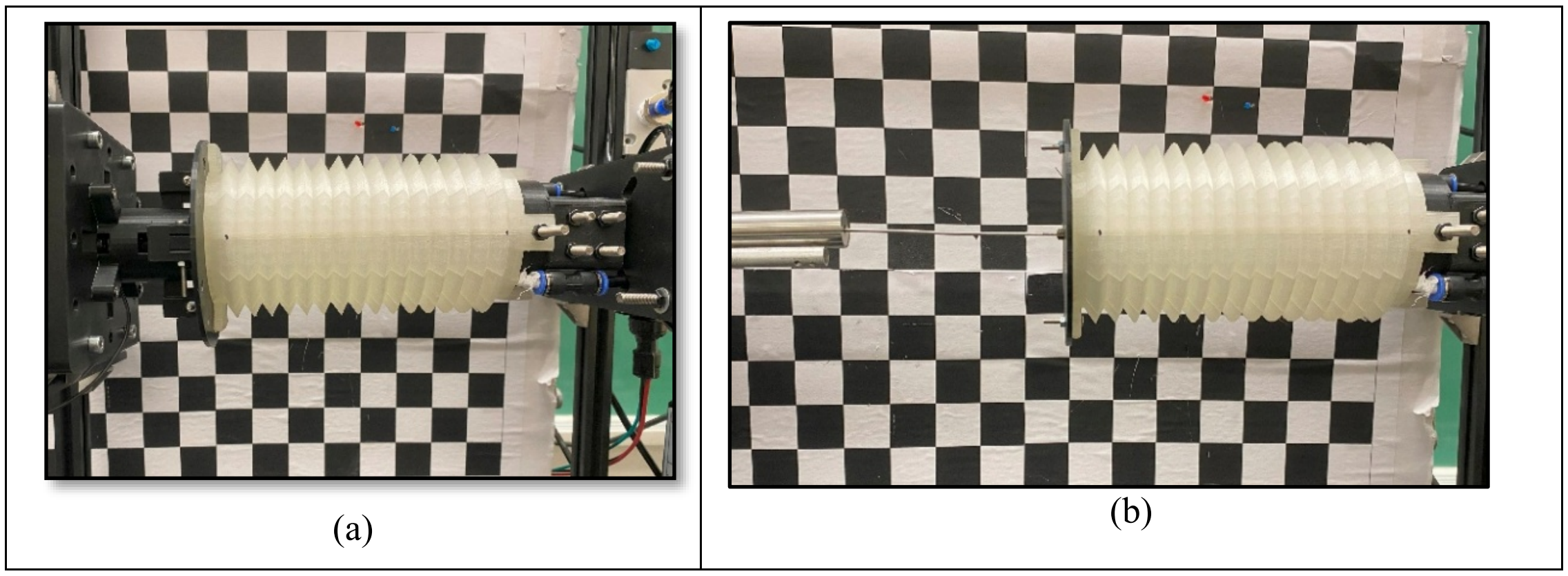

**Figure 6.19:** (a) Experimental setup for measuring force during the linear motion of the OSSA, (b) experimental setup for measuring the linear displacement of the OSSA

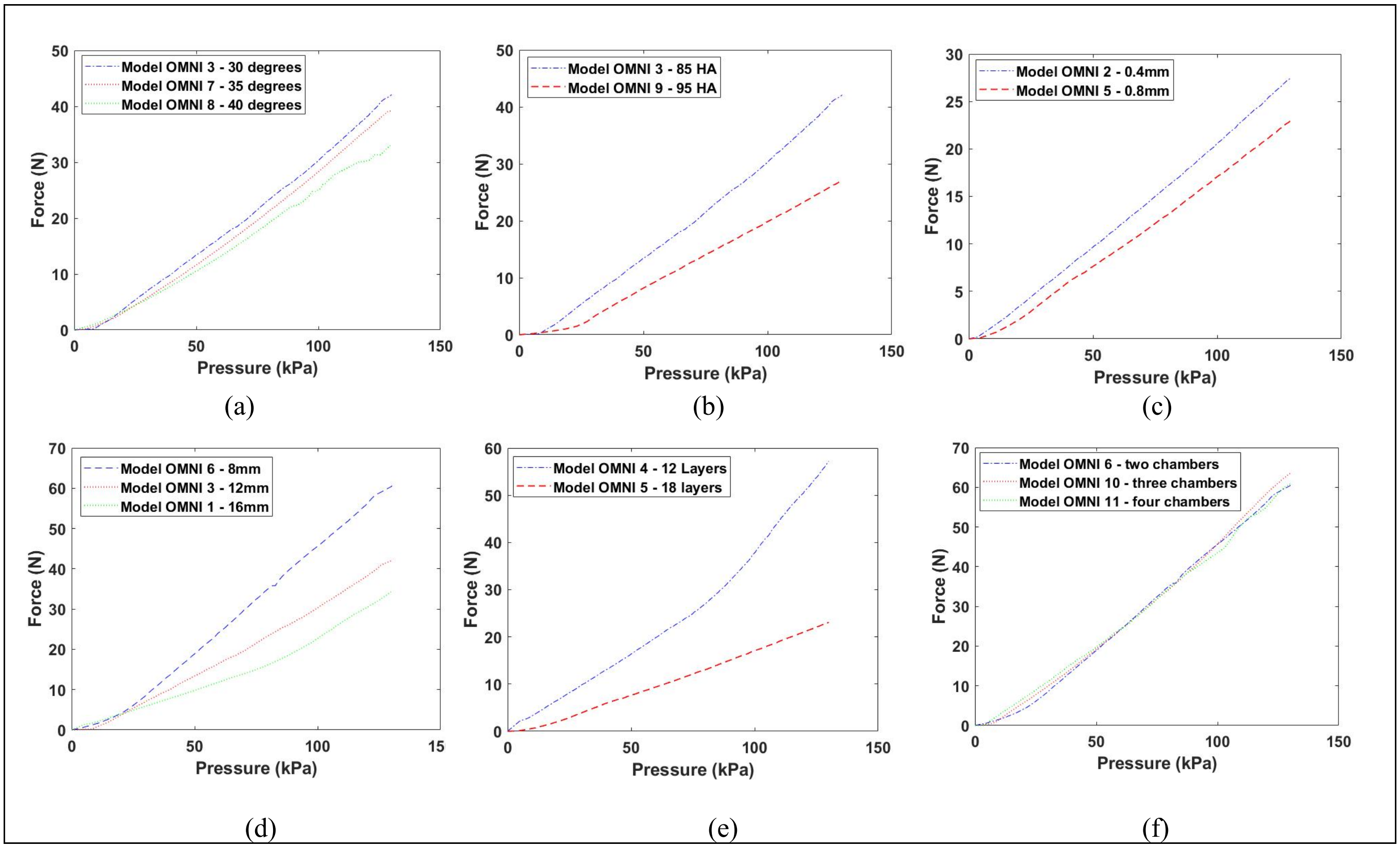

**Figure 6.20:** Impact of geometrical parameters and material stiffness on generated force of OSSA in extension motion (a) effect of fold angle (b) effect of material stiffness (c) effect of the tie restraining layers thickness (d) effect of fold width (e) effect of the number of tie-restraining layers (f) effect of the number of chambers.

**Effect of fold angle:** As depicted in Figure 6.20 (a), increasing the fold angle from 30º to 40º results in a decrease in the force generated by the OSSA, from 42 N to 38 N. This observation indicates that a larger fold angle negatively impacts the OSSA's force generation capacity.

**Effect of material stiffness:** Figure 6.20 (b) illustrates that increasing the material stiffness from 85 Ha to 95 Ha significantly reduces the force generated by the OSSA from 42 N to 27 N. This finding underscores the critical role of material flexibility in force generation, with stiffer materials limiting the OSSA's ability to convert internal pressure into mechanical force effectively.

**Effect of tie-restraining layer thickness:** The influence of tie-restraining layer thickness on force generation is evident in Figure 6.20 (c). Increasing the thickness of these layers from 0.4 mm to 0.8 mm reduces the force generated by the OSSA from 27 N to 23 N. This suggests that thicker restraining layers impose additional resistance, thereby diminishing the OSSA's overall force output.

**Effect of fold width:** As shown in Figure 6.20 (d), increasing the fold width from 8 mm to 16 mm significantly reduces the force generated by the OSSA from 60 N to 34 N.

**Effect of the number of tie-restraining layers:** The number of tie-restraining layers also plays a crucial role in force generation, as demonstrated in Figure 6.20 (e). Increasing the number of restraining layers from 12 to 18 results in a marked decrease in the force generated by the OSSA, from 57 N to 23 N.

**Effect of the number of air chambers:** As shown in Figure 6.20 (f), the number of air chambers within the actuator appears to have a negligible effect on the generated force.

### 6.3.2 OSSA Displacement Performance during Extension and Contraction

The displacement capabilities of the omnidirectional actuator, encompassing both extension and contraction motions, were evaluated using the experimental setup depicted in Figure 6.19 (b). In this setup, one end of the OSSA was securely fixed to a fixture on the platform while the opposite end was left free to move. To assess the OSSA's performance under varying conditions, all chambers of the OSSA were subjected to controlled pressurization. For extension motion, pressure levels were incrementally increased up to a maximum of 130 kPa, while for contraction motion, negative pressure was applied, incrementally decreasing to -90 kPa. To measure the displacement generated during both extension and contraction phases, LVDT was affixed to the free-moving end of the actuator using a designed 3D-printed coupling unit.

This investigation aimed to study the influence of geometric parameters and material stiffness on the displacement-generating capability of the OSSA during both extension and contraction. The models utilized for this analysis are detailed in Table 6.3, with the corresponding results summarized in Figure 6.21 and 6.22. The following subsections provide an in-depth analysis of how each parameter affects the actuator's performance.

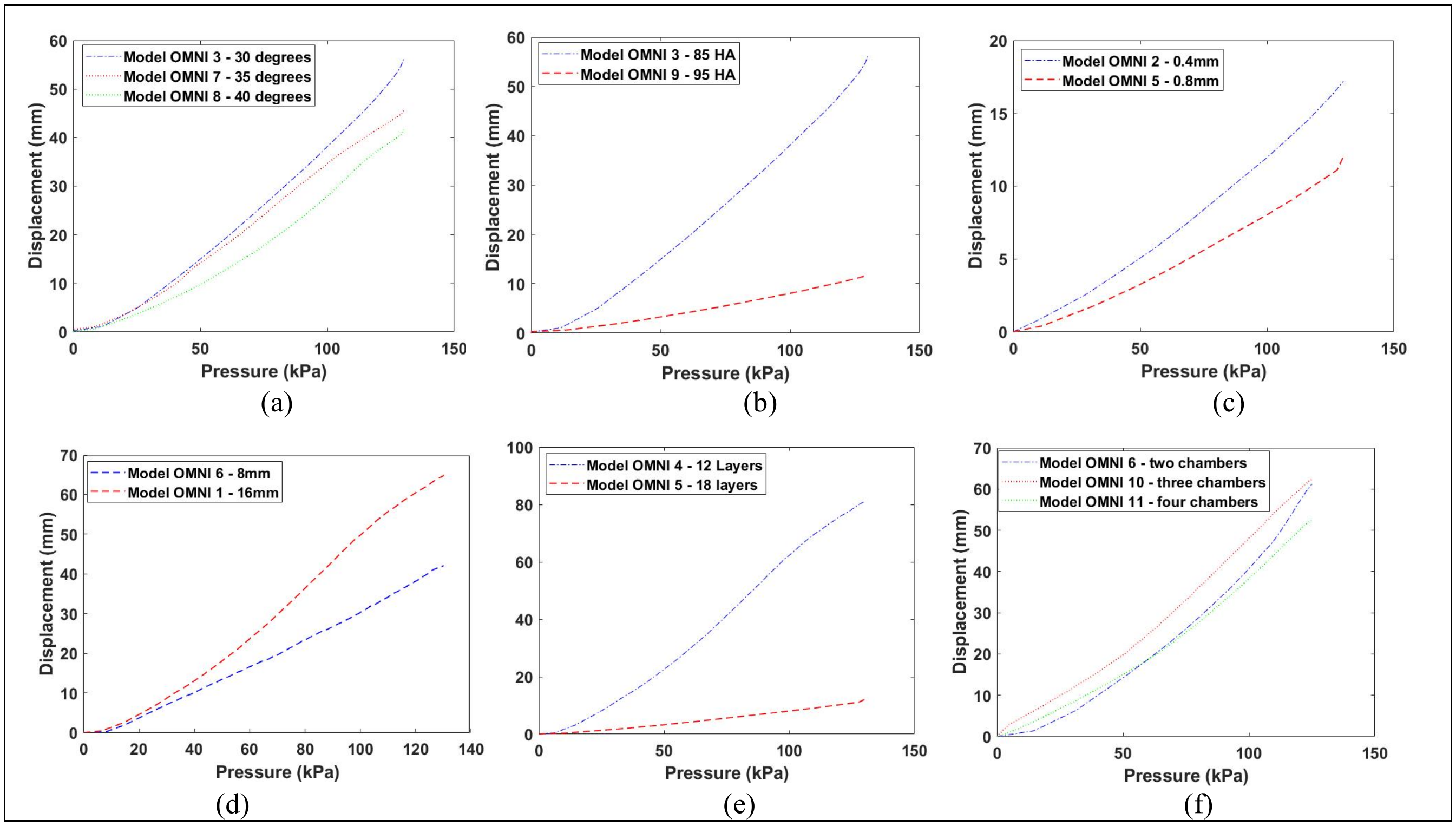

**Figure 6.21:** Impact of geometrical parameters and material stiffness on displacement of OSSA in extension motion (a) effect of fold angle (b) effect of material stiffness (c) effect of the tie restraining layers thickness (d) effect of fold width (e) effect of the number of tie-restraining layers (f) effect of the number of chambers.

**Effect of fold angle:** An increase in the fold angle from 30º to 40º reduces the extension displacement from 56 mm to 46 mm, as illustrated in Figure 6.21 (a). Similarly, this increase in fold angle also leads to decreased contraction displacement, reducing it from 16 mm to 9 mm, as shown in Figure 6.22 (a). These results suggest that larger fold angles hinder both the extension and contraction movements of the OSSA.

**Effect of material stiffness:** As depicted in Figure 6.21 (b), increasing the material stiffness from 85 Ha to 95 Ha significantly reduces the extension displacement from 56 mm to 11 mm. Likewise, this increase in material stiffness decreases the contraction displacement from 16 mm to 6 mm, as shown in Figure 6.22 (b). These findings indicate that higher material stiffness restricts the OSSA's ability to extend and contract.

**Effect of tie-restraining layer thickness:** The thickness of the tie-restraining layers also plays a significant role in displacement performance. As shown in Figure 6.21 (c), increasing the thickness from 0.4 mm to 0.8 mm results in a decrease in extension displacement from 17 mm to 11 mm. A similar reduction is observed in contraction displacement, which decreases from 8 mm to 6 mm, as presented in Figure 6.22 (c). These results suggest that thicker restraining layers limit the OSSA's displacement capabilities.

**Effect of fold width:** Increasing the fold width from 8 mm to 16 mm leads to an increase in extension displacement from 68 mm to 81 mm, as indicated in Figure 6.21 (d). However, this increase in fold width also causes a reduction in contraction displacement from 23 mm to 16 mm, as shown in Figure 6.22 (d). These findings imply that while a wider fold enhances extension, it may simultaneously constrain contraction.

**Effect of the number of tie-restraining layers:** As illustrated in Figure 6.21 (e), increasing the number of restraining layers from 12 to 18 leads to a reduction in extension displacement from 81 mm to 12 mm. Similarly, this increase in restraining layers reduces contraction displacement from 16 mm to 6 mm, as shown in Figure 6.22 (e). These results suggest that a higher number of restraining layers may severely limit the OSSA's ability to extend and contract.

**Effect of the number of air Chambers:** The number of air chambers within the actuator appears to have a negligible effect on both extension and contraction motions, as indicated in Figures 6.21 (f) and 6.22 (f). This suggests that other factors, such as material stiffness and geometric configuration, play a more critical role in determining the OSSA's performance.

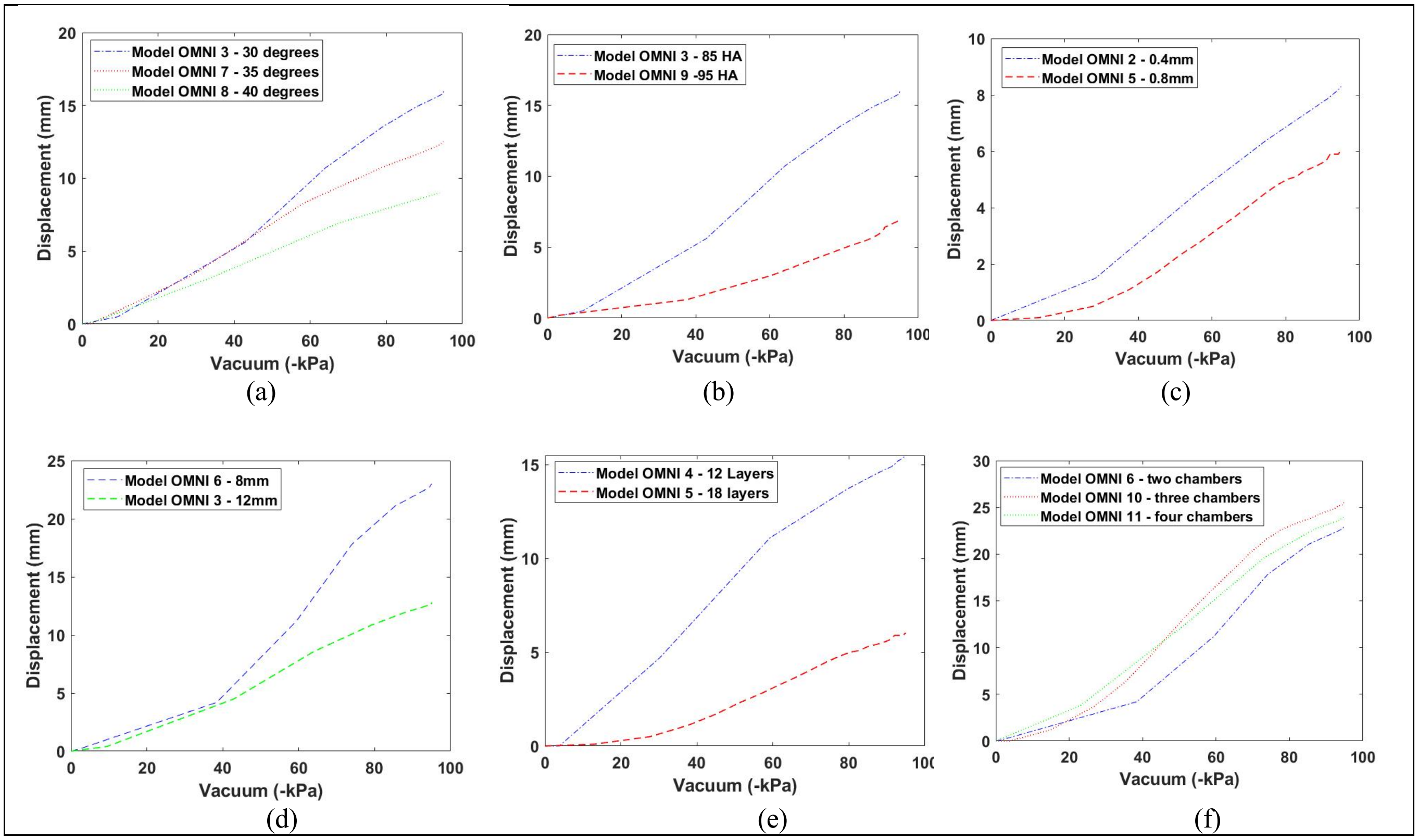

**Figure 6.22:** Impact of geometrical parameters and material stiffness on displacement of OSSA in contraction motion (a) effect of fold angle (b) effect of material stiffness (c) effect of the tie restraining layers thickness (d) effect of fold width (e) effect of the number of tie-restraining layers (f) effect of the number of chambers.

### 6.3.3 OSSA Force Performance during Bending

The force generated during the bending motion of the OSSA was conducted using the experimental setup illustrated in Figure 6.23 (a). The testing protocol employed in this experiment aligns with the procedures used in section 6.2.1. However, in this case, the experimental conditions involved subjecting the first half of the OSSA to a positive pressure of up to 130 kPa, while the second half was exposed to a negative pressure of up to -90 kPa.

The influence of geometric parameters and material stiffness on the force-generating capability of the omnidirectional actuator during bending motion was investigated using the models detailed in Table 6.3. The results of these experiments are summarized in Figure 6.24. The following subsections provide a comprehensive analysis of how variations in each parameter affected the OSSA's performance during bending motion.

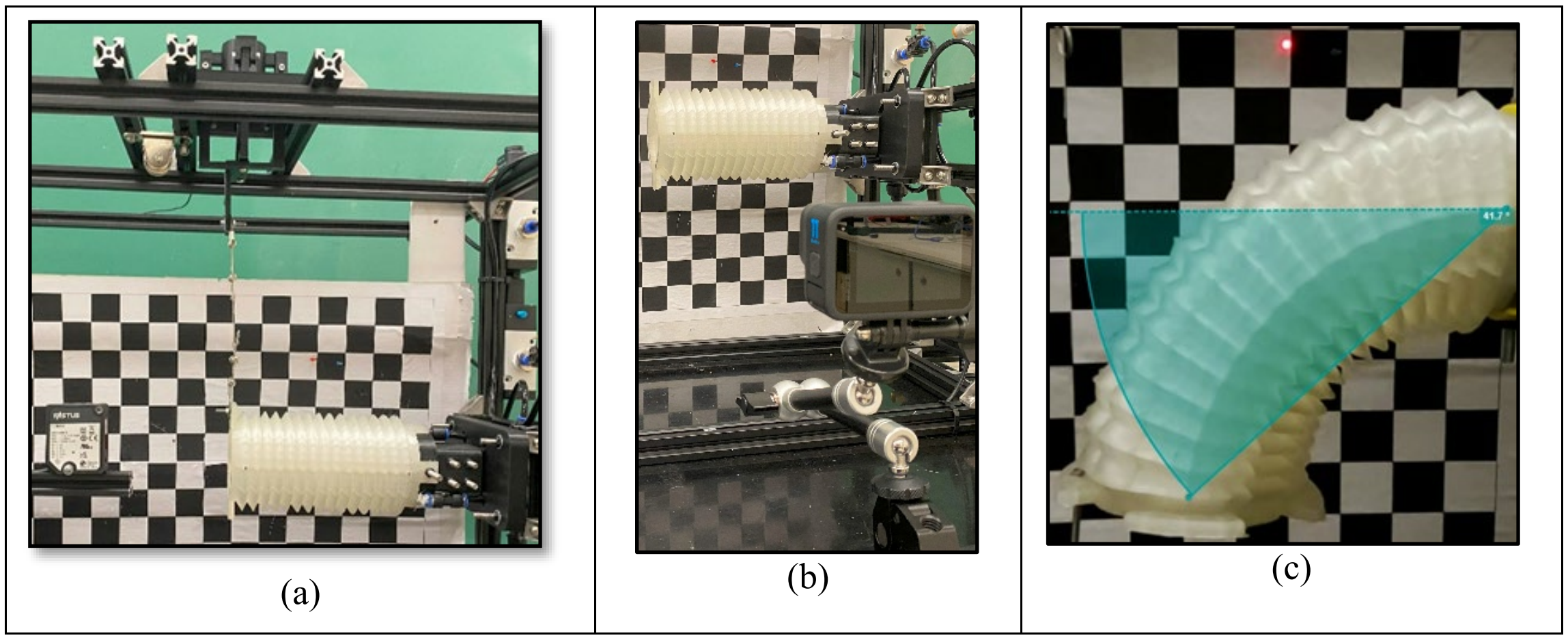

**Figure 6.23:** (a) Experimental setup for measuring force during the bending motion of the OSSA, (b) experimental setup for measuring the bending angle of the OSSA, (c) bending angle measurement using Kinovea software

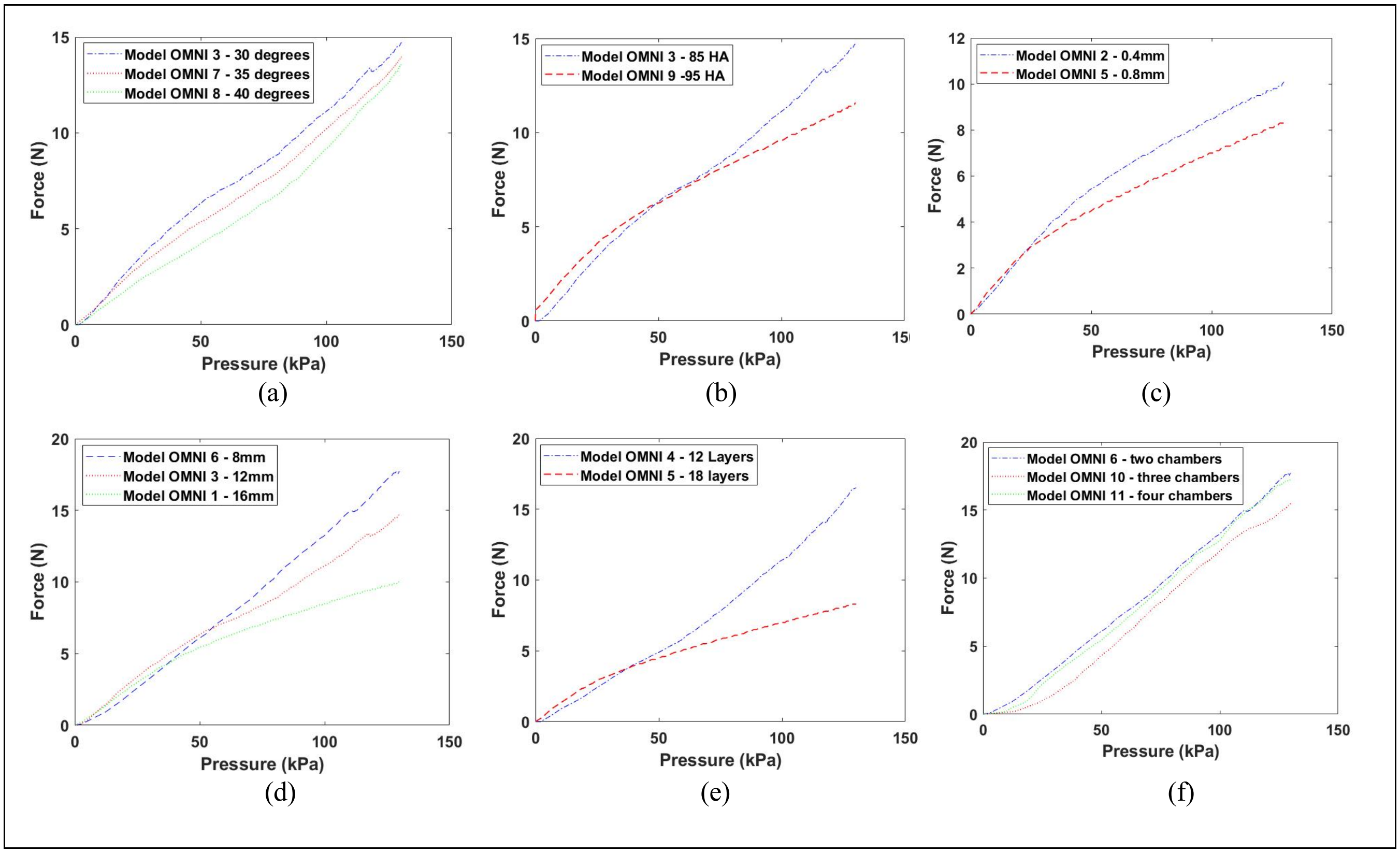

**Figure 6.24:** Impact of geometrical parameters and material stiffness on generated force of OSSA in bending motion (a) effect of fold angle (b) effect of material stiffness (c) effect of the tie restraining layers thickness (d) effect of fold width (e) effect of the number of tie-restraining layers (f) effect of the number of chambers.

**Effect of fold angle:** An increase in the fold angle from 30º to 40º resulted in a reduction in force generation, decreasing from 15 N to 13.5 N, as illustrated in Figure 6.24 (a). This suggests that a larger fold angle negatively impacts the OSSA's ability to generate force during bending motion.

**Effect of material stiffness:** As depicted in Figure 6.24 (b), increasing the material stiffness from 85 Ha to 95 Ha led to a reduction in force generation from 15 N to 12 N. This indicates that higher material stiffness reduces the flexibility of the OSSA, thereby diminishing its force output.

**Effect of tie-restraining layer thickness:** The influence of tie-restraining layer thickness on force generation is evident in Figure 6.24 (c). Increasing the thickness of these layers from 0.4 mm to 0.8 mm resulted in a decrease in force from 10 N to 8 N. This suggests that thicker restraining layers impose additional resistance, reducing the OSSA's overall force output.

**Effect of fold width:** As shown in Figure 6.24 (d), increasing the fold width from 8 mm to 16 mm significantly reduced the force generation from 18 N to 10 N. This reduction indicates that wider folds may distribute the force over a larger area, thus decreasing the effective force output during bending.

**Effect of the number of tie-restraining layers:** The number of tie-restraining layers also plays a crucial role in force generation. As illustrated in Figure 6.24 (e), increasing the number of restraining layers from 12 to 18 resulted in a significant reduction in force, from 16 N to 8 N. This suggests that a higher number of restraining layers may constrain the OSSA's motion, thereby reducing its force-generating capability.

**Effect of the Number of Air Chambers:** The number of air chambers within the actuator appears to have a negligible effect on the force generated, as indicated in Figure 6.24 (f).

### 6.3.4 OSSA Displacement Performance during Bending

The evaluation of bending angle generation during bending motion was performed using a testing protocol consistent with the methodology described in Section 6.2.2. This study investigated the impact of geometric parameters and material stiffness on the actuator's bending angle generation capability, utilizing the models outlined in Table 6.3. The results of these investigations are summarized in Figure 6.25, and the following subsections provide a comprehensive analysis of how variations in each parameter influenced the actuator's performance.

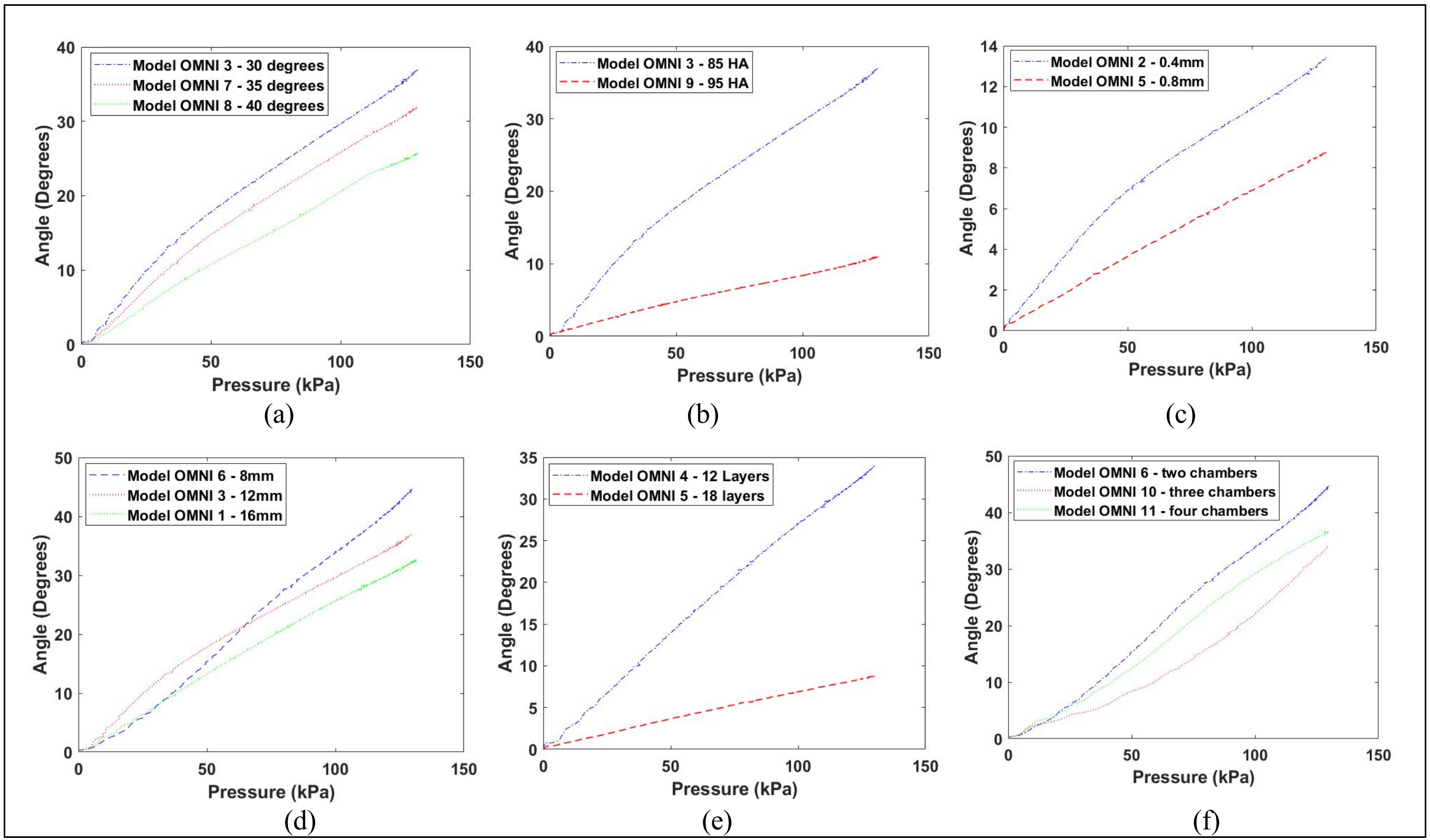

**Figure 6.25:** Impact of geometrical parameters and material stiffness on bending angle of OSSA (a) effect of fold angle (b) effect of material stiffness (c) effect of the tie restraining layers thickness (d) effect of fold width (e) effect of the number of tie-restraining layers (f) effect of the number of chambers.

**Effect of fold angle:** As depicted in Figure 6.25 (a), increasing the fold angle from 30º to 40º resulted in a significant reduction in the bending angle, decreasing from 37º to 25º. This suggests that a larger fold angle negatively impacts the actuator's ability to generate bending motion.

**Effect of material stiffness:** Figure 6.25 (b) illustrates the effect of material stiffness on the bending angle. Increasing the stiffness from 85 Ha to 95 Ha led to a dramatic reduction in the bending angle, from 37º to 11º, indicating that higher material stiffness substantially diminishes the actuator's flexibility and bending performance.

**Effect of tie-restraining layer thickness:** The influence of tie-restraining layer thickness on the bending angle is shown in Figure 6.25 (c). Increasing the thickness of these layers from 0.4 mm to 0.8 mm resulted in a decrease in the bending angle, from 13.5º to 8.7º. This finding suggests that thicker restraining layers constrain the actuator's bending motion, reducing its effectiveness.

**Effect of fold width:** As demonstrated in Figure 6.25 (d), increasing the fold width from 8 mm to 16 mm led to a reduction in the bending angle, from 45º to 32º. This result indicates that a wider fold may limit the actuator's ability to achieve larger bending angles.

**Effect of the number of tie-restraining layers:** The number of tie-restraining layers also significantly affects bending performance. As shown in Figure 6.25 (e), increasing the number of restraining layers from 12 to 18 caused a substantial reduction in the bending angle from 34º to 8.7º. This suggests that a higher number of restraining layers restricts the actuator's movement, thereby reducing its bending capability.

**Effect of the number of air chambers:** The impact of the number of air chambers on bending motion is illustrated in Figure 6.25 (f). Increasing the number of air chambers from 2 to 4 led to a decrease in the bending angle, from 45º to 37º. This finding indicates that while additional air chambers may provide structural support, they may also reduce the flexibility of the actuator, limiting its bending performance.

### 6.4 TSSA

This section presents the experimental results for the TSSA, highlighting its ability to generate rotary motion. The TSSA's performance depends on material stiffness as well as key geometric parameters. To analyze the effects of each parameter on the actuator's rotational behavior, a series of models was designed, constructed, and tested using the previously described experimental platform.

Table 6.4 provides a detailed summary of these models, outlining critical geometric parameters such as radius (*r*), actuator length (*l*), fold width (*fw*), fold angle ($\beta$), thickness of S-layers (*sl)*, twist angle ($\alpha$), wall thickness (*wt*), and Shore hardness (*sh*). These parameters, illustrated in Figure 3.13, were systematically varied to investigate their individual and collective effects on rotational performance. The following subsections discuss the experimental findings, offering insights into how each parameter influences the TSSA's capacity to generate controlled rotary motion.

**Table 6.4:** TSSA models (all dimensions in mm; all angles in degrees)

| *Model* | *r* | *l* | *fw* | *β* | *α* | *wt* | *sh* | *sl* |
|---|---|---|---|---|---|---|---|---|
| ***T1*** | 30 | 80 | 7 | 30 | 270 | 0.96 | 85 | 0.8 |
| ***T2*** | 30 | 80 | 7 | 30 | 270 | 0.96 | 85 | 1.6 |
| ***T3*** | 30 | 80 | 7 | 30 | 180 | 0.96 | 85 | 0.8 |
| ***T4*** | 30 | 80 | 7 | 35 | 270 | 0.96 | 85 | 0.8 |
| ***T5*** | 30 | 80 | 14 | 35 | 270 | 0.96 | 85 | 08 |
| ***T6*** | 30 | 80 | 14 | 25 | 270 | 0.96 | 85 | 0.8 |

### 6.4.1 TSSA Performance during Rotary Motion

The evaluation of the angle of rotation during the rotary motion of the TSSA was conducted using a testing protocol as outlined in Section 6.2.2. The experimental setup used for these measurements is shown in Figure 6.26. For clockwise rotary motion, the proximal end of the TSSA, equipped with an air inlet, was securely clamped within an actuator fixture, while the distal end remained unrestrained to allow free rotational movement. The TSSA was subjected to positive pressure, increasing to a maximum of 75 kPa. The same experimental procedure was followed for counterclockwise rotary motion, with negative pressure applied incrementally up to a maximum of -85 kPa. The models used in the analysis are presented in Table 6.4, and the results are summarized in Figures 6.27 and 6.28.

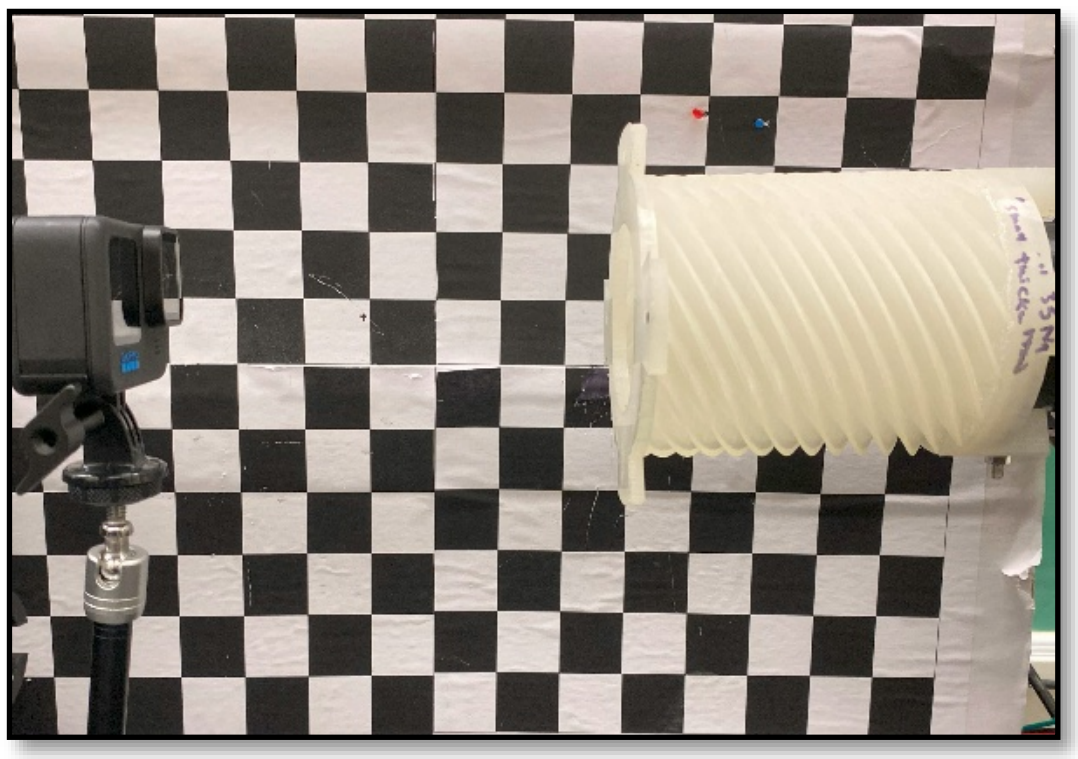

**(a)**

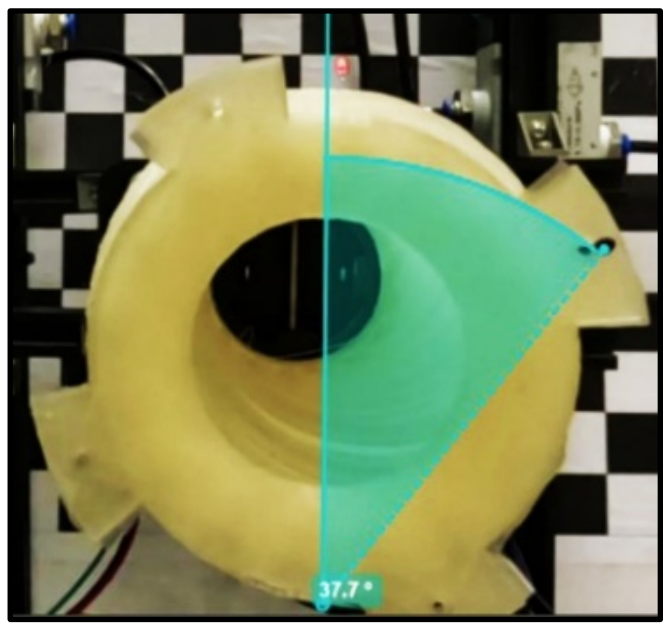

**(b)**

**Figure 6.26:** (a) Experimental setup for measuring angle during Rotary Motion(b) angle measurement using Kinovea software.

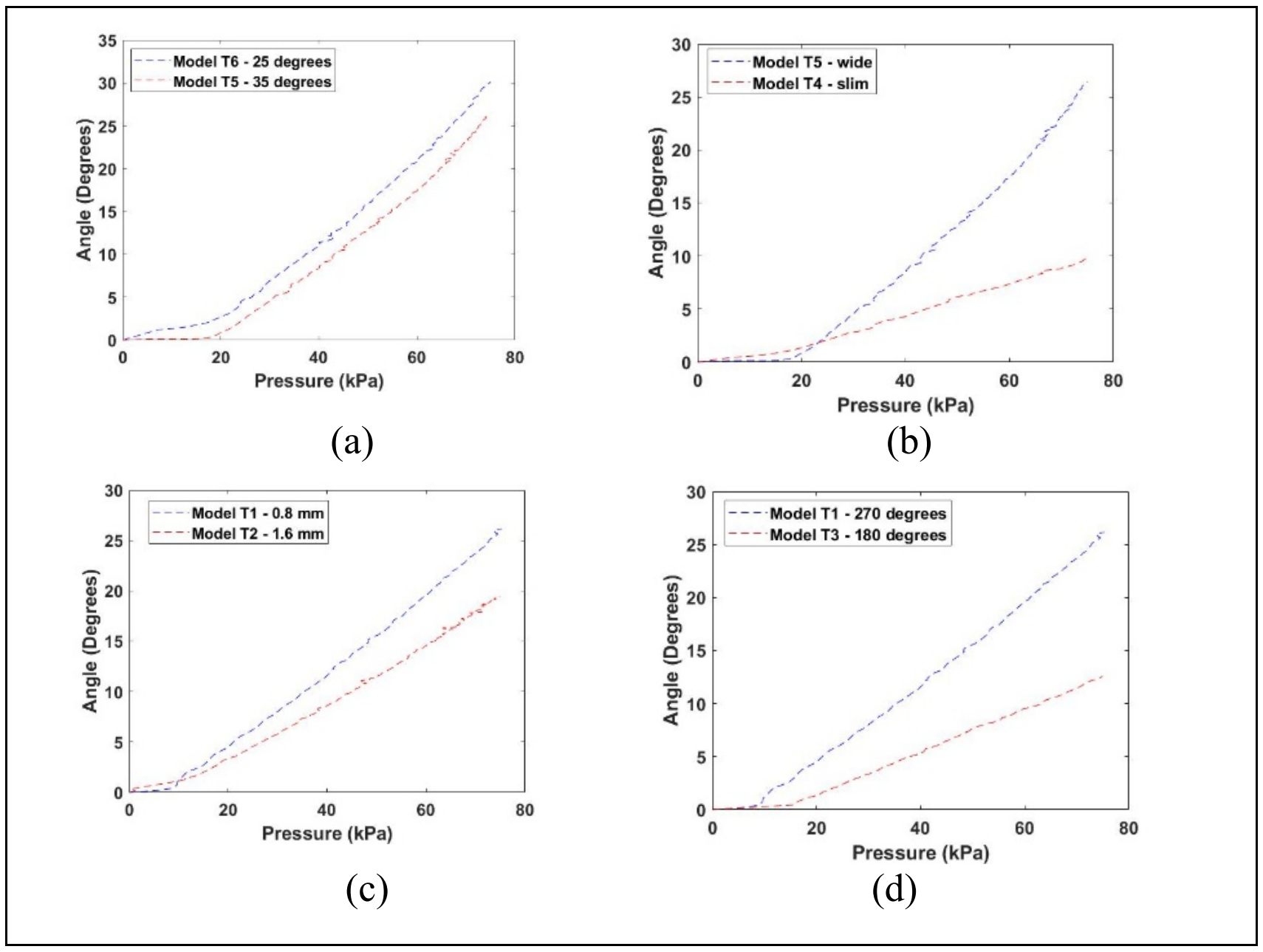

**Figure 6.27:** Impact of geometrical parameters on rotational angle in clockwise rotation (a) effect of fold angle (b) effect of fold width (c) effect of s-layer(d) Effect of twist angle.

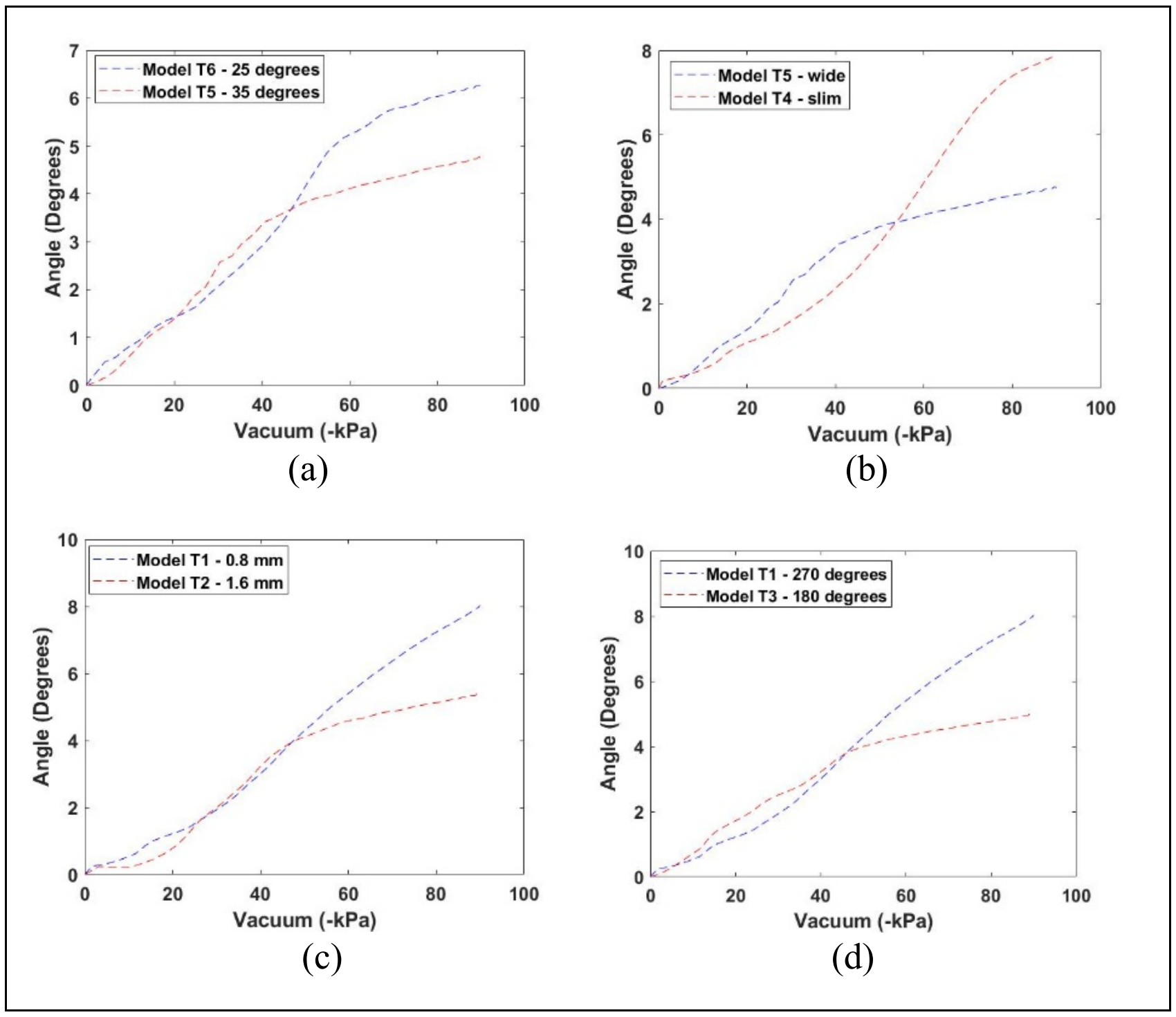

**Figure 6.28:** Impact of geometrical parameters on rotational angle in counterclockwise rotation (a) effect of fold angle (b) effect of fold width (c) effect of s-layer(d) Effect of twist angle.

**Effect of fold angle:** As illustrated in Figure 6.27 (a), increasing the fold angle from 25º to 35º led to a reduction in the rotational angle. For clockwise rotary motion, the angle decreased from 30º to 26º. In counterclockwise motion, a similar increase in the fold angle resulted in a reduction of the rotational angle from 6.2º to 4.5º. These results indicate that a larger fold angle diminishes the TSSA's ability to generate effective rotational motion in both directions.

**Effect of fold width:** As shown in Figure 6.27 (b), increasing the fold width from 7 mm to 14 mm significantly impacted the rotational motion. For clockwise rotary motion, the rotational angle decreased from 30º to 26º. Similarly, in counterclockwise motion, the same increase in fold width resulted in a reduction of the rotational angle from 8º to 5º. This suggests that while fold width is crucial in determining the range of rotational motion, larger fold widths may limit the overall performance of the TSSA.

**Effect of S-layer:** The influence of S-layer thickness on the rotational angle is depicted in Figure 6.27 (c). Increasing the thickness of the S-layer from 0.8 mm to 1.6 mm resulted in a decrease in the rotational angle from 26º to 19.5º during clockwise motion. For counterclockwise motion, this increase in S-layer thickness similarly reduced the rotational angle from 8º to 5.5º. These findings suggest that thicker S-layers reduce the flexibility of the TSSA, thereby constraining its rotational motion and reducing its effectiveness.

**Effect of twist angle:** As shown in Figure 6.27 (d), increasing the twist angle from 180º to 270º led to a significant reduction in the rotational angle. For clockwise rotary motion, the angle decreased from 26º to 12.5º, while for counterclockwise motion, the rotational angle decreased from 8º to 5º. This suggests that larger twist angles negatively impact the TSSA’s ability to generate rotational motion, likely due to increased internal resistance within the actuator’s structure.

## 6.5 Conclusion

This chapter has provided a comprehensive experimental evaluation of SSAs under extension, contraction, bending, and twisting conditions. The results confirm that key geometric parameters—such as fold angle, wall thickness, and the number of restraining layers—together with material stiffness, significantly influence force output and displacement. In particular, variations in these parameters modulate the stress and strain distributions within the actuator walls, thereby affecting both the magnitude and consistency of the generated forces and motions.

Additionally, testing the SSAs under varying static loads illustrates their operational versatility in realistic conditions. Observations regarding trade-offs between force production and displacement, as well as stiffness characteristics under different pressures, underscore the importance of optimizing design elements to achieve specific kinematic and kinetic outcomes. Overall, the analyses reported in this chapter offer a robust empirical basis that advances the understanding of SSA behavior, paving the way for more refined design and control strategies in future research and development.

# CHAPTER 7

# SSA MODELING

The newly developed SSAs exhibits unique mechanical behaviors that set it apart from existing soft actuators. These distinctive characteristics underscore the importance of gaining a thorough understanding of its operational principles and performance capabilities. To complement the valuable insights derived from experimental testing, preliminary analytical models were developed. These models aim to investigate the influence of critical design parameters on the SSAs' performance and behavior. The analytical framework addresses several key aspects of the SSAs, including its material properties, geometric configurations, static and dynamic responses. By integrating these factors, the models provide an in-depth understanding of the interplay between the actuator’s design features and its mechanical functionality. These models lay the groundwork for optimizing the SSAs' design, enhancing its efficiency, reliability, and adaptability.

## 7.1 Material Model

For the SSA fabrication, the TPU filament was selected from NinjaTek (NinjaTek, USA). TPU material exhibits an exceptional capacity to endure large elastic deformations capable of stretching up to 600% under relatively low mechanical loads. This behavior enables TPU to return to its original shape without undergoing permanent plastic deformation, thereby categorizing it as a hyperelastic material. The highly nonlinear stress-strain behavior of TPU necessitates the application of advanced constitutive material models to capture the mechanical response. Several Hyperelastic models, including the Neo-Hookean, Mooney-Rivlin, Yeoh, Ogden, and Arruda-Boyce models, are used to predict the behavior of Hyperelastic materials like TPU. These models express the strain energy potential $W$ as a function of strain invariants. $I_1$, $I_2$ and $I_3$ , which are mathematically represented in (7.1). Each model distinguishes itself by the specific formulation of this strain energy function:

$$W = W(I_1, I_2, I_3) \quad \textbf{(7.1)}$$

The strain invariants $I_1$, $I_2$ and $I_3$ are calculated through equations (7.3), (7.4), and (7.5), respectively.

$$\lambda = \frac{L}{Lo} = 1 + \epsilon_E \quad \textbf{(7.2)}$$

$$I_1 = \lambda_1^2 + \lambda_2^2 + \lambda_3^2 \quad \textbf{(7.3)}$$

$$I_2 = \lambda_1^2\,\lambda_2^2 + \lambda_2^2\,\lambda_3^2 + \lambda_3^2\,\lambda_1^2 \quad \textbf{(7.4)}$$

$$I_3 = \lambda_1^2\lambda_{21}^2\lambda_3^2 \quad \textbf{(7.5)}$$

Where $\lambda_1$, $\lambda_2$, $\lambda_3$ are the three principal stretches, each defined as the ratio of the deformed length $L$ to the undeformed reference length $Lo$ along the corresponding principal direction.

To determine the hyperelastic material model and its associated parameters for TPU 85 and TPU 95, a series of tensile tests was performed. Test specimens composed of each material were fabricated under the same conditions used in the production of the SSAs, ensuring that the results would be directly applicable. The specimen geometry followed the standardized dimensions prescribed by ASTM D638, as illustrated in Figure 7.1 (c).

These tensile tests were conducted using an Instron Universal Testing Machine (Model 4202). Each specimen was elongated at a controlled rate of 100 mm/min to approximately 660% of its original length under ambient conditions. The resulting stress-strain results as shown in Figure 7.1 (b), were subsequently analyzed using the finite element analysis software Ansys 2022. This platform provides multiple hyperelastic material models and offers advanced curve-fitting capabilities to align experimental data with theoretical predictions. The models listed in Table 7.1 were evaluated to determine the best fit for TPU 85 and TPU 95. The Mooney-Rivlin five-parameter model yielded the most accurate and stable predictions, as illustrated in Figure 7.2.

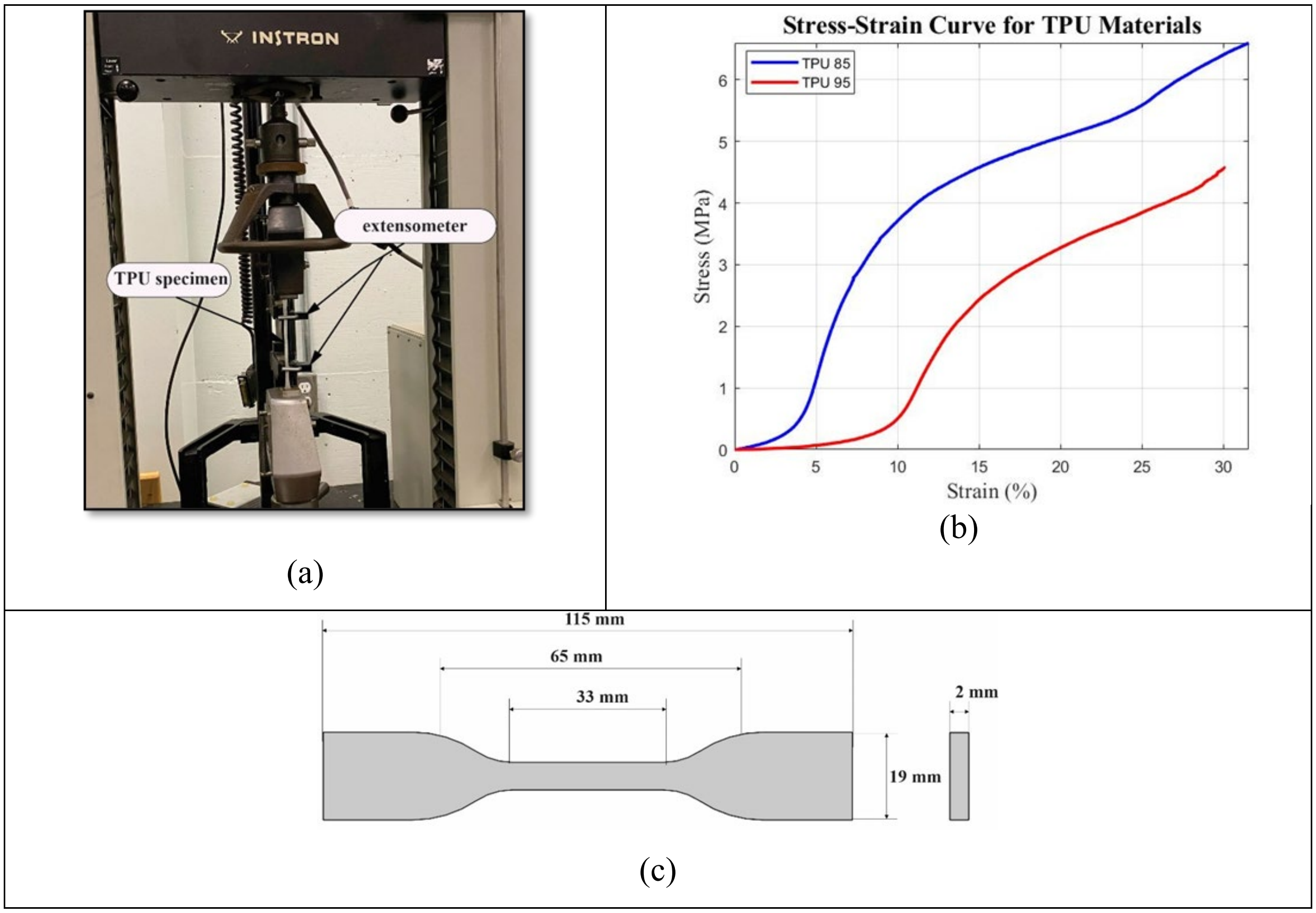

**Figure 7.1:** (a) Tensile test experimental setup (b) stress-strain curve (c) specimen design

**Table 7.1:** Hyperelastic models and their corresponding strain energy functions.

| **Hyperelastic Model** | Incompressible Strain Energy Function |
|---|---|
| **Mooney-Rivlin** | $W=\sum_{i=1}^{2} C_i \ (I_i-3)$ |
| **Yeoh** | $W=\sum_{i=1}^{3} C_i \ (I_1-3)^i$ |
| **Ogden** | $W=\sum_{i=1}^{N} \frac{\mu i}{\alpha i}(\lambda_1^{\alpha i}+\lambda_2^{\alpha i}+\lambda_3^{\alpha i}-3)$ |
| **Neo-Hookean** | $W=C_{10}(I_1-3)$ |

The curve-fitting process in Ansys was used to determine the parameters for the strain energy function, leveraging experimental data. The hyperelastic material model parameters, computed using Ansys Workbench, are listed in Table 7.2.

The Mooney-Rivlin model is particularly suited for capturing the nonlinear stress-strain response of hyperelastic materials. This model extends the Neo-Hookean approach by introducing additional terms that account for higher-order deformation effects, making it more flexible and accurate for materials that exhibit complex behavior under large strains. For the 5-parameter variant used in this study, the strain energy function expands to:

$$W = C_{10}(I_1 - 3) + C_{01}(I_2 - 3) + C_{11}(I_1 - 3)(I_2 - 3) + C_{10}(I_1 - 3) + C_{20}(I_1 - 3)^2 + C_{02}(I_1 - 3)^2 \quad \textbf{(7.6)}$$

where $C_{10}, C_{01}, C_{11}, C_{20}, C_{02}$ are material-specific constants that are determined through curve fitting to experimental data. These parameters reflect the material's response to different modes of deformation and are essential for accurately predicting the material's mechanical behavior. The parameters identified for TPU 85 and TPU 95 using the Mooney-Rivlin 5-parameter model are detailed in Table 7.2. The table also includes the incompressibility parameter $D_1$ which is set to zero, consistent with the assumption of incompressibility.

**Table 7.2:** Lists the hyperelastic material constants identified for TPU 85 and TPU 95 using the Mooney-Rivlin 5-parameter model.

| Material | Model | Parameters | Value (MPa) |
|---|---|---|---|
| TPU 85 | Mooney-Rivlin 5-parameter | C10 | -3.1992 |
| | | C01 | 6.977 |
| | | C20 | 0.0281 |
| | | C11 | -0.074972 |
| | | C02 | 0.92155 |
| | | Incompressible parameter D1 | 0 |
| TPU 95 | Mooney-Rivlin 5-parameter | C10 | -28.763 |
| | | C01 | 42.995 |
| | | C20 | 0.10499 |
| | | C11 | -6.6676 |
| | | C02 | 9.138 |
| | | Incompressible parameter D1 | 0 |

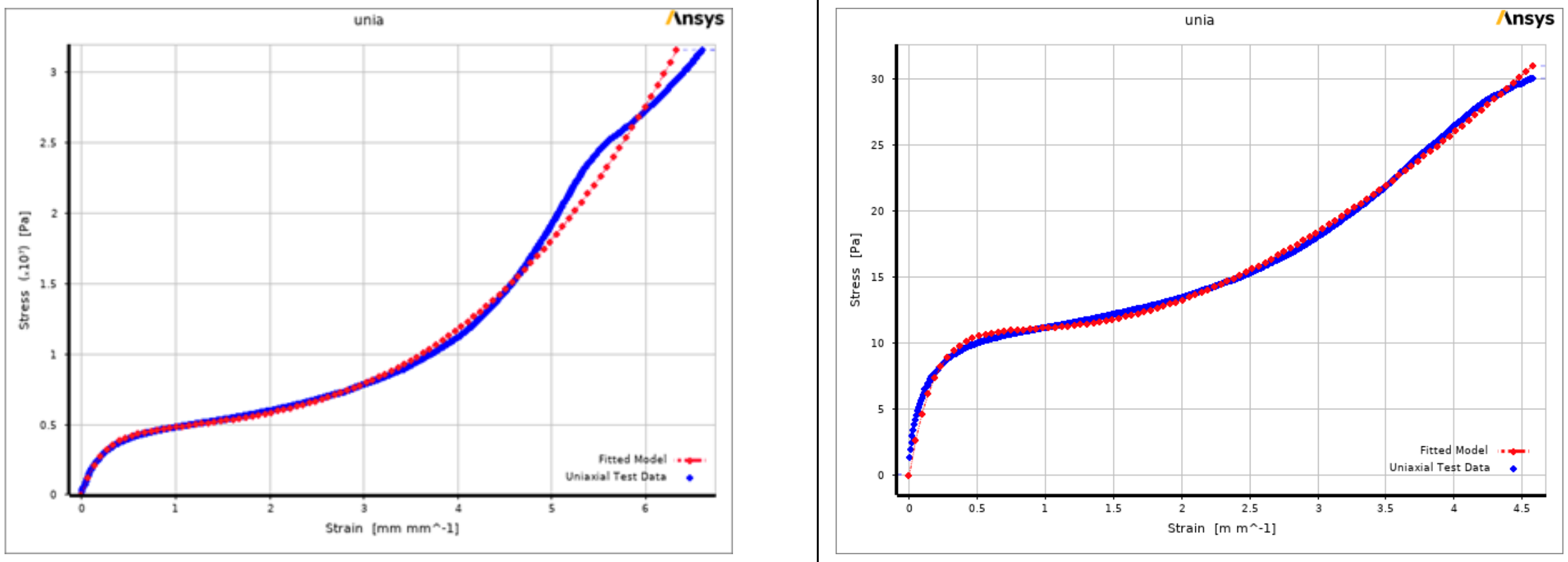

**(a)** **(b)**

**Figure 7.2:** Curve fitting of TPU material using Mooney Rivlin's 5-parameter material model (a) TPU 85 (b) TPU 95

## 7.2 Geometric Models

This section introduces geometric models for the extension, contraction, and bending motions of SSA. These models are formulated based on key parameters, including fold length, fold angle, and actuator width, which play a critical role in determining the actuator's displacement and force generation. The models assume that the SSA's motion arises exclusively from the folding and unfolding of its mechanisms, which function as hinges, while material deformation is omitted.

### 7.2.1 Extension Displacement Model

The actuator's extension displacement is influenced by its geometric parameters and axial stiffness. During extension, each fold contributes to the total displacement based on its initial geometry. As presented in Figure 7.3 (a), the displacement for a single fold during extension is denoted by $\delta_{ext_single}$ can be expressed as:

$$\delta_{ext_single} = 2S(1 - \sin(\theta)) \qquad \textbf{(7.7)}$$

where *S* is the length of the fold. Equation (7.7) captures the relationship between fold geometry and displacement by modeling each fold as part of a right triangle, where *S* is the hypotenuse. The term *2S(1−sin(θ))* represents the effective elongation of the fold when extended. An increase in fold length $S$ leads to a greater extension length, while a larger fold angle $\theta$ reduces the extension due to the resulting shorter hypotenuse.

For an actuator with $N$ folds, the total extension displacement $\delta_{ext}$ is given by:

$$\delta_{ext} = N\delta_{ext_single} \quad \textbf{(7.8)}$$

This relationship implies that extension is linearly proportional to the number of folds, providing a basis for tuning the actuator's design by adjusting the fold length and angle to achieve the desired displacement characteristics.

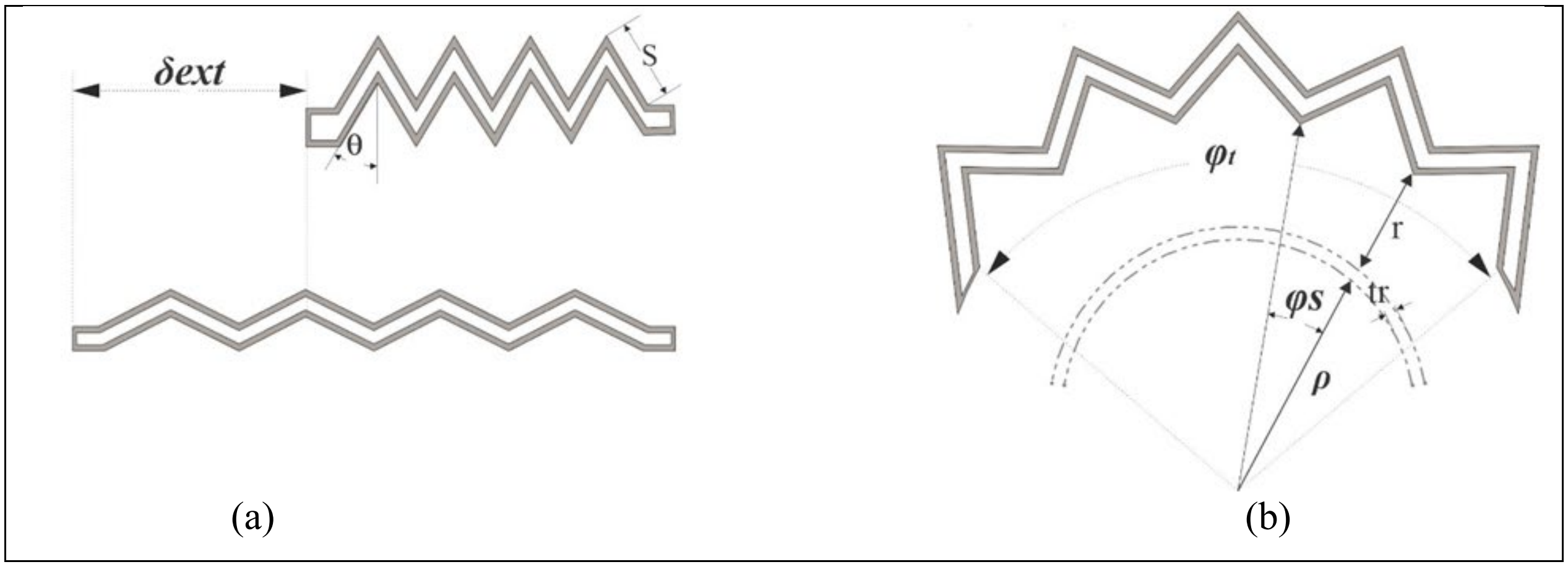

**Figure 7.3:** Two-dimensional wall representation illustrating (a) linear stroke parameters and (b) bending stroke parameters.

### 7.2.2 Contraction Displacement Model

During contraction, the actuator experiences inward motion, where each fold contributes to the total contraction displacement. The contraction displacement for the actuator, denoted $\delta_{Con}$, is given by:

$$\delta_{con} = 2 \times S \times \sin\theta \times N \quad \textbf{(7.9)}$$

This equation indicates that contraction displacement increases with both the fold angle $\theta$ and the number of folds $N$. Unlike the extension, increasing the fold angle in contraction allows for a larger effective displacement, as the fold becomes closer to fully collapsing, maximizing the effective length of each fold in contraction.

### 7.3.3 Bending Displacement Model

In bending motion, the actuator's folds create a curved profile, with extension on the outer side of each fold contributing to the bending angle of the entire structure. The bending displacement model considers the impact of the fold angle on the actuator's ability to achieve a desired bending radius and angle as presented in Figure 7.3 (b).

The bending angle for a single fold $\varphi_s$ can be derived based on the extension of a single fold during bending:

$$\varphi_s = \left(\frac{\delta_{ext_single}}{(\rho + r + tr)}\right)\left(\frac{180}{\pi}\right) \tag{7.10}$$

where $\rho$ is the radius of the central line (Figure 7.3 (b)), $r$ is the radius of the sleeve and $tr$ is the thickness of the concentrating layer. For an actuator with $N$ folds, the total bending angle $\varphi_t$ is:

$$\varphi_t = N\varphi_s \tag{7.11}$$

Determining the radius of curvature $\rho$ is essential for characterizing the actuator's bending deformation. By analyzing the geometry of the actuator during bending (see Figure 7.3 (b)), we can derive an expression for $\rho$ without explicitly using the bending angle $\varphi_t$.

**Inner Side (Constrained Side):** The total bending angle can be expressed in terms of the length of the constrained side $L$:

$$L = \rho\varphi_t$$

$$\varphi_t = \frac{L}{\rho} \tag{7.12}$$

Outer Side (Extended Side):

$$L + N\delta_{ext_single} = (\rho + r + tr)\ \varphi_t \tag{7.13}$$

By substituting (7.12) into (7.13), we get:

$$L + N\delta_{ext_single} = L + \frac{L}{\rho}r + \frac{L}{\rho}tr \tag{7.14}$$

Simplifying and solving for $\rho$:

$$\rho = \frac{L(r + tr)}{N\delta_{ext_single}} \tag{7.15}$$

### 7.3 Static Model

This section presents an analytical model describing the static forces of the LSSA. The model establishes the relationship between the internal pressure $P$ and the output force $Fy$ generated by the actuator. The analysis assumes quasi-static conditions, neglecting dynamic effects such as inertia and damping.

**LSSA output force:**

When a LSSA is subjected to internal pressure $P$, it extends by a displacement $y$ until constrained by external factors, such as physical boundaries or applied loads. At equilibrium, the LSSA generates a net axial force $Fy$, given by:

$$Fy = \sum (F_1 + F_{2y} - F_{3y}) - F_K \tag{7.16}$$

Where $F_1$ is the force generated by the cap, $F_{2y}$ is the axial component of the force generated by the external wall, $F_{3y}$ is the axial component of the force generated by the internal wall, and $F_K$ is the force due to the axial stiffness of the actuator. A force diagram of the LSSA is shown in Figure 7.4.

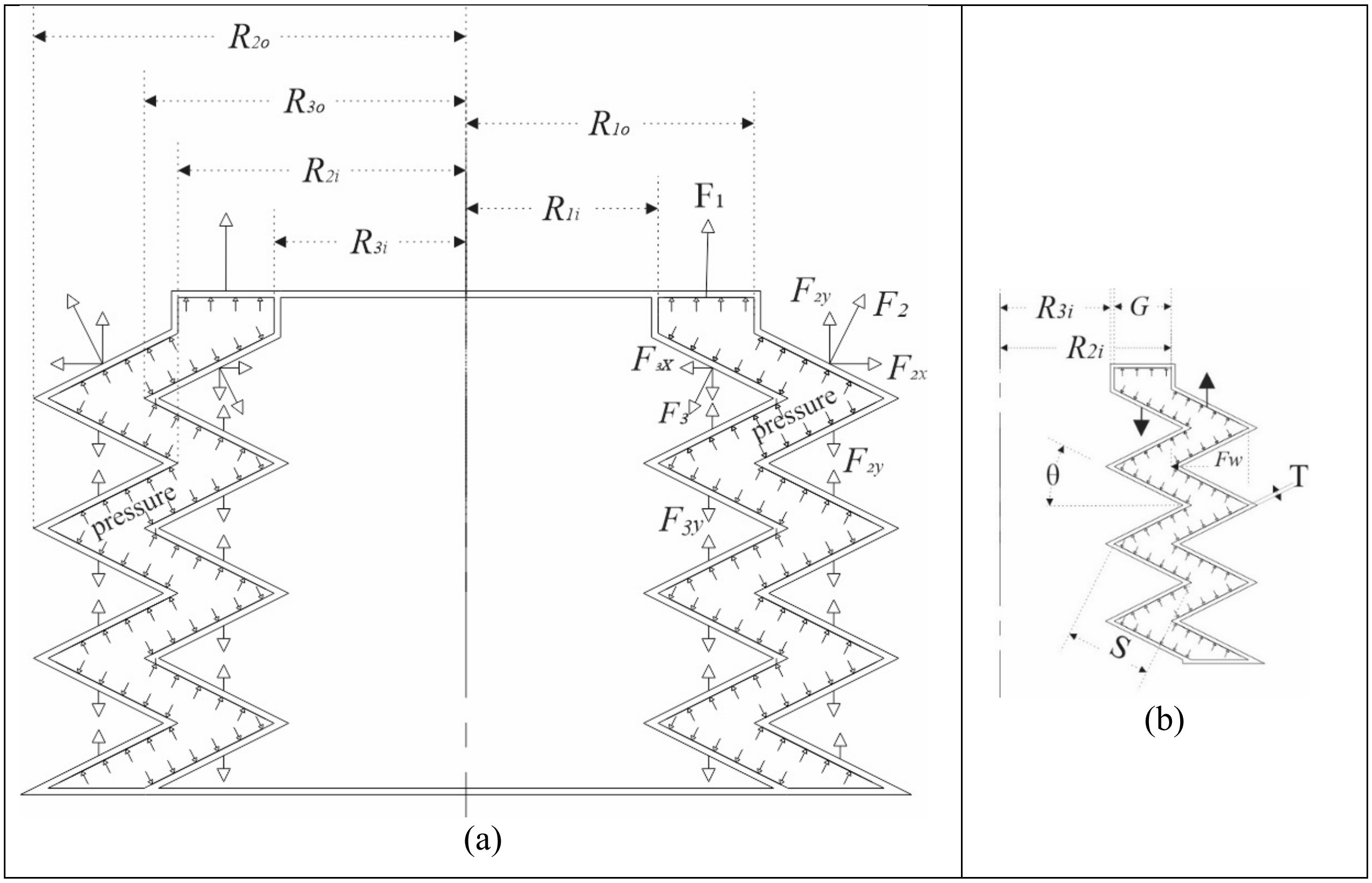

**Figure 7.4:** (a) Geometric representation of the LSSA and the forces acting on it, (b) Detailed wall geometrical parameters

The forces $F_1$, $F_{2y}$ and $F_{3y}$ are calculated as:

$$F_1 = P \times A_1 \quad \textbf{(7.17)}$$

$$F_{2y} = P \times A_2 \quad \textbf{(7.18)}$$

$$F_{3y} = P \times A_3 \quad \textbf{(7.19)}$$

where $A_1$, $A_2$, and $A_3$ represent the projected areas of the cap, external wall, and internal wall, respectively. The cap is considered an annular plate with an external radius $R_{1o}$ and an internal radius $R_{1i}$. The projected area is:

$$A_1 = \pi \times (R_{1o}^2 - R_{1i}^2) \quad \textbf{(7.20)}$$

where $R_{1i}$ is the internal diameter of the cap, and $R_{1o}$ is the external diameter of the cap.

**External Wall Force:**

The axial component of the force generated by the external wall is due to the pressure acting on the effective projected area $A_2$ of these folds. Considering the hollow nature of the actuator, the projected area $A_2$ is calculated as the difference between the areas of two concentric circles:

$$A_2 = \pi(R_{2i} + S \times cos(\theta))^2 - \pi\,(R_{2i})^2 \qquad \textbf{(7.21)}$$

where $R_{2i}$ is the internal radius of the external wall, *S* is the length of the fold (hypotenuse), $\theta$ is the fold angle relative to the horizontal direction.

**Internal Wall Force:**

Similarly, the internal wall contributes an axial force due to the internal pressure acting on its effective projected area $A_3$:

$$A_3 = \pi(R_{3i} + S \times cos(\theta))^2 - \pi\,(R_{3i})^2 \qquad \textbf{(7.22)}$$

where $R_{3i}$ is the external radius of the external wall

**Axial Stiffness Force:**

The actuator exhibits axial stiffness due to its material properties and geometric constraints. The force associated with axial stiffness is a function of the displacement $y$ and is predicted by an experimentally derived polynomial (Figure 7.5). The axial stiffness force $FK_{actuator}$ is given by:

$$FK_{actuator} = a\,y^3 + b\,y^2 + c\,y + d$$

where *a, b, c* and *d* are empirical constants determined through experiments. For LSSA Model L13:

$$FK_{actuator} = 4.1481 \times 10^{-4}\,y^3 + 1.2865 \times 10^{-2}\,y^2 + 2.0789\,y - 0.2246 \qquad \textbf{(7.23)}$$

The axial stiffness of the actuator, denoted as $K_{axial}$, can be expressed as the derivative of the axial stiffness force with respect to displacement ($y$):

$$K_{axial} = \frac{d}{dy} FK_{actuator}$$

$$K_{axial} = 1.24443 \times 10^{-3}\, y^2 + 2.5730 \times 10^{-2}\, y + 2.0789 \qquad \textbf{(7.24)}$$

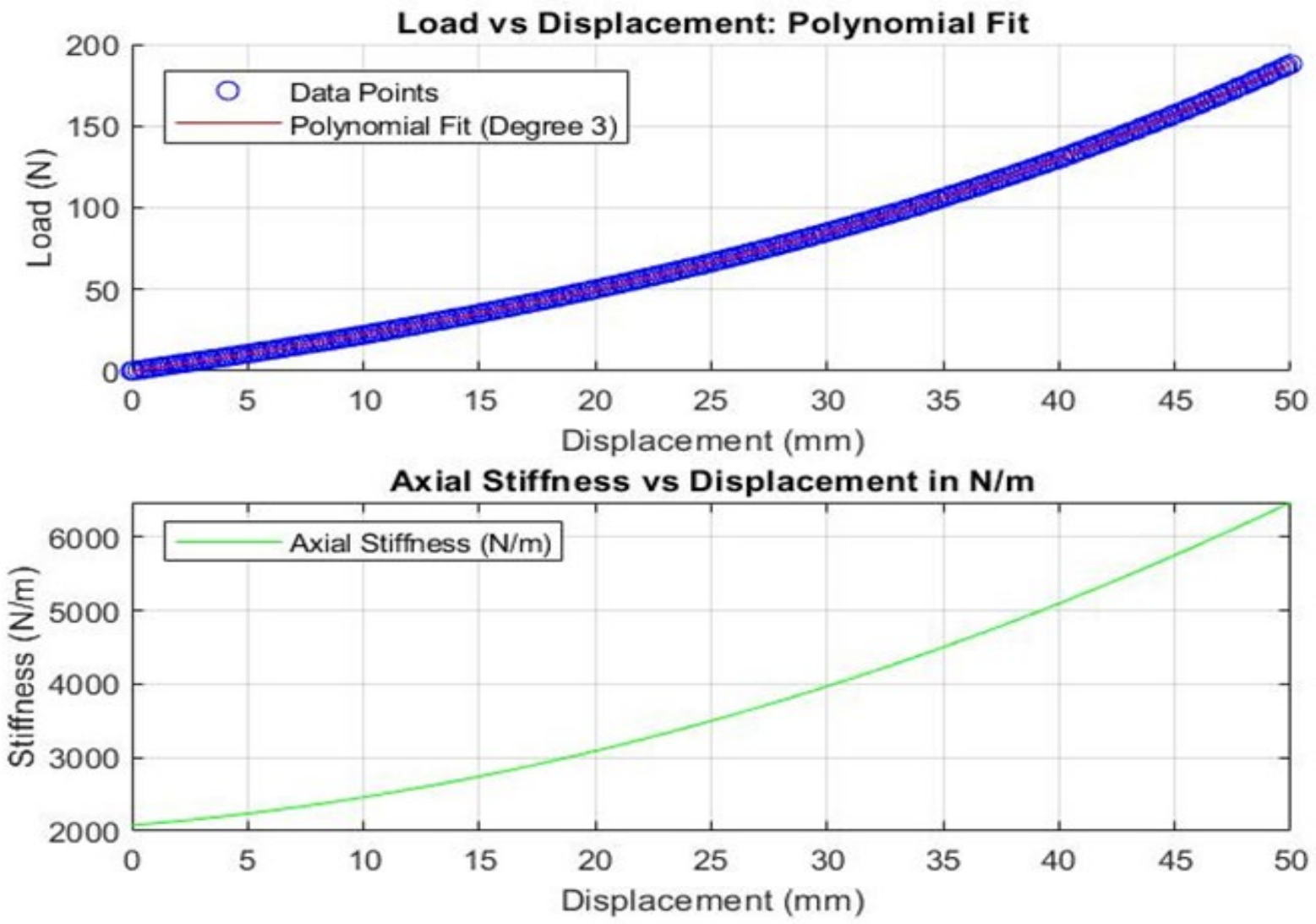

**Figure 7.5:** Axial stiffness polynomial model

By substituting (7.23) into (7.16), the LSSA static model is expressed as:

$$Fy = F_1 + F_{2y} - F_{3y} - 4.1481 \times 10^{-4}\, y^3 - 1.2865 \times 10^{-2}\, y^2 - 2.0789\, y + 0.2246 \qquad \textbf{(7.25)}$$

### 7.4 Dynamic Model

To derive the dynamic model of the system, the Newtonian approach was employed using the experimental setup depicted in Figure 7.6 (a). Using Newton's second law:

$$M\ddot{y} = \sum Fy$$

where $M$ is the external mass, $\ddot{y}$ is the acceleration and $\sum Fy$ represents the sum of forces acting along the line of action of the actuator. Substituting the forces acting on the LSSA:

$$M\ddot{y} = -F_d + F_1 + F_{2y} - F_{3y} - FK_{actuator} \quad \textbf{(7.26)}$$

where $F_d$ is the damping force due to the viscoelasticity, given by $b\dot{y}$, where $b$ is the damping coefficient and $\dot{y}$ is the velocity, and $FK_{actuator}$ is the force due to the axial stiffness of the actuator.

Since $F_1, F_{2y}$ and $F_{3y}$ represent the forces induced by the internal pressure, as discussed in Section 7.2, equation (7.26) can be expressed as

$$M\ddot{y} = -F_d + P\,(A_1 + A_2 - A_3) - FK_{actuator} \quad \textbf{(7.27)}$$

Substituting $F_d$ and $FK_{actuator}$ in equation 7.27 yields:

$$M\,\ddot{y} + b\dot{y} + (1.24443 \times 10^{-3}\,y^3 + 2.5730 \times 10^{-2}\,y^2 + 2.0789\,y) = (A_1 + A_2 - A_3)P$$

$$\ddot{y} = \frac{1}{M}\big((A_1 + A_2 - A_3)P - b\dot{y} - (1.24443 \times 10^{-3}\,y^3 + 2.5730 \times 10^{-2}\,y^2 + 2.0789\,y)\big) \quad \textbf{(7.28)}$$

where $A_1$, $A_2$ and , $A_3$ are the projected areas of the cap, external wall, and internal wall, respectively, $p$ is the pressure.

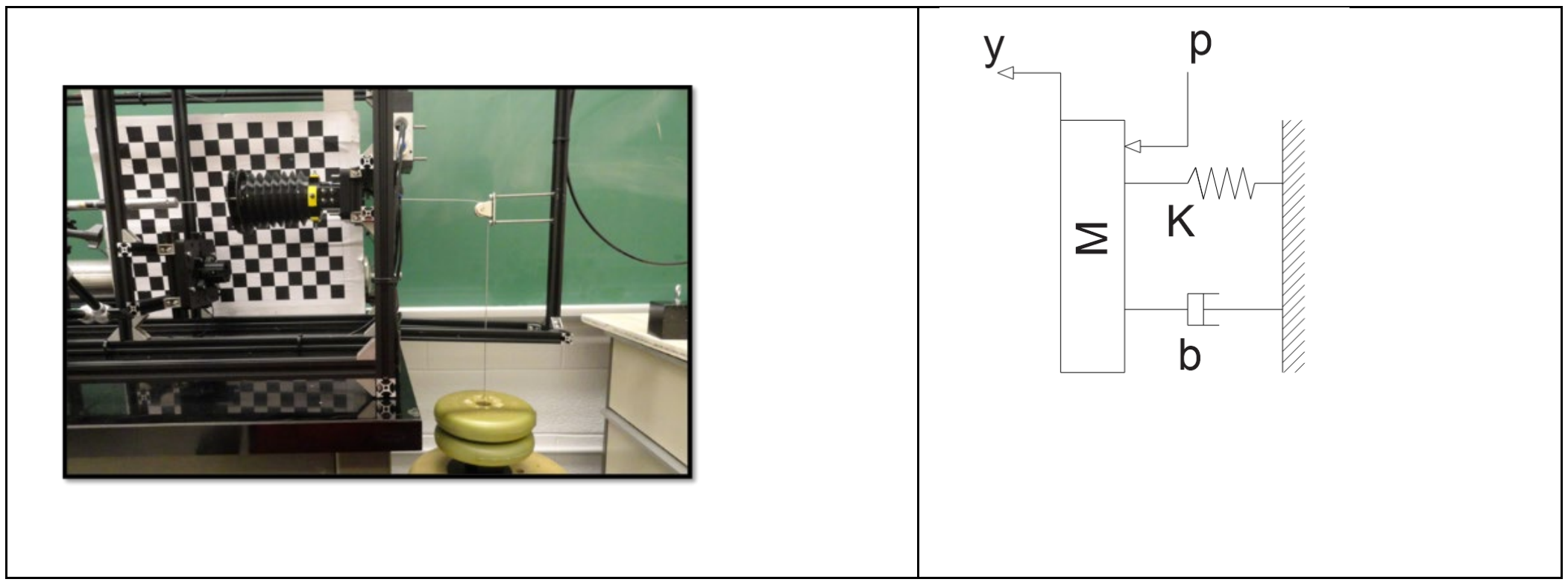

**Figure 7.6:** (a) Experimental setup (b) equivalent mechanical system

### 7.5 Conclusion

In conclusion, this chapter has established a comprehensive modeling framework for SSAs by presenting a hyperelastic material model, geometric models, a static model, and a dynamic model. The hyperelastic material model describes the nonlinear stress–strain behavior of the elastomeric components, while the geometric models illustrate how key structural parameters—such as fold angle, wall thickness, and restraining layers—govern the actuator's motion and force distribution. The static model predicts the SSAs' force–displacement relationships under steady loads, and the dynamic model extends this analysis to time-dependent behaviors, offering insights into the system's performance under varying pressures and control inputs.

By integrating these theoretical perspectives with the empirical findings from earlier chapters, this framework offers valuable guidance for refining both actuator design and control strategies. Collectively, these models provide essential predictive capabilities for optimizing SSA performance and lay a robust foundation for more advanced applications in wearable mobility assistance and other domains requiring compliant, high-performance actuation.

# CHAPTER 8

# LSSA DYNAMIC RESPONSE

The newly developed SSAs are innovative actuators that combine flexibility and adaptability to meet the demands of WMAD. Unlike traditional actuators, SSAs exhibit unique mechanical properties, and many aspects of their behavior and capabilities remain underexplored, necessitating further detailed investigation and analysis. This chapter highlights the SSA's capability to achieve dynamically controlled functions, such as maintaining position accuracy, following complex trajectories, and responding promptly to input signals. These attributes are essential for WMAD and soft robotics systems, where precision and responsiveness are critical.

## 8.1 Controller

In this research, a Proportional–Integral–Derivative (PID) controller was employed to regulate the displacement generated by the LSSA. The controller comprises three fundamental terms: the proportional term, which provides corrective action proportional to the instantaneous error; the integral term, which accumulates past errors to eliminate steady-state offsets; and the derivative term, which responds to the rate of change of the error to predict future trends and enhance dynamic stability. Mathematically, the PID control law is given by:

$$u(t) = K_p\, e(t) + K_i \int_0^t e(\tau)\, d\tau + K_d \frac{de(t)}{dt} \tag{8.1}$$

where $K_p$, $K_i$ , $K_d$ are the proportional, integral, and derivative gains, respectively, and $e(\tau)$ denotes the error signal at time $t$.

The PID controller was designed using the root-locus technique. MATLAB's Control System Designer tool was employed to define the design specifications—a rise time of 1 second and a settling time of 2 seconds—and to tune the individual PID components. This approach ensures that the controller achieves accurate, stable, and robust performance.

## 8.2 Accuracy of the LSSA

Actuator accuracy is important in applications that require precise motion control. In this study, the actuator's accuracy was assessed using an experimental setup for linear displacement testing, incorporating a high-precision laser sensor and a proportional valve for airflow regulation. A Proportional-Integral-Derivative (PID) controller was implemented.

The LSSA model 13 was commanded to move to four specific positions: 10 mm, 20 mm, 30 mm, and 40 mm. Accuracy was quantified using the root mean square error (RMSE) between the actuator's actual position and the commanded position. The results revealed a maximum RMSE of 51.9 µm, demonstrating the actuator's high accuracy and ability to achieve fine positioning.

A series of experiments were conducted to assess the actuator's stability in maintaining a specific position over an extended period. In the initial experiment, the actuator was instructed to move from its starting position to successive increments of 10 mm, holding each position for two minutes before returning to the initial point (Figure 8.2 (a)). This procedure was subsequently repeated using smaller displacement steps of 5 mm and 500 µm, as depicted in Figures 8.2 (b) and 8.2 (c), respectively. For each of the three displacement levels, the RMSE was calculated, yielding a value of 46.5 µm, which indicates minimal positional fluctuation during the tests. While this level of stability is adequate for many applications, it was influenced by the limitations of the proportional valve, which introduced some micro-level deviations. For applications requiring higher stability, particularly those involving micro-positioning, the implementation of servo valves, which offer greater control precision, would likely reduce the observed fluctuations, thereby providing a more stable positioning system.

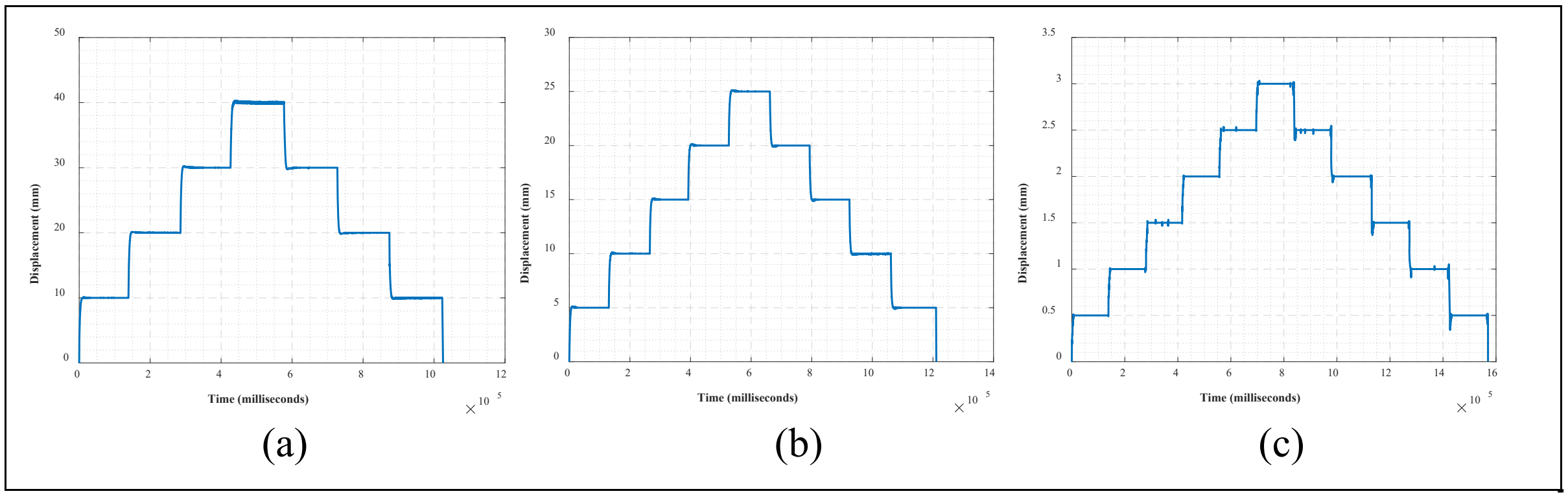

**Figure 8.2:** Position experiment - (a) 10 mm step, (b) 5 mm step, (c) 500 µm step

## 8.3 Frequency Response and Bandwidth Analysis of the LSSA

To characterize the frequency response and determine the bandwidth of the LSSA, an experimental setup was established utilizing precise control and measurement instrumentation. A LabVIEW-based program was developed to modulate the actuator's operating frequency and concurrently measure its displacement output in real-time. Control signals for a 12 V DC-powered solenoid valve were generated using a National Instruments data acquisition system, enabling accurate manipulation of the pressure supplied to the actuator. Actuation frequencies varied from 0.1 Hz to 60 Hz in increments of 0.1 Hz to evaluate the actuator's dynamic performance across a broad frequency spectrum. At the lowest tested frequency of 0.1 Hz, the LSSA model 13 achieved its maximum displacement of 57.5 mm, corresponding to the full stroke length of the actuator. As the frequency increased incrementally, a consistent reduction in maximum displacement was observed, with an approximate decrease of 3 mm for every 0.1 Hz increase. This inverse relationship between frequency and displacement amplitude persisted up to a frequency of approximately 5 Hz, where the displacement diminished to about 3 mm. The observed decrement in displacement with increasing frequency can be attributed to the inherent constraints associated with the pressurization and depressurization processes of the actuator. This includes the compressor, valves, pneumatic line, and SSA air orifices.

The actuator operates on the principle of cyclic filling and evacuation of pressurized air within its internal chambers. At lower frequencies, the duration of each cycle is sufficient for the air pressure to reach equilibrium, allowing the actuator to extend fully. However, as the frequency increases, the time allotted for each cycle decreases, preventing the pressure from building up to the levels required for full extension. This results in incomplete pressurization and reduced displacement amplitude, as the actuator cannot complete its stroke before the cycle reverses.

At frequencies exceeding 20 Hz, the solenoid valve's mechanical and electrical response limitations become significant. The pulse firing rate surpasses the valve's maximum excitation rate, causing it to remain continuously open due to its inability to respond to the rapid switching signals. Under these conditions, the actuator remains fully pressurized without exhibiting any dynamic movement, effectively behaving as if the valve is permanently activated. This highlights the actuator's inability to respond to high-frequency input signals when constrained by the mechanical and electrical limitations of the solenoid valve. Employing a valve with faster response characteristics or an always-open configuration in such scenarios would alter the system dynamics, potentially leading to continuous venting of the pressurized air and zero net displacement due to the absence of pressure differentials required for actuation.

The bandwidth of the actuator is a critical parameter defining the frequency range over which the actuator can effectively convert input signals into mechanical displacement. It is conventionally defined as the frequency at which the output displacement amplitude falls to 70.7% (or equivalently, -3 dB) of its maximum value. In this experimental investigation, it was observed that at a frequency of approximately 0.65 Hz, the actuator's displacement magnitude decreased from an initial value corresponding to -7.81 dB to -10.88 dB, indicating the -3 dB cutoff point. This signifies that beyond 0.65 Hz, the actuator's output is significantly attenuated, and its ability to produce meaningful displacement is compromised due to the limitations imposed by the dynamic response characteristics of the entire system, including the valves and pneumatic lines.

### 8.4 LSSA Time Response

To assess the time response characteristics of LSSA, an experimental setup was configured, incorporating a high-precision laser displacement sensor, solenoid and proportional valves, and a LabVIEW-based data acquisition system. The laser displacement sensor provided high temporal resolution measurements of the actuator's displacement, enabling precise tracking of its dynamic behavior. The valves were employed to accurately control the pressurization and depressurization cycles of the actuator. LabVIEW software facilitated the experimental protocol, acquisition of time-displacement data, and analysis of the time-displacement relationship, thereby allowing for a comprehensive evaluation of the actuator's dynamic performance.

The testing protocol involved pressurizing the L13 model of the LSSA and measuring the rise time, defined as the duration required for the actuator to transition from 10% to 90% of its maximum displacement. After achieving maximum displacement, the actuator was held in this position for a predetermined duration before being depressurized. The time required for the actuator to return to its initial position following depressurization, referred to as the decay time, was also recorded. Two principal experiments were conducted in this study:

**Influence of Valve Type on Time Response:** This experiment investigated the effect of valve type, namely, solenoid versus proportional valves, on the actuator's rise and decay times. The solenoid valve operates through rapid electromagnetic actuation, providing quick on-off control. In contrast, the proportional valve offers variable flow control via motorized adjustment, potentially influencing the actuator's responsiveness due to its ability to modulate airflow rates.

**Influence of Axial Stiffness on Time Response:** This experiment examined how variations in the actuator's axial stiffness affect its time response during pressurization and depressurization phases. Axial stiffness was modified by altering the actuator's structural parameters, such as increasing wall thickness or incorporating restraining layers, to assess how changes in mechanical properties influence the actuator's dynamic behavior.

The LSSA demonstrated a rapid response, achieving maximum displacement within a relatively short time frame and promptly returning to its initial position upon depressurization.

The type of valve utilized had a significant impact on both the rise and decay times of the actuator. Experimental results indicated that the solenoid valve achieved a minimum decay time of 0.73 seconds during depressurization. However, the rise time during pressurization with the solenoid valve was slightly longer than that with the proportional valve. In contrast, the proportional valve attained a minimum rise time of 0.8 seconds during pressurization but exhibited a considerably longer decay time during depressurization, likely due to its slower adjustment speed. Although the proportional valve slightly improved rise time, its extended decay time may be detrimental in applications requiring rapid responsiveness.

Increasing the axial stiffness of the actuator markedly influenced its time response characteristics. Actuators with higher axial stiffness exhibited longer rise times during pressurization, as evidenced by a reduced slope in the time-displacement curve. This behavior can be attributed to the stiffer actuator requiring more time to deform under pressurization due to its increased resistance to shape change, which is governed by the material's compliance properties. A higher stiffness corresponds to a greater modulus of elasticity, necessitating more energy or extended time under constant pressure to achieve the same level of deformation compared to a more compliant actuator.

Conversely, increased axial stiffness reduced the decay time, as the stiffer actuator returned to its original position more quickly upon depressurization. These actuators demonstrated shorter decay times, characterized by a steeper slope in the retraction phase of the time-displacement curve. Upon depressurization, the elastic energy stored in the stiffer actuator material facilitated a faster return to its original shape. This results in shorter decay times because the material's higher stiffness accelerates retraction, consistent with Hooke's Law, which states that the restorative force is directly proportional to the stiffness and displacement. The enhanced elastic recovery of the stiffer actuator suggests that it can release stored energy more rapidly, thereby increasing the speed at which it returns to its undeformed state.

## 8.5 Trajectory Experiments

Experiments involving step, ramp, and sinusoidal paths were conducted to evaluate the capability of the LSSA to follow specific trajectories. These tests aimed to assess the LSSA's dynamic response, tracking accuracy, and robustness under varying conditions. While the experimental setup was similar to that used for accuracy tests, distinct LabVIEW code and testing protocols were developed to address the specific requirements of trajectory experiments.

### 8.5.1 Step Trajectory

To analyze the dynamic response of the LSSA model L13 when subjected to a step input, a step amplitude of 30 mm was applied. The actuator exhibited a rise time of 0.8 seconds and a settling time of 1 second. As shown in Figure 8.3 (a), the steady-state error was zero, attributed to the integral component of the PID controller.

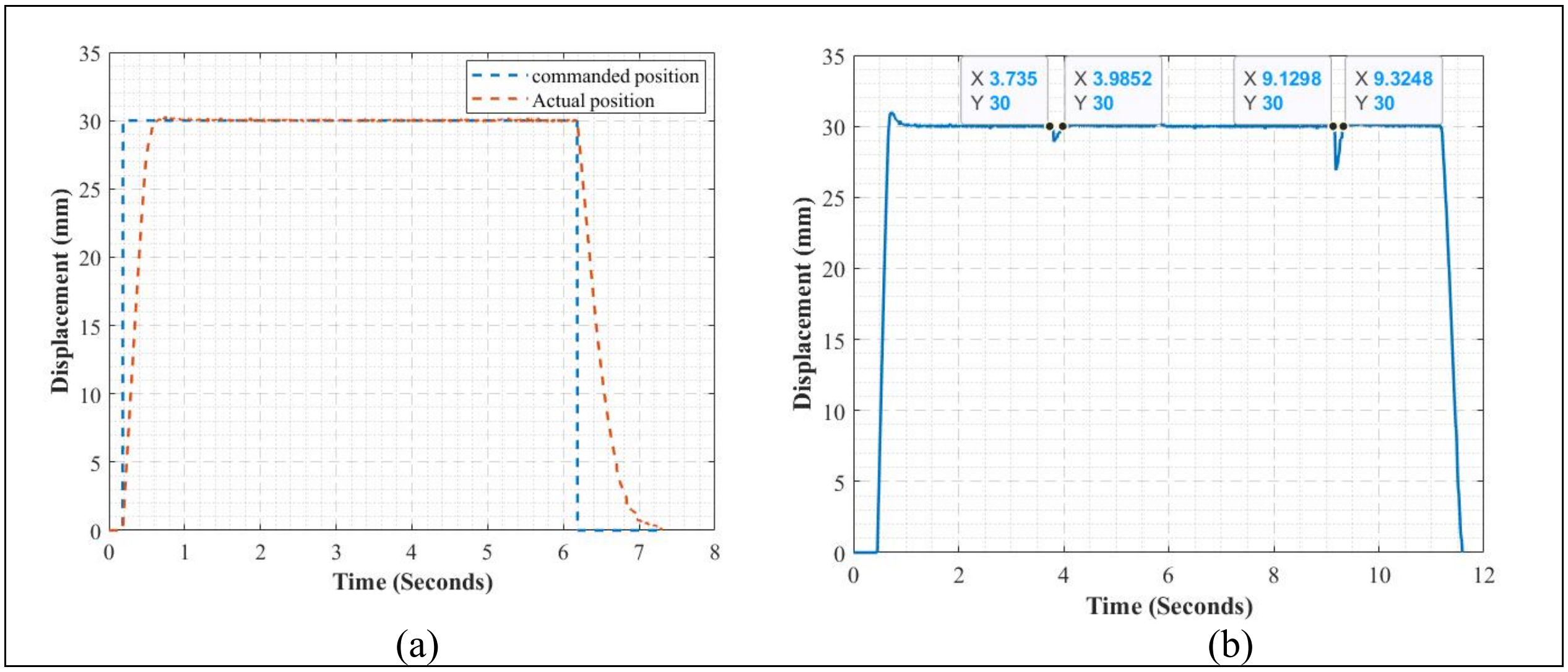

**Figure 8.3:** (a) Step trajectory: commanded vs. actual position (b) load disturbance response

To evaluate the actuator's performance under load disturbances, the actuator was controlled to maintain a constant displacement of 30 mm. Subsequently, two different masses, 2 kg and 4.5 kg were applied. As depicted in Figure 8.3(b), the actuator demonstrated a significant capability to recover the target position after applying the load, with a recovery time of approximately 0.2 seconds. The magnitude of the applied load had a minor impact on the recovery time due to its association with the valve's flow rate; however, it had a considerable effect on the displacement caused by the disturbance. Notably, the load disturbance did not induce steady-state error or instability in the actuator, as evidenced in Figure 8.3(b).

### 8.5.2 Ramp Trajectory

To study how the LSSA responds to a steadily increasing input, ramp trajectory experiments were conducted. Unlike step inputs, which introduce sudden changes, ramp inputs gradually change the control variable (e.g., position) over time. These experiments assess the actuator's tracking accuracy and ability to follow a continuously increasing displacement, represented as a linear slope in the desired trajectory.

The LSSA model 13 was commanded to follow a ramp trajectory. Figure 8.4(a) illustrates the actuator's ability to smoothly track the ramp trajectory with negligible steady-state error at a ramp slope of 1.5 mm/s. However, at higher speeds with ramp slopes of 3 mm/s and 6 mm/s, steady-state errors of 0.4 mm and 0.8 mm were observed, respectively.

While the PID controller effectively eliminates steady-state error for step inputs, it cannot do so for ramp inputs because ramp tracking requires an additional order of integration to achieve zero steady-state error.

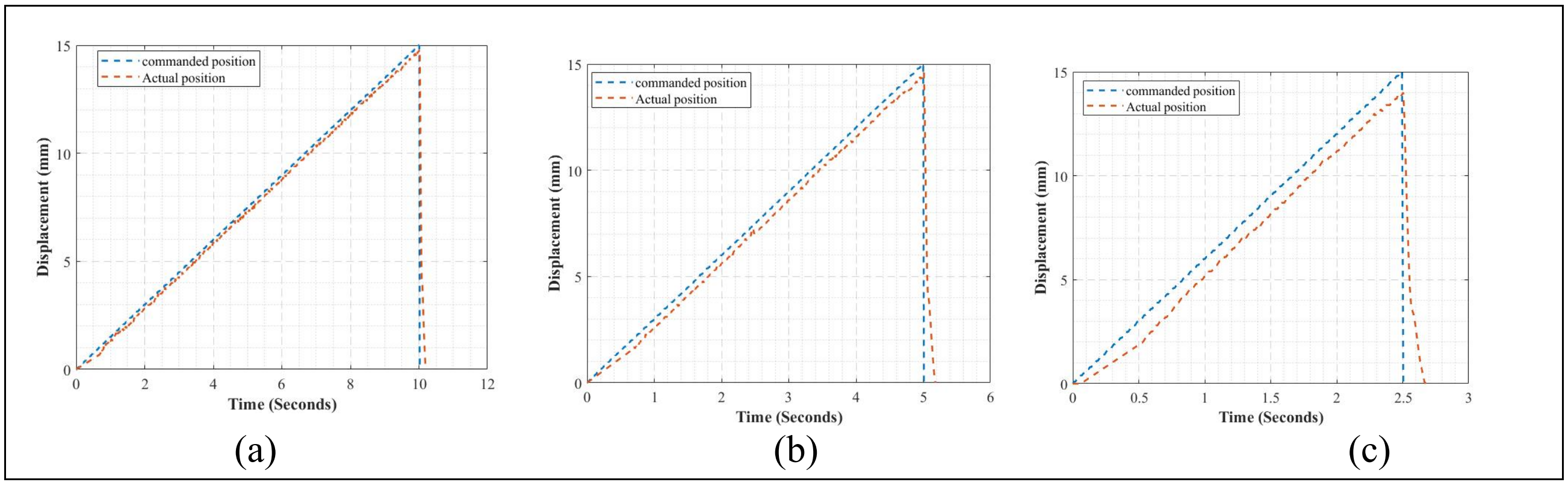

**Figure 8.4:** (a) Ramp trajectory tracking at 1.5 mm/s (b) ramp trajectory tracking at 3 mm/s (c) ramp trajectory tracking at 6 mm/s

### 8.5.3 Sinusoidal Trajectory

To evaluate the LSSA's response under sinusoidal trajectories, the actuator was commanded to move along a sinusoidal path with an amplitude of 10 mm and an offset of 15 mm at different speeds. As shown in Figure 8.5(a), at low speeds, the LSSA model 13 was capable of following the sinusoidal path with negligible steady-state error. However, as depicted in Figure 8.5(b), increasing the actuator speed by five times introduced a shift in amplitude of 0.7 mm and a phase shift of 19.5 degrees. Furthermore, Figures 8.5(b) and 8.5(c) demonstrate that the actuator required three cycles to reach steady-state performance. While this experiment highlights the limitations of the actuator in following sinusoidal paths at high speeds, these limitations are due to the frequency response of the valve rather than the actuator itself.

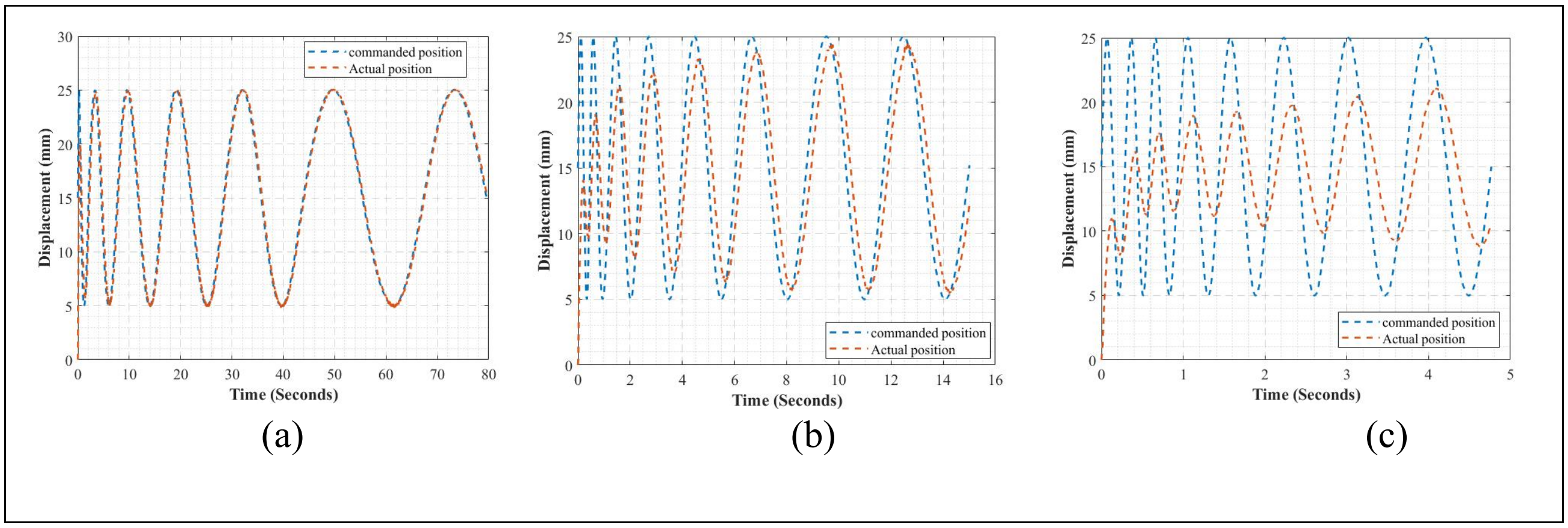

**Figure 8.5:** Tracking performance of the actuator under sinusoidal trajectory at different speeds (a) low speed, (b) moderate speed, (c) high speed

## 8.6 Conclusion

In conclusion, this chapter has presented a detailed examination of the SSAs dynamic performance and control methodologies, with particular emphasis on the LSSA. A series of experiments incorporating step, ramp, and sinusoidal trajectories revealed consistent and precise motion control across various operating speeds. Implementing a PID controller significantly reduced overshoot and steady-state error, thus affirming its suitability for real-time closed-loop applications.

Additionally, frequency response analyses offered valuable insights into the actuator's bandwidth and its capacity to address typical motion requirements in wearable mobility assistance. When interpreted alongside the modeling frameworks introduced in earlier chapters, these findings underscore the critical role of well-tuned control systems in optimizing both precision and responsiveness.

# CHAPTER 9

# DISCUSSION

## 9.1 LSSA Performance

The LSSA developed in this study demonstrates distinct performance characteristics compared to conventional linear soft actuators. One of its primary advantages is its bidirectional functionality, enabling both extension and contraction, whereas most existing designs provide only a single mode of actuation—either extension[71] or contraction [93]. Another notable strength arises from the actuator's folded bellows structure, which greatly reduces the volume of pressurized air required for operation relative to earlier systems [71], [93]. In extension mode, the L8 model achieves a force output 28 times higher than that documented in [94] and 12 times higher than that in [95]. With respect to extension displacement, the L8 design surpasses [94] by 2.63%, exceeds [96] by more than 239%, and outperforms [97] by 510%. Furthermore, in contraction motion, the L6 model generates a force 190% greater than that reported in [93].

Additionally, traditional soft linear actuators pose substantial challenges for wearable applications, primarily due to difficulties in achieving secure attachment. Typically, these actuators function as separate units that transmit force to a joint through cables [71], thereby limiting both portability and ease of use. In contrast, the LSSA can be easily attached to the wearer, owing to its hollow design, which simplifies alignment with the limb and effectively addresses the limitations observed in existing linear actuators.

### 9.1.1 Impact of Design Parameters Force Generation

The force output of the LSSA is linked to its geometric configurations and axial stiffness properties. As depicted in Figures 6.2 (a) and 6.5 (a), an increase in the fold angle adversely impacts the force generation capability of the actuator. A larger fold angle diminishes the projected area over which the internal pressure exerts force, thereby reducing the effective axial force produced by the external wall.

Axial stiffness, defined as the resistance to deformation along the longitudinal axis of the actuator, is a critical factor influencing force output. It is modulated by several parameters, including material stiffness, wall thickness, the number of restraining layers, fold width, and fold angle. An augmentation in any of these parameters increases the axial stiffness, as evidenced in Figure 6.11. This increase in stiffness inversely affects the actuator's net force output, as a stiffer actuator requires more force to achieve the same deformation, thereby reducing the net force available for external work. This inverse relationship is further elucidated in Sections 6.1.1 and 6.1.3. It is imperative to recognize that the force associated with axial stiffness is displacement-dependent, increasing proportionally with displacement, as described by (7.23). At the actuator's resting length and during minimal displacements, the influence of axial stiffness is negligible. However, axial stiffness becomes increasingly significant as the actuator extends, opposing the internal pressure-induced forces.

### 9.1.2 Impact of Design Parameters on Extension Displacement

The extension displacement of the LSSA is also contingent upon its geometric parameters and axial stiffness. Increasing the fold width while maintaining a constant fold angle elongates the fold length *S*, thereby enhancing the actuator's stroke length, as indicated by (7.8). Conversely, increasing the fold angle reduces both the range of motion and the projected area, significantly diminishing the actuator's displacement-generating capabilities.

Parameters such as wall thickness, material stiffness, and the number or thickness of restraining layers, although not directly altering the extension displacement, effectively increase the axial stiffness of the actuator. A heightened axial stiffness necessitates a greater actuation pressure to achieve an equivalent displacement, as illustrated in Figure 6.4. For example, increasing the material stiffness from 85 Shore A hardness to 95 Shore A results in a displacement reduction from 52.7 mm to 13.4 mm at the same pressure. This reduction indicates that while the actuator with higher stiffness can potentially achieve the same displacement, it requires substantially higher pressure to overcome the increased axial stiffness.

### 9.1.3 Impact of Design Parameters on Contraction Displacement

In contrast to extension motion, contraction displacement exhibits a different dependency on geometric parameters. Increasing the fold angle in contraction motion theoretically leads to an increased range of motion, as shown in (7.9). However, experimental observations presented in Figure 6.6 (a) reveal that an increased fold angle diminishes the contraction displacement. This counterintuitive result is primarily due to the larger fold width reducing the projected area of the actuator and simultaneously increasing its axial stiffness (Figure 6.11 (b)). The combined effect necessitates a higher actuation pressure to achieve the same level of contraction, thereby reducing the displacement efficiency.

Similar to extension motion, increasing the wall thickness, material stiffness, or the number and thickness of restraining layers elevates the axial stiffness. The heightened axial stiffness requires greater actuation pressure to induce contraction, ultimately leading to a reduced contraction displacement, as demonstrated in Figure 6.6.

### 9.1.4 Impact of Static Load on Actuator Performance

The presence of a static load influences the performance of the LSSA, particularly at low actuation pressures, as analyzed in Sections 6.1.5 and 6.1.6. At low pressures, the force generated by the actuator is relatively modest, rendering it more susceptible to the opposing effects of static loads. The actuator must overcome both its inherent axial stiffness and the external static load, which can markedly reduce displacement. Conversely, at higher actuation pressures, the generated force becomes substantial enough to mitigate the impact of static loads, allowing the actuator to overcome additional external resistance with minimal performance degradation.

Experimental data indicate that increasing the static load does not alter the slope of the force-pressure relationship curve (Figures 6.7 (b) and 6.9 (b)). This observation suggests that the relationship between force and pressure remains constant despite variations in static load, implying that the static load primarily affects the displacement magnitude rather than the fundamental force-generation mechanism.

### 9.1.5 Relationship Between Extension and Force Generation

The experiments detailed in Section 6.1.9 provide critical insights into the dynamic relationship between actuator extension and force generation. The findings reveal that the output force of the LSSA diminishes as the actuator extends. This phenomenon is attributable to two principal factors: the reduction in projected area and the increase in axial stiffness with displacement.

At the actuator's initial length or during minimal displacements, the projected area of the walls is maximal, and the axial stiffness influence is minimal. Under these conditions, the actuator can generate its peak force output. As the actuator extends, the projected area decreases due to geometric deformation, leading to a reduction in the axial force generated by the walls. Simultaneously, the axial stiffness increases with displacement, further diminishing the net output force.

Despite the reduction in wall-generated force, the actuator continues to produce force owing to the internal pressure acting on the end cap. The net output force reaches zero when the force generated by the cap balances the force required to overcome the axial stiffness of the actuator. This equilibrium point defines the maximum operational extension of the actuator under the specified conditions.

## 9.2 BSSA Performance

The BSSA developed in this study demonstrates notable functional advantages. Specifically, the B13 model produces forces exceeding 38 N, as shown in Figure 6.15(h), and achieves bending angles greater than 140 degrees. In contrast to most conventional soft bending actuators, which are predominantly unidirectional[98], [99] , the BSSA enables multi-directional bending—a capability particularly significant for wearable and biomedical applications.

Although traditional soft bending actuators can be attached more readily than soft linear actuators by mounting them onto a textile layer [61], [82], which is then affixed to the body, this approach still presents challenges. A significant portion of the actuation force is lost as the interface layer stretches rather than being converted into torque. Over time, these layers can also deform, altering the force profile and eventually failing. In contrast, the BSSA's self-contained design eliminates the need for additional layers, thereby enhancing overall stability and reliability.

Notably, the B13 model's force output is four times greater than that reported in [100], 12.3 times greater than that in [101], and 30 times greater than that in [102]. Furthermore, the BSSA's performance is highly customizable and can be enhanced through parameter optimization or by increasing the actuation pressure, offering considerable potential for further refinements.

### 9.2.1 Impact of Design Parameters on Force Generation

The force output of the BSSA is critically dependent on its geometric design parameters, notably the fold angle, tie-restraining layers, and material stiffness. Similar to the LSSA, increasing the fold angle leads to a reduction in the projected area over which the internal pressure acts, thereby diminishing the effective bending force generated by the actuator (Figure 6.15 (a)). Furthermore, an increased fold angle contributes to a rise in the axial stiffness of the actuator. Axial stiffness is influenced by material stiffness, wall thickness, and the geometric configuration of the tie-restraining layers. A higher axial stiffness necessitates a greater force to achieve the same bending displacement, thus reducing the net force output available for external work. This inverse relationship between axial stiffness and force generation highlights the importance of optimizing fold angles to strike a balance between flexibility and structural integrity.

The constraining layers have a negligible effect on force generation because they do not significantly alter the projected area or axial stiffness. This is attributed to their placement outside the primary line of action in the actuator's bending motion. In contrast, increasing the stiffness of tie-restraining layers through additional layers or increased thickness has a substantial impact on force generation, as demonstrated in Figures 6.15 (d) and 6.15 (e).

While material stiffness and wall thickness exhibit a minor impact on force generation at small displacements, their influence becomes more pronounced at larger bending angles. At greater displacements, the increased stiffness of the material and thicker walls contributes to higher axial stiffness, necessitating greater actuation pressure to produce the same force. This observation indicates that material properties become increasingly significant as the actuator operates beyond its initial range of motion.

### 9.2.2 Impact of Design Parameters on Bending Motion

The bending performance of the BSSA is significantly affected by its design parameters, as depicted in Figure 6.16. Increasing the fold angle has a considerable impact on the achievable bending angle. Specifically, a larger fold angle reduces the range of motion due to the geometric relationship between the fold angle and the effective fold length.

Increasing the thickness of the constraining layers has a minimal impact on bending motion because it does not directly influence the range of motion or the projected area. Its effect on axial stiffness is relatively minor compared to other factors. However, increasing either the number or the thickness of tie-restraining layers elevates the axial stiffness of the actuator. This increase in stiffness necessitates higher actuation pressure to achieve the same bending angle, thereby reducing bending efficiency, as shown in Figures 6.16 (e) and 6.16 (f).

Similarly, augmenting the material stiffness or wall thickness reduces the actuator's flexibility, leading to a decrease in the achievable bending angle. The actuator becomes more rigid, requiring greater force to bend, which can limit its applicability in scenarios where large bending motions are essential. In contrast, increasing the fold width enhances the fold length, thereby increasing the range of motion and the bending angle, as shown by (7.10). A larger fold width permits greater deformation under the same actuation pressure, improving the actuator's bending performance.

### 9.2.3 Relationship Between Bending Angle and Force Generation

Experimental results presented in Figure 6.18 indicate that the force generated by the BSSA decreases as the bending angle increases. This behavior can be attributed to two primary factors: the reduction in projected area and the increase in axial stiffness force with displacement.

At small bending angles, the projected area is at its maximum, enabling the internal pressure to generate the maximum possible force. Additionally, the axial stiffness is at its minimum due to the minimal deformation, resulting in lower resistance to bending. As the bending angle increases, the projected area diminishes because of the geometric reconfiguration of the folds, leading to a reduction in the force generated by the actuator. Simultaneously, the axial stiffness force escalates with greater displacement, further diminishing the net force output. This interplay between the decreasing projected area and increasing axial stiffness helps explain the observed decrease in force generation at higher bending angles.

## 9.3 OSSA Performance

Experimental findings confirm that the OSSA introduced in this study exhibits robust performance characteristics. Specifically, the OMNI 6 model is capable of generating forces exceeding 60 N at 130 kPa and achieves linear extensions greater than 81 mm. It can also contract by up to 25 mm while enabling bending motions in all directions within a 360° plane, producing bending forces above 18 N and angles exceeding 45°. In doing so, the OSSA addresses a key limitation of earlier BSSA systems, which typically restrict bending to one or two directions. Both force output and bending range can be further optimized by increasing the actuation pressure or refining key geometric parameters. Notably, in bending mode, the OSSA produces forces 36 times greater than those reported in [103] and 10 times greater than those in [104]. In extension, it achieves three times the displacement and 10 times the force documented in [105] and [106]. A defining advantage of the OSSA is its capacity for contraction, a capability absents in many existing actuator mechanisms.

### 9.3.1 Impact of Design Parameters on Linear Motion

The linear performance of the omnidirectional actuator is influenced by its geometric parameters, particularly the fold angle. An increase in the fold angle reduces the projected area over which the internal pressure acts, thereby diminishing the effective force generated, as depicted in Figure 6.20(a). Additionally, a larger fold angle decreases the range of motion, as shown by (7.8) and (7.9). Furthermore, increasing the fold angle elevates the axial stiffness of the actuator, thereby reducing its flexibility.

As previously discussed, increasing the material stiffness, the number of restraining layers, or the thickness of these layers enhances the axial stiffness of the actuator. An increased axial stiffness requires a higher actuation pressure to achieve the same displacement, leading to a reduction in both the generated force and displacement. This relationship is evident in Figures 6.20, 6.21, and 6.22.

The number of air chambers has a minimal impact on linear motion performance, as shown in Figures 6.20 (f), 6.21 (f), and 6.22 (f). The slight deviations observed are primarily attributed to changes in the number of structural layers (S-layers) required to accommodate additional chambers rather than the number of air chambers. This finding suggests that, for linear motion, the actuator's performance is relatively insensitive to the number of air chambers, provided that the overall structural integrity is preserved.

### 9.3.2 Impact of Design Parameters on Bending Motion

The omnidirectional actuator achieves bending motion through selective pressurization of specific chambers, causing differential extension and contraction, resulting in bending in the desired direction. The influence of geometric parameters on bending motion parallels that observed in dedicated bending actuators.

Increasing the fold angle reduces both the bending angle and the force generated during bending, as illustrated in Figures 6.24 and 6.25. A larger fold angle decreases the projected area available for force generation and increases the axial stiffness, which opposes bending. Consequently, the effectiveness of force generation diminishes, and more force is required to achieve the same bending displacement.

Similarly, increasing the axial stiffness of the actuator by raising material stiffness, adding restraining layers, or thickening existing layers, adversely affects both the bending angle and the force generated during bending, as shown in Figures 6.24 and 6.25. The increased stiffness opposes deformation, necessitating higher actuation pressures to attain comparable bending performance.

The number of air chambers plays a nuanced role in bending performance. Increasing the number of air chambers has a negligible impact on performance as long as the projected area remains constant. For example, an actuator with two air chambers can exhibit similar performance to one with four air chambers if the same proportion of the actuator's circumference is pressurized, pressurizing one chamber in the two-chamber actuator and two chambers in the four-chamber actuator. However, differences arise when only one chamber is pressurized in each actuator. In the two-chamber actuator, pressurizing one chamber affects half of the actuator's circumference, whereas in the four-chamber actuator, it affects only a quarter. This disparity leads to variations in the generated force and bending angle due to the unequal pressure distribution and resultant asymmetric deformation.

The minor deviations observed in actuator performance with varying numbers of air chambers are also attributable to changes in the number of structural layers required to accommodate the additional chambers. These structural modifications can influence the actuator's flexibility and stiffness, albeit to a lesser degree compared to other geometric parameters.

## 9.4 TSSA Performance

The TSSA developed in this study is a bidirectional actuator capable of generating rotary motion in both clockwise and counterclockwise directions. Under positive pressure, the T6 model achieves more than 37 degrees of clockwise rotation and up to 8 degrees of counterclockwise rotation. A noteworthy advantage of this design is its ability to function at relatively low actuation pressures, typically below 50 kPa.

Although numerous soft twisting actuators exist, none have been adopted for wearable applications because their center of rotation cannot coincide with the anatomical center of the biological joint [107], [108]. By contrast, the wearable nature of the TSSA allows it to be aligned with the joint's center of rotation, thereby resolving a critical limitation observed in traditional soft twisting mechanisms.

The actuator's performance is markedly influenced by its design parameters. An increase in the fold angle adversely affects rotational motion in both directions. In clockwise rotation, a larger fold angle reduces the range of motion due to geometric constraints. In counterclockwise rotation, the increased fold angle enhances axial stiffness and reduces the projected area, necessitating a higher actuation pressure to achieve the same rotational displacement.

Conversely, the twisting angle significantly enhances the range of motion without affecting the projected area or axial stiffness. Experimental results demonstrate that increasing the twisting angle by 30% effectively doubles the generated clockwise rotary angle. This indicates that the twisting angle is a critical parameter for optimizing rotational performance.

The thickness of the structural layers (S-layers) has a limited effect on the generated angle, as these layers are not directly involved in the actuator's primary line of action during twisting motion. However, increasing the fold width positively influences the range of motion, resulting in a larger rotary angle. A wider fold permits greater deformation under the same actuation pressure, enhancing the actuator's rotational capabilities.

The experimental findings suggest that the proposed twisting sleeve actuator can achieve higher rotary angles by adjusting key parameters such as the twisting angle and fold width or reducing the fold angle. These design modifications offer significant potential for augmenting the actuator's bending and rotational functionalities, rendering it suitable for applications necessitating an extensive range of motion and flexibility.

### 9.5 Conclusion

In conclusion, this chapter has integrated experimental findings, modeling insights, and control analyses from the preceding chapters to offer a holistic perspective on the Soft Sleeve Actuators (SSAs). By examining how design parameters, material properties, and fabrication methods jointly shape the actuators' performance, it has identified key trade-offs among force output, displacement range, and dynamic responsiveness. This integrated discussion underscores the alignment between theoretical predictions and empirical outcomes, highlighting the importance of tailoring SSA configurations for wearable mobility assistance.

Additionally, the comparative assessment of various actuator architectures has demonstrated the relevance of geometry and advanced control strategies in enhancing both precision and adaptability.

# CHAPTER 10

# CONCLUSION AND FUTURE WORK

## 10.1 Conclusion

This thesis has addressed a critical gap in WMAD by introducing the first soft sleeve actuators and providing a comprehensive study of their development and performance. Soft sleeve actuation mechanisms had not been explored before this work, making this the inaugural study on sleeve mechanisms. These novel actuators demonstrate significant potential for various applications in WMAD and soft robotics.

Several bidirectional mechanisms have been developed to achieve diverse motions, including linear, bending, twisting, and omnidirectional soft sleeve actuators. The linear actuators produce both extension and contraction motions, while the bending actuators enable movement in both clockwise and counterclockwise directions. The twisting actuator facilitates rotary motion in both rotational directions. Moreover, the omnidirectional actuator extends these capabilities, allowing for 360-degree bending and linear movements and showcasing unprecedented versatility in soft actuator design.

A notable characteristic of these actuators is their bidirectional self-restorative mechanism, which enables them to return to their original positions autonomously without external pressure modulation during bending, rotation, contraction, or extension. This self-restorative capability enhances the actuators' efficiency and simplifies control systems, improving their practicality in real-world applications.

The sleeve mechanism's self-contained design offers advantages, especially in wearable applications. Unlike traditional actuators that require secure attachment to the human body, the sleeve mechanism maintains structural stability, enhancing user comfort and ease of integration. Additionally, the proposed soft sleeve actuators exhibit exceptional force generation, with capabilities exceeding 210 N. This high force generation results from a folded bellows design, which enhances responsiveness while reducing energy consumption compared to conventional models and traditional soft actuators.

These benefits stem from key design innovations, including eliminating the central cavity within the actuator and reducing the gap between the internal and external walls. By decreasing the internal volume of the sleeve actuator, operational efficiency, and responsiveness have been improved. This reduction minimizes the compressed air required for actuation, thereby lowering energy consumption and boosting performance.

An experimental platform, as described in Chapter 5, was developed to evaluate the SSAs' performance. Systematic investigations were conducted into the effects of various design parameters on axial stiffness, force generation, and range of motion, as presented in Chapter 6. The findings from these studies provide valuable insights into how specific design modifications impact the actuator's operational capabilities, informing future design optimizations and applications.

Furthermore, this thesis explores diverse manufacturing techniques suitable for producing sleeve mechanisms, critically analyzing their limitations in Chapter 4. To overcome these limitations, a novel manufacturing framework was developed using FDM with Bowden-based 3D printers for fabricating soft pneumatic actuators. This framework has been validated and effectively addresses challenges associated with printing flexible materials and ensuring airtightness in pneumatic systems. The methodology presented offers a scalable and accessible approach for manufacturing soft robotic components, with the potential to accelerate advancements in the field.

## 10.2 Future Work

This thesis explored the development and production of innovative actuators utilizing a newly developed manufacturing process. In addition, it introduced the first mechanical characterization of these actuators alongside preliminary analytical models. This work represents a foundational effort, with many aspects initiated from the ground up. Consequently, there is significant potential for improvement and further research to refine the actuators, enhance their performance, and fully realize their capabilities. The findings of this study provide a strong starting point, paving the way for future advancements and optimization in the field. Some specific future work is noted in this section.

The current methodology for producing these actuators has certain limitations, particularly regarding the flexibility of materials used, as the material stiffness cannot currently be reduced below 85 Shore A. Investigating alternative manufacturing methods to lower material stiffness could enhance the energy efficiency of the actuators without compromising their operational capacity. Additionally, due to inherent manufacturing constraints, the current setup limits the achievable fold angle to 30°. Reducing this angle could proportionally increase the actuator's displacement performance.

Future work should also focus on integrating FEA and numerical simulations to improve understanding and optimization of SLA. These simulations would provide valuable insights into the actuator's behavior under various load conditions, enabling a comprehensive assessment of its structural integrity and operational limits. Moreover, numerical analyses could be applied to optimize design parameters, further enhancing the efficiency and applicability of these actuators.

This research primarily concentrated on the design, fabrication, and mechanical evaluation of novel sleeve actuators. Future research could shift towards adapting these sleeve mechanisms for mobility assistive devices, requiring modifications to the sleeve geometry to accommodate human anatomical specifications. Another potential area for development is the integration of sensory feedback systems within the SLA design, allowing for real-time monitoring and adjustment of actuator behavior. Such enhancements would improve user experience and safety in assistive applications.